%% file: root.tex
\documentclass[letterpaper, 10 pt, journal, twoside]{IEEEtran}

\include{includes}
\title{
MoRe-ERL: Learning Motion Residuals using \\ Episodic Reinforcement Learning 
}

\author{Xi Huang, Hongyi Zhou, Ge Li, Yucheng Tang, Weiran Liao, Björn Hein, \\ Tamim Asfour and Rudolf Lioutikov%
\thanks{The authors are with the Institute for Anthropomatics
and Robotics, Karlsruhe Institute of Technology, Germany
        {\tt\footnotesize \{x.huang, hongyi.zhou, ge.li, weiran.liao, bjoern.hein, asfour, lioutikov\}@kit.edu, yucheng.tang@partner.kit.edu}}%
\thanks{Digital Object Identifier (DOI): see top of this page.}
}

\IEEEpeerreviewmaketitle

\input{tex/99_math_commands}

\begin{document}

\maketitle

\begin{abstract}
We propose MoRe-ERL, a framework that combines Episodic Reinforcement Learning (ERL) and residual learning, which refines preplanned reference trajectories into safe, feasible, and efficient task-specific trajectories.
This framework is general enough to incorporate into arbitrary ERL methods and motion generators seamlessly. 
MoRe-ERL identifies trajectory segments requiring modification while preserving critical task-related maneuvers. Then it generates smooth residual adjustments using B-Spline-based movement primitives to ensure adaptability to dynamic task contexts and smoothness in trajectory refinement. 
Experimental results demonstrate that residual learning significantly outperforms training from scratch using ERL methods, achieving superior sample efficiency and task performance. Hardware evaluations further validate the framework, showing that policies trained in simulation can be directly deployed in real-world systems, exhibiting a minimal sim-to-real gap.

\end{abstract}

\begin{IEEEkeywords}
Motion and Path Planning; Reinforcement Learning; Integrated Planning and Learning
\end{IEEEkeywords}

\input{tex/0_introduction}
\input{tex/1_related_work}
\input{tex/2_preliminaries}
\input{tex/3_method}
\input{tex/4_experiments}

\input{tex/5_conclusions}


\bibliographystyle{IEEEtran}
\bibliography{IEEEabrv, main, motion_planning}

\end{document}

%% file: includes.tex
\usepackage{booktabs} 
\usepackage{tabularx}

\usepackage{amsmath}
\usepackage{bm}
\usepackage{hyperref}
\usepackage{url}
\usepackage{cite}
\usepackage{amssymb}  
\usepackage[]{todonotes}
\usepackage{caption}
\usepackage{subcaption}
\usepackage{cite}
\usepackage{tikz}
\usepackage{pgfplots}
\usepackage{xcolor}
\usepackage{tikz}
\usepackage{multicol}
\usepackage{multirow}
\usepackage{dashrule}
\usepackage{svg}
\usepackage{fix-cm}

\usetikzlibrary{positioning, arrows, automata, matrix}
\usepackage{pgfplots}
\pgfplotsset{compat=1.8}
\usepackage{pgfplotstable}
\usepgfplotslibrary{groupplots}
\usepgfplotslibrary{colorbrewer}
\usetikzlibrary{external}
\usetikzlibrary{backgrounds, calc}
\usetikzlibrary{spy}
\usetikzlibrary{fit}
\usetikzlibrary{arrows.meta}
\usepackage{adjustbox}

\definecolor{C0}{HTML}{1f77b4}  
\definecolor{C1}{HTML}{ff7f0e}  
\definecolor{C2}{HTML}{2ca02c}  
\definecolor{C3}{HTML}{d62728}  
\definecolor{C4}{HTML}{9467bd}  
\definecolor{C5}{HTML}{8c564b}  
\definecolor{C6}{HTML}{e377c2}  
\definecolor{C7}{HTML}{7f7f7f}  
\definecolor{C8}{HTML}{bcbd22}  
\definecolor{C9}{HTML}{17becf}  
\definecolor{C10}{HTML}{CCCC00} 
\definecolor{plot_background}{HTML}{EBF0F0}

\definecolor{GreenAccent}{HTML}{92D050}
\definecolor{OrangeAccent}{HTML}{C45818}
\definecolor{BlueAccent}{HTML}{366092}
\definecolor{LightBlue}{HTML}{ADD8E6}

\definecolor{B0}{HTML}{f29a2e}
\definecolor{B1}{HTML}{fe9aa2}
\definecolor{B2}{HTML}{866abf}
\definecolor{B3}{HTML}{b35ea2}
\definecolor{B4}{HTML}{ac571e}

\newcommand{\centeredinline}[3][0.2cm]{%
    \raisebox{0.5ex}{\textcolor{#2}{\rule{#1}{#3}}}%
}
\newcommand{\centereddashed}[3][0.2cm]{%
    \centeredinline[#1]{#2}{#3}\hspace{0.05cm}\centeredinline[#1]{#2}{#3}
}

\usepackage[normalem]{ulem}
\definecolor{OliveGreen}{RGB}{0,200,25}
\newcommand{\red}[1]{\textcolor{red}{#1}}
\newcommand{\darkgreen}[1]{\textcolor{OliveGreen}{#1}}

\newcommand{\added}[1]{\darkgreen{#1}}
\newcommand{\replaced}[2]{\red{\ifmmode\text{\sout{\ensuremath{#1}}}\else\sout{#1}\fi} \darkgreen{#2}}
\newcommand{\deleted}[1]{\red{\ifmmode\text{\sout{\ensuremath{#1}}}\else\sout{#1}\fi}}

\renewcommand{\added}[1]{\textcolor{OliveGreen}{#1}}
\renewcommand{\replaced}[1]{\textcolor{C1}{#1}}
\renewcommand{\added}[1]{\textcolor{black}{#1}}
\renewcommand{\replaced}[1]{\textcolor{black}{#1}}
\renewcommand{\deleted}[1]{\color{blue} }

\usepackage{xparse}
\usepackage{soul}

\NewDocumentCommand{\rudi}{om}{\bgroup\color{red}RUDI: \IfNoValueF{#1}{\st{#1}} #2\egroup}

\usepackage{graphicx} 

%% file: tex/99_math_commands.tex
\usepackage{amsmath,amsfonts,bm}

\def\eqref#1{equation~\ref{#1}}

\def\1{\bm{1}}

\def\vs{{\bm{s}}}

\def\vw{{\bm{w}}}

\def\vy{{\bm{y}}}

\DeclareMathAlphabet{\mathsfit}{\encodingdefault}{\sfdefault}{m}{sl}
\SetMathAlphabet{\mathsfit}{bold}{\encodingdefault}{\sfdefault}{bx}{n}

\newcommand{\E}{\mathbb{E}}



%% file: tex/0_introduction.tex
\section{Introduction}
\label{sec:introduction}

\IEEEPARstart{R}{obotic} applications, such as multi-arm cooperation, often require frequent motion adaptation to ensure safety and task efficiency. 
The adaptation must be reactive, smooth, and feasible, with algorithms responding rapidly. Moreover, it is desirable that the new trajectories consider system dynamics, such as potential environmental changes. Taking these changes into account reduces the need for repeated online adaptations.
Episode-based Reinforcement Learning (ERL) methods \cite{williams1992simple, kober2008policy, otto2023mp3, Otto_bbrl_2022} have shown success in addressing complex tasks. Based on the initial observations, the ERL methods generate full trajectories in deterministic computation time by parameterizing the movement primitives (MPs), accounting for the possible changes and interactions in the environment, ensuring smooth execution across an episode. Recent developments in MPs \cite{li2023prodmp, liao2024bmp} support various boundary conditions, enabling smooth transitions during trajectory switches. Replanning techniques \cite{otto2023mp3} further allow trajectory updates as the episode progresses.
This paper proposes MoRe-ERL (\textbf{Mo}tion \textbf{Re}siduals using \textbf{ERL}), a general framework that generates motion residuals to refine previously planned task-related reference trajectories. Given a reference trajectory, MoRe-ERL identifies the trajectory segments that require modification while preserving critical task-related behaviors. Trajectory refinements for these segments are generated using B-Spline-based movement primitives to ensure smooth transitions. Three refinement strategies are investigated, and their performance is evaluated in multiple simulation tasks. The segment identification and the trajectory refinements are jointly learned as a correlated policy. Having the reference trajectory as prior knowledge, MoRe-ERL significantly outperforms training from scratch using ERL methods, achieving superior sample efficiency and task performance.
Our main contributions are
\begin{itemize}
    \item To the best of our knowledge, the first RL algorithm that combines ERL and residual learning, achieving superior sample efficiency and task performance.
    \item An end-to-end policy that identifies the segments of reference trajectories needing modification and parameterizes movement primitives as residuals.
    \item Three novel trajectory refinement strategies using B-spline-based movement primitives, which enforce smooth transition between the reference and the fixed trajectories.
\end{itemize}

\begin{figure}
    \centering
    \includegraphics[width=0.92\linewidth]{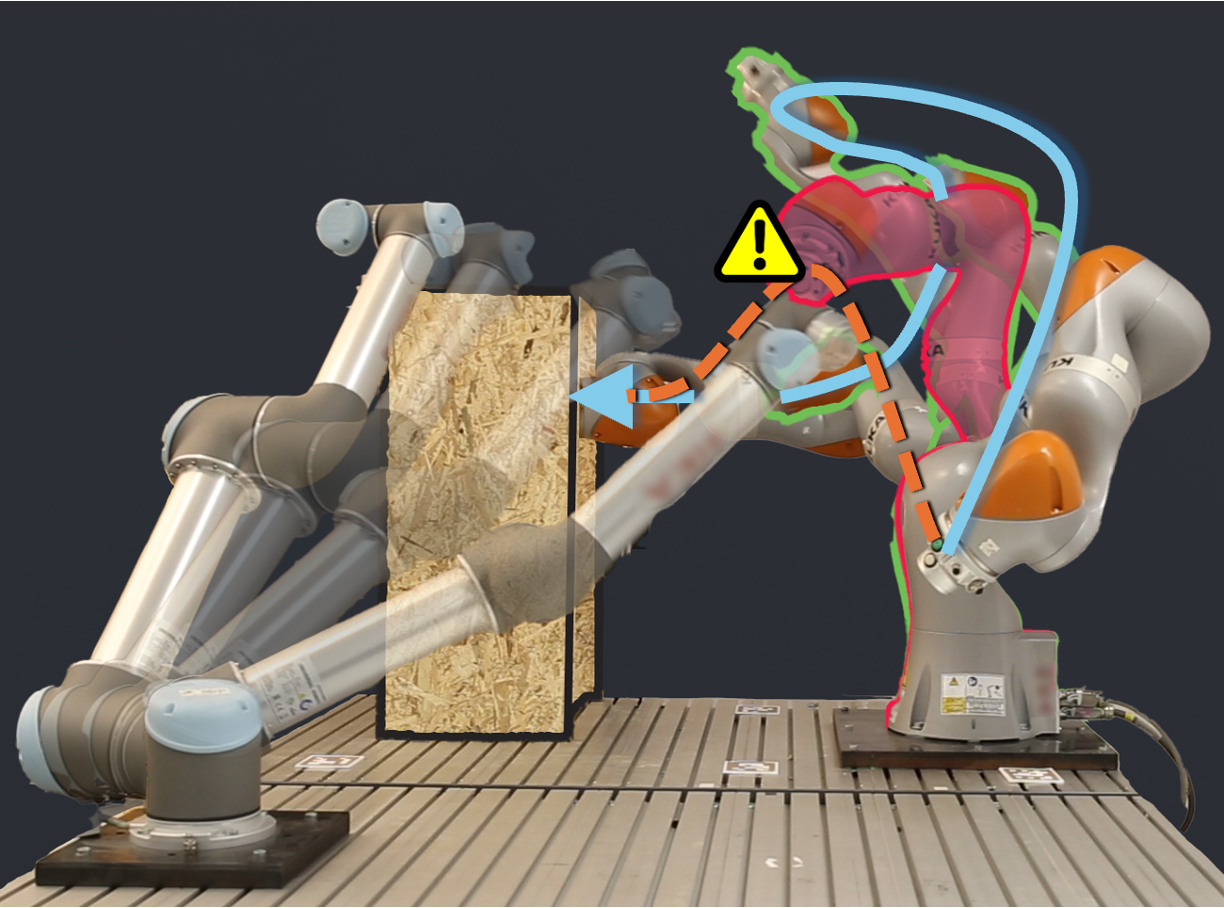}
    \caption{
    MoRe-ERL applies residuals to adjust the trajectory \centereddashed{B0}{2pt}of a KUKA iiwa robot into a safe and feasible motion \centeredinline{LightBlue}{2.5pt}, effectively avoiding the moving UR5 robot. Snapshots of the KUKA iiwa executing the adjusted motion are outlined in green, while the red marker highlights the frame on the reference trajectory where the robots would have collided.}
    \label{fig:first_page}
\end{figure}

%% file: tex/1_related_work.tex
\section{Related Works}

\begin{figure*}[t]
    \centering
    \includegraphics[width=0.836\linewidth]{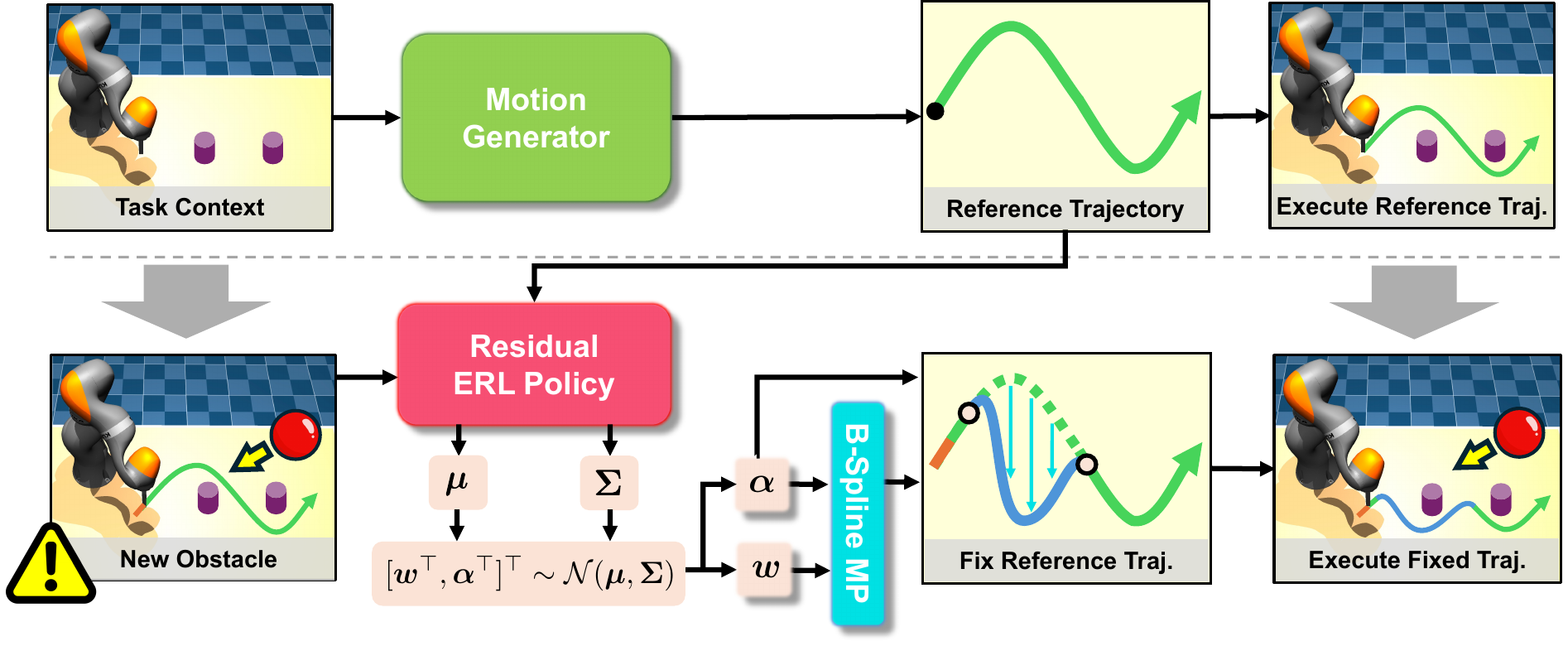}
    \caption{Illustration of the MoRe-ERL pipeline. The robot follows the reference trajectory provided by a motion generator, with the executed segment shown in \centeredinline{B0}{2pt} and the remainder in \centeredinline{GreenAccent}{2pt}, based on the task context. When the task context changes, such as the appearance of new obstacles, MoRe-ERL identifies critical segments on the remaining reference trajectory (\centeredinline{GreenAccent}{2pt}) using learned parameters $\bm \alpha = [\alpha_s, \alpha_e]^\top$ and parameterize residuals $f(\bm{w})$ for the selected segments using B-spline-based movement primitives. The adjusted trajectory, after applying these residuals, is shown by the solid blue-green curve (\centeredinline{BlueAccent}{2pt}\centeredinline{GreenAccent}{2pt}).}
    \label{fig:pipeline}
\end{figure*}

\subsection{Episodic Reinforcement Learning}
ERL is a distinct family of RL that emphasizes the maximization of returns over entire episodes, typically lasting several seconds, rather than optimizing each step in the episodes while interacting with the environment
\cite{peters2008reinforcement}.
Unlike Step-based RL, such as PPO \cite{schulman2017proximal} and SAC \cite{haarnoja2018soft}, ERL shifts the solution search from per-step actions to a parameterized trajectory space, leveraging techniques like MPs \cite{schaal2006dynamic, paraschos2013probabilistic} for generating action sequences. This approach enables a broader exploration horizon \cite{kober2008policy}.
Recent advances in \textit{Deep Black-Box Reinforcement Learning} (BBRL)\cite{Otto_bbrl_2022} have integrated ERL with deep neural networks.
MoRe-ERL employs ERL to learn motion refinements and improve the sample efficiency by leveraging prior knowledge from pre-planned reference trajectories.

\subsection{Learning Motion Residual with RL}
The idea of using RL to learn motion residual \cite{silver2018residual, johannink2019residual} leverages the generalization capability of RL to handle complex dynamics like contacts and friction. At the same time, learning motion residuals simplifies the RL problem, thereby reducing the demand for samples and making real-world applications more feasible~\cite{johannink2019residual,ranjbar2021residual}. 
Recent works have also explored incorporating MPs into residual RL \cite{carvalho2022adapt, davchev2022residual}, where MPs serve as a base motion generator, and step-based RLs are used to learn the motion residual for every control step. \added{For example, model‑free RL generates step‑based residuals for a DMP‑based policy using only task observations \cite{davchev2022residual}.  Similarly, ProMP‑based policies serve as fixed nominal policies, with step‑based agents computing residuals in a low‑frequency control loop \cite{davchev2022residual}.  } However, step-based residual RL suffers from similar limitations to step-based RL, including a lack of smoothness in generated motions and a heavy reliance on dense Markovian reward.  \replaced{In contrast, MoRe-ERL takes reference trajectories from arbitrary base policies as input and uses BMPs to
parameterize residual sequences. It fully leverages BMPs by allowing partial refinement of the reference
trajectory rather than modifying it entirely, thereby preserving critical task behaviors. Consequently, the
length of the residual sequence is flexible and determined by the policy. 
}\added{
Compared to DMPs \cite{schaal2006dynamic} and ProMPs \cite{paraschos2013probabilistic}, BMPs mathematically guarantee enforcement of initial and terminal conditions, making them well-suited for generating motion residuals. To our knowledge, MoRe‑ERL is the first framework to combine BMPs with ERL for parameterizing \textit{residual sequences}. } 


\subsection{Motion Planning in Robotics}
Two important categories in motion planning methods are sampling-based and optimization-based motion planning. 
In a dynamic environment, sampling-based methods usually address problems using reactive methods or fully taking into account the states of the environment in the future. 
\replaced{Reactive methods \cite{huang2022hiro, fishman2023motion, huang2024planning} enable fast replanning but neglect temporal aspects, often producing solutions that soon become invalid.}
On the other hand, methods such as ST-RRT* \cite{grothe2022st} extend the spatial space and \replaced{take into account the temporal aspect.} They assume full knowledge of all paths of participants in the environment. In real-world scenarios, however, full knowledge of the environment in the future is usually not accessible.
Optimization-based methods \cite{ratliff2009chomp, schulman2013finding} directly output a trajectory with parameterized velocity and acceleration profile regarding user-defined constraints. Methods of this category can naturally support the temporal aspect when providing full knowledge of how the environment evolves.
In MoRe-ERL, motion generators serve the role of base planner and provide reference trajectories in an offline manner. MoRe-ERL is base-planner-agnostic and refines trajectories during execution. 

%% file: tex/2_preliminaries.tex
\section{PRELIMINARIES}

\subsection{Episodic Reinforcement Learning (ERL)}
\label{subsec:erl}
ERL \cite{whitley1993genetic, kober2008policy} predicts an entire sequence of actions to accomplish a task by optimizing cumulative rewards without explicitly modeling detailed state transitions within an episode. 
ERL methods usually predict a weight vector $\vw$ given the task context. This vector is then used to parameterize a complete trajectory $\vy(t) = f(\vw)$ for $\vy(t) \in \mathbb{R}^D$ and $t\in[0,T]$, where $D$ corresponds to the dimensionality of the trajectory space, such as the DoFs in a robotic system, $T$ represents the trajectory duration, and $f[\cdot]$ indicates a generic function for trajectory parameterization using a motion generator. The predicted trajectory can either be directly utilized as per-step actions or serve as input to a trajectory tracking controller for computing lower-level motor commands.
Given the initial state $\vs_0 \sim p(\vs_0)$ specifying the starting configuration and task context, the goal of ERL is to find a weight vector $\vw$ that maximizes the return $R(\bm{s}_0, f(\vw))$ after executing the trajectory $\vy(t) = f(\vw)$.   
The ERL learning objective is generally expressed as: 
\begin{equation}
    J = \E_{p(\vs_0),{\pi_{\theta}}(\vw|\vs_0)} ~\left[ R(\bm{s}_0, f(\vw)) - V_{\phi}(\bm{s}_0) \right], \label{eq:traj_rl}
\end{equation}

\noindent
where $\pi_{\theta}$ denotes the policy parameterized by $\theta$, often implemented using a neural network. 
The return $R(\bm{s}_0, f(\vw)) = \sum_{t=0}^{T} \gamma^t r_t$ is the cumulative reward obtained by following the trajectory,
where $\gamma$ is the discount factor, and $r_t$ is the reward at time step $t$. The term $V_{\phi}(\vs_0)$ represents a value estimator of the state $\vs_0$, parameterized by $\phi$, and acts as a baseline to stabilize training \cite{sutton2018reinforcement}. 


When compared to traditional step-based RL (SRL) methods like PPO \cite{schulman2017proximal}, ERL shifts the solution search from the per-step action space $\mathcal{A}$ to a parameterized trajectory space $\mathcal{W}$, predicting trajectory parameters as $\pi(\vw|\vs)$. This often facilitates broader exploration and results in smooth, correlated motion trajectories. 
Additionally, the learning objective in Eq.~(\ref{eq:traj_rl}) relaxes the requirement for Markovian rewards \cite{sutton2018reinforcement}, which enforces that the reward $r_t$ at a given time step $t$ depends only on the current state $\bm s$ and action $\bm a$. 
Step-based RL methods such as SAC \cite{haarnoja2018soft} rely on temporal difference (TD) learning. This requires Markovian rewards in order to assign value credits to per-step actions and states properly.
In contrast, ERL assigns task credit to the entire trajectory episode parameterized by $\vw$ by aggregating per-step rewards. This removes the requirement for rewards to be Markovian, allowing for the use of delayed or history-dependent rewards, referred to as \textit{non-Markovian rewards} \cite{sutton2018reinforcement}. Intuitively, non-Markovian rewards offer greater flexibility and simplicity in task design \cite{Otto_bbrl_2022}, as they rely on fewer assumptions compared to their Markovian counterparts. 

Usually, ERL methods use MPs as the motion generator. MPs can encapsulate trajectories from a lower-dimensional parameter space, thereby reducing the problem complexity.

\subsection{Using Movement Primitives in ERL}
\label{subsec:mp_erl}
Parameterizing trajectories using MPs \cite{schaal2006dynamic, paraschos2013probabilistic, li2023prodmp} is central to ERL methods.
We first describe Probabilistic Movement Primitives (ProMPs) and then introduce BMPs using the same formalism as ProMPs. 

Probabilistic Movement Primitives (ProMPs) \cite{paraschos2013probabilistic} represent a trajectory $\bm{y}(t)$ using a linear basis function:

\begin{equation}
    \bm{y}(t) =f(\vw)= \bm{\Phi}(\frac{t}{T})^\top \bm{\omega} = \bm{\Phi}(u)^\top \bm{\omega}, \label{eq:prompy}
\end{equation}
where $u = t/T \in [0, 1]$ denotes the normalized time, also called the phase variable. The term $\bm{\Phi}(u) = [\Phi_1(u), \Phi_2(u), ..., \Phi_N(u)]^\top$ represents $N$ basis functions for each DoF, evaluated at $u$. The weight vector $\vw = [w_1, w_2,...,w_N]^\top$ controls the trajectory shape by scaling the basis functions. Typically, a neural network is used to predict the mean $\bm{\mu}_{\bm{w}}$ and covariance matrix $\bm{\Sigma}_{\bm{w}}$ and the weight vector is sampled from the distribution $\bm{w} \sim \mathcal{N}(\bm{w} | \bm{\mu}_{\bm{w}}, \bm{\Sigma}_{\bm{w}})$.
For non-periodic trajectories, ProMPs often utilize radial basis functions (RBF) as the basis functions, with their centers uniformly distributed in the phase space $[0, 1]$. 

ProMPs are advantageous due to their simple linear representation, enabling fast computation and probabilistic modeling\replaced{\cite{paraschos2013probabilistic}}. However, ProMPs lack mathematical support for enforcing specific boundary conditions at the trajectory's start and end points. This limitation restricts their ability to generate new trajectories that seamlessly transition from an existing one.
However, this is a critical requirement for real-world scenarios, where frequent trajectory switching is necessary.
Recent works \cite{kicki2024bridging,liao2024bmp} address these inherent limitations by replacing the RBF functions in ProMPs with B-splines.
The resulting model B-spline-based Movement Primitives (BMP) retains the linear basis function representation of ProMPs while supporting an arbitrary number and order of trajectory transition conditions. 
Mathematically, these conditions are known as boundary conditions.

\textbf{Definition of BMP.} The basis functions of BMP, $\bm{\Phi}^P(u) = [\Phi^P_1(u), \Phi^P_2(u), ..., \Phi^P_N(u)]^\top$, are defined as $P$-th order polynomial functions, where $0 \leq P < N$. These basis functions are constructed over $M$ definition intervals, equally divided by $M+1$ knots $u_0, ..., u_M$, with $u_0 = 0$ and $u_M = 1$. Typically, $M = N + P$ \cite{prautzsch2002bezier} and the intervals between two adjacent knots have the same length $\delta$. 
In the context of B-splines, the weights $\vw$ are also interpreted as \textit{control points}, which define a convex hull that bounds the trajectory, see Fig. \ref{subfig:bp-2d}. 
\input{images/tex/0_b_spline_illustration}
Each basis function $\Phi^P_n$, where $n \in [1, N]$, is defined recursively from order $0$ to order $P$ \cite{prautzsch2002bezier}. To illustrate this recursive process, we denote intermediate orders with the index $p$, where $p \in [0, P]$.
For $p=0$, the basis functions are piecewise constant:
\begin{equation}
    \Phi^{p=0}_{n}(u) = 
    \begin{cases} 
    1 & \text{if } u_n \leq u < u_{n+1}, \\
    0 & \text{otherwise}.
    \end{cases}
    \label{eq:deboor_constant}
\end{equation}
For $p>0$, each basis function $\Phi^{p}_{n}(u)$ is computed by interpolating between two corresponding lower-order basis functions, $\Phi^{p-1}_{n}(u)$ and $\Phi^{p-1}_{n+1}(u)$, using coefficients $j_{n}$ and $j_{n+1}$: 
\begin{align}
    &\Phi^p_{n}(u) = j_{n}~ \Phi^{p-1}_{n}(u) ~+~ j_{n+1}~ \Phi^{p-1}_{n+1}(u).    
    \label{eq:deboor_recursive}
    \\
    &j_{n} = \frac{u - u_n}{p~\delta}, ~~j_{n+1} = \frac{u_{n+p+1} - u}{p~\delta}
\end{align}
By recursively
applying Eq.~(\ref{eq:deboor_recursive}) until order $P$, the $P$-th order basis function is obtained, see Fig. \ref{subfig:bp-recursive}.
By substituting the resulting BMP basis functions into Eq.~(\ref{eq:prompy}), the trajectory of BMP can be computed using the linear basis function representation of ProMPs.

\textbf{Derivative of B-Splines.}
It is worth noting that the $i$-th order derivative of a $P$-th order B-spline remains a $(P-i)$-th order B-spline: \begin{equation} \bm{y}^{(i)}(t) = \bm{\Phi}^{P-i}(u)^{\top} \bm{\omega}^{(i)}, \label{eq:bmp_derivative} \end{equation} where $\bm{\omega}^{(i)} = [w^{(i)}_1, ..., w^{(i)}_{N-i}]^{\top}$ represents the control points of the B-spline's derivatives. These control points are computed recursively: 
\begin{equation} 
w^{(i)}_n = \frac{P-i}{\delta} [w_{n+1}^{(i-1)} - w_n^{(i-1)}]. \label{eq:bmp_derivative_weights} \end{equation}

\textbf{Enforcing boundary conditions of arbitrary orders.} To ensure the trajectory passes through given starting and ending positions, BMP employs clamped B-splines \cite{prautzsch2002bezier} where the trajectory goes through the first and the last control points. Thus, we directly align these two control points with the given position values. 
Similarly, control points derived from Eq.~(\ref{eq:bmp_derivative}) and (\ref{eq:bmp_derivative_weights}) enforce higher-order conditions, such as velocity and acceleration. In Table \ref{table:bmp_init_cond_example}, we summarize the mapping between boundary conditions at $t_0$ and the corresponding control points. Ending conditions follow similar mappings.

%% file: images/tex/0_b_spline_illustration.tex
\begin{figure}[!]
    \centering

    \begin{subfigure}{0.235\textwidth}    
        \resizebox{\textwidth}{!}{\input{images/tex/bspline_2d}}%
        \caption{B-Spline in 2D}
        \label{subfig:bp-2d}
    \end{subfigure}
    \begin{subfigure}{0.235\textwidth}    
        \resizebox{\textwidth}{!}{\input{images/tex/bspline_basis}}%
        \caption{Basis functions}
        \label{subfig:bp-recursive}
    \end{subfigure}
    
    \caption{Illustration of BMPs: (a) A clamped B-spline curve in 2D parameterized with 6 control points. (b) Basis function of different orders using recursive formulation, where $\Phi_0^p$ denotes the basis function of $p^{\mathrm{th}}$ order for the $0^{\mathrm{th}}$ control point. The knots $u$ represent the change of time.}
    \label{fig:b-spine_illustration}
\end{figure}
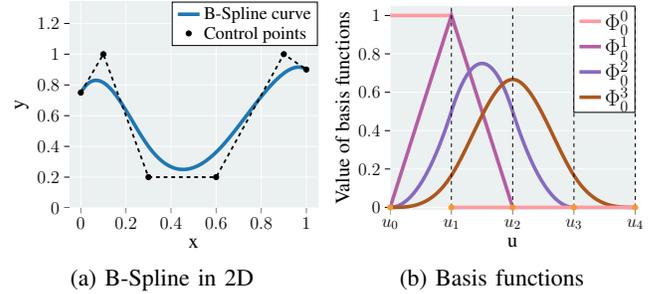

\begin{table}[!]
\centering
\resizebox{0.9\linewidth}{!} {
\small
\renewcommand{\arraystretch}{2}
\setlength{\aboverulesep}{0pt}
\setlength{\belowrulesep}{0pt}
\begin{tabular}{ccc}
\toprule
\textbf{Type} & \textbf{Condition} & \textbf{$w_1^{(i)}$ = ?} \\ 
\hline
\textbf{Position} & $w_{1}^{(0)} = y(t_0)$ & $w_1^{(0)}$\\
\textbf{Velocity} & $w_{1}^{(1)}=\dot{y}(t_0)$ & $\frac{P}{\delta}(w_2^{(0)} - w_1^{(0)})$\\
\textbf{Acceleration} & $w_{1}^{(2)}=\ddot{y}(t_0)$ &$\frac{P(P-1)}{\delta^2}(w_3^{(0)} - 2w_2^{(0)} + w_1^{(0)})$\\
\bottomrule
\end{tabular}
}
\caption{Mapping between boundary conditions and control points at $t_0$. We present the control point $w_1^{(i)}, i=0,1,2$ given initial position $y(t_0)$, velocity $\dot{y}(t_0)$ and acceleration $\ddot{y}(t_0)$ conditions, respectively. By recursively applying Eq.~(\ref{eq:bmp_derivative}) and (\ref{eq:bmp_derivative_weights}), the higher-order control point $w_1^{(i)}$ can be eventually represented using the 0-th order control points.}
\label{table:bmp_init_cond_example}
\end{table}

%% file: images/tex/bspline_2d.tex
\def\linewidthtop{1mm}
\def\linewidthothers{0.5mm}

\begin{tikzpicture}
\definecolor{darkgray176}{RGB}{176,176,176}
\definecolor{darkorange25512714}{RGB}{255,127,14}
\definecolor{steelblue31119180}{RGB}{31,119,180}
\begin{axis}[
legend cell align={left},
tick align=outside,
tick pos=left,
xmin=-0.05, xmax=1.05,
ymin=0, ymax=1.34,
ytick style={color=black},
legend style={at={(1.0, 1.0)}, anchor=north east, legend columns=1, font=\Large},
xlabel={x},
ylabel={y},
xlabel style={font=\fontsize{16}{18}\selectfont},
ylabel style={font=\fontsize{16}{18}\selectfont},
xticklabel style={font=\fontsize{14}{16}\selectfont},
yticklabel style={font=\fontsize{14}{16}\selectfont},
y grid style={white},
ymajorgrids,
x grid style={white},
xmajorgrids,
axis background/.style={fill=plot_background},
x axis line style={draw=none},
y axis line style={draw=none},
]
\addplot [line width=\linewidthtop, C0]
table {%
0 0.75
0.00180360721442886 0.754438721097441
0.00360721442885772 0.758737587141705
0.00541082164328657 0.762897706368919
0.00721442885771543 0.766920187015209
0.00901803607214429 0.770806137316701
0.0108216432865731 0.774556665509525
0.012625250501002 0.778172879829804
0.0144288577154309 0.781655888513667
0.0162324649298597 0.78500679979724
0.0180360721442886 0.788226721916651
0.0198396793587174 0.791316763108025
0.0216432865731463 0.79427803160749
0.0234468937875751 0.797111635651172
0.025250501002004 0.799818683475199
0.0270541082164329 0.802400283315697
0.0288577154308617 0.804857543408792
0.0306613226452906 0.807191571990613
0.0324649298597194 0.809403477297284
0.0342685370741483 0.811494367564934
0.0360721442885772 0.813465351029689
0.037875751503006 0.815317535927675
0.0396793587174349 0.817052030495021
0.0414829659318637 0.818669942967851
0.0432865731462926 0.820172381582294
0.0450901803607214 0.821560454574475
0.0468937875751503 0.822835270180523
0.0486973947895792 0.823997936636563
0.050501002004008 0.825049562178723
0.0523046092184369 0.825991255043128
0.0541082164328657 0.826824123465907
0.0559118236472946 0.827549275683185
0.0577154308617234 0.82816781993109
0.0595190380761523 0.828680864445748
0.0613226452905812 0.829089517463286
0.06312625250501 0.829394887219831
0.0649298597194389 0.82959808195151
0.0667334669338677 0.829700209894449
0.0685370741482966 0.829702379284776
0.0703406813627254 0.829605698358617
0.0721442885771543 0.829411275352098
0.0739478957915832 0.829120218501348
0.075751503006012 0.828733636042491
0.0775551102204409 0.828252636211657
0.0793587174348697 0.827678327244969
0.0811623246492986 0.827011817378557
0.0829659318637275 0.826254214848547
0.0847695390781563 0.825406627891065
0.0865731462925852 0.824470164742238
0.088376753507014 0.823445933638193
0.0901803607214429 0.822335042815057
0.0919839679358717 0.821138600508956
0.0937875751503006 0.819857714956018
0.0955911823647295 0.818493494392369
0.0973947895791583 0.817047047054137
0.0991983967935871 0.815519481177446
0.101002004008016 0.813911904998426
0.102805611222445 0.812225426753201
0.104609218436874 0.8104611546779
0.106412825651303 0.808620197008649
0.108216432865731 0.806703661981575
0.11002004008016 0.804712657832804
0.111823647294589 0.802648292798464
0.113627254509018 0.80051167511468
0.115430861723447 0.79830391301758
0.117234468937876 0.796026114743292
0.119038076152305 0.79367938852794
0.120841683366733 0.791264842607653
0.122645290581162 0.788783585218557
0.124448897795591 0.786236724596779
0.12625250501002 0.783625368978446
0.128056112224449 0.780950626599684
0.129859719438878 0.77821360569662
0.131663326653307 0.775415414505382
0.133466933867735 0.772557161262095
0.135270541082164 0.769639954202887
0.137074148296593 0.766664901563884
0.138877755511022 0.763633111581213
0.140681362725451 0.760545692491002
0.14248496993988 0.757403752529376
0.144288577154309 0.754208399932463
0.146092184368737 0.750960742936389
0.147895791583166 0.747661889777281
0.149699398797595 0.744312948691267
0.151503006012024 0.740915027914472
0.153306613226453 0.737469235683024
0.155110220440882 0.733976680233049
0.156913827655311 0.730438469800674
0.158717434869739 0.726855712622026
0.160521042084168 0.723229516933232
0.162324649298597 0.719560990970419
0.164128256513026 0.715851242969713
0.165931863727455 0.71210138116724
0.167735470941884 0.708312513799129
0.169539078156313 0.704485749101506
0.171342685370741 0.700622195310497
0.17314629258517 0.696722960662229
0.174949899799599 0.692789153392829
0.176753507014028 0.688821881738425
0.178557114228457 0.684822253935142
0.180360721442886 0.680791378219107
0.182164328657315 0.676730362826448
0.183967935871743 0.67264031599329
0.185771543086172 0.668522345955762
0.187575150300601 0.664377560949989
0.18937875751503 0.660207069212099
0.191182364729459 0.656011978978217
0.192985971943888 0.651793398484472
0.194789579158317 0.64755243596699
0.196593186372745 0.643290199661897
0.198396793587174 0.63900779780532
0.200200400801603 0.634706338633387
0.202004008016032 0.630386930382224
0.203807615230461 0.626050681287958
0.20561122244489 0.621698699586715
0.207414829659319 0.617332093514622
0.209218436873747 0.612951971307807
0.211022044088176 0.608559441202395
0.212825651302605 0.604155611434515
0.214629258517034 0.599741590240291
0.216432865731463 0.595318485855853
0.218236472945892 0.590887406517325
0.220040080160321 0.586449460460835
0.221843687374749 0.58200575592251
0.223647294589178 0.577557401138476
0.225450901803607 0.573105504344861
0.227254509018036 0.568651173777791
0.229058116232465 0.564195517673392
0.230861723446894 0.559739644267793
0.232665330661323 0.555284661797118
0.234468937875751 0.550831678497496
0.23627254509018 0.546381802605054
0.238076152304609 0.541936142355916
0.239879759519038 0.537495805986212
0.241683366733467 0.533061901732067
0.243486973947896 0.528635537829608
0.245290581162325 0.524217822514962
0.247094188376754 0.519809864024256
0.248897795591182 0.515412770593617
0.250701402805611 0.511027650459171
0.25250501002004 0.506655611857045
0.254308617234469 0.502297763023366
0.256112224448898 0.497955212194261
0.257915831663327 0.493629067605856
0.259719438877756 0.489320437494279
0.261523046092184 0.485030430095656
0.263326653306613 0.480760153646114
0.265130260521042 0.47651071638178
0.266933867735471 0.47228322653878
0.2687374749499 0.468078792353241
0.270541082164329 0.463898522061291
0.272344689378757 0.459743523899056
0.274148296593186 0.455614906102662
0.275951903807615 0.451513776908237
0.277755511022044 0.447441244551907
0.279559118236473 0.443398417269799
0.281362725450902 0.43938640329804
0.283166332665331 0.435406310872756
0.284969939879759 0.431459248230076
0.286773547094188 0.427546323606124
0.288577154308617 0.423668645237028
0.290380761523046 0.419827321358916
0.292184368737475 0.416023460207913
0.293987975951904 0.412258170020146
0.295791583166333 0.408532559031743
0.297595190380761 0.404847735478829
0.29939879759519 0.401204807597532
0.301202406419258 0.39760482889627
0.30300603717465 0.394048216673829
0.304809722255343 0.390534977771174
0.306613494256516 0.387065112188304
0.30841738577335 0.383638619925221
0.310221429401025 0.380255500981924
0.312025657734721 0.376915755358412
0.313830103369618 0.373619383054687
0.315634798900897 0.370366384070747
0.317439776923738 0.367156758406593
0.31924507003332 0.363990506062225
0.321050710824825 0.360867627037642
0.322856731893432 0.357788121332846
0.324663165834321 0.354751988947836
0.326470045242673 0.351759229882611
0.328277402713669 0.348809844137172
0.330085270842487 0.345903831711519
0.331893682224309 0.343041192605652
0.333702669454314 0.340221926819571
0.335512265127683 0.337446034353276
0.337322501839595 0.334713515206766
0.339133412185233 0.332024369380043
0.340945028759774 0.329378596873105
0.3427573841584 0.326776197685953
0.34457051097629 0.324217171818587
0.346384441808626 0.321701519271007
0.348199209250586 0.319229240043213
0.350014845897352 0.316800334135204
0.351831384344103 0.314414801546982
0.353648857186021 0.312072642278545
0.355467297018284 0.309773856329894
0.357286736436073 0.307518443701029
0.359107208034569 0.30530640439195
0.360928744408951 0.303137738402657
0.362751378154399 0.30101244573315
0.364575141866095 0.298930526383428
0.366400068139218 0.296891980353493
0.368226189568948 0.294896807643343
0.370053538750466 0.292945008252979
0.371882148278951 0.291036582182401
0.373712050749585 0.289171529431609
0.375543278757546 0.287349850000602
0.377375864898016 0.285571543889382
0.379209841766175 0.283836611097947
0.381045241957202 0.282145051626299
0.382882098066278 0.280496865474436
0.384720442688583 0.278892052642359
0.386560308419297 0.277330613130068
0.388401727853601 0.275812546937562
0.390244733586675 0.274337854064843
0.392089358213698 0.27290653451191
0.393935634329852 0.271518588278762
0.395783594530316 0.2701740153654
0.397633271410271 0.268872815771824
0.399484697564896 0.267614989498034
0.401337905589372 0.26640053654403
0.40319292807888 0.265229456909812
0.405049797628598 0.264101750595379
0.406908546833709 0.263017417600733
0.40876920828939 0.261976457925872
0.410631814590824 0.260978871570797
0.41249639833319 0.260024658535508
0.414362992111669 0.259113818820005
0.416231628521439 0.258246352424287
0.418102340157683 0.257422259348356
0.419975159615579 0.25664153959221
0.421850119490309 0.255904193155851
0.423727252377052 0.255210220039277
0.425606590870988 0.254559620242489
0.427488167567298 0.253952393765487
0.429372015061162 0.253388540608271
0.43125816594776 0.25286806077084
0.433146652822273 0.252390954253196
0.43503750827988 0.251957221055337
0.436930764915762 0.251566861177264
0.438826455325098 0.251219874618977
0.44072461210307 0.250916261380476
0.442625267844857 0.250656021461761
0.44452845514564 0.250439154862832
0.446434206600598 0.250265661583688
0.448342554804912 0.250135541624331
0.450253532353762 0.250048794984759
0.452167171842329 0.250005421664973
0.454083505865792 0.250005421664973
0.456002567019332 0.250048794984759
0.457924387898129 0.250135541624331
0.459849001097363 0.250265661583688
0.461776439212214 0.250439154862832
0.463706734837863 0.250656021461761
0.465639920569489 0.250916261380476
0.467576029002274 0.251219874618977
0.469515092731396 0.251566861177264
0.471457144352037 0.251957221055337
0.473402216459376 0.252390954253196
0.475350341648594 0.25286806077084
0.477301552514871 0.253388540608271
0.479255881653388 0.253952393765487
0.481213361659323 0.254559620242489
0.483174025127858 0.255210220039277
0.485137904654172 0.255904193155851
0.487105032833447 0.25664153959221
0.489075442260861 0.257422259348356
0.491049165531596 0.258246352424287
0.493026235240832 0.259113818820005
0.495006683983748 0.260024658535508
0.496990544355525 0.260978871570797
0.498977848951343 0.261976457925872
0.500968630366383 0.263017417600732
0.502962921195824 0.264101750595379
0.504960754034847 0.265229456909812
0.506962161478631 0.26640053654403
0.508967176122358 0.267614989498034
0.510975830561207 0.268872815771824
0.512988157390359 0.2701740153654
0.515004189204993 0.271518588278762
0.51702395860029 0.272906534511909
0.51904749817143 0.274337854064843
0.521074840513594 0.275812546937562
0.523106018221961 0.277330613130068
0.525141063891712 0.278892052642359
0.527180010118027 0.280496865474436
0.529222889496086 0.282145051626299
0.531269734621069 0.283836611097947
0.533320578088156 0.285571543889382
0.535375452492529 0.287349850000602
0.537434390429366 0.289171529431609
0.539497424493848 0.291036582182401
0.541564587281156 0.292945008252979
0.543635911386469 0.294896807643343
0.545711429404968 0.296891980353492
0.547791173931833 0.298930526383428
0.549875177562244 0.30101244573315
0.551963472891381 0.303137738402657
0.554056092514425 0.30530640439195
0.556153069026555 0.307518443701029
0.558254435022953 0.309773856329894
0.560360223098797 0.312072642278545
0.562470465849269 0.314414801546982
0.564585195869548 0.316800334135204
0.566704445754815 0.319229240043213
0.56882824810025 0.321701519271007
0.570956635501033 0.324217171818587
0.573089640552344 0.326776197685953
0.575227295849364 0.329378596873105
0.577369633987273 0.332024369380043
0.57951668756125 0.334713515206766
0.581668489166477 0.337446034353276
0.583825071398132 0.340221926819571
0.585986466851398 0.343041192605652
0.588152708121453 0.345903831711519
0.590323827803478 0.348809844137172
0.592499858492653 0.351759229882611
0.594680832784158 0.354751988947835
0.596866783273174 0.357788121332846
0.59905774255488 0.360867627037642
0.601253743224458 0.363990506062225
0.603454817877086 0.367156758406593
0.605660999107946 0.370366384070747
0.607872319512218 0.373619383054686
0.610088811685081 0.376915755358412
0.612310508221716 0.380255500981924
0.614537441717303 0.383638619925221
0.616769644767022 0.387065112188304
0.619007149966054 0.390534977771174
0.621249989909578 0.394048216673829
0.623498197192776 0.397604828896269
0.625751803606007 0.401204808804761
0.628010792650477 0.404847812741478
0.6302750729792 0.408532973111254
0.632544546806634 0.412259377249042
0.63481911634724 0.416026112489798
0.637098683815476 0.419832266168475
0.639383151425803 0.423676925620028
0.641672421392679 0.427559178179411
0.643966395930563 0.43147811118158
0.646264977253916 0.435432811961488
0.648568067577197 0.439422367854089
0.650875569114864 0.443445866194338
0.653187384081378 0.44750239431719
0.655503414691198 0.451591039557599
0.657823563158783 0.455710889250519
0.660147731698593 0.459861030730905
0.662475822525087 0.464040551333711
0.664807737852724 0.468248538393891
0.667143379895964 0.4724840792464
0.669482650869266 0.476746261226192
0.67182545298709 0.481034171668223
0.674171688463895 0.485346897907445
0.676521259514141 0.489683527278814
0.678874068352286 0.494043147117283
0.681230017192791 0.498424844757808
0.683589008250114 0.502827707535343
0.685950943738715 0.507250822784842
0.688315725873054 0.51169327784126
0.690683256867589 0.51615416003955
0.693053438936781 0.520632556714668
0.695426174295088 0.525127555201567
0.697801365156971 0.529638242835203
0.700178913736888 0.534163706950529
0.702558722249298 0.538703034882501
0.704940692908662 0.543255313966071
0.707324727929439 0.547819631536196
0.709710729526088 0.552395074927828
0.712098599913068 0.556980731475924
0.714488241304839 0.561575688515436
0.716879555915861 0.566179033381319
0.719272445960592 0.570789853408529
0.721666813653492 0.575407235932019
0.72406256120902 0.580030268286743
0.726459590841636 0.584658037807656
0.7288578047658 0.589289631829713
0.73125710519597 0.593924137687868
0.733657394346606 0.598560642717075
0.736058574432168 0.603198234252288
0.738460547667115 0.607835999628463
0.740863216265906 0.612473026180553
0.743266482443 0.617108401243513
0.745670248412858 0.621741212152297
0.748074416389938 0.62637054624186
0.7504788885887 0.630995490847156
0.752883567223603 0.63561513330314
0.755288354509107 0.640228560944766
0.757693152659671 0.644834861106987
0.760097863889755 0.64943312112476
0.762502390413817 0.654022428333037
0.764906634446317 0.658601870066775
0.767310498201716 0.663170533660926
0.769713883894471 0.667727506450445
0.772116693739043 0.672271875770287
0.77451882994989 0.676802728955406
0.776920194741473 0.681319153340757
0.779320690328251 0.685820236261294
0.781720218924683 0.690305065051971
0.784118682745228 0.694772727047744
0.786515984004346 0.699222309583565
0.788912024916496 0.70365289999439
0.791306707696138 0.708063585615172
0.793699934557731 0.712453453780867
0.796091607715734 0.716821591826429
0.798481629384608 0.721167087086812
0.80086990177881 0.725489026896971
0.803256327112802 0.72978649859186
0.805640807601041 0.734058589506433
0.808023245457988 0.738304386975645
0.810403542898102 0.74252297833445
0.812781602135842 0.746713450917803
0.815157325385668 0.750874892060658
0.817530614862039 0.75500638909797
0.819901372779414 0.759107029364692
0.822269501352253 0.763175900195779
0.824634902795016 0.767212088926186
0.826997479322161 0.771214682890868
0.829357133148148 0.775182769424777
0.831713766487437 0.77911543586287
0.834067281554486 0.7830117695401
0.836417580563756 0.786870857791422
0.838764565729706 0.790691787951789
0.841108139266795 0.794473647356158
0.843448203389482 0.798215523339481
0.845784660312227 0.801916503236713
0.848117412249489 0.80557567438281
0.850446361415728 0.809192124112724
0.852771410025403 0.812764939761411
0.855092460292974 0.816293208663825
0.857409414432899 0.81977601815492
0.859722174659639 0.823212455569651
0.862030643187653 0.826601608242972
0.864334722231399 0.829942563509837
0.866634314005338 0.833234408705202
0.868929320723929 0.83647623116402
0.871219644601631 0.839667118221246
0.873505187852904 0.842806157211833
0.875785852692208 0.845892435470738
0.878061541334 0.848925040332914
0.880332155992742 0.851903059133315
0.882597598882891 0.854825579206895
0.884857772218909 0.85769168788861
0.887112578215253 0.860500472513414
0.889361919086384 0.86325102041626
0.891605697046761 0.865942418932105
0.893843814310844 0.868573755395901
0.896076173093091 0.871144117142603
0.898302675607962 0.873652591507165
0.900523224069916 0.876098265824543
0.902737720693414 0.878480227429691
0.904946067692914 0.880797563657562
0.907148167282875 0.883049361843111
0.909343921677758 0.885234709321293
0.911533233092021 0.887352693427063
0.913716003740124 0.889402401495373
0.915892135836526 0.89138292086118
0.918061531595687 0.893293338859437
0.920224093232066 0.895132742825099
0.922379722960123 0.896900220093119
0.924528322994316 0.898594857998454
0.926669795549106 0.900215743876056
0.928804042838952 0.90176196506088
0.930930967078313 0.903232608887881
0.933050470481648 0.904626762692014
0.935162455263417 0.905943513808232
0.937266823638079 0.90718194957149
0.939363477820094 0.908341157316742
0.941452320023922 0.909420224378943
0.943533252464021 0.910418238093047
0.94560617735485 0.911334285794009
0.94767099691087 0.912167454816782
0.94972761334654 0.912916832496323
0.951775928876319 0.913581506167583
0.953815845714666 0.914160563165519
0.955847266076041 0.914653090825085
0.957870092174904 0.915058176481235
0.959884226225713 0.915374907468923
0.961889570442929 0.915602371123104
0.96388602704101 0.915739654778732
0.965873498234416 0.915785845770762
0.967851886237606 0.915740031434148
0.96982109326504 0.915601299103844
0.971781021531177 0.915368736114805
0.973731573250476 0.915041429801986
0.975672650637398 0.91461846750034
0.9776041559064 0.914098936544822
0.979525991271944 0.913481924270386
0.981438058948488 0.912766518011988
0.983340261150491 0.911951805104581
0.985232500092413 0.911036872883119
0.987114677988714 0.910020808682558
0.988986697053852 0.908902699837851
0.990848459502287 0.907681633683953
0.992699867548479 0.906356697555818
0.994540823406887 0.904926978788401
0.99637122929197 0.903391564716656
0.998190987418188 0.901749542675538
1 0.9
};

\addplot [mark=*, only marks,]
table{%
x  y
0 0.75
0.1 1
0.3 0.2
0.6 0.2
0.9 1
1 0.9
};

\addlegendentry{B-Spline curve}
\addlegendentry{Control points}

\addplot [line width=\linewidthothers, dashed]
table {%
0 0.75
0.1 1
0.3 0.2
0.6 0.2
0.9 1
1 0.9
};

\end{axis}
\end{tikzpicture}

%% file: images/tex/bspline_basis.tex
\begin{tikzpicture}

\definecolor{crimson2143940}{RGB}{214,39,40}
\definecolor{darkgray176}{RGB}{176,176,176}
\definecolor{darkorange25512714}{RGB}{255,127,14}
\definecolor{forestgreen4416044}{RGB}{44,160,44}
\definecolor{steelblue31119180}{RGB}{31,119,180}

\def\linewidthtop{1mm}
\def\linewidthothers{0.5mm}

\begin{axis}[
tick align=outside,
tick pos=left,
xlabel={u},
xticklabels={,$u_0$,$u_1$,$u_2$,$u_3$,$u_4$},
xticklabel style={font=\fontsize{14}{16}\selectfont},
yticklabel style={font=\fontsize{14}{16}\selectfont},
xlabel style={font=\fontsize{16}{18}\selectfont},
xmin=0, xmax=0.4,
y grid style={darkgray176},
ymin=-0.01, ymax=1.05,
ytick style={color=black},
ylabel={Value of basis functions},
ylabel style={font=\fontsize{16}{18}\selectfont},
legend style={at={(1.0, 1.0)}, anchor=north east, legend columns=1, font=\LARGE},
y grid style={white},
ymajorgrids,
x grid style={white},
xmajorgrids,
axis background/.style={fill=plot_background},
x axis line style={draw=none},
y axis line style={draw=none},
]
\addplot [line width=\linewidthtop, B1, mark=*, mark size=0, mark options={solid}]
table {%
0 1
0.001 1
0.002 1
0.003 1
0.004 1
0.005 1
0.006 1
0.007 1
0.008 1
0.009 1
0.01 1
0.011 1
0.012 1
0.013 1
0.014 1
0.015 1
0.016 1
0.017 1
0.018 1
0.019 1
0.02 1
0.021 1
0.022 1
0.023 1
0.024 1
0.025 1
0.026 1
0.027 1
0.028 1
0.029 1
0.03 1
0.031 1
0.032 1
0.033 1
0.034 1
0.035 1
0.036 1
0.037 1
0.038 1
0.039 1
0.04 1
0.041 1
0.042 1
0.043 1
0.044 1
0.045 1
0.046 1
0.047 1
0.048 1
0.049 1
0.05 1
0.051 1
0.052 1
0.053 1
0.054 1
0.055 1
0.056 1
0.057 1
0.058 1
0.059 1
0.06 1
0.061 1
0.062 1
0.063 1
0.064 1
0.065 1
0.066 1
0.067 1
0.068 1
0.069 1
0.07 1
0.071 1
0.072 1
0.073 1
0.074 1
0.075 1
0.076 1
0.077 1
0.078 1
0.079 1
0.08 1
0.081 1
0.082 1
0.083 1
0.084 1
0.085 1
0.086 1
0.087 1
0.088 1
0.089 1
0.09 1
0.091 1
0.092 1
0.093 1
0.094 1
0.095 1
0.096 1
0.097 1
0.098 1
0.099 1
};

\addplot [line width=\linewidthtop, B3, mark=*, mark size=0, mark options={solid}]
table {%
0 0
0.001 0.01
0.002 0.02
0.003 0.03
0.004 0.04
0.005 0.05
0.006 0.06
0.007 0.07
0.008 0.08
0.009 0.09
0.01 0.1
0.011 0.11
0.012 0.12
0.013 0.13
0.014 0.14
0.015 0.15
0.016 0.16
0.017 0.17
0.018 0.18
0.019 0.19
0.02 0.2
0.021 0.21
0.022 0.22
0.023 0.23
0.024 0.24
0.025 0.25
0.026 0.26
0.027 0.27
0.028 0.28
0.029 0.29
0.03 0.3
0.031 0.31
0.032 0.32
0.033 0.33
0.034 0.34
0.035 0.35
0.036 0.36
0.037 0.37
0.038 0.38
0.039 0.39
0.04 0.4
0.041 0.41
0.042 0.42
0.043 0.43
0.044 0.44
0.045 0.45
0.046 0.46
0.047 0.47
0.048 0.48
0.049 0.49
0.05 0.5
0.051 0.51
0.052 0.52
0.053 0.53
0.054 0.54
0.055 0.55
0.056 0.56
0.057 0.57
0.058 0.58
0.059 0.59
0.06 0.6
0.061 0.61
0.062 0.62
0.063 0.63
0.064 0.64
0.065 0.65
0.066 0.66
0.067 0.67
0.068 0.68
0.069 0.69
0.07 0.7
0.071 0.71
0.072 0.72
0.073 0.73
0.074 0.74
0.075 0.75
0.076 0.76
0.077 0.77
0.078 0.78
0.079 0.79
0.08 0.8
0.081 0.81
0.082 0.82
0.083 0.83
0.084 0.84
0.085 0.85
0.086 0.86
0.087 0.87
0.088 0.88
0.089 0.89
0.09 0.9
0.091 0.91
0.092 0.92
0.093 0.93
0.094 0.94
0.095 0.95
0.096 0.96
0.097 0.97
0.098 0.98
0.099 0.99
0.1 1
0.101 0.99
0.102 0.98
0.103 0.97
0.104 0.96
0.105 0.95
0.106 0.94
0.107 0.93
0.108 0.92
0.109 0.91
0.11 0.9
0.111 0.89
0.112 0.88
0.113 0.87
0.114 0.86
0.115 0.85
0.116 0.84
0.117 0.83
0.118 0.82
0.119 0.81
0.12 0.8
0.121 0.79
0.122 0.78
0.123 0.77
0.124 0.76
0.125 0.75
0.126 0.74
0.127 0.73
0.128 0.72
0.129 0.71
0.13 0.7
0.131 0.69
0.132 0.68
0.133 0.67
0.134 0.66
0.135 0.65
0.136 0.64
0.137 0.63
0.138 0.62
0.139 0.61
0.14 0.6
0.141 0.59
0.142 0.58
0.143 0.57
0.144 0.56
0.145 0.55
0.146 0.54
0.147 0.53
0.148 0.52
0.149 0.51
0.15 0.5
0.151 0.49
0.152 0.48
0.153 0.47
0.154 0.46
0.155 0.45
0.156 0.44
0.157 0.43
0.158 0.42
0.159 0.41
0.16 0.4
0.161 0.39
0.162 0.38
0.163 0.37
0.164 0.36
0.165 0.35
0.166 0.34
0.167 0.33
0.168 0.32
0.169 0.31
0.17 0.3
0.171 0.29
0.172 0.28
0.173 0.27
0.174 0.26
0.175 0.25
0.176 0.24
0.177 0.23
0.178 0.22
0.179 0.21
0.18 0.2
0.181 0.19
0.182 0.18
0.183 0.17
0.184 0.16
0.185 0.15
0.186 0.14
0.187 0.13
0.188 0.12
0.189 0.11
0.19 0.1
0.191 0.0900000000000001
0.192 0.0800000000000001
0.193 0.0700000000000001
0.194 0.0600000000000001
0.195 0.05
0.196 0.04
0.197 0.03
0.198 0.02
0.199 0.01
};

\addplot [
line width=\linewidthtop, B2, mark=*, mark size=0, mark options={solid}]
table {%
0 0
0.001 5e-05
0.002 0.0002
0.003 0.00045
0.004 0.0008
0.005 0.00125
0.006 0.0018
0.007 0.00245
0.008 0.0032
0.009 0.00405
0.01 0.005
0.011 0.00605
0.012 0.0072
0.013 0.00845
0.014 0.0098
0.015 0.01125
0.016 0.0128
0.017 0.01445
0.018 0.0162
0.019 0.01805
0.02 0.02
0.021 0.02205
0.022 0.0242
0.023 0.02645
0.024 0.0288
0.025 0.03125
0.026 0.0338
0.027 0.03645
0.028 0.0392
0.029 0.04205
0.03 0.045
0.031 0.04805
0.032 0.0512
0.033 0.05445
0.034 0.0578
0.035 0.06125
0.036 0.0648
0.037 0.06845
0.038 0.0722
0.039 0.07605
0.04 0.08
0.041 0.08405
0.042 0.0882
0.043 0.09245
0.044 0.0968
0.045 0.10125
0.046 0.1058
0.047 0.11045
0.048 0.1152
0.049 0.12005
0.05 0.125
0.051 0.13005
0.052 0.1352
0.053 0.14045
0.054 0.1458
0.055 0.15125
0.056 0.1568
0.057 0.16245
0.058 0.1682
0.059 0.17405
0.06 0.18
0.061 0.18605
0.062 0.1922
0.063 0.19845
0.064 0.2048
0.065 0.21125
0.066 0.2178
0.067 0.22445
0.068 0.2312
0.069 0.23805
0.07 0.245
0.071 0.25205
0.072 0.2592
0.073 0.26645
0.074 0.2738
0.075 0.28125
0.076 0.2888
0.077 0.29645
0.078 0.3042
0.079 0.31205
0.08 0.32
0.081 0.32805
0.082 0.3362
0.083 0.34445
0.084 0.3528
0.085 0.36125
0.086 0.3698
0.087 0.37845
0.088 0.3872
0.089 0.39605
0.09 0.405
0.091 0.41405
0.092 0.4232
0.093 0.43245
0.094 0.4418
0.095 0.45125
0.096 0.4608
0.097 0.47045
0.098 0.4802
0.099 0.49005
0.1 0.5
0.101 0.5099
0.102 0.5196
0.103 0.5291
0.104 0.5384
0.105 0.5475
0.106 0.5564
0.107 0.5651
0.108 0.5736
0.109 0.5819
0.11 0.59
0.111 0.5979
0.112 0.6056
0.113 0.6131
0.114 0.6204
0.115 0.6275
0.116 0.6344
0.117 0.6411
0.118 0.6476
0.119 0.6539
0.12 0.66
0.121 0.6659
0.122 0.6716
0.123 0.6771
0.124 0.6824
0.125 0.6875
0.126 0.6924
0.127 0.6971
0.128 0.7016
0.129 0.7059
0.13 0.71
0.131 0.7139
0.132 0.7176
0.133 0.7211
0.134 0.7244
0.135 0.7275
0.136 0.7304
0.137 0.7331
0.138 0.7356
0.139 0.7379
0.14 0.74
0.141 0.7419
0.142 0.7436
0.143 0.7451
0.144 0.7464
0.145 0.7475
0.146 0.7484
0.147 0.7491
0.148 0.7496
0.149 0.7499
0.15 0.75
0.151 0.7499
0.152 0.7496
0.153 0.7491
0.154 0.7484
0.155 0.7475
0.156 0.7464
0.157 0.7451
0.158 0.7436
0.159 0.7419
0.16 0.74
0.161 0.7379
0.162 0.7356
0.163 0.7331
0.164 0.7304
0.165 0.7275
0.166 0.7244
0.167 0.7211
0.168 0.7176
0.169 0.7139
0.17 0.71
0.171 0.7059
0.172 0.7016
0.173 0.6971
0.174 0.6924
0.175 0.6875
0.176 0.6824
0.177 0.6771
0.178 0.6716
0.179 0.6659
0.18 0.66
0.181 0.6539
0.182 0.6476
0.183 0.6411
0.184 0.6344
0.185 0.6275
0.186 0.6204
0.187 0.6131
0.188 0.6056
0.189 0.5979
0.19 0.59
0.191 0.5819
0.192 0.5736
0.193 0.5651
0.194 0.5564
0.195 0.5475
0.196 0.5384
0.197 0.5291
0.198 0.5196
0.199 0.5099
0.2 0.5
0.201 0.49005
0.202 0.4802
0.203 0.47045
0.204 0.4608
0.205 0.45125
0.206 0.4418
0.207 0.43245
0.208 0.4232
0.209 0.41405
0.21 0.405
0.211 0.39605
0.212 0.3872
0.213 0.37845
0.214 0.3698
0.215 0.36125
0.216 0.3528
0.217 0.34445
0.218 0.3362
0.219 0.32805
0.22 0.32
0.221 0.31205
0.222 0.3042
0.223 0.29645
0.224 0.2888
0.225 0.28125
0.226 0.2738
0.227 0.26645
0.228 0.2592
0.229 0.25205
0.23 0.245
0.231 0.23805
0.232 0.2312
0.233 0.22445
0.234 0.2178
0.235 0.21125
0.236 0.2048
0.237 0.19845
0.238 0.1922
0.239 0.18605
0.24 0.18
0.241 0.17405
0.242 0.1682
0.243 0.16245
0.244 0.1568
0.245 0.15125
0.246 0.1458
0.247 0.14045
0.248 0.1352
0.249 0.13005
0.25 0.125
0.251 0.12005
0.252 0.1152
0.253 0.11045
0.254 0.1058
0.255 0.10125
0.256 0.0968000000000001
0.257 0.0924500000000001
0.258 0.0882000000000001
0.259 0.0840500000000001
0.26 0.0800000000000001
0.261 0.0760500000000001
0.262 0.0722000000000001
0.263 0.0684500000000001
0.264 0.0648000000000001
0.265 0.0612500000000001
0.266 0.0578000000000001
0.267 0.0544500000000001
0.268 0.0512000000000001
0.269 0.0480500000000001
0.27 0.0450000000000001
0.271 0.0420500000000001
0.272 0.0392
0.273 0.03645
0.274 0.0338
0.275 0.03125
0.276 0.0288
0.277 0.02645
0.278 0.0242
0.279 0.02205
0.28 0.02
0.281 0.01805
0.282 0.0162
0.283 0.01445
0.284 0.0128
0.285 0.01125
0.286 0.00980000000000001
0.287 0.00845000000000001
0.288 0.00720000000000001
0.289 0.00605000000000007
0.29 0.00500000000000006
0.291 0.00405000000000006
0.292 0.00320000000000005
0.293 0.00245000000000004
0.294 0.00180000000000004
0.295 0.00125000000000003
0.296 0.000800000000000023
0.297 0.000450000000000017
0.298 0.000200000000000011
0.299 5.00000000000056e-05
};

\addplot [
line width=\linewidthtop, B4, mark=*, mark size=0, mark options={solid}]
table {%
0 0
0.001 1.66666666666667e-07
0.002 1.33333333333333e-06
0.003 4.5e-06
0.004 1.06666666666667e-05
0.005 2.08333333333333e-05
0.006 3.6e-05
0.007 5.71666666666666e-05
0.008 8.53333333333333e-05
0.009 0.0001215
0.01 0.000166666666666667
0.011 0.000221833333333333
0.012 0.000288
0.013 0.000366166666666667
0.014 0.000457333333333333
0.015 0.0005625
0.016 0.000682666666666667
0.017 0.000818833333333333
0.018 0.000972
0.019 0.00114316666666667
0.02 0.00133333333333333
0.021 0.0015435
0.022 0.00177466666666667
0.023 0.00202783333333333
0.024 0.002304
0.025 0.00260416666666667
0.026 0.00292933333333333
0.027 0.0032805
0.028 0.00365866666666667
0.029 0.00406483333333333
0.03 0.0045
0.031 0.00496516666666667
0.032 0.00546133333333333
0.033 0.0059895
0.034 0.00655066666666667
0.035 0.00714583333333333
0.036 0.007776
0.037 0.00844216666666666
0.038 0.00914533333333333
0.039 0.0098865
0.04 0.0106666666666667
0.041 0.0114868333333333
0.042 0.012348
0.043 0.0132511666666667
0.044 0.0141973333333333
0.045 0.0151875
0.046 0.0162226666666667
0.047 0.0173038333333333
0.048 0.018432
0.049 0.0196081666666667
0.05 0.0208333333333333
0.051 0.0221085
0.052 0.0234346666666667
0.053 0.0248128333333333
0.054 0.026244
0.055 0.0277291666666667
0.056 0.0292693333333333
0.057 0.0308655
0.058 0.0325186666666667
0.059 0.0342298333333333
0.06 0.036
0.061 0.0378301666666667
0.062 0.0397213333333333
0.063 0.0416745
0.064 0.0436906666666667
0.065 0.0457708333333333
0.066 0.047916
0.067 0.0501271666666667
0.068 0.0524053333333333
0.069 0.0547515
0.07 0.0571666666666667
0.071 0.0596518333333333
0.072 0.062208
0.073 0.0648361666666666
0.074 0.0675373333333333
0.075 0.0703125
0.076 0.0731626666666666
0.077 0.0760888333333333
0.078 0.079092
0.079 0.0821731666666666
0.08 0.0853333333333333
0.081 0.0885735
0.082 0.0918946666666667
0.083 0.0952978333333333
0.084 0.098784
0.085 0.102354166666667
0.086 0.106009333333333
0.087 0.1097505
0.088 0.113578666666667
0.089 0.117494833333333
0.09 0.1215
0.091 0.125595166666667
0.092 0.129781333333333
0.093 0.1340595
0.094 0.138430666666667
0.095 0.142895833333333
0.096 0.147456
0.097 0.152112166666667
0.098 0.156865333333333
0.099 0.1617165
0.1 0.166666666666667
0.101 0.171716166666667
0.102 0.176862666666667
0.103 0.182103166666667
0.104 0.187434666666667
0.105 0.192854166666667
0.106 0.198358666666667
0.107 0.203945166666667
0.108 0.209610666666667
0.109 0.215352166666667
0.11 0.221166666666667
0.111 0.227051166666667
0.112 0.233002666666667
0.113 0.239018166666667
0.114 0.245094666666667
0.115 0.251229166666667
0.116 0.257418666666667
0.117 0.263660166666667
0.118 0.269950666666667
0.119 0.276287166666667
0.12 0.282666666666667
0.121 0.289086166666667
0.122 0.295542666666667
0.123 0.302033166666667
0.124 0.308554666666667
0.125 0.315104166666667
0.126 0.321678666666667
0.127 0.328275166666667
0.128 0.334890666666667
0.129 0.341522166666667
0.13 0.348166666666667
0.131 0.354821166666667
0.132 0.361482666666667
0.133 0.368148166666667
0.134 0.374814666666667
0.135 0.381479166666667
0.136 0.388138666666667
0.137 0.394790166666667
0.138 0.401430666666667
0.139 0.408057166666667
0.14 0.414666666666667
0.141 0.421256166666667
0.142 0.427822666666667
0.143 0.434363166666667
0.144 0.440874666666667
0.145 0.447354166666666
0.146 0.453798666666666
0.147 0.460205166666667
0.148 0.466570666666667
0.149 0.472892166666667
0.15 0.479166666666666
0.151 0.485391166666666
0.152 0.491562666666667
0.153 0.497678166666667
0.154 0.503734666666667
0.155 0.509729166666667
0.156 0.515658666666667
0.157 0.521520166666667
0.158 0.527310666666667
0.159 0.533027166666666
0.16 0.538666666666667
0.161 0.544226166666667
0.162 0.549702666666667
0.163 0.555093166666667
0.164 0.560394666666667
0.165 0.565604166666667
0.166 0.570718666666667
0.167 0.575735166666667
0.168 0.580650666666667
0.169 0.585462166666667
0.17 0.590166666666667
0.171 0.594761166666667
0.172 0.599242666666667
0.173 0.603608166666667
0.174 0.607854666666667
0.175 0.611979166666667
0.176 0.615978666666666
0.177 0.619850166666666
0.178 0.623590666666667
0.179 0.627197166666667
0.18 0.630666666666666
0.181 0.633996166666667
0.182 0.637182666666666
0.183 0.640223166666666
0.184 0.643114666666667
0.185 0.645854166666667
0.186 0.648438666666667
0.187 0.650865166666667
0.188 0.653130666666667
0.189 0.655232166666667
0.19 0.657166666666667
0.191 0.658931166666667
0.192 0.660522666666667
0.193 0.661938166666667
0.194 0.663174666666667
0.195 0.664229166666667
0.196 0.665098666666667
0.197 0.665780166666667
0.198 0.666270666666667
0.199 0.666567166666667
0.2 0.666666666666667
0.201 0.666567166666667
0.202 0.666270666666666
0.203 0.665780166666667
0.204 0.665098666666667
0.205 0.664229166666667
0.206 0.663174666666667
0.207 0.661938166666667
0.208 0.660522666666666
0.209 0.658931166666667
0.21 0.657166666666667
0.211 0.655232166666667
0.212 0.653130666666667
0.213 0.650865166666667
0.214 0.648438666666667
0.215 0.645854166666667
0.216 0.643114666666667
0.217 0.640223166666667
0.218 0.637182666666667
0.219 0.633996166666667
0.22 0.630666666666667
0.221 0.627197166666667
0.222 0.623590666666667
0.223 0.619850166666667
0.224 0.615978666666667
0.225 0.611979166666667
0.226 0.607854666666667
0.227 0.603608166666667
0.228 0.599242666666667
0.229 0.594761166666667
0.23 0.590166666666667
0.231 0.585462166666667
0.232 0.580650666666667
0.233 0.575735166666667
0.234 0.570718666666667
0.235 0.565604166666667
0.236 0.560394666666667
0.237 0.555093166666667
0.238 0.549702666666667
0.239 0.544226166666667
0.24 0.538666666666667
0.241 0.533027166666667
0.242 0.527310666666667
0.243 0.521520166666667
0.244 0.515658666666667
0.245 0.509729166666667
0.246 0.503734666666667
0.247 0.497678166666667
0.248 0.491562666666667
0.249 0.485391166666667
0.25 0.479166666666667
0.251 0.472892166666667
0.252 0.466570666666667
0.253 0.460205166666667
0.254 0.453798666666667
0.255 0.447354166666667
0.256 0.440874666666667
0.257 0.434363166666667
0.258 0.427822666666667
0.259 0.421256166666667
0.26 0.414666666666667
0.261 0.408057166666667
0.262 0.401430666666667
0.263 0.394790166666667
0.264 0.388138666666667
0.265 0.381479166666667
0.266 0.374814666666667
0.267 0.368148166666667
0.268 0.361482666666667
0.269 0.354821166666667
0.27 0.348166666666667
0.271 0.341522166666667
0.272 0.334890666666667
0.273 0.328275166666667
0.274 0.321678666666667
0.275 0.315104166666667
0.276 0.308554666666667
0.277 0.302033166666667
0.278 0.295542666666667
0.279 0.289086166666667
0.28 0.282666666666667
0.281 0.276287166666667
0.282 0.269950666666667
0.283 0.263660166666667
0.284 0.257418666666667
0.285 0.251229166666667
0.286 0.245094666666667
0.287 0.239018166666667
0.288 0.233002666666667
0.289 0.227051166666667
0.29 0.221166666666667
0.291 0.215352166666667
0.292 0.209610666666667
0.293 0.203945166666667
0.294 0.198358666666667
0.295 0.192854166666667
0.296 0.187434666666667
0.297 0.182103166666667
0.298 0.176862666666667
0.299 0.171716166666667
0.3 0.166666666666667
0.301 0.1617165
0.302 0.156865333333333
0.303 0.152112166666667
0.304 0.147456
0.305 0.142895833333333
0.306 0.138430666666667
0.307 0.1340595
0.308 0.129781333333333
0.309 0.125595166666667
0.31 0.1215
0.311 0.117494833333333
0.312 0.113578666666667
0.313 0.1097505
0.314 0.106009333333333
0.315 0.102354166666667
0.316 0.0987840000000001
0.317 0.0952978333333334
0.318 0.0918946666666667
0.319 0.0885735000000001
0.32 0.0853333333333334
0.321 0.0821731666666667
0.322 0.0790920000000001
0.323 0.0760888333333334
0.324 0.0731626666666667
0.325 0.0703125
0.326 0.0675373333333334
0.327 0.0648361666666667
0.328 0.062208
0.329 0.0596518333333334
0.33 0.0571666666666667
0.331 0.0547515
0.332 0.0524053333333333
0.333 0.0501271666666667
0.334 0.047916
0.335 0.0457708333333333
0.336 0.0436906666666667
0.337 0.0416745
0.338 0.0397213333333333
0.339 0.0378301666666667
0.34 0.036
0.341 0.0342298333333333
0.342 0.0325186666666667
0.343 0.0308655
0.344 0.0292693333333333
0.345 0.0277291666666667
0.346 0.026244
0.347 0.0248128333333333
0.348 0.0234346666666667
0.349 0.0221085
0.35 0.0208333333333333
0.351 0.0196081666666667
0.352 0.018432
0.353 0.0173038333333334
0.354 0.0162226666666667
0.355 0.0151875
0.356 0.0141973333333334
0.357 0.0132511666666667
0.358 0.012348
0.359 0.0114868333333334
0.36 0.0106666666666667
0.361 0.00988650000000003
0.362 0.00914533333333336
0.363 0.00844216666666669
0.364 0.00777600000000002
0.365 0.00714583333333335
0.366 0.00655066666666668
0.367 0.00598950000000002
0.368 0.00546133333333335
0.369 0.00496516666666668
0.37 0.00450000000000001
0.371 0.00406483333333334
0.372 0.00365866666666668
0.373 0.00328050000000001
0.374 0.00292933333333334
0.375 0.00260416666666667
0.376 0.00230400000000001
0.377 0.00202783333333334
0.378 0.00177466666666667
0.379 0.0015435
0.38 0.00133333333333334
0.381 0.00114316666666667
0.382 0.000972000000000003
0.383 0.000818833333333335
0.384 0.000682666666666668
0.385 0.000562500000000001
0.386 0.000457333333333335
0.387 0.000366166666666668
0.388 0.000288000000000001
0.389 0.000221833333333334
0.39 0.000166666666666667
0.391 0.0001215
0.392 8.53333333333336e-05
0.393 5.71666666666668e-05
0.394 3.60000000000001e-05
0.395 2.08333333333334e-05
0.396 1.06666666666667e-05
0.397 4.50000000000001e-06
0.398 1.33333333333334e-06
0.399 1.66666666666667e-07
};

\addlegendentry{$\Phi_0^0$}
\addlegendentry{$\Phi_0^1$}
\addlegendentry{$\Phi_0^2$}
\addlegendentry{$\Phi_0^3$}

\addplot [semithick, dashed]
table {%
0.1 0
0.1 1.05
};
\addplot [semithick, dashed]
table {%
0.2 0
0.2 1.05
};
\addplot [semithick, dashed]
table {%
0.3 0
0.3 1.05
};
\addplot [semithick, dashed]
table {%
0.4 0
0.4 1.05
};
\addplot [draw=B0, fill=B0, mark=*, only marks]
table{%
x  y
0 0
0.1 0
0.2 0
0.3 0
0.4 0
};

\addplot [line width=\linewidthtop, B1, mark=*, mark size=0, mark options={solid}]
table{%
x  y
0.1 0
0.1 0
0.2 0
0.3 0
0.4 0
};

\end{axis}

\end{tikzpicture}

%% file: tex/3_method.tex
\section{Method}
\subsection{Problem Definition}
For an task context $\bm{c}_0$, a robot trajectory $\bm{y}_{r, 0:T} = \{\bm{y}_{r, 0}, \dots, \bm{y}_{r, T}\}$ is generated solve this task. As the task context changed to $\bm{c}_{\tau}$ at time $\tau \in (0, T]$, this work aims to find a policy that generates trajectory-level refinements, $\Delta \bm{y}_{\tau:T}$, based on the current context $\bm{c}_{\tau}$ and original robot trajectory $\bm{y}_{r, \tau:T}$. 
A state encoder $s(\cdot)$ is used to encode $\bm{c}_{\tau}$ and $\bm{y}_{r, \tau:T}$ into a state vector $\bm{s}_{\tau} = s(\bm{c}_{\tau}, \bm{y}_{r, \tau:T})$. Following the previous section \ref{subsec:erl}, an ERL problem is formulated to optimize the policy distribution $\pi(\bm{w}|\bm{s}_{\tau})$ by maximizing the expected roll-out return $R(\bm{s}_{\tau}, \bm{w})$ of an episode using the following objective function
\begin{equation}
    J = \E_{p(\vs_{\tau}),{\pi_{\theta}}(\vw|\vs_{\tau})} ~\left[ R(\bm{s}_{\tau}, \bm{w}) - V_{\phi}(\bm{s}_{\tau}) \right], \label{eq:traj_morerl}
\end{equation}
where 
the vector $\bm{w}$ determines the start and the end of the residual action as well as the parameterization of the action sequence using BMPs. 
While MoRe-ERL adopts BBRL \cite{Otto_bbrl_2022} for the optimization, the framework is modular enough to plug in any other ERL algorithms. 


\subsection{Learning Residuals for Reference Trajectories}
\label{subsec:learn_residual}
Reference trajectories contain information for completing the given tasks and can be accessed from various sources, such as sampling-based motion planning or other learned policies. 
This information serves as a valuable prior in two key aspects: (a) it reduces the complexity of learning, thereby improving sample efficiency, and (b) it preserves dexterous robot behaviors that are challenging to learn from scratch.

\begin{figure}[t]
    \includegraphics[width=\linewidth]{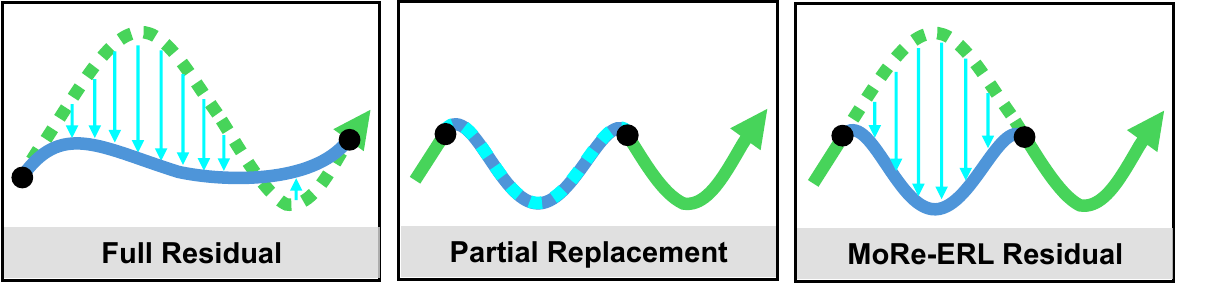}
    \vspace{0.05cm}

    \resizebox{\linewidth}{!} {
        \added{
        \renewcommand{\arraystretch}{2}
        \setlength{\aboverulesep}{0pt}
        \setlength{\belowrulesep}{0pt}
        \begin{tabular}{ccc}
        \toprule
        \textbf{Type} & \textbf{Learned Parameters} & \textbf{How to Refine} \\ 
        \hline
        Full Residual & $\bm w$ & Residuals on top of the reference\\
        Partial Replacement & $\alpha_s, \alpha_e, \bm w$ & Replace the reference\\
        \textbf{MoRe-ERL Residual} & $\alpha_s, \alpha_e, \bm w$ & Residuals on top of the reference\\
        \bottomrule
        \end{tabular}
        }
        \hspace{0.3cm}
    }
    \caption{MoRe-ERL residuals and two ablation variants. The reference trajectory is shown in green, with bold solid points indicating the timing variables $\alpha_s$ and $\alpha_e$.  
    Cyan sections show learned residuals or replacements, and the solid blue-green curve denotes the adjusted trajectory.}
    \label{fig:enter-ablation_variants}
\end{figure}

When the task context changes to $\bm{c}_{\tau}$ at time $\tau$ unexpectedly, e.g., the environment didn't evolve as anticipated, the policy generates
trajectory-level refinements $\Delta \bm{y}_{\tau:T}$ based on the context $\bm{c}_{\tau}$ and the reference trajectory $\bm{y}_{r, \tau:T}$, see Fig. \ref{fig:pipeline}. 
The reference trajectory $\bm{y}_{r, \tau:T}$ is a partially executed trajectory grounded on the previous context $\bm{c}_t$ at time $t < \tau$. 
The refinements are applied on top of the reference trajectory,
\begin{equation}
    \bm{y}_{\tau:T} = \bm{y}_{r, \tau:T} + \Delta \bm{y}_{\tau:T}.
    \label{eq:resiual_general}
\end{equation} 
To allow for greater flexibility in modifying the reference trajectory, we introduce two additional timing variables, $\alpha_s, \alpha_e \in [\tau, T]$ with $\alpha_s \leq \alpha_e$, marking the refinement's start and end on the reference trajectory.  Using these variables, the refinements at time $k \in [\tau, T]$ are defined as:
\begin{equation}
    \Delta \bm{y}_{\tau:T} = 
    \begin{cases} 
    \Delta \bm{y}_{k} & \text{for} ~ k \in [\alpha_s, \alpha_e] \\
    0 & \text{otherwise}.
    \end{cases}
    \label{eq:partial_residual}
\end{equation}
 Fig. \ref{fig:enter-ablation_variants} (right) visualizes the refinements described in Eq. (\ref{eq:partial_residual}), termed as \textit{MoRe-ERL residuals}. The reference trajectory is partially modified by the learned residuals $\Delta \bm{y}_{k}$ between $\alpha_s$ and $\alpha_e$. These two timing variables are represented by bold solid points.
To ensure continuity and smoothness at $\alpha_s$ and $\alpha_e$ during trajectory switching, we use BMPs to parameterize the residual. BMPs enforce boundary conditions up to arbitrary orders.
In this case, the boundary conditions are set to be $\bm{0}$. This ensures that the position and velocity of the refined trajectory align with the reference trajectory $\alpha_s$ and $\alpha_e$, guaranteeing the continuity and smoothness on trajectory switches. 
To be more specific, to parameterize a residual trajectory $\Delta \bm{y}_{\alpha_s:\alpha_e}$ using $N$ control points $\bm{w}_{1:N} = [w_1, ..., w_{N}]^T$, BMPs use $w_1$ and $w_2$ to enforce the boundary conditions at the start of the refinement, and use $w_{N-1}$ and $w_{N}$ at the end. The remaining control points $\bm{w}_{3:N-2} = [w_3, ..., w_{N-2}]^T$ parametrize the transition behavior $\Delta \bm{y}_{\alpha_s:\alpha_e}$.
The control points $\bm{w}_{3:N-2}$ are jointly learned with $\alpha_s$ and $\alpha_e$. 
Given the encoded state $\bm{s}_{\tau}$, the policy  $\pi(\bm{w}, \bm{\alpha}|\bm{s}_{\tau})$ 
returns the mean $\bm{\mu}_{\bm{w}, \bm{\alpha}}$ 
and the covariance matrix $\bm{\Sigma}_{\bm{w}, \bm{\alpha}}$ of a single Gaussian distribution. 
Sampling from this Gaussian distribution $[\bm w^\top_{3:N-2}, \alpha_s, \alpha_e]^\top \sim \mathcal{N}(\bm{\mu}_{\bm{w}, \bm{\alpha}}, \bm{\Sigma}_{\bm{w}, \bm{\alpha}})$, the residual sequence can be expressed as 
\begin{equation}
    \Delta \bm{y}_{\alpha_s:\alpha_e} = \bm{\Phi}_{\alpha_s:\alpha_e}^\top \bm{w}_{3:N-2}. 
    \label{eq:delta_bmp}
\end{equation}
Fig. \ref{fig:residual_diff} illustrates how the MoRe-ERL residual is applied to the reference trajectory. Different from the jerky motions produced by step-based methods, BMPs ensure smooth transitions between the reference trajectory and the refinements. 

\textbf{Refinement Variants.} In addition to partially applying residuals, we consider two alternative approaches, termed \textit{full residual} and \textit{partial replacement}. The full residual variant is a special case of Eq. (\ref{eq:partial_residual}) with $\alpha_s = \tau$ and $\alpha_e = T$, applying residuals to the entire trajectory (Fig. \ref{fig:enter-ablation_variants}, left). The partial replacement variant partially replaces segments between $\alpha_s$ and $\alpha_e$ with the refinement sequences (see Fig. \ref{fig:enter-ablation_variants}, middle).
The partial replacement modifies the reference trajectory as:
\begin{equation}
    \bm{y}_{\tau:T} = 
    \begin{cases} 
    \Delta \bm{y}_{k} & \text{for} ~ k \in [\alpha_s, \alpha_e] \\
    \bm{y}_{r,k} & \text{otherwise}.
    \end{cases}
    \label{eq:partial_replacement}
\end{equation}
When parameterizing trajectories with BMPs in partial replacement, boundary conditions are set to match the position and velocity of the reference trajectory at $\alpha_s$ and $\alpha_e$. We exclude full replacement from consideration as it equates to learning the trajectory from scratch.

Among MoRe-ERL residuals and these two variants, MoRe-ERL residuals demonstrate the best overall performance across various scenarios. Learning residuals leverages prior knowledge embedded in reference trajectories, preserving critical maneuvers and enhancing task completion. The identification of $\alpha_s$ and $\alpha_e$ contributes to retaining essential behaviors. Further details are provided in Section \ref{sec:exp}.

\input{images/tex/0_residual_plot}

%% file: images/tex/0_residual_plot.tex
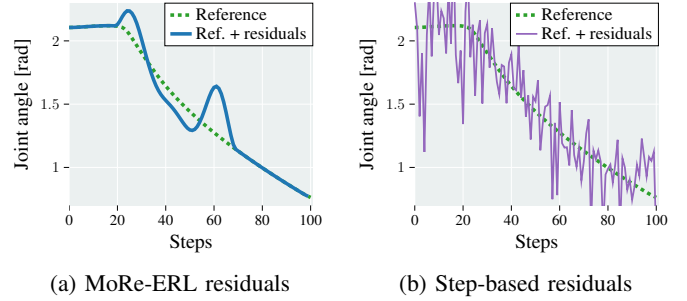
\begin{figure}[t]
    \centering
    \begin{subfigure}{0.235\textwidth}
        \resizebox{\textwidth}{!}{\input{images/tex/residual_data}}
        \caption{\centering MoRe-ERL residuals}
    \end{subfigure}
    \hfill
    \begin{subfigure}{0.235\textwidth}
        \resizebox{\textwidth}{!}{\input{images/tex/step_based_data}}
        \caption{\centering Step-based residuals}
    \end{subfigure}
    \caption{Random roll-out using MoRe-ERL and step-based residual method. In the demonstrated case, the trajectory with MoRe-ERL residuals (\centeredinline{BlueAccent}{1.5pt}) deviates from the reference trajectory at $\alpha_s = 20$ and converges back at $\alpha_e = 70$. }
    \label{fig:residual_diff}
\end{figure}

%% file: images/tex/residual_data.tex
%
%

\begin{tikzpicture}
\def\res_color{C0}
\def\replace_color{C1}
\def\whole_color{C3}
\def\bbrl_color{C9}
\def\ppo_color{C2}
\def\sac_color{C4}
\def\linewidthtop{1mm}
\def\linewidthothers{0.5mm}

\begin{axis}[
legend cell align={left},
legend style={at={(1.0, 1.0)}, anchor=north east, legend columns=1, font=\Large},
tick align=outside,
tick pos=left,
x grid style={white},
scaled x ticks=false,
xticklabels={,0,20,40,60,80,100},
xlabel={Steps},
xmajorgrids,
xmin=0, xmax=101,
xtick style={color=black},
y grid style={white},
ylabel={Joint angle [rad]},
ymajorgrids,
ymin=0.7, ymax=2.3,
ytick style={color=black},
axis background/.style={fill=plot_background},
label style={font=\large},
tick label style={font=\large},
x axis line style={draw=none},
y axis line style={draw=none},
xlabel style={font=\fontsize{16}{18}\selectfont},
ylabel style={font=\fontsize{16}{18}\selectfont},
]

\addplot [line width=\linewidthtop, \ppo_color, mark=*, mark size=0, mark options={solid}, dashed]
table {%
0   2.1058063247107244
1   2.1060561994196316
2   2.106510118459146
3   2.107243653803568
4   2.108354324547985
5   2.1095752891565063
6   2.110908097108268
7   2.1121979261983554
8   2.1134591203194053
9   2.114646221583195
10   2.115773968159425
11   2.1168493429525146
12   2.117779106037891
13   2.118621354888194
14   2.1193296820659198
15   2.1197897507371035
16   2.120082205021909
17   2.1200020421414303
18   2.119416204596885
19   2.1184538392893817
20   2.1160986248327047
21   2.112037822663548
22   2.1061613948649773
23   2.0936375627459003
24   2.07629587850355
25   2.0504971393508935
26   2.021949352718131
27   1.9925359861707592
28   1.9633672350786249
29   1.934056445729833
30   1.9046616290536034
31   1.876095687180513
32   1.8479072775701213
33   1.8202042665831997
34   1.7931864168071117
35   1.7665275944135173
36   1.7404997644809572
37   1.7152354183821288
38   1.6904345410521229
39   1.6666214262175103
40   1.643396220160548
41   1.6206581385193526
42   1.598874964533733
43   1.5774696147777054
44   1.556605617402336
45   1.5363887751189576
46   1.5164885169265097
47   1.4971345780540601
48   1.4782017447561853
49   1.4595345542915181
50   1.4413991157549098
51   1.4235250553258316
52   1.4059123730042842
53   1.3887385099586695
54   1.3717510883547186
55   1.3550486049976826
56   1.338649447545745
57   1.3224115034758732
58   1.3064697796296847
59   1.2907329499491675
60   1.2751371969426948
61   1.2598432060009437
62   1.2446803710487313
63   1.2296758278080309
64   1.2149072820825437
65   1.2002472709756342
66   1.18576446676096
67   1.1714513891242917
68   1.1572359196520319
69   1.1432110766138983
70   1.1293037097346905
71   1.115487713201314
72   1.1018656681792902
73   1.0883223461409184
74   1.0748895812853498
75   1.06160285033059
76   1.0483882988585633
77   1.0352993590663073
78   1.0223172432647931
79   1.0094014942063385
80   0.9966229817233381
81   0.9839189193711506
82   0.971284795020636
83   0.958779802367938
84   0.9463307791688418
85   0.9339661815185989
86   0.921700071072884
87   0.9094860712065187
88   0.8973683207452448
89   0.8853230826685854
90   0.8733265649965871
91   0.8614360444603938
92   0.8495957059275074
93   0.8378122403317027
94   0.8261205672375316
95   0.8144711402787885
96   0.8028898547385984
97   0.7914924757903323
98   0.7805253463172289
99   0.7715223592198192
100  0.7628769680013242
};

\addplot [line width=\linewidthtop, \res_color, mark=*, mark size=0, mark options={solid}]
table {%
0 2.1058063247107244
1 2.1060561994196316
2 2.106510118459146
3 2.107243653803568
4 2.108354324547985
5 2.1095752891565063
6 2.110908097108268
7 2.1121979261983554
8 2.1134591203194053
9 2.114646221583195
10 2.115773968159425
11 2.1168493429525146
12 2.117779106037891
13 2.118621354888194
14 2.1193296820659198
15 2.1197897507371035
16 2.120082205021909
17 2.1200020421414303
18 2.119416204596885
19 2.1184538392893817
20 2.129129045821503
21 2.156712859928372
22 2.1912214697904
23 2.219677380675168
24 2.237492870304812
25 2.236339400069239
26 2.218963922954337
27 2.1860171334600356
28 2.139104586257748
29 2.0800631193042807
30 2.012799519750129
31 1.9425681132373749
32 1.8726187970498729
33 1.806017839586015
34 1.745180135550436
35 1.6912519646546824
36 1.645234368541607
37 1.6072461593179312
38 1.5762304993922909
39 1.5512118657159348
40 1.5296038901339215
41 1.5090530828891455
42 1.4883283458337055
43 1.4657641327716324
44 1.4410457861407011
45 1.4144113266555176
46 1.3862726783749884
47 1.3582125274587766
48 1.332068798741781
49 1.3102596388347463
50 1.296161669923611
51 1.2923446497090714
52 1.3005710726359752
53 1.3216299086269587
54 1.3547288660742163
55 1.3982758305397511
56 1.4494477170560143
57 1.504109080713956
58 1.5572514051389468
59 1.6024879216917975
60 1.6324824203695136
61 1.6410505827194648
62 1.6245106718566842
63 1.5823853893944109
64 1.51725518056582
65 1.4345020782094446
66 1.3427111407502559
67 1.253397829165357
68 1.1810212518698444
69 1.1432110766138983
70 1.1293037097346905
71 1.115487713201314
72 1.1018656681792902
73 1.0883223461409184
74 1.0748895812853498
75 1.06160285033059
76 1.0483882988585633
77 1.0352993590663073
78 1.0223172432647931
79 1.0094014942063385
80 0.9966229817233381
81 0.9839189193711506
82 0.971284795020636
83 0.958779802367938
84 0.9463307791688418
85 0.9339661815185989
86 0.921700071072884
87 0.9094860712065187
88 0.8973683207452448
89 0.8853230826685854
90 0.8733265649965871
91 0.8614360444603938
92 0.8495957059275074
93 0.8378122403317027
94 0.8261205672375316
95 0.8144711402787885
96 0.8028898547385984
97 0.7914924757903323
98 0.7805253463172289
99 0.7715223592198192
};

\addlegendentry{Reference}
\addlegendentry{Ref. + residuals}

\end{axis}
\end{tikzpicture}

%% file: images/tex/step_based_data.tex
%
%

\begin{tikzpicture}
\def\res_color{C0}
\def\replace_color{C1}
\def\whole_color{C3}
\def\bbrl_color{C9}
\def\ppo_color{C2}
\def\sac_color{C4}
\def\linewidthtop{1mm}
\def\linewidthothers{0.5mm}

\begin{axis}[
legend cell align={left},
legend style={at={(1.0, 1.0)}, anchor=north east, legend columns=1, font=\Large},
tick align=outside,
tick pos=left,
x grid style={white},
scaled x ticks=false,
xticklabels={,0,20,40,60,80,100},
xlabel={Steps},
xmajorgrids,
xmin=0, xmax=101,
xtick style={color=black},
y grid style={white},
ymajorgrids,
ymin=0.7, ymax=2.3,
ytick style={color=black},
axis background/.style={fill=plot_background},
label style={font=\large},
tick label style={font=\large},
x axis line style={draw=none},
y axis line style={draw=none},
ylabel={Joint angle [rad]},
xlabel style={font=\fontsize{16}{18}\selectfont},
ylabel style={font=\fontsize{16}{18}\selectfont},
]

\addplot [line width=\linewidthtop, \ppo_color, mark=*, mark size=0, mark options={solid}, dashed]
table {%

0   2.1058063247107244
1   2.1060561994196316
2   2.106510118459146
3   2.107243653803568
4   2.108354324547985
5   2.1095752891565063
6   2.110908097108268
7   2.1121979261983554
8   2.1134591203194053
9   2.114646221583195
10   2.115773968159425
11   2.1168493429525146
12   2.117779106037891
13   2.118621354888194
14   2.1193296820659198
15   2.1197897507371035
16   2.120082205021909
17   2.1200020421414303
18   2.119416204596885
19   2.1184538392893817
20   2.1160986248327047
21   2.112037822663548
22   2.1061613948649773
23   2.0936375627459003
24   2.07629587850355
25   2.0504971393508935
26   2.021949352718131
27   1.9925359861707592
28   1.9633672350786249
29   1.934056445729833
30   1.9046616290536034
31   1.876095687180513
32   1.8479072775701213
33   1.8202042665831997
34   1.7931864168071117
35   1.7665275944135173
36   1.7404997644809572
37   1.7152354183821288
38   1.6904345410521229
39   1.6666214262175103
40   1.643396220160548
41   1.6206581385193526
42   1.598874964533733
43   1.5774696147777054
44   1.556605617402336
45   1.5363887751189576
46   1.5164885169265097
47   1.4971345780540601
48   1.4782017447561853
49   1.4595345542915181
50   1.4413991157549098
51   1.4235250553258316
52   1.4059123730042842
53   1.3887385099586695
54   1.3717510883547186
55   1.3550486049976826
56   1.338649447545745
57   1.3224115034758732
58   1.3064697796296847
59   1.2907329499491675
60   1.2751371969426948
61   1.2598432060009437
62   1.2446803710487313
63   1.2296758278080309
64   1.2149072820825437
65   1.2002472709756342
66   1.18576446676096
67   1.1714513891242917
68   1.1572359196520319
69   1.1432110766138983
70   1.1293037097346905
71   1.115487713201314
72   1.1018656681792902
73   1.0883223461409184
74   1.0748895812853498
75   1.06160285033059
76   1.0483882988585633
77   1.0352993590663073
78   1.0223172432647931
79   1.0094014942063385
80   0.9966229817233381
81   0.9839189193711506
82   0.971284795020636
83   0.958779802367938
84   0.9463307791688418
85   0.9339661815185989
86   0.921700071072884
87   0.9094860712065187
88   0.8973683207452448
89   0.8853230826685854
90   0.8733265649965871
91   0.8614360444603938
92   0.8495957059275074
93   0.8378122403317027
94   0.8261205672375316
95   0.8144711402787885
96   0.8028898547385984
97   0.7914924757903323
98   0.7805253463172289
99   0.7715223592198192
100  0.7628769680013242        

};

\addplot [line width=\linewidthothers, \sac_color, mark=*, mark size=0, mark options={solid}]
table {%
0   2.305403
1   2.097606
2   1.403605
3   1.902015
4   1.122178
5   2.047042
6   2.425503
7   1.877845
8   2.390017
9   2.493560
10  2.046225
11  2.236660
12  2.178521
13  1.901195
14  2.437204
15  1.977808
16  2.711627
17  1.888706
18  1.665924
19  1.923723
20  1.342568
21  2.232383
22  2.511718
23  1.870906
24  2.127689
25  1.566972
26  2.050978
27  2.163979
28  1.505166
29  2.061042
30  2.133729
31  1.788569
32  2.098760
33  1.848841
34  1.676739
35  1.578031
36  1.809765
37  1.797277
38  1.507183
39  1.827557
40  1.894313
41  1.624097
42  1.610570
43  1.965309
44  1.550590
45  1.768903
46  1.400506
47  1.569620
48  1.221014
49  1.559359
50  1.522366
51  1.689641
52  1.135378
53  1.348779
54  1.480572
55  1.492231
56  1.566600
57  0.748742
58  1.352000
59  0.781125
60  1.516102
61  1.191078
62  1.156291
63  1.159707
64  1.015986
65  1.452930
66  1.097788
67  1.004043
68  1.153182
69  0.896139
70  1.046036
71  1.268439
72  1.380126
73  0.525875
74  1.133827
75  1.064611
76  1.134199
77  1.045934
78  0.507602
79  0.943780
80  0.989093
81  1.022743
82  0.879055
83  1.022861
84  1.218144
85  0.499786
86  1.072439
87  1.000273
88  0.823186
89  0.902682
90  0.742618
91  0.848687
92  1.021871
93  0.934042
94  1.227562
95  1.080736
96  0.928052
97  1.083065
98  1.052763
99  1.123361
100 0.424376

};

\addlegendentry{Reference}
\addlegendentry{Ref. + residuals}

\end{axis}
\end{tikzpicture}

%% file: tex/4_experiments.tex
\section{Experiments}
\label{sec:exp}

\begin{figure*}[h]
    \centering
    \begin{subfigure}{0.45\textwidth}
        \includegraphics[width=\textwidth]{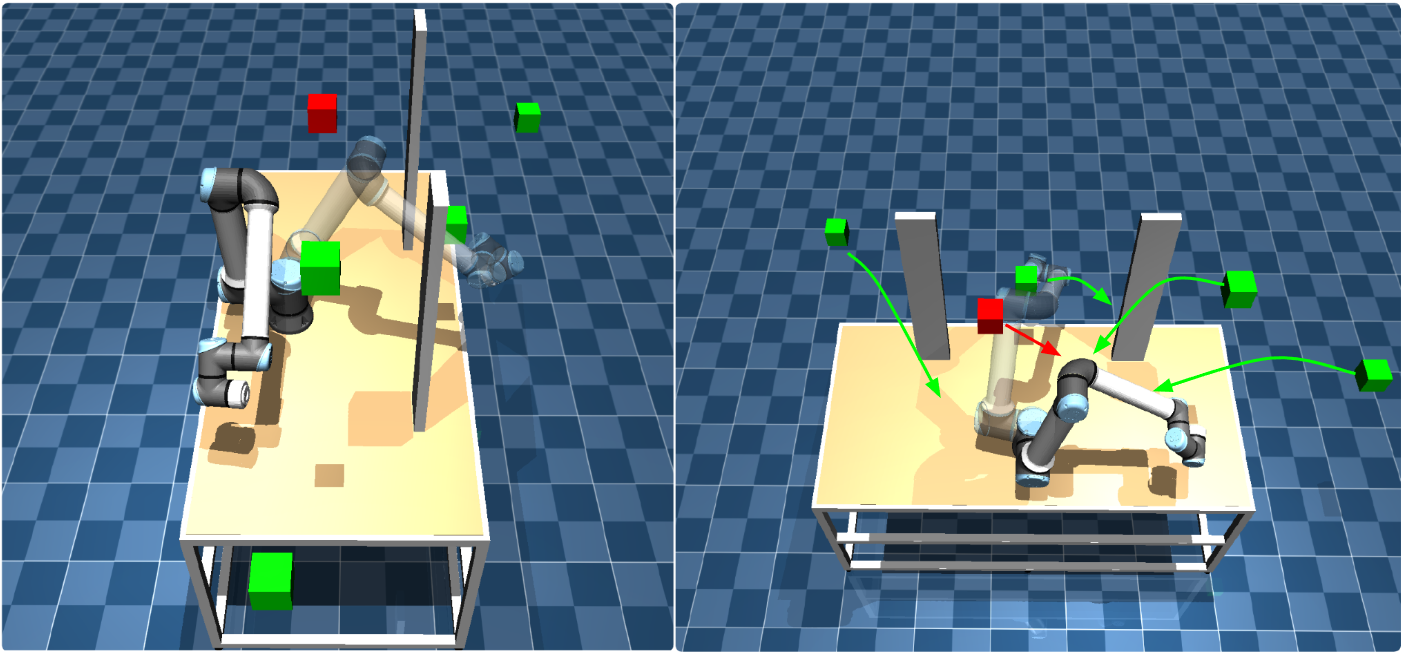}
        \caption{Multi-box}
        \label{subfig:multi-box}
    \end{subfigure}
    \begin{subfigure}{0.266\textwidth}
        \includegraphics[width=\textwidth]{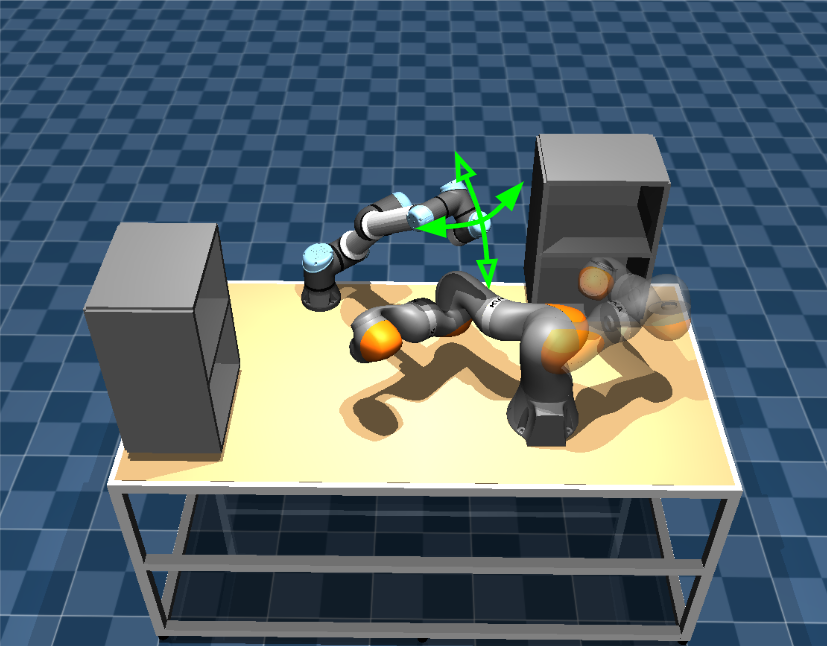}
        \caption{Dual-arm}
        \label{subfig:dual-arm}
    \end{subfigure}
    \begin{subfigure}{0.222\textwidth}
        \includegraphics[width=\textwidth]{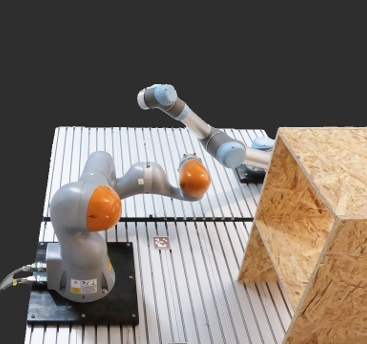}
        \caption{Real-world hardware setup}
        \label{subfig:real-world-above}
    \end{subfigure}
    \caption{Simulation experiments in Mujoco and real-world experiment setup. The robot with reduced opacity in simulation indicates the desired goal. In (a), green boxes follow parabolic trajectories, while the red box moves at a constant velocity. Each box is released at a different timestamp. In (b), the UR5 robot moves in the directions shown by the green arrows.}
    \label{fig:experiment_setup}
\end{figure*}

\input{images/tex/0_multi_box}

More-ERL is evaluated in two simulation experiments in MuJoCo\cite{todorov2012mujoco} and an experiment with real-world hardware. In both simulation scenarios, baseline methods include sampling-based motion planning methods, episode-based RL methods, and step-based RL methods. While the episode-based methods work perfectly with non-Markovian rewards, the step-based methods perform poorly in such settings \cite{Otto_bbrl_2022}.
For an \textit{unfavourable} comparison against our method, we shape a Markovian reward grounded on the performance of step-based methods and evaluate MoRe-ERL on both Markovian and non-Markovian rewards. Despite this \textit{unfavourable} setting, MoRe-ERL shows comparable performance with Markovian rewards in both tasks and achieves superior performance with non-Markovian rewards.

The Markovian returns of an episode with $n_e$ steps summarize rewards from each step with a discount factor $\gamma$
\begin{equation}
    R_M = \sum_{t=\tau}^{n_e} \gamma^t( \beta_{c}r_{c, t} + \beta_{g}r_{g, t} + \beta_{l}r_{l, t}),
\end{equation}
where $r_{c, t}$, $r_{g, t}$, and $r_{l, t}$ indicate the reward terms at $t$ regarding collision, goal reaching, and joint limit violation, respectively. The collision reward is assigned $r_{c, t} = -1$ when a collision occurs. The other two terms are both computed using L2-Norm.
These reward terms are weighted by corresponding coefficients $\beta_{[\cdot]}$. The non-Markovian return $R_{NM}$ does not collect rewards regarding collision and goal reaching at every step, but at the end of the episode
\begin{equation}
    R_{NM} = \underbrace{\beta_{c}r_{c} +  \beta_{g}r_{g}}_{\mathrm{Non-Markovian}} + \beta_{l}\sum_i^{n_e} \gamma^ir_{l, t}.
\end{equation}
This setting links the reward closer to the definition of success and avoids potential reward hacking. 
The results of both simulation scenarios show that MoRe-ERL with non-Markovian rewards achieved a significantly higher performance than baselines, see Fig. \ref{fig:learning_curve} and TABLE \ref{table:multi-box}. A video for simulation and real-world experiments is attached to the multimedia materials. Parameters for RL and BMPs are selected based on a grid search over the parameter space.

\subsection{Multi-Box}
\label{subsec:multi-box}
The multi-box scenario has an UR10e robot mounted on a table, which travels among three regions separated by two bars to complete arbitrary tasks, see Fig. \ref{subfig:multi-box}. While the robot is operating, dynamic obstacles enter the robot's working space, either moving with constant velocity or following parabolic paths.
In every episode, the initial joint configurations and goal
are randomly selected. The obstacles are released to move at different time stamps. The trajectories of the obstacles are designed to be \textit{adversarial} to our method that at least one of the obstacles will hit the robot if the robot follows the reference trajectory.
The episode terminates when the goal or the pre-defined maximum duration of 3 seconds is reached. For episode-based methods, the observation includes (a) the current robot configurations and velocity, (b) the goal of the robot, and (c) the parameters from which the agent can infer the trajectories of the dynamic obstacles, such as their position, velocity, and end position. The end positions of the obstacles serve the purpose of distinguishing the parabolic trajectories from the ones with constant velocity. On the other hand, the step-based methods receive a new observation at every simulation step and return the next joint position as an action. The observation for step-based methods additionally includes the current timestamp.

The residual version of both episode-based and step-based methods must be aware of the reference, on which the residual acts. Reference trajectories is generated by RRT-Connect\cite{kuffner2000rrt}.
A representation of the reference is included in the observation of the residual methods. The ERL residual methods use five intermediate waypoints of the reference trajectory as the representation, while the step-based methods use the next reference action. The reference trajectories are generated using a sampling-based motion planner. Note that this approach is agnostic to the choice of planner, allowing an alternative motion generator to be seamlessly integrated.


The results of baseline methods and ablations are summarized in TABLE \ref{table:multi-box}, collected with a single core on an Intel i9-9900K CPU. 
Sampling-based methods were allocated 1 second of planning time for each problem in a space-time state space ($\mathbb{R}^{6+1}$), with 6 DoFs for robot joints and 1 DoF for time. 
For ST-RRT*, the maximum arrival time was set to 3 seconds, while RRT-Connect\cite{kuffner2000rrt} used a fixed arrival time of 3 seconds due to its inability to handle unknown arrival times.
For RL methods, the Markovian reward uses $\beta_c = 5$, $\beta_g=20$ and $\beta_l = 0$, and the non-Markovian reward uses $\beta_c = 10$, $\beta_g=40$ and $\beta_l = 1$. 

The learning curves in Fig. \ref{subfig:learning_curve_multi_box} demonstrate that MoRe-ERL achieves higher sample efficiency and yields better performance compared to ERL trained from scratch and step-based RL approaches, under both Markovian and non-Markovian reward settings. 
However, with a sufficient planning time budget, ST-RRT* achieves a success rate of 92.8\% in this task. When the budget is reduced to 100 ms, the success rate of ST-RRT* drops to below 70\%. In contrast, MoRe-ERL maintains a significantly faster inference time of 10.1 ms while achieving a competitive success rate of 88.9\%. 
Note that the scenarios are initialized to be \textit{adversarial} to residual methods. ST-RRT* does not suffer from such a disadvantage, which reduces the task complexity for ST-RRT*. A common solution is to wait for the boxes to settle and approach the goal. For a more complicated task where the environment is constantly moving, such as the dual-arm task, MoRe-ERL significantly outperforms ST-RRT*.

\subsection{Dual-Arm}
\label{subsec:dual_arm}
The dual-arm scenario involves a UR5 and a KUKA iiwa 14 robot, mounted on a shared workspace, as shown in Fig. \ref{subfig:dual-arm}. The UR5 follows pre-defined trajectories, acting as a dynamic part of the environment, while the KUKA iiwa 14 actively avoids collisions and moves toward the goal pose. 

Similar to the multi-box environment, episodes are randomly initialized and terminate when either the goal is reached or the pre-defined maximum duration of 5 seconds elapses.
For step-based methods, the observation includes (a) the position and velocity of both robots, (b) the goal of the iiwa robot and (c) the current timestamp. As in \ref{subsec:multi-box}, the step-based residual method includes the next reference action in the observation, while MoRe-ERL incorporates 5 intermediate waypoints from the reference trajectories.
Sampling-based baselines reported in TABLE \ref{table:multi-box} are given 5 seconds planning time budgets for each problem in a space-time state space $\mathbb R^{7+1}$. The maximum arrival time of ST-RRT* is set to 5 seconds, and RRT-Connect has a fixed arrival time. The Markovian reward is weighted by $\beta_c = 5$, $\beta_g=20$ and $\beta_l = 0$ and the non-Markovian setting is same as \ref{subsec:multi-box}.

MoRe-ERL achieves a 76.7\% success rate under the non-Markovian reward, outperforming ST-RRT*, which succeeds in only 39.2\% of cases after 5 seconds of planning. 
It is worth mentioning that the test cases are randomly generated, and some of them may not be solvable. To approximate the upper bound of success rate, we increased the planning budget to 120 seconds, where ST-RRT* achieved success in 86\% of the test cases. \added{No significant performance gain is shown if the planning budget increases further.} Fig. \ref{subfig:learning_curve_dual_arm} shows that step-based methods failed to learn a reasonable policy in both Markovian and non-Markovian reward settings. 
MoRe-ERL’s superior performance stems from its ability to identify the residual intervals while retaining critical behaviors, e.g., retracting from a shelf. These behaviors are usually difficult to learn from scratch.

\input{images/tex/0_result_table}

\subsection{Ablation Studies}
\textbf{Refinement Variants.} 
We conducted ablation studies on different trajectory refinement strategies illustrated in Fig. \ref{fig:enter-ablation_variants}. 
The results in Fig. \ref{subfig:learning_curve_multi_box} demonstrate that full residual achieves slightly better sample efficiency compared to MoRe-ERL in the multi-box task. This behavior can be attributed to the fact that full residual is a special case of MoRe-ERL with $\alpha_s=\tau$ and $\alpha_e=T$, leading to a reduced dimension of the searching space.
However, in scenarios that require dexterous robot behaviors, such as the dual-arm task,
identifying the residual interval using $\alpha_s$ and $\alpha_e$ shows clear advantages in both convergence speed and final performance.
Meanwhile, partial replacement underperforms in both tasks, highlighting the effectiveness of learning motion residuals. 

\textbf{Movement Primitives.} \added{Prior work \cite{tan2011potential, park2008movement} has modeled obstacles in the task space using potential fields, which are explicitly incorporated into the DMP formulation as coupling terms. However, these methods are limited to simple obstacle geometries, such as a single moving sphere, and therefore do not scale to our experimental setup, which involves complex static and dynamic structures like bookshelves and moving robot arms. To demonstrate the benefits of BMPs, we replace them with DMPs in both the learning-from-scratch (ERL + DMPs) and residual settings. Table \ref{table:multi-box} shows that BMPs consistently outperform DMPs, as DMPs achieve high collision-free rates but often fail to reach the goal.}

\subsection{Real-World Experiment}
The hardware setup aligns with \ref{subsec:dual_arm}, with modifications with respect to real-world calibration between two robots. Fig. \ref{fig:first_page} and Fig. \ref{subfig:real-world-above} provide two distinct views of the setup. Initially, the policy is trained in simulation using the procedure outlined in \ref{subsec:dual_arm}, achieving a success rate of approximately 95\%. The trained policy is subsequently deployed on the real-world hardware. Episodes are designed such that the goal of the current episode becomes the starting point of the next, allowing for seamless sequential rollouts.
The policy learned by MoRe-ERL demonstrates a minimal gap in transferring from simulation to the real world. 

%% file: images/tex/0_multi_box.tex
\newcommand\subfigwidth{0.235}
\newcommand\multiboxxlable{{,0,0.5,1,1.5,2, 2.5}}
\newcommand\multiboxxmax{26000000.05}
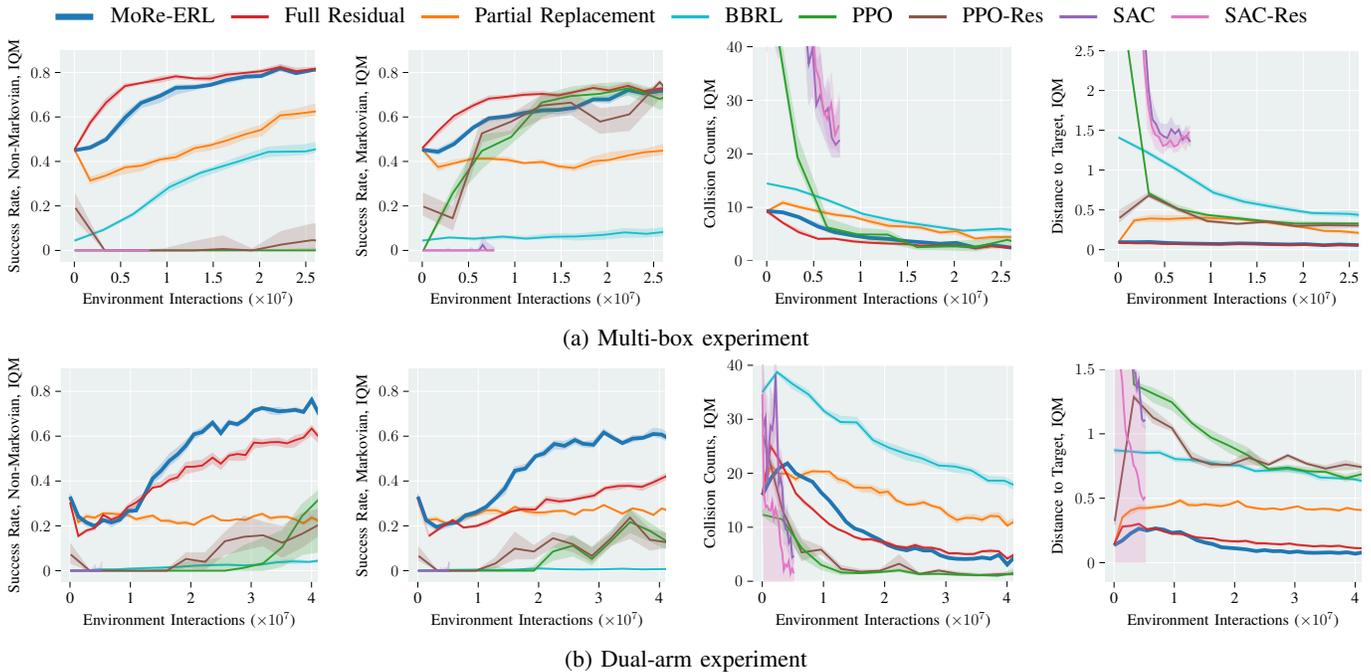
\begin{figure*}[h]
    \centering
    \hspace*{\fill}%
    \resizebox{0.95\textwidth}{!}{
        \input{images/tex/legend}
    }%
    \hspace*{\fill}%

    \begin{subfigure}{\textwidth}        
        \resizebox{\subfigwidth\textwidth}{!}{\input{images/tex/multi_box_nm_success}}%
        \hfill
        \resizebox{\subfigwidth\textwidth}{!}{\input{images/tex/multi_box_m_success}}%
        \hfill
        \resizebox{\subfigwidth\textwidth}{!}{\input{images/tex/multi_box_nm_collision_counts}}%
        \hfill
        \resizebox{\subfigwidth\textwidth}{!}{\input{images/tex/multi_box_nm_dist2target}}%
        \caption{Multi-box experiment}
        \label{subfig:learning_curve_multi_box}
    \end{subfigure}
    \begin{subfigure}{\textwidth}        
        \resizebox{\subfigwidth\textwidth}{!}{\input{images/tex/dual_arm_nm_success}}%
        \hfill
        \resizebox{\subfigwidth\textwidth}{!}{\input{images/tex/dual_arm_m_success}}%
        \hfill
        \resizebox{\subfigwidth\textwidth}{!}{\input{images/tex/dual_arm_nm_collision_counts}}%
        \hfill
        \resizebox{\subfigwidth\textwidth}{!}{\input{images/tex/dual_arm_nm_dist2target}}%
        \caption{Dual-arm experiment}
        \label{subfig:learning_curve_dual_arm}
    \end{subfigure}

          
    \caption{From left to right: (1) Success rate with non-Markovian reward; (2) Success rate with Markovian reward; (3) Collision counts; (4) Distance to target. Better results from the two reward settings are shown in (3) and (4).
    MoRe-ERL and its variant full residual show an advantage with respect to sample efficiency and task performance compared to baseline methods in both tasks, even with Markovian rewards. In the multi-box experiment, the full residual variant achieves better sample efficiency due to its smaller search space. In the scenarios requiring more dexterous behaviors, such as the dual-arm task, MoRe-ERL's ability to identify the residual interval shows clear advantages compared to applying residual to the entire trajectory.    
    }
    \label{fig:learning_curve}
\end{figure*}

%% file: images/tex/legend.tex
\begin{tikzpicture} 
\def\linewidthtcp{1mm}
\def\linewidthothers{0.5mm}

    \begin{axis}[%
    hide axis,
    xmin=10,
    xmax=50,
    ymin=0,
    ymax=0.4,
    legend style={
        draw=white!15!black,
        legend cell align=left,
        legend columns=-1, 
        legend style={
            draw=none,
            column sep=1ex,
            line width=1pt
        }
    },
    ]
    \addlegendimage{C0, line width=\linewidthtcp}
    \addlegendentry{MoRe-ERL};
    \addlegendimage{C3, line width=\linewidthothers}
    \addlegendentry{Full Residual};
    \addlegendimage{C1, line width=\linewidthothers}
    \addlegendentry{Partial Replacement};
    \addlegendimage{C9, line width=\linewidthothers}
    \addlegendentry{BBRL};
    \addlegendimage{C2, line width=\linewidthothers}
    \addlegendentry{PPO};
    \addlegendimage{C5, line width=\linewidthothers}
    \addlegendentry{PPO-Res};
    \addlegendimage{C4, line width=\linewidthothers}
    \addlegendentry{SAC}; 
    \addlegendimage{C6, line width=\linewidthothers}
    \addlegendentry{SAC-Res}; 
    \end{axis}
\end{tikzpicture}

%% file: images/tex/multi_box_nm_success.tex
%
%

\begin{tikzpicture}
\def\res_color{C0}
\def\replace_color{C1}
\def\whole_color{C3}
\def\bbrl_color{C9}
\def\ppo_color{C2}
\def\sac_color{C4}
\def\linewidthtop{1mm}
\def\linewidthothers{0.5mm}

\begin{axis}[
legend cell align={left},
legend style={fill opacity=0.8, draw opacity=1, text opacity=1, draw=lightgray204, at={(0.03,0.03)},  anchor=north west},
tick align=outside,
tick pos=left,
x grid style={white},
scaled x ticks=false,
xticklabels={,0,0.5,1,1.5,2, 2.5},
xlabel={Environment Interactions ($\times 10^7$)},
xmajorgrids,
xmin=-1500000, xmax=\multiboxxmax,
xtick style={color=black},
y grid style={white},
ylabel={Success Rate, Non-Markovian, IQM},
ymajorgrids,
ymin=-0.05, ymax=0.9,
ytick style={color=black},
axis background/.style={fill=plot_background},
label style={font=\large},
tick label style={font=\large},
x axis line style={draw=none},
y axis line style={draw=none},
]
\path [draw=\bbrl_color, fill=\bbrl_color, opacity=0.2]
(axis cs:19400,0.0501644741743803)
--(axis cs:19400,0.0411184206604958)
--(axis cs:3112700,0.087376644165488)
--(axis cs:6333000,0.153009697143011)
--(axis cs:10275100,0.266533724119128)
--(axis cs:13589000,0.326957372266501)
--(axis cs:17586100,0.385593492121226)
--(axis cs:20931700,0.421702201356736)
--(axis cs:24961300,0.431794933037173)
--(axis cs:28323300,0.462639258875239)
--(axis cs:31683900,0.487030760883715)
--(axis cs:35726000,0.494124078995261)
--(axis cs:39105500,0.534032461108462)
--(axis cs:43159900,0.553763902915667)
--(axis cs:46545200,0.572451512301561)
--(axis cs:50609500,0.590895931275429)
--(axis cs:53996700,0.587042996984665)
--(axis cs:57383400,0.586545426038551)
--(axis cs:61450800,0.611338981036432)
--(axis cs:64838500,0.608155447960251)
--(axis cs:68906600,0.631808492658727)
--(axis cs:72298600,0.6099247620358)
--(axis cs:76370300,0.615746444207579)
--(axis cs:79765700,0.625929788011846)
--(axis cs:83157500,0.636696395006606)
--(axis cs:87230500,0.647192355471969)
--(axis cs:90624600,0.631105935187626)
--(axis cs:94697100,0.642386149531503)
--(axis cs:98094600,0.639785041718468)
--(axis cs:102170100,0.665844444006954)
--(axis cs:105564600,0.6594182173037)
--(axis cs:108958800,0.65618356657487)
--(axis cs:113034000,0.666369519569232)
--(axis cs:116432700,0.660508553776649)
--(axis cs:120507600,0.666298346258253)
--(axis cs:123901200,0.667565900310935)
--(axis cs:127975600,0.688924587665817)
--(axis cs:131373700,0.664028207623597)
--(axis cs:134769900,0.683449970578908)
--(axis cs:138848400,0.685097579113087)
--(axis cs:142246700,0.672745880118075)
--(axis cs:146320600,0.678962066274258)
--(axis cs:149719000,0.681939422888057)
--(axis cs:153801300,0.686754842781838)
--(axis cs:157203500,0.682585504143578)
--(axis cs:160601700,0.682012104136625)
--(axis cs:164680900,0.685734218990237)
--(axis cs:168080600,0.693252362091123)
--(axis cs:172161200,0.691586085941003)
--(axis cs:175565900,0.698290287141868)
--(axis cs:179652200,0.69058605474341)
--(axis cs:179652200,0.758524981791922)
--(axis cs:179652200,0.758524981791922)
--(axis cs:175565900,0.743724100763666)
--(axis cs:172161200,0.756559417484055)
--(axis cs:168080600,0.753970121978221)
--(axis cs:164680900,0.735605425327413)
--(axis cs:160601700,0.750873712343695)
--(axis cs:157203500,0.729088805252804)
--(axis cs:153801300,0.753307229352952)
--(axis cs:149719000,0.749683059473615)
--(axis cs:146320600,0.754229077981894)
--(axis cs:142246700,0.722963178522424)
--(axis cs:138848400,0.746446171309167)
--(axis cs:134769900,0.742742606313139)
--(axis cs:131373700,0.737869519079703)
--(axis cs:127975600,0.748337664921386)
--(axis cs:123901200,0.727491873537888)
--(axis cs:120507600,0.730980225474352)
--(axis cs:116432700,0.719228257252392)
--(axis cs:113034000,0.719006893817944)
--(axis cs:108958800,0.727789547418166)
--(axis cs:105564600,0.727872217243342)
--(axis cs:102170100,0.733017870253567)
--(axis cs:98094600,0.711119894067682)
--(axis cs:94697100,0.706363905114389)
--(axis cs:90624600,0.701801343352195)
--(axis cs:87230500,0.695217436568613)
--(axis cs:83157500,0.693098022866802)
--(axis cs:79765700,0.702524747522393)
--(axis cs:76370300,0.709608794581357)
--(axis cs:72298600,0.659870195243857)
--(axis cs:68906600,0.69743245553476)
--(axis cs:64838500,0.667947732007942)
--(axis cs:61450800,0.667316587336521)
--(axis cs:57383400,0.646344173107912)
--(axis cs:53996700,0.638834002116574)
--(axis cs:50609500,0.634093588189865)
--(axis cs:46545200,0.626104711899272)
--(axis cs:43159900,0.61415411272715)
--(axis cs:39105500,0.587694834821119)
--(axis cs:35726000,0.551811292532743)
--(axis cs:31683900,0.543121570844176)
--(axis cs:28323300,0.503500853501829)
--(axis cs:24961300,0.475931311740249)
--(axis cs:20931700,0.464139684125195)
--(axis cs:17586100,0.417842951944482)
--(axis cs:13589000,0.356100725549792)
--(axis cs:10275100,0.300329009145543)
--(axis cs:6333000,0.169536390370922)
--(axis cs:3112700,0.0976048514712602)
--(axis cs:19400,0.0501644741743803)
--cycle;

\addplot [line width=\linewidthothers, \bbrl_color, mark=*, mark size=0, mark options={solid}]
table {%
19400 0.0435855267569423
3112700 0.0918482734414283
6333000 0.162894801636867
10275100 0.284707545847482
13589000 0.348214474323413
17586100 0.398289411575927
20931700 0.442910153935117
24961300 0.445100891674329
28323300 0.474998844382975
31683900 0.515622783586232
35726000 0.53000355739123
39105500 0.565005250578972
43159900 0.583406749302612
46545200 0.609617149082772
50609500 0.623557795575807
53996700 0.618521710370148
57383400 0.611537707060482
61450800 0.638011799849103
64838500 0.633893353721838
68906600 0.66400870444845
72298600 0.637372397743933
76370300 0.665219534551155
79765700 0.658457725436583
83157500 0.662049868584645
87230500 0.668971688391656
90624600 0.655413999714016
94697100 0.672414469758632
98094600 0.660455796961982
102170100 0.692217237608585
105564600 0.685899882324566
108958800 0.682534262170404
113034000 0.681355290346132
116432700 0.674248777007671
120507600 0.697596320820299
123901200 0.691386838682949
127975600 0.704897198613044
131373700 0.687010139052713
134769900 0.70481822248839
138848400 0.709063130813373
142246700 0.702364823874909
146320600 0.707830534017691
149719000 0.705236999996136
153801300 0.70871337070691
157203500 0.702028591262719
160601700 0.710315865073454
164680900 0.704390164595467
168080600 0.721627954907422
172161200 0.717614891219217
175565900 0.714990710790899
179652200 0.729300846729738
};

\path [draw=\replace_color, fill=\replace_color, opacity=0.2]
(axis cs:11400,0.454769738018513)
--(axis cs:11400,0.449013151228428)
--(axis cs:1721400,0.297954357229173)
--(axis cs:3431400,0.324666881220764)
--(axis cs:5483400,0.345370904149641)
--(axis cs:7193400,0.357943139468485)
--(axis cs:9245400,0.397550651690776)
--(axis cs:10955400,0.403961496557158)
--(axis cs:13007400,0.441975294020512)
--(axis cs:14717400,0.451751728141884)
--(axis cs:16427400,0.469471255547694)
--(axis cs:18479400,0.502280281079922)
--(axis cs:20189400,0.513704897129865)
--(axis cs:22241400,0.588306126616664)
--(axis cs:23951400,0.593024205518623)
--(axis cs:26003400,0.606261002472909)
--(axis cs:27713400,0.607476492662874)
--(axis cs:29423400,0.611886889801468)
--(axis cs:31475400,0.662979507977833)
--(axis cs:33185400,0.650575235782038)
--(axis cs:35237400,0.667733046281164)
--(axis cs:36947400,0.667853808093985)
--(axis cs:38999400,0.687104122991439)
--(axis cs:40709400,0.691428974927799)
--(axis cs:42419400,0.690367386195614)
--(axis cs:44471400,0.721918661070189)
--(axis cs:46181400,0.700130935580037)
--(axis cs:48233400,0.696553282589975)
--(axis cs:49943400,0.716921545044363)
--(axis cs:51995400,0.757837450399312)
--(axis cs:53705400,0.720099721716795)
--(axis cs:55415400,0.744726511522594)
--(axis cs:57467400,0.734747337401034)
--(axis cs:59177400,0.725596432814921)
--(axis cs:61229400,0.737931400750006)
--(axis cs:62939400,0.734699187261056)
--(axis cs:64991400,0.764211760386049)
--(axis cs:66701400,0.74305061227034)
--(axis cs:68411400,0.744146621207449)
--(axis cs:70463400,0.756052000101936)
--(axis cs:72173400,0.768783504192239)
--(axis cs:74225400,0.759016424580811)
--(axis cs:75935400,0.752873059391598)
--(axis cs:77987400,0.763584090409424)
--(axis cs:79697400,0.747840111317597)
--(axis cs:81407400,0.750321148229314)
--(axis cs:83459400,0.749055215234213)
--(axis cs:85169400,0.759401761822918)
--(axis cs:87221400,0.766224637753355)
--(axis cs:88931400,0.776275027557451)
--(axis cs:90983400,0.78748611883502)
--(axis cs:90983400,0.804215357672551)
--(axis cs:90983400,0.804215357672551)
--(axis cs:88931400,0.785078771391416)
--(axis cs:87221400,0.784955035263478)
--(axis cs:85169400,0.774125169100638)
--(axis cs:83459400,0.776823188868245)
--(axis cs:81407400,0.784856575051966)
--(axis cs:79697400,0.773848205894953)
--(axis cs:77987400,0.780286208824839)
--(axis cs:75935400,0.77157436700417)
--(axis cs:74225400,0.791139571998845)
--(axis cs:72173400,0.787534171935888)
--(axis cs:70463400,0.783145624527949)
--(axis cs:68411400,0.768732797733262)
--(axis cs:66701400,0.772957026460752)
--(axis cs:64991400,0.779453203986609)
--(axis cs:62939400,0.754433318847505)
--(axis cs:61229400,0.768467727428108)
--(axis cs:59177400,0.757129319803653)
--(axis cs:57467400,0.765756673373179)
--(axis cs:55415400,0.769699635901823)
--(axis cs:53705400,0.759941622690463)
--(axis cs:51995400,0.773581669673108)
--(axis cs:49943400,0.737431251225877)
--(axis cs:48233400,0.727013256679564)
--(axis cs:46181400,0.729100155762164)
--(axis cs:44471400,0.74145536599994)
--(axis cs:42419400,0.718560322468954)
--(axis cs:40709400,0.726812965795231)
--(axis cs:38999400,0.734988558287053)
--(axis cs:36947400,0.713563979270262)
--(axis cs:35237400,0.705696817064944)
--(axis cs:33185400,0.699957865256526)
--(axis cs:31475400,0.705428528380836)
--(axis cs:29423400,0.658480298008539)
--(axis cs:27713400,0.654937573955162)
--(axis cs:26003400,0.657920897158082)
--(axis cs:23951400,0.634942341583181)
--(axis cs:22241400,0.624688742587546)
--(axis cs:20189400,0.563765302796689)
--(axis cs:18479400,0.541005365880121)
--(axis cs:16427400,0.514626490990366)
--(axis cs:14717400,0.480974011568297)
--(axis cs:13007400,0.473614587439618)
--(axis cs:10955400,0.436528165328337)
--(axis cs:9245400,0.418949627208462)
--(axis cs:7193400,0.392804144016836)
--(axis cs:5483400,0.384033842298436)
--(axis cs:3431400,0.352604917017743)
--(axis cs:1721400,0.328795104840537)
--(axis cs:11400,0.454769738018513)
--cycle;

\addplot [line width=\linewidthothers, \replace_color, mark=*, mark size=0, mark options={solid}]
table {%
11400 0.453947365283966
1721400 0.314196137012914
3431400 0.336895594788075
5483400 0.373387500246167
7193400 0.380653400157673
9245400 0.408669040132702
10955400 0.418981591215045
13007400 0.459862344164162
14717400 0.473319299640886
16427400 0.493713364322068
18479400 0.522462076200781
20189400 0.54201485423194
22241400 0.607111641356542
23951400 0.613816928241766
26003400 0.624543974048779
27713400 0.631876563066166
29423400 0.627328898700682
31475400 0.677566724109017
33185400 0.676538232904034
35237400 0.69095823987854
36947400 0.693947634065966
38999400 0.710010283952033
40709400 0.707324324056713
42419400 0.705244720883045
44471400 0.728432405328896
46181400 0.712858705791064
48233400 0.70287418318539
49943400 0.728141477972652
51995400 0.762846604670401
53705400 0.742971040145517
55415400 0.75607968732406
57467400 0.747823616603313
59177400 0.742127977063835
61229400 0.752767850898576
62939400 0.744381315228308
64991400 0.773924894208587
66701400 0.755960862684393
68411400 0.75502794765617
70463400 0.764858328915867
72173400 0.775538009025376
74225400 0.773730229639649
75935400 0.764002259705949
77987400 0.77153746914026
79697400 0.757998627867156
81407400 0.773741805189911
83459400 0.76052069130744
85169400 0.767333034481544
87221400 0.770560858458247
88931400 0.780764736985142
90983400 0.798510216717685
};

\path [draw=\res_color, fill=\res_color, opacity=0.2]
(axis cs:11400,0.453947372734547)
--(axis cs:11400,0.445723682641983)
--(axis cs:1721400,0.453253493644297)
--(axis cs:3431400,0.489068282069638)
--(axis cs:5483400,0.572920882735843)
--(axis cs:7193400,0.641743428450408)
--(axis cs:9245400,0.665496288476507)
--(axis cs:10955400,0.702217181034394)
--(axis cs:13007400,0.721697540774946)
--(axis cs:14717400,0.738654355979361)
--(axis cs:16427400,0.751768779691115)
--(axis cs:18479400,0.771587367326683)
--(axis cs:20189400,0.777165369305129)
--(axis cs:22241400,0.804902159490885)
--(axis cs:23951400,0.79045807988587)
--(axis cs:26003400,0.809070018799577)
--(axis cs:27713400,0.781709047688992)
--(axis cs:29423400,0.799342699127919)
--(axis cs:31475400,0.793876122624942)
--(axis cs:33185400,0.792348832183595)
--(axis cs:35237400,0.806295023556063)
--(axis cs:36947400,0.821207724052873)
--(axis cs:38999400,0.813464323190555)
--(axis cs:40709400,0.800998743906677)
--(axis cs:42419400,0.814164784262706)
--(axis cs:44471400,0.804168082943992)
--(axis cs:46181400,0.805592277922184)
--(axis cs:48233400,0.820630690978034)
--(axis cs:49943400,0.800737962545854)
--(axis cs:51995400,0.833882818986981)
--(axis cs:53705400,0.810294626168454)
--(axis cs:55415400,0.830530719627225)
--(axis cs:57467400,0.808224041673926)
--(axis cs:59177400,0.801116059848828)
--(axis cs:61229400,0.810478040793438)
--(axis cs:62939400,0.811300067295365)
--(axis cs:64991400,0.827403987209807)
--(axis cs:66701400,0.823830590733558)
--(axis cs:68411400,0.828471028220063)
--(axis cs:70463400,0.83492412348361)
--(axis cs:72173400,0.831388785164394)
--(axis cs:74225400,0.817781855757593)
--(axis cs:75935400,0.816768986942161)
--(axis cs:77987400,0.82654713296494)
--(axis cs:79697400,0.826927570007668)
--(axis cs:81407400,0.821276875469954)
--(axis cs:83459400,0.825918036370127)
--(axis cs:85169400,0.81963596766109)
--(axis cs:87221400,0.827933046362246)
--(axis cs:88931400,0.821007161506761)
--(axis cs:90983400,0.840285693456291)
--(axis cs:90983400,0.857474202699879)
--(axis cs:90983400,0.857474202699879)
--(axis cs:88931400,0.84150007994903)
--(axis cs:87221400,0.848249619747544)
--(axis cs:85169400,0.828437954914417)
--(axis cs:83459400,0.847764154094701)
--(axis cs:81407400,0.836130304394984)
--(axis cs:79697400,0.838920695659872)
--(axis cs:77987400,0.842008264764144)
--(axis cs:75935400,0.829353356715215)
--(axis cs:74225400,0.835811103783562)
--(axis cs:72173400,0.85372783256049)
--(axis cs:70463400,0.854264429767784)
--(axis cs:68411400,0.840155978325618)
--(axis cs:66701400,0.838888889625447)
--(axis cs:64991400,0.840875586848008)
--(axis cs:62939400,0.824423302062409)
--(axis cs:61229400,0.835116292906395)
--(axis cs:59177400,0.812916427105389)
--(axis cs:57467400,0.823079272404505)
--(axis cs:55415400,0.852983996900813)
--(axis cs:53705400,0.831605792452367)
--(axis cs:51995400,0.847080699710011)
--(axis cs:49943400,0.825774901106606)
--(axis cs:48233400,0.835541061475517)
--(axis cs:46181400,0.829817443501416)
--(axis cs:44471400,0.816306375528688)
--(axis cs:42419400,0.822283244266353)
--(axis cs:40709400,0.820417117629037)
--(axis cs:38999400,0.830314677446556)
--(axis cs:36947400,0.838994844994795)
--(axis cs:35237400,0.821378298786921)
--(axis cs:33185400,0.818697742883311)
--(axis cs:31475400,0.814252574426596)
--(axis cs:29423400,0.819368538381808)
--(axis cs:27713400,0.80221279646981)
--(axis cs:26003400,0.817302173316757)
--(axis cs:23951400,0.808956728789058)
--(axis cs:22241400,0.833706362357808)
--(axis cs:20189400,0.791591872787656)
--(axis cs:18479400,0.790066315066979)
--(axis cs:16427400,0.772241114071395)
--(axis cs:14717400,0.754150957505168)
--(axis cs:13007400,0.745163634599906)
--(axis cs:10955400,0.743868616203775)
--(axis cs:9245400,0.705180769108122)
--(axis cs:7193400,0.675809371076536)
--(axis cs:5483400,0.618476276828758)
--(axis cs:3431400,0.511107598700619)
--(axis cs:1721400,0.471988072618842)
--(axis cs:11400,0.453947372734547)
--cycle;

\addplot [line width=\linewidthtop, \res_color, mark=*, mark size=0, mark options={solid}]
table {%
11400 0.449835531413555
1721400 0.462710730032995
3431400 0.49894152165507
5483400 0.597238498170896
7193400 0.66281920801865
9245400 0.694680099415942
10955400 0.731381060243348
13007400 0.734168224159609
14717400 0.743924673198564
16427400 0.765615577976429
18479400 0.780955893891305
20189400 0.78419918067142
22241400 0.818413477254798
23951400 0.796689964606504
26003400 0.812933333264633
27713400 0.791776537008614
29423400 0.808032830995398
31475400 0.799463459409277
33185400 0.803991943766999
35237400 0.81384151949277
36947400 0.830702975330549
38999400 0.819657073583967
40709400 0.812413571710437
42419400 0.81873041096298
44471400 0.809189562116689
46181400 0.821615929809452
48233400 0.830563806664675
49943400 0.810080749096918
51995400 0.841026463648472
53705400 0.820571242995649
55415400 0.83858906567285
57467400 0.817106374152159
59177400 0.806699986828051
61229400 0.821898012148426
62939400 0.81951974567623
64991400 0.833417621719016
66701400 0.83014125672809
68411400 0.83294901481844
70463400 0.84737223671577
72173400 0.844077558809055
74225400 0.827255782179777
75935400 0.823493062246533
77987400 0.83313580370568
79697400 0.832661064231838
81407400 0.830525881345198
83459400 0.83785437406385
85169400 0.824310153243899
87221400 0.839010748924667
88931400 0.833702491943434
90983400 0.847643000956835
};

\path [draw=\whole_color, fill=\whole_color, opacity=0.2]
(axis cs:11400,0.459703952074051)
--(axis cs:11400,0.445723682641983)
--(axis cs:1721400,0.562782686436549)
--(axis cs:3431400,0.649149052565917)
--(axis cs:5483400,0.729344927497777)
--(axis cs:7193400,0.747622954348419)
--(axis cs:9245400,0.755481197921708)
--(axis cs:10955400,0.772033459298051)
--(axis cs:13007400,0.76646875615239)
--(axis cs:14717400,0.761473645063483)
--(axis cs:16427400,0.778760487308262)
--(axis cs:18479400,0.792537551435295)
--(axis cs:20189400,0.793089947492147)
--(axis cs:22241400,0.811173887431923)
--(axis cs:23951400,0.797445137884438)
--(axis cs:26003400,0.812765086737382)
--(axis cs:27713400,0.802484325711104)
--(axis cs:29423400,0.794344636409189)
--(axis cs:31475400,0.803066987751723)
--(axis cs:33185400,0.801784008357247)
--(axis cs:35237400,0.811531879275534)
--(axis cs:36947400,0.817624720174042)
--(axis cs:38999400,0.81905858063183)
--(axis cs:40709400,0.808100527271269)
--(axis cs:42419400,0.809617957615333)
--(axis cs:44471400,0.808918953201802)
--(axis cs:46181400,0.80861400428626)
--(axis cs:48233400,0.817233995026695)
--(axis cs:49943400,0.793377303635925)
--(axis cs:51995400,0.828007241039519)
--(axis cs:53705400,0.801318575696448)
--(axis cs:55415400,0.829090435370868)
--(axis cs:57467400,0.805485960821477)
--(axis cs:59177400,0.795393227987566)
--(axis cs:61229400,0.81461676211379)
--(axis cs:62939400,0.808429509067457)
--(axis cs:64991400,0.829062814120868)
--(axis cs:66701400,0.819342733359753)
--(axis cs:68411400,0.824820266284906)
--(axis cs:70463400,0.831792855672607)
--(axis cs:72173400,0.828916316891158)
--(axis cs:74225400,0.818924959989288)
--(axis cs:75935400,0.814187990157567)
--(axis cs:77987400,0.813052249866825)
--(axis cs:79697400,0.830783187033379)
--(axis cs:81407400,0.82315020689366)
--(axis cs:83459400,0.812746152634228)
--(axis cs:85169400,0.808933117439324)
--(axis cs:87221400,0.814046342751093)
--(axis cs:88931400,0.814136702537599)
--(axis cs:90983400,0.832837822291023)
--(axis cs:90983400,0.855969015033816)
--(axis cs:90983400,0.855969015033816)
--(axis cs:88931400,0.839470336334541)
--(axis cs:87221400,0.834194690295707)
--(axis cs:85169400,0.826264402680674)
--(axis cs:83459400,0.839441130548587)
--(axis cs:81407400,0.836440110652608)
--(axis cs:79697400,0.839213372343349)
--(axis cs:77987400,0.832704190206919)
--(axis cs:75935400,0.825829427708528)
--(axis cs:74225400,0.83566308366079)
--(axis cs:72173400,0.842972571380493)
--(axis cs:70463400,0.845194152626418)
--(axis cs:68411400,0.845978645197413)
--(axis cs:66701400,0.828904647138932)
--(axis cs:64991400,0.856843998488249)
--(axis cs:62939400,0.825845073214902)
--(axis cs:61229400,0.826714414673432)
--(axis cs:59177400,0.821827361897017)
--(axis cs:57467400,0.825055279678115)
--(axis cs:55415400,0.846593063748015)
--(axis cs:53705400,0.820851616090645)
--(axis cs:51995400,0.840159759678637)
--(axis cs:49943400,0.811883150788275)
--(axis cs:48233400,0.835236179049244)
--(axis cs:46181400,0.827140162842513)
--(axis cs:44471400,0.821873649561652)
--(axis cs:42419400,0.823907690750914)
--(axis cs:40709400,0.81894946609492)
--(axis cs:38999400,0.836837529141066)
--(axis cs:36947400,0.837354634775578)
--(axis cs:35237400,0.832376440171283)
--(axis cs:33185400,0.821214584137214)
--(axis cs:31475400,0.81872365033134)
--(axis cs:29423400,0.81653116981982)
--(axis cs:27713400,0.815232998288821)
--(axis cs:26003400,0.823934769850254)
--(axis cs:23951400,0.819116845024299)
--(axis cs:22241400,0.835179901988963)
--(axis cs:20189400,0.813389450237306)
--(axis cs:18479400,0.803538665703657)
--(axis cs:16427400,0.800556368238969)
--(axis cs:14717400,0.786596878899016)
--(axis cs:13007400,0.776686263074973)
--(axis cs:10955400,0.790241434214528)
--(axis cs:9245400,0.781275394783189)
--(axis cs:7193400,0.756112853054859)
--(axis cs:5483400,0.749527829048589)
--(axis cs:3431400,0.682157020761224)
--(axis cs:1721400,0.588713000062853)
--(axis cs:11400,0.459703952074051)
--cycle;

\addplot [line width=\linewidthothers, \whole_color, mark=*, mark size=0, mark options={solid}]
table {%
11400 0.453125007450581
1721400 0.574732737615705
3431400 0.66477485728683
5483400 0.739277195566842
7193400 0.752523065540942
9245400 0.768311627642676
10955400 0.78270589311642
13007400 0.773060018564262
14717400 0.772574207345142
16427400 0.790096864525418
18479400 0.798327475499112
20189400 0.805523363952291
22241400 0.823873372179643
23951400 0.805655909472879
26003400 0.815890366683939
27713400 0.806800905903335
29423400 0.807156647156024
31475400 0.811320794567367
33185400 0.809330929956174
35237400 0.81957689223186
36947400 0.827338140105095
38999400 0.830263435811356
40709400 0.813690050966463
42419400 0.818728791645141
44471400 0.814237917248349
46181400 0.816893086888727
48233400 0.825691285810769
49943400 0.803929637991821
51995400 0.835343138874094
53705400 0.812682945907704
55415400 0.836229205170456
57467400 0.818579480216878
59177400 0.806211903185452
61229400 0.819548180341554
62939400 0.81945298869374
64991400 0.84299432112889
66701400 0.825661501679707
68411400 0.83499246252745
70463400 0.836810952061591
72173400 0.83724250117449
74225400 0.827278923134339
75935400 0.818424641723584
77987400 0.823258436232049
79697400 0.832935161453126
81407400 0.832118491253456
83459400 0.8262072202362
85169400 0.816232408879905
87221400 0.826075701888729
88931400 0.825035435446652
90983400 0.847610621600657
};

\path [draw=C2, fill=C2, opacity=0.2]
(axis cs:121600,0)
--(axis cs:121600,0)
--(axis cs:3283200,0)
--(axis cs:6444800,0)
--(axis cs:9606400,0)
--(axis cs:12889600,0)
--(axis cs:16051200,0)
--(axis cs:19212800,0)
--(axis cs:22374400,0)
--(axis cs:25657600,0)
--(axis cs:28819200,0)
--(axis cs:31980800,0)
--(axis cs:35142400,0)
--(axis cs:38425600,0)
--(axis cs:41587200,0)
--(axis cs:44748800,0.0131578947368421)
--(axis cs:47910400,0.0131578947368421)
--(axis cs:51193600,0.0197368421052632)
--(axis cs:54355200,0.00657894736842105)
--(axis cs:57516800,0.0131578947368421)
--(axis cs:60800000,0.0263157894736842)
--(axis cs:60800000,0.164473684210526)
--(axis cs:60800000,0.164473684210526)
--(axis cs:57516800,0.171052631578947)
--(axis cs:54355200,0.151398026315788)
--(axis cs:51193600,0.148026315789474)
--(axis cs:47910400,0.144736842105263)
--(axis cs:44748800,0.121710526315789)
--(axis cs:41587200,0.101973684210526)
--(axis cs:38425600,0.108552631578947)
--(axis cs:35142400,0.0657894736842105)
--(axis cs:31980800,0.0330592105263134)
--(axis cs:28819200,0.00986842105263158)
--(axis cs:25657600,0.00986842105263158)
--(axis cs:22374400,0.00328947368421053)
--(axis cs:19212800,0.00328947368421053)
--(axis cs:16051200,0)
--(axis cs:12889600,0)
--(axis cs:9606400,0)
--(axis cs:6444800,0)
--(axis cs:3283200,0)
--(axis cs:121600,0)
--cycle;

\addplot [line width=\linewidthothers, C2, mark=*, mark size=0, mark options={solid}]
table {%
121600 0
3283200 0
6444800 0
9606400 0
12889600 0
16051200 0
19212800 0
22374400 0
25657600 0
28819200 0
31980800 0
35142400 0
38425600 0.0197368421052632
41587200 0.0131578947368421
44748800 0.0493421052631579
47910400 0.0328947368421053
51193600 0.0493421052631579
54355200 0.0657894736842105
57516800 0.0592105263157895
60800000 0.0789473684210526
};

\path [draw=C5, fill=C5, opacity=0.2]
(axis cs:121600,0.25)
--(axis cs:121600,0.138157894736842)
--(axis cs:3283200,0)
--(axis cs:6444800,0)
--(axis cs:9606400,0)
--(axis cs:12889600,0)
--(axis cs:16051200,0)
--(axis cs:19212800,0)
--(axis cs:22374400,0)
--(axis cs:25657600,0)
--(axis cs:28819200,0.00657894736842105)
--(axis cs:31980800,0.0789473684210526)
--(axis cs:35142400,0.105263157894737)
--(axis cs:38425600,0.0921052631578947)
--(axis cs:41587200,0.0592105263157895)
--(axis cs:44748800,0.138157894736842)
--(axis cs:47910400,0.118421052631579)
--(axis cs:51193600,0.171052631578947)
--(axis cs:54355200,0.131578947368421)
--(axis cs:57516800,0.157894736842105)
--(axis cs:60800000,0.269736842105263)
--(axis cs:60800000,0.421052631578947)
--(axis cs:60800000,0.421052631578947)
--(axis cs:57516800,0.302631578947368)
--(axis cs:54355200,0.342105263157895)
--(axis cs:51193600,0.25)
--(axis cs:47910400,0.243421052631579)
--(axis cs:44748800,0.276315789473684)
--(axis cs:41587200,0.210526315789474)
--(axis cs:38425600,0.256578947368421)
--(axis cs:35142400,0.223684210526316)
--(axis cs:31980800,0.217105263157895)
--(axis cs:28819200,0.125)
--(axis cs:25657600,0.118421052631579)
--(axis cs:22374400,0.0855263157894737)
--(axis cs:19212800,0.00657894736842105)
--(axis cs:16051200,0.0657894736842105)
--(axis cs:12889600,0.0394736842105263)
--(axis cs:9606400,0.00657894736842105)
--(axis cs:6444800,0)
--(axis cs:3283200,0)
--(axis cs:121600,0.25)
--cycle;

\addplot [line width=\linewidthothers, C5, mark=*, mark size=0, mark options={solid}]
table {%
121600 0.190789473684211
3283200 0
6444800 0
9606400 0
12889600 0
16051200 0.00657894736842105
19212800 0
22374400 0.0263157894736842
25657600 0.0460526315789474
28819200 0.0394736842105263
31980800 0.118421052631579
35142400 0.157894736842105
38425600 0.164473684210526
41587200 0.118421052631579
44748800 0.190789473684211
47910400 0.190789473684211
51193600 0.197368421052632
54355200 0.236842105263158
57516800 0.25
60800000 0.368421052631579
};

\path [draw=C4, fill=C4, opacity=0.2]
(axis cs:10400,0)
--(axis cs:10400,0)
--(axis cs:443200,0)
--(axis cs:858000,0)
--(axis cs:1267200,0)
--(axis cs:1696800,0)
--(axis cs:2136800,0)
--(axis cs:2572000,0)
--(axis cs:3002400,0)
--(axis cs:3442800,0)
--(axis cs:3881200,0)
--(axis cs:4316400,0)
--(axis cs:4736000,0)
--(axis cs:5170800,0)
--(axis cs:5590000,0)
--(axis cs:6018400,0)
--(axis cs:6452000,0)
--(axis cs:6879600,0)
--(axis cs:7319600,0)
--(axis cs:7767600,0)
--(axis cs:8193600,0)
--(axis cs:8193600,0)
--(axis cs:8193600,0)
--(axis cs:7767600,0)
--(axis cs:7319600,0)
--(axis cs:6879600,0)
--(axis cs:6452000,0)
--(axis cs:6018400,0)
--(axis cs:5590000,0)
--(axis cs:5170800,0)
--(axis cs:4736000,0)
--(axis cs:4316400,0)
--(axis cs:3881200,0)
--(axis cs:3442800,0)
--(axis cs:3002400,0)
--(axis cs:2572000,0)
--(axis cs:2136800,0)
--(axis cs:1696800,0)
--(axis cs:1267200,0)
--(axis cs:858000,0)
--(axis cs:443200,0)
--(axis cs:10400,0)
--cycle;

\addplot [line width=\linewidthothers, C4, mark=*, mark size=0, mark options={solid}]
table {%
10400 0
443200 0
858000 0
1267200 0
1696800 0
2136800 0
2572000 0
3002400 0
3442800 0
3881200 0
4316400 0
4736000 0
5170800 0
5590000 0
6018400 0
6452000 0
6879600 0
7319600 0
7767600 0
8193600 0
};

\path [draw=C6, fill=C6, opacity=0.2]
(axis cs:10400,0)
--(axis cs:10400,0)
--(axis cs:446800,0)
--(axis cs:880400,0)
--(axis cs:1283200,0)
--(axis cs:1724800,0)
--(axis cs:2165600,0)
--(axis cs:2587200,0)
--(axis cs:2999600,0)
--(axis cs:3418400,0)
--(axis cs:3839600,0)
--(axis cs:4258400,0)
--(axis cs:4696800,0)
--(axis cs:5106800,0)
--(axis cs:5524000,0)
--(axis cs:5952800,0)
--(axis cs:6373600,0)
--(axis cs:6799600,0)
--(axis cs:7217600,0)
--(axis cs:7649600,0)
--(axis cs:8070000,0)
--(axis cs:8070000,0)
--(axis cs:8070000,0)
--(axis cs:7649600,0)
--(axis cs:7217600,0)
--(axis cs:6799600,0)
--(axis cs:6373600,0)
--(axis cs:5952800,0)
--(axis cs:5524000,0)
--(axis cs:5106800,0)
--(axis cs:4696800,0)
--(axis cs:4258400,0)
--(axis cs:3839600,0)
--(axis cs:3418400,0)
--(axis cs:2999600,0)
--(axis cs:2587200,0)
--(axis cs:2165600,0)
--(axis cs:1724800,0)
--(axis cs:1283200,0)
--(axis cs:880400,0)
--(axis cs:446800,0)
--(axis cs:10400,0)
--cycle;

\addplot [line width=\linewidthothers, C6, mark=*, mark size=0, mark options={solid}]
table {%
10400 0
446800 0
880400 0
1283200 0
1724800 0
2165600 0
2587200 0
2999600 0
3418400 0
3839600 0
4258400 0
4696800 0
5106800 0
5524000 0
5952800 0
6373600 0
6799600 0
7217600 0
7649600 0
8070000 0
};

\end{axis}

\end{tikzpicture}

%% file: images/tex/multi_box_m_success.tex
%
%

\begin{tikzpicture}
\def\res_color{C0}
\def\replace_color{C1}
\def\whole_color{C3}
\def\bbrl_color{C9}
\def\ppo_color{C2}
\def\sac_color{C4}
\def\linewidthtop{1mm}
\def\linewidthothers{0.5mm}

\begin{axis}[
legend cell align={left},
legend style={fill opacity=0.8, draw opacity=1, text opacity=1, draw=lightgray204, at={(0.03,0.03)},  anchor=north west},
tick align=outside,
tick pos=left,
x grid style={white},
scaled x ticks=false,
xticklabels={,0,0.5,1,1.5,2, 2.5},
xlabel={Environment Interactions ($\times 10^7$)},
xmajorgrids,
xmin=-1500000, xmax=\multiboxxmax,
xtick style={color=black},
y grid style={white},
ylabel={Success Rate, Markovian, IQM},
ymajorgrids,
ymin=-0.05, ymax=0.9,
ytick style={color=black},
axis background/.style={fill=plot_background},
label style={font=\large},
tick label style={font=\large},
x axis line style={draw=none},
y axis line style={draw=none},
]
\path [draw=\bbrl_color, fill=\bbrl_color, opacity=0.2]
(axis cs:19700,0.0485197370871902)
--(axis cs:19700,0.0419407896697521)
--(axis cs:2906000,0.0554584717901889)
--(axis cs:5598400,0.046138562991473)
--(axis cs:8701100,0.0523703978627879)
--(axis cs:11233200,0.0487698399622176)
--(axis cs:14188700,0.053741369956741)
--(axis cs:16611600,0.0466215963300421)
--(axis cs:19497800,0.0503807978960129)
--(axis cs:21878400,0.0593795628307806)
--(axis cs:24230600,0.0599321630125603)
--(axis cs:27033000,0.0695975762380512)
--(axis cs:29369400,0.0698584104203909)
--(axis cs:32173600,0.0757001428835565)
--(axis cs:34513800,0.0779899675544233)
--(axis cs:37323700,0.085500252153595)
--(axis cs:39651700,0.0894851019138585)
--(axis cs:41975000,0.0912818952944719)
--(axis cs:44768000,0.0930861621890451)
--(axis cs:47086700,0.0983871306510337)
--(axis cs:49875500,0.102259909410326)
--(axis cs:52181500,0.0919803183891822)
--(axis cs:54947700,0.0907072178397679)
--(axis cs:57234000,0.0971976086005155)
--(axis cs:59521900,0.0793255828062962)
--(axis cs:62230100,0.0896522553066706)
--(axis cs:64498900,0.0960294051406135)
--(axis cs:67195300,0.0898662936160408)
--(axis cs:69441300,0.0954421070330694)
--(axis cs:72163200,0.0930567971621818)
--(axis cs:74421800,0.0995501380167528)
--(axis cs:76676300,0.0906705066740275)
--(axis cs:79386700,0.0979333260936537)
--(axis cs:81641700,0.105356074693226)
--(axis cs:84360500,0.0972645549539564)
--(axis cs:86617200,0.102419311780513)
--(axis cs:89310400,0.104304071338168)
--(axis cs:91567100,0.105284624493651)
--(axis cs:93796600,0.0949190288071552)
--(axis cs:96465800,0.0989313381207249)
--(axis cs:98693400,0.101933346926935)
--(axis cs:101372900,0.0960831531168562)
--(axis cs:103615300,0.0998803792824929)
--(axis cs:106317200,0.108471391647681)
--(axis cs:108549100,0.106949204847607)
--(axis cs:110771700,0.0931223438882801)
--(axis cs:113442100,0.116030704582305)
--(axis cs:115683200,0.103666780443954)
--(axis cs:118368300,0.102925978027272)
--(axis cs:120617400,0.126084179909804)
--(axis cs:123323400,0.121944847600403)
--(axis cs:123323400,0.145752037421277)
--(axis cs:123323400,0.145752037421277)
--(axis cs:120617400,0.158385131763798)
--(axis cs:118368300,0.139258976722746)
--(axis cs:115683200,0.135556934410541)
--(axis cs:113442100,0.140563958745014)
--(axis cs:110771700,0.121939425657059)
--(axis cs:108549100,0.131565742881413)
--(axis cs:106317200,0.126222205904225)
--(axis cs:103615300,0.121586197495445)
--(axis cs:101372900,0.120027729328795)
--(axis cs:98693400,0.132555434954837)
--(axis cs:96465800,0.12877220107772)
--(axis cs:93796600,0.120439423404524)
--(axis cs:91567100,0.12314842658479)
--(axis cs:89310400,0.128924930213114)
--(axis cs:86617200,0.130051968660822)
--(axis cs:84360500,0.123588905358867)
--(axis cs:81641700,0.129541515728762)
--(axis cs:79386700,0.121827458021168)
--(axis cs:76676300,0.119359830253707)
--(axis cs:74421800,0.118394274374711)
--(axis cs:72163200,0.114764697632413)
--(axis cs:69441300,0.117627608077173)
--(axis cs:67195300,0.11212125136757)
--(axis cs:64498900,0.116730175804728)
--(axis cs:62230100,0.106689708179093)
--(axis cs:59521900,0.10806435838498)
--(axis cs:57234000,0.118403722470386)
--(axis cs:54947700,0.10355961938651)
--(axis cs:52181500,0.117229516035267)
--(axis cs:49875500,0.125229174467145)
--(axis cs:47086700,0.118847924962593)
--(axis cs:44768000,0.12366248840401)
--(axis cs:41975000,0.116585534707874)
--(axis cs:39651700,0.104977571775892)
--(axis cs:37323700,0.11829244127556)
--(axis cs:34513800,0.109367539436085)
--(axis cs:32173600,0.101154518618073)
--(axis cs:29369400,0.09602322600807)
--(axis cs:27033000,0.105703092750803)
--(axis cs:24230600,0.094159634397543)
--(axis cs:21878400,0.09807298411304)
--(axis cs:19497800,0.0848195057845864)
--(axis cs:16611600,0.0731852282983289)
--(axis cs:14188700,0.0747571802759778)
--(axis cs:11233200,0.0627219509989048)
--(axis cs:8701100,0.0672299983105233)
--(axis cs:5598400,0.0571489837720947)
--(axis cs:2906000,0.0609323613171)
--(axis cs:19700,0.0485197370871902)
--cycle;

\addplot [line width=\linewidthothers, \bbrl_color, mark=*, mark size=0, mark options={solid}]
table {%
19700 0.0444078948348761
2906000 0.0581568674824666
5598400 0.0527118878972033
8701100 0.0620226109933526
11233200 0.0530877884633005
14188700 0.0615538756646679
16611600 0.0627058183358573
19497800 0.0696218423800799
21878400 0.0811079818488838
24230600 0.0746460546323616
27033000 0.0870138977604126
29369400 0.0787980566325592
32173600 0.0895463523500311
34513800 0.0904777435799428
37323700 0.105048504439289
39651700 0.0962173844083172
41975000 0.1015419810509
44768000 0.108899818549341
47086700 0.108586432529472
49875500 0.107807889873596
52181500 0.10190514712624
54947700 0.0987869181377871
57234000 0.10570024624058
59521900 0.0948263143598759
62230100 0.09648113607145
64498900 0.10556583141552
67195300 0.1039197676921
69441300 0.113083018534624
72163200 0.103099021504357
74421800 0.109659473861036
76676300 0.107036178247425
79386700 0.109828888209214
81641700 0.11815327864308
84360500 0.109800421064908
86617200 0.116574646618212
89310400 0.11728219940187
91567100 0.113858874933845
93796600 0.106863132255223
96465800 0.118155983397598
98693400 0.114054877128545
101372900 0.107122238648646
103615300 0.104841706886173
106317200 0.115386054390201
108549100 0.118027547411564
110771700 0.106478246965181
113442100 0.132645962404569
115683200 0.115051112089728
118368300 0.126379691076417
120617400 0.145643287963406
123323400 0.130570130925668
};

\path [draw=\replace_color, fill=\replace_color, opacity=0.2]
(axis cs:11400,0.456414483487606)
--(axis cs:11400,0.449013151228428)
--(axis cs:1721400,0.359477788675576)
--(axis cs:3431400,0.385488407751836)
--(axis cs:5483400,0.407121671458071)
--(axis cs:7193400,0.405369251115037)
--(axis cs:9245400,0.398498612669115)
--(axis cs:10955400,0.387040677428678)
--(axis cs:13007400,0.388862458419736)
--(axis cs:14717400,0.369458125043648)
--(axis cs:16427400,0.358189670012507)
--(axis cs:18479400,0.380820307781448)
--(axis cs:20189400,0.398908995236861)
--(axis cs:22241400,0.40716017879696)
--(axis cs:23951400,0.418184318962399)
--(axis cs:26003400,0.43201881009892)
--(axis cs:27713400,0.425699443251094)
--(axis cs:29423400,0.420904457011016)
--(axis cs:31475400,0.423372085104248)
--(axis cs:33185400,0.408277738926427)
--(axis cs:35237400,0.454469770321486)
--(axis cs:36947400,0.460279640268529)
--(axis cs:38999400,0.474721165672119)
--(axis cs:40709400,0.47775135754841)
--(axis cs:42419400,0.469853657940185)
--(axis cs:44471400,0.472977417232412)
--(axis cs:46181400,0.471925829483631)
--(axis cs:48233400,0.498755939891484)
--(axis cs:49943400,0.494300506131727)
--(axis cs:51995400,0.511607641954013)
--(axis cs:53705400,0.516204992010701)
--(axis cs:55415400,0.530898812628657)
--(axis cs:57467400,0.532222600062085)
--(axis cs:59177400,0.551329627617647)
--(axis cs:61229400,0.527352717201303)
--(axis cs:62939400,0.509828490368099)
--(axis cs:64991400,0.53948153984127)
--(axis cs:66701400,0.544603528650938)
--(axis cs:68411400,0.554564674203693)
--(axis cs:70463400,0.553038164285391)
--(axis cs:72173400,0.540711487882172)
--(axis cs:74225400,0.542549534042151)
--(axis cs:75935400,0.542445443556937)
--(axis cs:77987400,0.535674896086234)
--(axis cs:79697400,0.568409740418592)
--(axis cs:81407400,0.563182752825666)
--(axis cs:83459400,0.553605162086293)
--(axis cs:85169400,0.551859026736378)
--(axis cs:87221400,0.563128629024558)
--(axis cs:88931400,0.57811260253537)
--(axis cs:90983400,0.555392300703967)
--(axis cs:90983400,0.60120128903336)
--(axis cs:90983400,0.60120128903336)
--(axis cs:88931400,0.596441275424378)
--(axis cs:87221400,0.600566060178865)
--(axis cs:85169400,0.589744809752651)
--(axis cs:83459400,0.589901387795102)
--(axis cs:81407400,0.588122673601294)
--(axis cs:79697400,0.59092148598931)
--(axis cs:77987400,0.588565569193341)
--(axis cs:75935400,0.572653876799748)
--(axis cs:74225400,0.581339174569441)
--(axis cs:72173400,0.571212957581242)
--(axis cs:70463400,0.575490292474951)
--(axis cs:68411400,0.589590260773186)
--(axis cs:66701400,0.560868184034333)
--(axis cs:64991400,0.574871671292439)
--(axis cs:62939400,0.538860966191961)
--(axis cs:61229400,0.556384976796838)
--(axis cs:59177400,0.566582157250948)
--(axis cs:57467400,0.564121823884075)
--(axis cs:55415400,0.569597302107986)
--(axis cs:53705400,0.542729609475884)
--(axis cs:51995400,0.543180722602861)
--(axis cs:49943400,0.528802933378601)
--(axis cs:48233400,0.519438208507308)
--(axis cs:46181400,0.499067153167849)
--(axis cs:44471400,0.507251649488458)
--(axis cs:42419400,0.51044242426596)
--(axis cs:40709400,0.504523376686954)
--(axis cs:38999400,0.506370040739528)
--(axis cs:36947400,0.48838253907353)
--(axis cs:35237400,0.488000158996803)
--(axis cs:33185400,0.448812319268624)
--(axis cs:31475400,0.470592896567394)
--(axis cs:29423400,0.44555154819301)
--(axis cs:27713400,0.451728576206772)
--(axis cs:26003400,0.476971425908873)
--(axis cs:23951400,0.453774632455393)
--(axis cs:22241400,0.449331312822237)
--(axis cs:20189400,0.439479645127861)
--(axis cs:18479400,0.421314112665057)
--(axis cs:16427400,0.38301425495574)
--(axis cs:14717400,0.387142632096582)
--(axis cs:13007400,0.406933911169056)
--(axis cs:10955400,0.397519810502411)
--(axis cs:9245400,0.41873171689469)
--(axis cs:7193400,0.421051284459686)
--(axis cs:5483400,0.417679072747205)
--(axis cs:3431400,0.406066888470377)
--(axis cs:1721400,0.390098815027158)
--(axis cs:11400,0.456414483487606)
--cycle;

\addplot [line width=\linewidthothers, \replace_color, mark=*, mark size=0, mark options={solid}]
table {%
11400 0.453947365283966
1721400 0.374614517902955
3431400 0.391901915478229
5483400 0.41260423266715
7193400 0.41295152905743
9245400 0.40695577354517
10955400 0.392901859224571
13007400 0.397714767674361
14717400 0.378206911819873
16427400 0.370425675543133
18479400 0.400843275810019
20189400 0.40672579563773
22241400 0.426956816681117
23951400 0.441096240636826
26003400 0.448462630993444
27713400 0.438586174588194
29423400 0.433216505191624
31475400 0.446127101812727
33185400 0.426955055570133
35237400 0.466863400828645
36947400 0.469368779223685
38999400 0.488213747042748
40709400 0.485919950382874
42419400 0.486511275095601
44471400 0.491432502382046
46181400 0.481384832379
48233400 0.510905050543631
49943400 0.510124781857282
51995400 0.523127553216437
53705400 0.529334003670921
55415400 0.550907244586411
57467400 0.543670077060882
59177400 0.558329500667995
61229400 0.540047141122586
62939400 0.527692374839874
64991400 0.556938264777153
66701400 0.550025939871057
68411400 0.569151966552687
70463400 0.562868935227422
72173400 0.553444745098522
74225400 0.564339319165889
75935400 0.55880173802732
77987400 0.561495131917081
79697400 0.579252260728681
81407400 0.575025145062283
83459400 0.565218845797106
85169400 0.566760026106564
87221400 0.581155497349462
88931400 0.587895311854932
90983400 0.581907695145792
};

\path [draw=\res_color, fill=\res_color, opacity=0.2]
(axis cs:11400,0.458881586790085)
--(axis cs:11400,0.447368413209915)
--(axis cs:1721400,0.430315195064759)
--(axis cs:3431400,0.47113920164702)
--(axis cs:5483400,0.543034652644792)
--(axis cs:7193400,0.580933961813933)
--(axis cs:9245400,0.592260177772764)
--(axis cs:10955400,0.606459853810992)
--(axis cs:13007400,0.619908472834007)
--(axis cs:14717400,0.6124903620361)
--(axis cs:16427400,0.625531756253338)
--(axis cs:18479400,0.667835867883805)
--(axis cs:20189400,0.662445489007477)
--(axis cs:22241400,0.705802971362209)
--(axis cs:23951400,0.692894624007219)
--(axis cs:26003400,0.699832976648056)
--(axis cs:27713400,0.695716666371794)
--(axis cs:29423400,0.686040760852747)
--(axis cs:31475400,0.688458565108626)
--(axis cs:33185400,0.702530489598829)
--(axis cs:35237400,0.707685954640795)
--(axis cs:36947400,0.728300098861207)
--(axis cs:38999400,0.718390260704129)
--(axis cs:40709400,0.700144999540563)
--(axis cs:42419400,0.710713346528247)
--(axis cs:44471400,0.726737845032869)
--(axis cs:46181400,0.69896196514849)
--(axis cs:48233400,0.72201461980615)
--(axis cs:49943400,0.692561115390882)
--(axis cs:51995400,0.710875969045345)
--(axis cs:53705400,0.714462878378373)
--(axis cs:55415400,0.72169713984146)
--(axis cs:57467400,0.710781990628842)
--(axis cs:59177400,0.703116994781711)
--(axis cs:61229400,0.727305532926529)
--(axis cs:62939400,0.706237612929665)
--(axis cs:64991400,0.71089789959388)
--(axis cs:66701400,0.69945381094791)
--(axis cs:68411400,0.721831369211036)
--(axis cs:70463400,0.72081276311405)
--(axis cs:72173400,0.724368931161892)
--(axis cs:74225400,0.701141535419098)
--(axis cs:75935400,0.702495241702593)
--(axis cs:77987400,0.706403781200577)
--(axis cs:79697400,0.709585825754357)
--(axis cs:81407400,0.712191866093565)
--(axis cs:83459400,0.707436380396701)
--(axis cs:85169400,0.705399225017907)
--(axis cs:87221400,0.712240240032493)
--(axis cs:88931400,0.704006191648912)
--(axis cs:90983400,0.708239497759582)
--(axis cs:90983400,0.732300658627951)
--(axis cs:90983400,0.732300658627951)
--(axis cs:88931400,0.730083424947556)
--(axis cs:87221400,0.741361211368276)
--(axis cs:85169400,0.732812654596651)
--(axis cs:83459400,0.72441521791334)
--(axis cs:81407400,0.733141825280088)
--(axis cs:79697400,0.737356200614308)
--(axis cs:77987400,0.729271140931268)
--(axis cs:75935400,0.725216527848261)
--(axis cs:74225400,0.729988354915914)
--(axis cs:72173400,0.736320243076202)
--(axis cs:70463400,0.741403435505746)
--(axis cs:68411400,0.737663964355698)
--(axis cs:66701400,0.738827120406598)
--(axis cs:64991400,0.732104367728262)
--(axis cs:62939400,0.742871091624911)
--(axis cs:61229400,0.750427083897701)
--(axis cs:59177400,0.733001411306982)
--(axis cs:57467400,0.736650504943324)
--(axis cs:55415400,0.733572103917424)
--(axis cs:53705400,0.740930523087066)
--(axis cs:51995400,0.731480589284818)
--(axis cs:49943400,0.709541045050473)
--(axis cs:48233400,0.738362213778369)
--(axis cs:46181400,0.725712780893212)
--(axis cs:44471400,0.749591146064105)
--(axis cs:42419400,0.742916129334969)
--(axis cs:40709400,0.723575155864947)
--(axis cs:38999400,0.729083040570658)
--(axis cs:36947400,0.742522067211515)
--(axis cs:35237400,0.725272589237914)
--(axis cs:33185400,0.726260726594613)
--(axis cs:31475400,0.716108180158173)
--(axis cs:29423400,0.706583717534068)
--(axis cs:27713400,0.721010300981187)
--(axis cs:26003400,0.728737281059306)
--(axis cs:23951400,0.713902602297014)
--(axis cs:22241400,0.726079456283115)
--(axis cs:20189400,0.697163732803456)
--(axis cs:18479400,0.689128117104071)
--(axis cs:16427400,0.648089579755656)
--(axis cs:14717400,0.638333437367928)
--(axis cs:13007400,0.640853188778413)
--(axis cs:10955400,0.631494985810145)
--(axis cs:9245400,0.611470084988653)
--(axis cs:7193400,0.602831453707861)
--(axis cs:5483400,0.56072067194581)
--(axis cs:3431400,0.493871590333583)
--(axis cs:1721400,0.457031254656613)
--(axis cs:11400,0.458881586790085)
--cycle;

\addplot [line width=\linewidthtop, \res_color, mark=*, mark size=0, mark options={solid}]
table {%
11400 0.452302634716034
1721400 0.443410779116675
3431400 0.478083557696664
5483400 0.551326526464436
7193400 0.59361788046634
9245400 0.602749377387494
10955400 0.617771443424016
13007400 0.62978112180254
14717400 0.631794475021687
16427400 0.641029921712331
18479400 0.678810198716668
20189400 0.678862803274357
22241400 0.720895358674298
23951400 0.706368397251692
26003400 0.718625916526519
27713400 0.710775501862761
29423400 0.695364598161977
31475400 0.70973148334869
33185400 0.718895572946359
35237400 0.714683151174487
36947400 0.736098246832965
38999400 0.722120247833189
40709400 0.712719658589358
42419400 0.724647276576023
44471400 0.739671589107544
46181400 0.715156275725494
48233400 0.733373869780306
49943400 0.701191920687871
51995400 0.722337273525517
53705400 0.727872053250711
55415400 0.725731533631514
57467400 0.725627889129976
59177400 0.718015246938984
61229400 0.736956152895942
62939400 0.722387357489635
64991400 0.723218095601574
66701400 0.72681334207741
68411400 0.729598884038456
70463400 0.733241092405823
72173400 0.731135222129778
74225400 0.714469798717589
75935400 0.71257517489874
77987400 0.715310130028787
79697400 0.728407158616397
81407400 0.725661511148812
83459400 0.717848046602926
85169400 0.718876344827266
87221400 0.727730587310472
88931400 0.715077120999444
90983400 0.723725441259722
};

\path [draw=\whole_color, fill=\whole_color, opacity=0.2]
(axis cs:11400,0.463815778493881)
--(axis cs:11400,0.450637330114842)
--(axis cs:1721400,0.521561473142356)
--(axis cs:3431400,0.586867487698328)
--(axis cs:5483400,0.639998152722228)
--(axis cs:7193400,0.675001496668173)
--(axis cs:9245400,0.679362538362308)
--(axis cs:10955400,0.693133647394175)
--(axis cs:13007400,0.697366020870196)
--(axis cs:14717400,0.688595309024494)
--(axis cs:16427400,0.705061794440332)
--(axis cs:18479400,0.723350397447544)
--(axis cs:20189400,0.706582197405649)
--(axis cs:22241400,0.727960218493641)
--(axis cs:23951400,0.696481473913278)
--(axis cs:26003400,0.71413557615525)
--(axis cs:27713400,0.712899961330136)
--(axis cs:29423400,0.712080261821414)
--(axis cs:31475400,0.719310467857898)
--(axis cs:33185400,0.712321019005108)
--(axis cs:35237400,0.71631863662311)
--(axis cs:36947400,0.72901385430008)
--(axis cs:38999400,0.722360704875144)
--(axis cs:40709400,0.718697882832184)
--(axis cs:42419400,0.706576845516623)
--(axis cs:44471400,0.720468664148699)
--(axis cs:46181400,0.69563576589034)
--(axis cs:48233400,0.715984548600361)
--(axis cs:49943400,0.695848572036314)
--(axis cs:51995400,0.701951327737401)
--(axis cs:53705400,0.720367345569924)
--(axis cs:55415400,0.708060701715835)
--(axis cs:57467400,0.726375170755039)
--(axis cs:59177400,0.709693266635608)
--(axis cs:61229400,0.728457612373979)
--(axis cs:62939400,0.716990083829871)
--(axis cs:64991400,0.716279974622224)
--(axis cs:66701400,0.711574274563291)
--(axis cs:68411400,0.702587882412888)
--(axis cs:70463400,0.718512933895536)
--(axis cs:72173400,0.73070177208461)
--(axis cs:74225400,0.702865980127284)
--(axis cs:75935400,0.71787199426978)
--(axis cs:77987400,0.722911532704905)
--(axis cs:79697400,0.709230104206792)
--(axis cs:81407400,0.722412090325321)
--(axis cs:83459400,0.709032333024611)
--(axis cs:85169400,0.70683354473763)
--(axis cs:87221400,0.719662370852647)
--(axis cs:88931400,0.710096593387047)
--(axis cs:90983400,0.711880641251565)
--(axis cs:90983400,0.725831804804146)
--(axis cs:90983400,0.725831804804146)
--(axis cs:88931400,0.737041331607071)
--(axis cs:87221400,0.742357402974368)
--(axis cs:85169400,0.733632148084524)
--(axis cs:83459400,0.722781940106155)
--(axis cs:81407400,0.749572588717715)
--(axis cs:79697400,0.737494831803013)
--(axis cs:77987400,0.74190669237617)
--(axis cs:75935400,0.730960472119781)
--(axis cs:74225400,0.724881716088517)
--(axis cs:72173400,0.750181093355017)
--(axis cs:70463400,0.749390752868472)
--(axis cs:68411400,0.731780325834093)
--(axis cs:66701400,0.734653104563982)
--(axis cs:64991400,0.735608583247875)
--(axis cs:62939400,0.734194592711927)
--(axis cs:61229400,0.748147928632053)
--(axis cs:59177400,0.731761224521983)
--(axis cs:57467400,0.737529854142215)
--(axis cs:55415400,0.726990707486308)
--(axis cs:53705400,0.746816948366641)
--(axis cs:51995400,0.730766743528918)
--(axis cs:49943400,0.721312482740568)
--(axis cs:48233400,0.73456673121223)
--(axis cs:46181400,0.725009542828461)
--(axis cs:44471400,0.734020209792423)
--(axis cs:42419400,0.730049421364988)
--(axis cs:40709400,0.737605575820767)
--(axis cs:38999400,0.744956641523595)
--(axis cs:36947400,0.748396595410127)
--(axis cs:35237400,0.727984286896274)
--(axis cs:33185400,0.735171429024962)
--(axis cs:31475400,0.737417253125397)
--(axis cs:29423400,0.727061049746887)
--(axis cs:27713400,0.741987122370525)
--(axis cs:26003400,0.735107442464801)
--(axis cs:23951400,0.726121370579711)
--(axis cs:22241400,0.754085445402848)
--(axis cs:20189400,0.734061521149329)
--(axis cs:18479400,0.742902226791554)
--(axis cs:16427400,0.718964986338881)
--(axis cs:14717400,0.705772713945834)
--(axis cs:13007400,0.714693717971427)
--(axis cs:10955400,0.710805061405022)
--(axis cs:9245400,0.696897878176604)
--(axis cs:7193400,0.694100283983929)
--(axis cs:5483400,0.661149872973169)
--(axis cs:3431400,0.60854379965167)
--(axis cs:1721400,0.542429071152583)
--(axis cs:11400,0.463815778493881)
--cycle;

\addplot [line width=\linewidthothers, \whole_color, mark=*, mark size=0, mark options={solid}]
table {%
11400 0.459703952074051
1721400 0.53728926833719
3431400 0.605032672981906
5483400 0.651076380542918
7193400 0.68231074386652
9245400 0.691227339712634
10955400 0.699530772252226
13007400 0.703899050624328
14717400 0.696768147129425
16427400 0.712323623791331
18479400 0.730566175757112
20189400 0.71959508526777
22241400 0.740363702097606
23951400 0.716721440128853
26003400 0.729800105907833
27713400 0.727121558983253
29423400 0.719202272088444
31475400 0.729281455390467
33185400 0.721335144295627
35237400 0.719822819529132
36947400 0.738211986469048
38999400 0.731883642907189
40709400 0.729067522106045
42419400 0.720872581693589
44471400 0.731274739639516
46181400 0.712819428428661
48233400 0.722428698712701
49943400 0.70724493281522
51995400 0.722151843766197
53705400 0.734046446698145
55415400 0.717213856095289
57467400 0.730757461723149
59177400 0.722869025142357
61229400 0.734265307202977
62939400 0.728531678636378
64991400 0.725144840531266
66701400 0.728561264687923
68411400 0.716844212988855
70463400 0.734992869354468
72173400 0.737933164231964
74225400 0.71774726531499
75935400 0.725849849103406
77987400 0.732452024407519
79697400 0.72518446721156
81407400 0.736180998531613
83459400 0.717230724156982
85169400 0.720429350919532
87221400 0.731634565834619
88931400 0.722805423824005
90983400 0.719638144446183
};

\path [draw=C2, fill=C2, opacity=0.2]
(axis cs:121600,0)
--(axis cs:121600,0)
--(axis cs:3283200,0.223684210526316)
--(axis cs:6444800,0.368421052631579)
--(axis cs:9606400,0.480263157894737)
--(axis cs:12889600,0.598684210526316)
--(axis cs:16051200,0.651315789473684)
--(axis cs:19212800,0.680921052631579)
--(axis cs:22374400,0.697368421052632)
--(axis cs:25657600,0.634868421052632)
--(axis cs:28819200,0.694078947368421)
--(axis cs:31980800,0.697368421052632)
--(axis cs:35142400,0.595394736842105)
--(axis cs:38425600,0.585526315789474)
--(axis cs:41587200,0.601973684210526)
--(axis cs:44748800,0.615131578947368)
--(axis cs:47910400,0.625)
--(axis cs:51193600,0.634868421052632)
--(axis cs:54355200,0.651315789473684)
--(axis cs:57516800,0.559210526315789)
--(axis cs:60800000,0.539473684210526)
--(axis cs:60800000,0.618421052631579)
--(axis cs:60800000,0.618421052631579)
--(axis cs:57516800,0.657894736842105)
--(axis cs:54355200,0.710526315789474)
--(axis cs:51193600,0.700657894736842)
--(axis cs:47910400,0.717105263157895)
--(axis cs:44748800,0.707236842105263)
--(axis cs:41587200,0.713815789473684)
--(axis cs:38425600,0.677631578947368)
--(axis cs:35142400,0.6875)
--(axis cs:31980800,0.743421052631579)
--(axis cs:28819200,0.75)
--(axis cs:25657600,0.717105263157895)
--(axis cs:22374400,0.756578947368421)
--(axis cs:19212800,0.75)
--(axis cs:16051200,0.733552631578947)
--(axis cs:12889600,0.697368421052632)
--(axis cs:9606400,0.536184210526316)
--(axis cs:6444800,0.519736842105263)
--(axis cs:3283200,0.282894736842105)
--(axis cs:121600,0)
--cycle;

\addplot [line width=\linewidthothers, C2, mark=*, mark size=0, mark options={solid}]
table {%
121600 0
3283200 0.256578947368421
6444800 0.447368421052632
9606400 0.509868421052632
12889600 0.664473684210526
16051200 0.694078947368421
19212800 0.703947368421053
22374400 0.730263157894737
25657600 0.680921052631579
28819200 0.726973684210526
31980800 0.720394736842105
35142400 0.648026315789474
38425600 0.631578947368421
41587200 0.661184210526316
44748800 0.674342105263158
47910400 0.684210526315789
51193600 0.667763157894737
54355200 0.684210526315789
57516800 0.615131578947368
60800000 0.572368421052632
};

\path [draw=C5, fill=C5, opacity=0.2]
(axis cs:121600,0.256578947368421)
--(axis cs:121600,0.157894736842105)
--(axis cs:3283200,0.0921052631578947)
--(axis cs:6444800,0.486842105263158)
--(axis cs:9606400,0.526315789473684)
--(axis cs:12889600,0.592105263157895)
--(axis cs:16051200,0.611842105263158)
--(axis cs:19212800,0.526315789473684)
--(axis cs:22374400,0.539473684210526)
--(axis cs:25657600,0.684210526315789)
--(axis cs:28819200,0.565789473684211)
--(axis cs:31980800,0.605263157894737)
--(axis cs:35142400,0.559210526315789)
--(axis cs:38425600,0.631578947368421)
--(axis cs:41587200,0.671052631578947)
--(axis cs:44748800,0.572368421052632)
--(axis cs:47910400,0.526315789473684)
--(axis cs:51193600,0.519736842105263)
--(axis cs:54355200,0.532894736842105)
--(axis cs:57516800,0.592105263157895)
--(axis cs:60800000,0.592105263157895)
--(axis cs:60800000,0.684210526315789)
--(axis cs:60800000,0.684210526315789)
--(axis cs:57516800,0.736842105263158)
--(axis cs:54355200,0.611842105263158)
--(axis cs:51193600,0.611842105263158)
--(axis cs:47910400,0.618421052631579)
--(axis cs:44748800,0.684210526315789)
--(axis cs:41587200,0.730263157894737)
--(axis cs:38425600,0.776315789473684)
--(axis cs:35142400,0.671052631578947)
--(axis cs:31980800,0.802631578947368)
--(axis cs:28819200,0.631578947368421)
--(axis cs:25657600,0.763157894736842)
--(axis cs:22374400,0.651315789473684)
--(axis cs:19212800,0.638157894736842)
--(axis cs:16051200,0.710526315789474)
--(axis cs:12889600,0.697368421052632)
--(axis cs:9606400,0.611842105263158)
--(axis cs:6444800,0.592105263157895)
--(axis cs:3283200,0.197368421052632)
--(axis cs:121600,0.256578947368421)
--cycle;

\addplot [line width=\linewidthothers, C5, mark=*, mark size=0, mark options={solid}]
table {%
121600 0.197368421052632
3283200 0.144736842105263
6444800 0.526315789473684
9606400 0.578947368421053
12889600 0.651315789473684
16051200 0.664473684210526
19212800 0.578947368421053
22374400 0.611842105263158
25657600 0.756578947368421
28819200 0.605263157894737
31980800 0.690789473684211
35142400 0.605263157894737
38425600 0.717105263157895
41587200 0.703947368421053
44748800 0.638157894736842
47910400 0.572368421052632
51193600 0.559210526315789
54355200 0.559210526315789
57516800 0.690789473684211
60800000 0.651315789473684
};

\path [draw=C4, fill=C4, opacity=0.2]
(axis cs:10400,0)
--(axis cs:10400,0)
--(axis cs:416400,0)
--(axis cs:833200,0)
--(axis cs:1240800,0)
--(axis cs:1645200,0)
--(axis cs:2037200,0)
--(axis cs:2449600,0)
--(axis cs:2872400,0)
--(axis cs:3272000,0)
--(axis cs:3672400,0)
--(axis cs:4081200,0)
--(axis cs:4482400,0)
--(axis cs:4892800,0)
--(axis cs:5296400,0)
--(axis cs:5721200,0)
--(axis cs:6139200,0)
--(axis cs:6531200,0)
--(axis cs:6947200,0)
--(axis cs:7351600,0)
--(axis cs:7776400,0)
--(axis cs:7776400,0.0125)
--(axis cs:7776400,0.0125)
--(axis cs:7351600,0.0253124999999955)
--(axis cs:6947200,0.0375)
--(axis cs:6531200,0.0625)
--(axis cs:6139200,0.0125)
--(axis cs:5721200,0)
--(axis cs:5296400,0.0125)
--(axis cs:4892800,0)
--(axis cs:4482400,0)
--(axis cs:4081200,0)
--(axis cs:3672400,0)
--(axis cs:3272000,0)
--(axis cs:2872400,0)
--(axis cs:2449600,0)
--(axis cs:2037200,0)
--(axis cs:1645200,0)
--(axis cs:1240800,0)
--(axis cs:833200,0)
--(axis cs:416400,0)
--(axis cs:10400,0)
--cycle;

\addplot [line width=\linewidthothers, C4, mark=*, mark size=0, mark options={solid}]
table {%
10400 0
416400 0
833200 0
1240800 0
1645200 0
2037200 0
2449600 0
2872400 0
3272000 0
3672400 0
4081200 0
4482400 0
4892800 0
5296400 0
5721200 0
6139200 0
6531200 0.025
6947200 0
7351600 0
7776400 0
};

\path [draw=C6, fill=C6, opacity=0.2]
(axis cs:10400,0)
--(axis cs:10400,0)
--(axis cs:397600,0)
--(axis cs:802800,0)
--(axis cs:1204800,0)
--(axis cs:1610800,0)
--(axis cs:1994000,0)
--(axis cs:2397200,0)
--(axis cs:2798000,0)
--(axis cs:3194800,0)
--(axis cs:3606800,0)
--(axis cs:3992400,0)
--(axis cs:4410800,0)
--(axis cs:4816000,0)
--(axis cs:5220000,0)
--(axis cs:5627200,0)
--(axis cs:6026800,0)
--(axis cs:6442000,0)
--(axis cs:6892400,0)
--(axis cs:7280400,0)
--(axis cs:7691200,0)
--(axis cs:7691200,0.0125)
--(axis cs:7691200,0.0125)
--(axis cs:7280400,0)
--(axis cs:6892400,0)
--(axis cs:6442000,0)
--(axis cs:6026800,0)
--(axis cs:5627200,0)
--(axis cs:5220000,0)
--(axis cs:4816000,0)
--(axis cs:4410800,0)
--(axis cs:3992400,0)
--(axis cs:3606800,0)
--(axis cs:3194800,0)
--(axis cs:2798000,0.0125)
--(axis cs:2397200,0)
--(axis cs:1994000,0)
--(axis cs:1610800,0)
--(axis cs:1204800,0)
--(axis cs:802800,0)
--(axis cs:397600,0)
--(axis cs:10400,0)
--cycle;

\addplot [line width=\linewidthothers, C6, mark=*, mark size=0, mark options={solid}]
table {%
10400 0
397600 0
802800 0
1204800 0
1610800 0
1994000 0
2397200 0
2798000 0
3194800 0
3606800 0
3992400 0
4410800 0
4816000 0
5220000 0
5627200 0
6026800 0
6442000 0
6892400 0
7280400 0
7691200 0
};

\end{axis}

\end{tikzpicture}

%% file: images/tex/multi_box_nm_collision_counts.tex
%
%

\begin{tikzpicture}
\def\res_color{C0}
\def\replace_color{C1}
\def\whole_color{C3}
\def\bbrl_color{C9}
\def\ppo_color{C2}
\def\sac_color{C4}
\def\linewidthtop{1mm}
\def\linewidthothers{0.5mm}

\begin{axis}[
legend cell align={left},
legend style={fill opacity=0.8, draw opacity=1, text opacity=1, draw=lightgray204, at={(0.03,0.03)},  anchor=north west},
tick align=outside,
tick pos=left,
x grid style={white},
scaled x ticks=false,
xticklabels={,0,0.5,1,1.5,2, 2.5},
xlabel={Environment Interactions ($\times 10^7$)},
xmajorgrids,
xmin=-1500000, xmax=\multiboxxmax,
xtick style={color=black},
y grid style={white},
ylabel={Collision Counts, IQM},
ymajorgrids,
ymin=0., ymax=40.0,
ytick style={color=black},
axis background/.style={fill=plot_background},
label style={font=\large},
tick label style={font=\large},
x axis line style={draw=none},
y axis line style={draw=none},
]
\path [draw=\bbrl_color, fill=\bbrl_color, opacity=0.2]
(axis cs:19400,14.5444078445435)
--(axis cs:19400,14.336328214407)
--(axis cs:3112700,13.0826478898525)
--(axis cs:6333000,11.2209167573601)
--(axis cs:10275100,8.4352208749915)
--(axis cs:13589000,7.12839487601795)
--(axis cs:17586100,6.27049025081376)
--(axis cs:20931700,5.51100217274718)
--(axis cs:24961300,5.47941446041812)
--(axis cs:28323300,5.10364684398734)
--(axis cs:31683900,5.07486652440921)
--(axis cs:35726000,4.53659259033695)
--(axis cs:39105500,4.63816337181532)
--(axis cs:43159900,3.83209855377161)
--(axis cs:46545200,3.78177462425685)
--(axis cs:50609500,4.09824769132421)
--(axis cs:53996700,3.78752722052953)
--(axis cs:57383400,3.95515705856702)
--(axis cs:61450800,3.68586611983809)
--(axis cs:64838500,3.64843136335022)
--(axis cs:68906600,3.39931119354716)
--(axis cs:72298600,3.75964747955341)
--(axis cs:76370300,3.12176943776458)
--(axis cs:79765700,3.34424298922333)
--(axis cs:83157500,3.42843584452043)
--(axis cs:87230500,3.43315063621388)
--(axis cs:90624600,3.50369904278651)
--(axis cs:94697100,3.07934443574472)
--(axis cs:98094600,3.23295610169996)
--(axis cs:102170100,2.84097944043701)
--(axis cs:105564600,3.00980755821573)
--(axis cs:108958800,2.96010435982385)
--(axis cs:113034000,2.8339959991834)
--(axis cs:116432700,3.05889447779423)
--(axis cs:120507600,3.07392891065632)
--(axis cs:123901200,3.06548715998315)
--(axis cs:127975600,3.0137016979909)
--(axis cs:131373700,3.15205227720959)
--(axis cs:134769900,2.78008515857814)
--(axis cs:138848400,2.86122494926948)
--(axis cs:142246700,3.20050916824487)
--(axis cs:146320600,2.78247597919228)
--(axis cs:149719000,2.86150260814327)
--(axis cs:153801300,2.6201502935812)
--(axis cs:157203500,2.99394795425383)
--(axis cs:160601700,2.82829796380825)
--(axis cs:164680900,2.91467080359571)
--(axis cs:168080600,2.77310536780553)
--(axis cs:172161200,2.71052868497418)
--(axis cs:175565900,2.79699124988961)
--(axis cs:179652200,2.76100591505184)
--(axis cs:179652200,3.25176218292644)
--(axis cs:179652200,3.25176218292644)
--(axis cs:175565900,3.16582530373222)
--(axis cs:172161200,3.10167155241801)
--(axis cs:168080600,3.07602729894517)
--(axis cs:164680900,3.35612004778369)
--(axis cs:160601700,3.43277002817512)
--(axis cs:157203500,3.29271934703856)
--(axis cs:153801300,3.14869127463465)
--(axis cs:149719000,3.2939650326311)
--(axis cs:146320600,3.29253860637364)
--(axis cs:142246700,3.54715061168828)
--(axis cs:138848400,3.52151444491327)
--(axis cs:134769900,3.30039237207285)
--(axis cs:131373700,3.57623757490301)
--(axis cs:127975600,3.52072152709214)
--(axis cs:123901200,3.65240772912072)
--(axis cs:120507600,3.53989978718888)
--(axis cs:116432700,3.55078257376763)
--(axis cs:113034000,3.34755401571868)
--(axis cs:108958800,3.51653613655924)
--(axis cs:105564600,3.48945521027055)
--(axis cs:102170100,3.38889960393297)
--(axis cs:98094600,3.93817728935841)
--(axis cs:94697100,3.81386707093351)
--(axis cs:90624600,3.96689904656226)
--(axis cs:87230500,3.89384390195938)
--(axis cs:83157500,3.91199546577329)
--(axis cs:79765700,4.08829127152772)
--(axis cs:76370300,3.67197250088359)
--(axis cs:72298600,4.20476503348887)
--(axis cs:68906600,4.03227642991144)
--(axis cs:64838500,4.04120410025319)
--(axis cs:61450800,4.07492782932681)
--(axis cs:57383400,4.46882013897684)
--(axis cs:53996700,4.37070146145142)
--(axis cs:50609500,4.55382106948276)
--(axis cs:46545200,4.27993011544428)
--(axis cs:43159900,4.28482154431166)
--(axis cs:39105500,5.12027062887938)
--(axis cs:35726000,5.07815505068478)
--(axis cs:31683900,5.50863722320888)
--(axis cs:28323300,5.59280290066693)
--(axis cs:24961300,6.20625623083729)
--(axis cs:20931700,5.84330864580006)
--(axis cs:17586100,6.79829133833542)
--(axis cs:13589000,7.75262829575672)
--(axis cs:10275100,8.91305032255332)
--(axis cs:6333000,11.6979964543134)
--(axis cs:3112700,13.5328690037131)
--(axis cs:19400,14.5444078445435)
--cycle;

\addplot [line width=\linewidthothers, \bbrl_color, mark=*, mark size=0, mark options={solid}]
table {%
19400 14.4621710777283
3112700 13.3768502920866
6333000 11.4508658170234
10275100 8.77061347536437
13589000 7.50123485887866
17586100 6.48686256067183
20931700 5.64244475308168
24961300 6.00774062599956
28323300 5.32077558109452
31683900 5.34818430286264
35726000 4.68549605868314
39105500 4.90474109844854
43159900 4.07903008371636
46545200 3.91524724063465
50609500 4.2227766158731
53996700 3.97464523352368
57383400 4.19504622343318
61450800 3.85324805660222
64838500 3.75943470171806
68906600 3.66340256231676
72298600 3.88198191676106
76370300 3.39852399737418
79765700 3.76971466934641
83157500 3.64195839991381
87230500 3.69389725078737
90624600 3.7248366660165
94697100 3.48663619147748
98094600 3.74318329523846
102170100 3.1519273188058
105564600 3.27123630728196
108958800 3.22433286983259
113034000 3.11113234681951
116432700 3.29432400631145
120507600 3.29862109659423
123901200 3.4244832648351
127975600 3.40580677842254
131373700 3.4295010845214
134769900 3.13534032733165
138848400 3.14563743576905
142246700 3.30422563413614
146320600 3.01747930353247
149719000 3.09967860937442
153801300 2.98535423032944
157203500 3.13934345848675
160601700 3.20961063314851
164680900 3.15460733123861
168080600 2.93841741282628
172161200 2.98164399623991
175565900 2.98253434579625
179652200 2.98398962899203
};

\path [draw=\replace_color, fill=\replace_color, opacity=0.2]
(axis cs:11400,9.38898038864136)
--(axis cs:11400,9.09703946113586)
--(axis cs:1721400,10.6096319332719)
--(axis cs:3431400,9.80723555968143)
--(axis cs:5483400,8.98717709349148)
--(axis cs:7193400,8.33443576113825)
--(axis cs:9245400,8.08212916442976)
--(axis cs:10955400,7.07160557356781)
--(axis cs:13007400,6.14489593288777)
--(axis cs:14717400,6.22016647744126)
--(axis cs:16427400,5.85633015287203)
--(axis cs:18479400,5.04111171218821)
--(axis cs:20189400,4.9466056158819)
--(axis cs:22241400,3.86910205125775)
--(axis cs:23951400,4.12132882572482)
--(axis cs:26003400,3.96143629503496)
--(axis cs:27713400,3.6910508136052)
--(axis cs:29423400,3.91113011577593)
--(axis cs:31475400,3.4543667302104)
--(axis cs:33185400,3.35172676179036)
--(axis cs:35237400,3.34896179624437)
--(axis cs:36947400,3.319599422078)
--(axis cs:38999400,2.91599746985551)
--(axis cs:40709400,3.16237176692594)
--(axis cs:42419400,3.19295827800538)
--(axis cs:44471400,2.76585870659305)
--(axis cs:46181400,3.11045181009986)
--(axis cs:48233400,3.04398730591429)
--(axis cs:49943400,3.02703599287174)
--(axis cs:51995400,2.7571499552913)
--(axis cs:53705400,2.71155016573218)
--(axis cs:55415400,2.63523064249944)
--(axis cs:57467400,2.7703164474545)
--(axis cs:59177400,2.62272286987577)
--(axis cs:61229400,2.63008269505148)
--(axis cs:62939400,2.85900234093628)
--(axis cs:64991400,2.43015990065662)
--(axis cs:66701400,2.49051817448153)
--(axis cs:68411400,2.84319937851594)
--(axis cs:70463400,2.4385031845181)
--(axis cs:72173400,2.50115696081486)
--(axis cs:74225400,2.41565222372129)
--(axis cs:75935400,2.54362736618826)
--(axis cs:77987400,2.43998758063271)
--(axis cs:79697400,2.65273556951484)
--(axis cs:81407400,2.37317093952334)
--(axis cs:83459400,2.58245100080809)
--(axis cs:85169400,2.48491853445151)
--(axis cs:87221400,2.30244378188151)
--(axis cs:88931400,2.60113368008055)
--(axis cs:90983400,2.16405654185216)
--(axis cs:90983400,2.46716005616032)
--(axis cs:90983400,2.46716005616032)
--(axis cs:88931400,2.90744378103744)
--(axis cs:87221400,2.69439632215557)
--(axis cs:85169400,2.8202483981985)
--(axis cs:83459400,2.86395486166876)
--(axis cs:81407400,2.91390465389056)
--(axis cs:79697400,2.85801078464043)
--(axis cs:77987400,2.73454711502037)
--(axis cs:75935400,2.85927196513252)
--(axis cs:74225400,3.04189392575775)
--(axis cs:72173400,2.81831001244357)
--(axis cs:70463400,2.8773444003951)
--(axis cs:68411400,3.06059678177971)
--(axis cs:66701400,3.05217372431196)
--(axis cs:64991400,2.60404189751581)
--(axis cs:62939400,3.25612403670895)
--(axis cs:61229400,2.93342824998041)
--(axis cs:59177400,3.06358104878402)
--(axis cs:57467400,3.17954116310303)
--(axis cs:55415400,2.85665654203594)
--(axis cs:53705400,3.25514033505638)
--(axis cs:51995400,3.06526614862226)
--(axis cs:49943400,3.39397803263239)
--(axis cs:48233400,3.7005114334016)
--(axis cs:46181400,3.64403414365922)
--(axis cs:44471400,3.34334579761399)
--(axis cs:42419400,3.5520957585226)
--(axis cs:40709400,3.6508583237078)
--(axis cs:38999400,3.39340466386146)
--(axis cs:36947400,4.03854413396412)
--(axis cs:35237400,3.6638395303792)
--(axis cs:33185400,4.15888433703567)
--(axis cs:31475400,4.00167049462717)
--(axis cs:29423400,4.56554257585259)
--(axis cs:27713400,4.41494102538684)
--(axis cs:26003400,4.74853609723475)
--(axis cs:23951400,4.68025294549365)
--(axis cs:22241400,4.38782092911369)
--(axis cs:20189400,5.92453823003025)
--(axis cs:18479400,5.62494832607286)
--(axis cs:16427400,6.44245639011021)
--(axis cs:14717400,6.6893151170755)
--(axis cs:13007400,6.94147532938524)
--(axis cs:10955400,7.97243899199788)
--(axis cs:9245400,8.46113272581082)
--(axis cs:7193400,8.94656959674126)
--(axis cs:5483400,9.71261813394449)
--(axis cs:3431400,10.337150105508)
--(axis cs:1721400,11.2688629627228)
--(axis cs:11400,9.38898038864136)
--cycle;

\addplot [line width=\linewidthothers, \replace_color, mark=*, mark size=0, mark options={solid}]
table {%
11400 9.28371715545654
1721400 10.8791374862194
3431400 10.0667611586396
5483400 9.26880886760046
7193400 8.64535753748532
9245400 8.19617873274625
10955400 7.38425489588778
13007400 6.54479285762213
14717400 6.42685771395946
16427400 6.22155033208478
18479400 5.2987151426863
20189400 5.53783508105449
22241400 4.15733311765273
23951400 4.45262520133905
26003400 4.48884784416019
27713400 4.18143605623147
29423400 4.33521355237574
31475400 3.78189037036742
33185400 3.85317538125336
35237400 3.53850886783743
36947400 3.6779362855092
38999400 3.16876068618728
40709400 3.36022727344232
42419400 3.43341225217236
44471400 3.15796305218689
46181400 3.38483029189387
48233400 3.41064270963869
49943400 3.23672363374794
51995400 2.95757095556054
53705400 2.96251123353408
55415400 2.78518846167266
57467400 2.95572105591086
59177400 2.85974734296512
61229400 2.81113794483111
62939400 3.0676364024699
64991400 2.50135397061366
66701400 2.66859461700162
68411400 2.91924474779179
70463400 2.7131674234325
72173400 2.71536053975674
74225400 2.72299329466429
75935400 2.65961929781728
77987400 2.607634655092
79697400 2.78056094219295
81407400 2.59253026259489
83459400 2.77164023231217
85169400 2.60980700821297
87221400 2.58994490018978
88931400 2.74961793064417
90983400 2.27168711091212
};

\path [draw=\res_color, fill=\res_color, opacity=0.2]
(axis cs:11400,9.48766446113586)
--(axis cs:11400,9.09375023841858)
--(axis cs:1721400,8.7636204957962)
--(axis cs:3431400,7.91143151046708)
--(axis cs:5483400,6.15192585561817)
--(axis cs:7193400,4.96754264474805)
--(axis cs:9245400,4.46282436536015)
--(axis cs:10955400,4.06866770429806)
--(axis cs:13007400,3.87102081705856)
--(axis cs:14717400,3.59948567513413)
--(axis cs:16427400,3.34698984416036)
--(axis cs:18479400,2.98229345166853)
--(axis cs:20189400,3.17275207667036)
--(axis cs:22241400,2.23839242191802)
--(axis cs:23951400,2.7497716238612)
--(axis cs:26003400,2.38072289331761)
--(axis cs:27713400,2.56717081132671)
--(axis cs:29423400,2.5236108858543)
--(axis cs:31475400,2.67324749779999)
--(axis cs:33185400,2.39737320677427)
--(axis cs:35237400,2.21344513341631)
--(axis cs:36947400,2.20473022994662)
--(axis cs:38999400,2.22683074587411)
--(axis cs:40709400,2.45652433497144)
--(axis cs:42419400,2.2417306339549)
--(axis cs:44471400,2.31685399357065)
--(axis cs:46181400,2.20903943840442)
--(axis cs:48233400,2.18149810199536)
--(axis cs:49943400,2.21297595207976)
--(axis cs:51995400,1.95660094400965)
--(axis cs:53705400,2.31724828363971)
--(axis cs:55415400,1.89446292395922)
--(axis cs:57467400,2.32886410185501)
--(axis cs:59177400,2.41149210334163)
--(axis cs:61229400,2.07145093401639)
--(axis cs:62939400,2.3145270797433)
--(axis cs:64991400,1.91994541845287)
--(axis cs:66701400,2.06689063706254)
--(axis cs:68411400,1.9360849310677)
--(axis cs:70463400,1.97074827799005)
--(axis cs:72173400,1.82948229554206)
--(axis cs:74225400,2.15333542443487)
--(axis cs:75935400,2.1771436607862)
--(axis cs:77987400,2.0178605893621)
--(axis cs:79697400,2.11167473440618)
--(axis cs:81407400,2.0234181575753)
--(axis cs:83459400,1.79173433910888)
--(axis cs:85169400,2.11511742083698)
--(axis cs:87221400,2.02373529660322)
--(axis cs:88931400,2.09485218676695)
--(axis cs:90983400,1.84208258456199)
--(axis cs:90983400,2.13157765077504)
--(axis cs:90983400,2.13157765077504)
--(axis cs:88931400,2.41670228922579)
--(axis cs:87221400,2.25902834273542)
--(axis cs:85169400,2.32001067203085)
--(axis cs:83459400,2.09011112236428)
--(axis cs:81407400,2.30496476988124)
--(axis cs:79697400,2.49191248771937)
--(axis cs:77987400,2.1428451965063)
--(axis cs:75935400,2.46877665980721)
--(axis cs:74225400,2.5876423064922)
--(axis cs:72173400,2.25702185454226)
--(axis cs:70463400,2.14588959750191)
--(axis cs:68411400,2.15793227111824)
--(axis cs:66701400,2.34481512661045)
--(axis cs:64991400,2.27170640828983)
--(axis cs:62939400,2.43041477221083)
--(axis cs:61229400,2.45364355683378)
--(axis cs:59177400,2.63876006662402)
--(axis cs:57467400,2.69124855972605)
--(axis cs:55415400,2.44364533871505)
--(axis cs:53705400,2.57317733331421)
--(axis cs:51995400,2.29839433414515)
--(axis cs:49943400,2.59948802620078)
--(axis cs:48233400,2.56458374953573)
--(axis cs:46181400,2.65894371195897)
--(axis cs:44471400,2.55305022522555)
--(axis cs:42419400,2.38787969423891)
--(axis cs:40709400,2.77600517585667)
--(axis cs:38999400,2.57167331941007)
--(axis cs:36947400,2.46627435673131)
--(axis cs:35237400,2.35208066688695)
--(axis cs:33185400,2.90015152700179)
--(axis cs:31475400,3.04256439200462)
--(axis cs:29423400,3.03981697332457)
--(axis cs:27713400,3.11817064760649)
--(axis cs:26003400,2.67485050531576)
--(axis cs:23951400,3.09088005292923)
--(axis cs:22241400,2.78043285767991)
--(axis cs:20189400,3.46980479640041)
--(axis cs:18479400,3.38999547793121)
--(axis cs:16427400,3.69955733235221)
--(axis cs:14717400,3.76231603304373)
--(axis cs:13007400,4.39918571447959)
--(axis cs:10955400,4.53418835481755)
--(axis cs:9245400,5.30814606220218)
--(axis cs:7193400,5.78896179326523)
--(axis cs:5483400,6.97863128323661)
--(axis cs:3431400,8.4454480391054)
--(axis cs:1721400,9.29258324205875)
--(axis cs:11400,9.48766446113586)
--cycle;

\addplot [line width=\linewidthtop, \res_color, mark=*, mark size=0, mark options={solid}]
table {%
11400 9.30674362182617
1721400 9.06522405147552
3431400 8.16910031880252
5483400 6.40852432995234
7193400 5.41291303158027
9245400 4.73589685599989
10955400 4.30361418181291
13007400 4.09527264192102
14717400 3.71578793373803
16427400 3.48787546396441
18479400 3.20668137192665
20189400 3.36062864863074
22241400 2.4814232546343
23951400 2.98025024574672
26003400 2.49140293402151
27713400 2.9432026228668
29423400 2.75522313710393
31475400 2.83594294062132
33185400 2.65502350367307
35237400 2.29669469955168
36947400 2.33697017358936
38999400 2.38418682507027
40709400 2.55856162661589
42419400 2.30886668546761
44471400 2.42126842670559
46181400 2.39459818645688
48233400 2.35443524838857
49943400 2.38587654047369
51995400 2.09608299247622
53705400 2.43503326815234
55415400 2.20958526564174
57467400 2.49996194499192
59177400 2.54157566334915
61229400 2.17889607174822
62939400 2.36804155039604
64991400 2.04372561342321
66701400 2.22002034003895
68411400 2.07075544443359
70463400 2.03479643141175
72173400 2.01830534097714
74225400 2.37107094121296
75935400 2.27839945397397
77987400 2.09712627917233
79697400 2.35701340914695
81407400 2.18421276117633
83459400 1.96900371893531
85169400 2.23406405427622
87221400 2.13375118215203
88931400 2.21329490631345
90983400 1.98510031365427
};

\path [draw=\whole_color, fill=\whole_color, opacity=0.2]
(axis cs:11400,9.58470368385315)
--(axis cs:11400,8.96940807104111)
--(axis cs:1721400,6.85377260297537)
--(axis cs:3431400,5.11878077907022)
--(axis cs:5483400,4.04113185846836)
--(axis cs:7193400,4.02634311219936)
--(axis cs:9245400,3.39785746536723)
--(axis cs:10955400,3.27125796154578)
--(axis cs:13007400,3.08479161925456)
--(axis cs:14717400,2.90612800485882)
--(axis cs:16427400,2.68355708578357)
--(axis cs:18479400,2.5348575263072)
--(axis cs:20189400,2.67352241702106)
--(axis cs:22241400,2.13609249054078)
--(axis cs:23951400,2.41209561460123)
--(axis cs:26003400,2.30994334285804)
--(axis cs:27713400,2.42214525412634)
--(axis cs:29423400,2.55535056477591)
--(axis cs:31475400,2.53707848500693)
--(axis cs:33185400,2.38073133745752)
--(axis cs:35237400,2.14915705352334)
--(axis cs:36947400,2.12011030902642)
--(axis cs:38999400,2.12883289683949)
--(axis cs:40709400,2.39241610199088)
--(axis cs:42419400,2.30132082678713)
--(axis cs:44471400,2.23312649189337)
--(axis cs:46181400,2.18477871322844)
--(axis cs:48233400,2.11629884738681)
--(axis cs:49943400,2.31704791118051)
--(axis cs:51995400,2.13039867101265)
--(axis cs:53705400,2.31072684347758)
--(axis cs:55415400,2.05204012762796)
--(axis cs:57467400,2.31054513393606)
--(axis cs:59177400,2.30684241362912)
--(axis cs:61229400,2.18938530819403)
--(axis cs:62939400,2.3490169750067)
--(axis cs:64991400,1.80484424292109)
--(axis cs:66701400,2.06944525239807)
--(axis cs:68411400,1.83548354061487)
--(axis cs:70463400,2.08082097480772)
--(axis cs:72173400,2.05171929788573)
--(axis cs:74225400,2.1882639696743)
--(axis cs:75935400,2.19552448797346)
--(axis cs:77987400,2.10817658325074)
--(axis cs:79697400,2.12962106697392)
--(axis cs:81407400,2.0391891645731)
--(axis cs:83459400,1.89178197337183)
--(axis cs:85169400,2.24908843680113)
--(axis cs:87221400,2.05081516778713)
--(axis cs:88931400,1.99935994780306)
--(axis cs:90983400,1.88775786372703)
--(axis cs:90983400,2.18459309881988)
--(axis cs:90983400,2.18459309881988)
--(axis cs:88931400,2.46450546768346)
--(axis cs:87221400,2.31278758546842)
--(axis cs:85169400,2.68860711828499)
--(axis cs:83459400,2.2873326520098)
--(axis cs:81407400,2.30953031775964)
--(axis cs:79697400,2.37979812379558)
--(axis cs:77987400,2.50335562463652)
--(axis cs:75935400,2.36321036756021)
--(axis cs:74225400,2.5359742982453)
--(axis cs:72173400,2.25646980719707)
--(axis cs:70463400,2.33528694190062)
--(axis cs:68411400,2.23509122902497)
--(axis cs:66701400,2.31523817231518)
--(axis cs:64991400,2.34638120155154)
--(axis cs:62939400,2.6218833126113)
--(axis cs:61229400,2.4801106012025)
--(axis cs:59177400,2.65405439250749)
--(axis cs:57467400,2.81398501617625)
--(axis cs:55415400,2.31697394445123)
--(axis cs:53705400,2.76849124148881)
--(axis cs:51995400,2.60642600804576)
--(axis cs:49943400,2.63842030126081)
--(axis cs:48233400,2.395192500322)
--(axis cs:46181400,2.60892067148363)
--(axis cs:44471400,2.42412547180807)
--(axis cs:42419400,2.55851425174941)
--(axis cs:40709400,2.57690829055238)
--(axis cs:38999400,2.37202047183003)
--(axis cs:36947400,2.70853666543045)
--(axis cs:35237400,2.55355488331126)
--(axis cs:33185400,2.83226412090539)
--(axis cs:31475400,2.90987900937477)
--(axis cs:29423400,2.85451969026171)
--(axis cs:27713400,2.69758767468824)
--(axis cs:26003400,2.67242930378012)
--(axis cs:23951400,2.95412697921868)
--(axis cs:22241400,2.6491297465709)
--(axis cs:20189400,2.9574081865113)
--(axis cs:18479400,2.99437823804845)
--(axis cs:16427400,2.97324771443433)
--(axis cs:14717400,3.33547422693575)
--(axis cs:13007400,3.50292356339308)
--(axis cs:10955400,3.71984427501756)
--(axis cs:9245400,3.90635149476573)
--(axis cs:7193400,4.30961407425605)
--(axis cs:5483400,4.34873773682193)
--(axis cs:3431400,5.81415198277682)
--(axis cs:1721400,7.30347973480821)
--(axis cs:11400,9.58470368385315)
--cycle;

\addplot [line width=\linewidthothers, \whole_color, mark=*, mark size=0, mark options={solid}]
table {%
11400 9.27138161659241
1721400 7.12394631654024
3431400 5.34233489225153
5483400 4.10472500135438
7193400 4.1501234869167
9245400 3.635581223159
10955400 3.39022885626475
13007400 3.22731510038542
14717400 3.15651519749016
16427400 2.8578856194092
18479400 2.83372846597622
20189400 2.81262162758851
22241400 2.41345677689867
23951400 2.72726451360719
26003400 2.49571905894407
27713400 2.53680575137303
29423400 2.67213076041246
31475400 2.7374260027265
33185400 2.59228257058897
35237400 2.36759056622877
36947400 2.34830570633738
38999400 2.2631891586796
40709400 2.46789885243533
42419400 2.41701365582003
44471400 2.36522826750824
46181400 2.4365711033408
48233400 2.27422412643964
49943400 2.50031661301646
51995400 2.32554819910187
53705400 2.5767046151029
55415400 2.14771122907549
57467400 2.53239396301311
59177400 2.4197974221267
61229400 2.27319822636313
62939400 2.49164800515577
64991400 2.05594510861485
66701400 2.23297184119841
68411400 2.0798224179001
70463400 2.19256508653126
72173400 2.1410674670377
74225400 2.37320676298129
75935400 2.26350793427071
77987400 2.24194863061497
79697400 2.28449931553316
81407400 2.13199078575261
83459400 2.14234636940286
85169400 2.47079075714491
87221400 2.13954315014686
88931400 2.24932943912179
90983400 2.0743061902788
};

\path [draw=C2, fill=C2, opacity=0.2]
(axis cs:121600,55.7444901315789)
--(axis cs:121600,52.4407894736842)
--(axis cs:3283200,18.0427631578947)
--(axis cs:6444800,5.35855263157895)
--(axis cs:9606400,4.36513157894737)
--(axis cs:12889600,3.28618421052632)
--(axis cs:16051200,2.15131578947368)
--(axis cs:19212800,2.03947368421053)
--(axis cs:22374400,1.86184210526316)
--(axis cs:25657600,3.1436677631579)
--(axis cs:28819200,1.94736842105263)
--(axis cs:31980800,2.10855263157895)
--(axis cs:35142400,2.76644736842105)
--(axis cs:38425600,2.5)
--(axis cs:41587200,2.68421052631579)
--(axis cs:44748800,2.33223684210526)
--(axis cs:47910400,2.28939144736842)
--(axis cs:51193600,1.98684210526316)
--(axis cs:54355200,1.70394736842105)
--(axis cs:57516800,2.67763157894737)
--(axis cs:60800000,2.55921052631579)
--(axis cs:60800000,4.63815789473684)
--(axis cs:60800000,4.63815789473684)
--(axis cs:57516800,4.06578947368421)
--(axis cs:54355200,4.61184210526316)
--(axis cs:51193600,2.83560855263158)
--(axis cs:47910400,3.64473684210526)
--(axis cs:44748800,4)
--(axis cs:41587200,5.28947368421053)
--(axis cs:38425600,4.22368421052632)
--(axis cs:35142400,4.61184210526316)
--(axis cs:31980800,3.32894736842105)
--(axis cs:28819200,2.71710526315789)
--(axis cs:25657600,4.48026315789474)
--(axis cs:22374400,3.83881578947368)
--(axis cs:19212800,3.59325657894735)
--(axis cs:16051200,2.81907894736842)
--(axis cs:12889600,5.70394736842105)
--(axis cs:9606400,6.13157894736842)
--(axis cs:6444800,7.90460526315789)
--(axis cs:3283200,23.2861842105263)
--(axis cs:121600,55.7444901315789)
--cycle;

\addplot [line width=\linewidthothers, C2, mark=*, mark size=0, mark options={solid}]
table {%
121600 53.3651315789474
3283200 19.3717105263158
6444800 6.25986842105263
9606400 4.95394736842105
12889600 4.90460526315789
16051200 2.54934210526316
19212800 2.70394736842105
22374400 2.62171052631579
25657600 3.92105263157895
28819200 2.28289473684211
31980800 2.50986842105263
35142400 3.46052631578947
38425600 3.45065789473684
41587200 3.65460526315789
44748800 2.79605263157895
47910400 2.98684210526316
51193600 2.32236842105263
54355200 2.81907894736842
57516800 3.43092105263158
60800000 3.49342105263158
};

\path [draw=C4, fill=C4, opacity=0.2]
(axis cs:10400,92.99)
--(axis cs:10400,70.575)
--(axis cs:416400,54.35)
--(axis cs:833200,40.4353125)
--(axis cs:1240800,38.0875)
--(axis cs:1645200,49.5125)
--(axis cs:2037200,75.125)
--(axis cs:2449600,70.4625)
--(axis cs:2872400,70.3125)
--(axis cs:3272000,67.5)
--(axis cs:3672400,47.5053125)
--(axis cs:4081200,40.3)
--(axis cs:4482400,29.3)
--(axis cs:4892800,31.00875)
--(axis cs:5296400,26.75)
--(axis cs:5721200,25.0984375)
--(axis cs:6139200,23.0625)
--(axis cs:6531200,21.2625)
--(axis cs:6947200,19.4625)
--(axis cs:7351600,19.3475)
--(axis cs:7776400,19.425)
--(axis cs:7776400,29.825)
--(axis cs:7776400,29.825)
--(axis cs:7351600,26.8875)
--(axis cs:6947200,27.95)
--(axis cs:6531200,33.95)
--(axis cs:6139200,31.475)
--(axis cs:5721200,38.6125)
--(axis cs:5296400,37.3125)
--(axis cs:4892800,45.175)
--(axis cs:4482400,44.575)
--(axis cs:4081200,50.675)
--(axis cs:3672400,62.725)
--(axis cs:3272000,77.3)
--(axis cs:2872400,77.9628125)
--(axis cs:2449600,81.6)
--(axis cs:2037200,84.2875)
--(axis cs:1645200,83.25)
--(axis cs:1240800,62.875)
--(axis cs:833200,53.8125)
--(axis cs:416400,66.925)
--(axis cs:10400,92.99)
--cycle;

\addplot [line width=\linewidthothers, C4, mark=*, mark size=0, mark options={solid}]
table {%
10400 90.775
416400 59.55
833200 47.0125
1240800 50.1125
1645200 70.6625
2037200 79.8875
2449600 79.225
2872400 74.8375
3272000 72.425
3672400 57.1625
4081200 45.4
4482400 36.8125
4892800 39.775
5296400 29.125
5721200 31.0125
6139200 27.5375
6531200 28.4875
6947200 22.95
7351600 21.6625
7776400 22.55
};

\path [draw=C6, fill=C6, opacity=0.2]
(axis cs:10400,91.2)
--(axis cs:10400,38.95)
--(axis cs:397600,50.2375)
--(axis cs:802800,49.3)
--(axis cs:1204800,43.025)
--(axis cs:1610800,61.9125)
--(axis cs:1994000,55.6)
--(axis cs:2397200,64.3)
--(axis cs:2798000,61.3)
--(axis cs:3194800,70.075)
--(axis cs:3606800,49.625)
--(axis cs:3992400,41.1)
--(axis cs:4410800,41.0875)
--(axis cs:4816000,38.4)
--(axis cs:5220000,29.2475)
--(axis cs:5627200,28.125)
--(axis cs:6026800,26.375)
--(axis cs:6442000,22.75)
--(axis cs:6892400,21.0375)
--(axis cs:7280400,21.0625)
--(axis cs:7691200,20.25)
--(axis cs:7691200,30.7625)
--(axis cs:7691200,30.7625)
--(axis cs:7280400,27.5125)
--(axis cs:6892400,34.775)
--(axis cs:6442000,30.5875)
--(axis cs:6026800,32.9)
--(axis cs:5627200,40.225)
--(axis cs:5220000,40.0875)
--(axis cs:4816000,47.1625)
--(axis cs:4410800,51.7125)
--(axis cs:3992400,47.15)
--(axis cs:3606800,72.675)
--(axis cs:3194800,76.825)
--(axis cs:2798000,76.2875)
--(axis cs:2397200,80.500625)
--(axis cs:1994000,76.3625)
--(axis cs:1610800,83.475)
--(axis cs:1204800,70.55)
--(axis cs:802800,59.60125)
--(axis cs:397600,59.5625)
--(axis cs:10400,91.2)
--cycle;

\addplot [line width=\linewidthothers, C6, mark=*, mark size=0, mark options={solid}]
table {%
10400 75.4875
397600 53.2
802800 54.075
1204800 56.6125
1610800 72.575
1994000 70.125
2397200 74.7125
2798000 69.975
3194800 73.9625
3606800 61.75
3992400 44.2125
4410800 49.5875
4816000 44.8625
5220000 36.05
5627200 33.75
6026800 28.2125
6442000 26.4875
6892400 29.375
7280400 23.4875
7691200 25.3125
};

\end{axis}

\end{tikzpicture}

%% file: images/tex/multi_box_nm_dist2target.tex
%
%

\begin{tikzpicture}
\def\res_color{C0}
\def\replace_color{C1}
\def\whole_color{C3}
\def\bbrl_color{C9}
\def\ppo_color{C2}
\def\ppo_res{C5}
\def\sac_color{C4}
\def\sac_res{C6}
\def\linewidthtop{1mm}
\def\linewidthothers{0.5mm}

\begin{axis}[
legend cell align={left},
legend style={fill opacity=0.8, draw opacity=1, text opacity=1, draw=lightgray204, at={(0.03,0.03)},  anchor=north west},
tick align=outside,
tick pos=left,
x grid style={white},
scaled x ticks=false,
xticklabels={,0,0.5,1,1.5,2, 2.5},
xlabel={Environment Interactions ($\times 10^7$)},
xmajorgrids,
xmin=-1500000, xmax=\multiboxxmax,
xtick style={color=black},
y grid style={white},
ylabel={Distance to Target, IQM},
ymajorgrids,
ymin=-0.15, ymax=2.5,
ytick style={color=black},
axis background/.style={fill=plot_background},
label style={font=\large},
tick label style={font=\large},
x axis line style={draw=none},
y axis line style={draw=none},
]
\path [draw=\bbrl_color, fill=\bbrl_color, opacity=0.2]
(axis cs:19400,1.42098593711853)
--(axis cs:19400,1.40161168575287)
--(axis cs:3112700,1.1972370063886)
--(axis cs:6333000,0.980561612988822)
--(axis cs:10275100,0.691272585931529)
--(axis cs:13589000,0.5765210501566)
--(axis cs:17586100,0.501264818833866)
--(axis cs:20931700,0.434731312489742)
--(axis cs:24961300,0.419493656668551)
--(axis cs:28323300,0.379701378909507)
--(axis cs:31683900,0.350531310223443)
--(axis cs:35726000,0.341006032617393)
--(axis cs:39105500,0.323786955100572)
--(axis cs:43159900,0.294470995546964)
--(axis cs:46545200,0.290795614322534)
--(axis cs:50609500,0.267166840433344)
--(axis cs:53996700,0.254525328737948)
--(axis cs:57383400,0.293608189627895)
--(axis cs:61450800,0.234600023119847)
--(axis cs:64838500,0.205389026033463)
--(axis cs:68906600,0.225364206965723)
--(axis cs:72298600,0.262397583330399)
--(axis cs:76370300,0.20253130205025)
--(axis cs:79765700,0.215215189647233)
--(axis cs:83157500,0.209751529787973)
--(axis cs:87230500,0.199607557815116)
--(axis cs:90624600,0.198021717694521)
--(axis cs:94697100,0.216448825536115)
--(axis cs:98094600,0.201841878872791)
--(axis cs:102170100,0.161720642519728)
--(axis cs:105564600,0.160415327621809)
--(axis cs:108958800,0.159252856267547)
--(axis cs:113034000,0.19293811835921)
--(axis cs:116432700,0.168272085960534)
--(axis cs:120507600,0.163960366140007)
--(axis cs:123901200,0.151654149067291)
--(axis cs:127975600,0.163346617936715)
--(axis cs:131373700,0.196689104064356)
--(axis cs:134769900,0.18580451890035)
--(axis cs:138848400,0.139367259311785)
--(axis cs:142246700,0.189253729232859)
--(axis cs:146320600,0.13229898334117)
--(axis cs:149719000,0.155134069155854)
--(axis cs:153801300,0.165028729836049)
--(axis cs:157203500,0.158152332158684)
--(axis cs:160601700,0.156349014544188)
--(axis cs:164680900,0.155193147994611)
--(axis cs:168080600,0.134925147919196)
--(axis cs:172161200,0.141161093539897)
--(axis cs:175565900,0.156966715085488)
--(axis cs:179652200,0.165956417986884)
--(axis cs:179652200,0.21583639357543)
--(axis cs:179652200,0.21583639357543)
--(axis cs:175565900,0.216047079753802)
--(axis cs:172161200,0.214195217311159)
--(axis cs:168080600,0.186959274204201)
--(axis cs:164680900,0.193486416627292)
--(axis cs:160601700,0.203645820868849)
--(axis cs:157203500,0.214053268034528)
--(axis cs:153801300,0.249407536132396)
--(axis cs:149719000,0.230511061144508)
--(axis cs:146320600,0.182065064945215)
--(axis cs:142246700,0.228401460553824)
--(axis cs:138848400,0.208275537501855)
--(axis cs:134769900,0.246433693497741)
--(axis cs:131373700,0.26471159555001)
--(axis cs:127975600,0.213945919299848)
--(axis cs:123901200,0.22338921661166)
--(axis cs:120507600,0.238821568226504)
--(axis cs:116432700,0.215149587774538)
--(axis cs:113034000,0.251134355808073)
--(axis cs:108958800,0.243112612423742)
--(axis cs:105564600,0.230536967552843)
--(axis cs:102170100,0.220996411266718)
--(axis cs:98094600,0.282978536958506)
--(axis cs:94697100,0.264732602174597)
--(axis cs:90624600,0.257741354941898)
--(axis cs:87230500,0.249257351622799)
--(axis cs:83157500,0.269484594483673)
--(axis cs:79765700,0.299084951170888)
--(axis cs:76370300,0.271591652228931)
--(axis cs:72298600,0.309276214235037)
--(axis cs:68906600,0.273419737846402)
--(axis cs:64838500,0.266458070571342)
--(axis cs:61450800,0.287090317066598)
--(axis cs:57383400,0.323025905193305)
--(axis cs:53996700,0.302992961067998)
--(axis cs:50609500,0.298800631362779)
--(axis cs:46545200,0.334240635656806)
--(axis cs:43159900,0.361797441681751)
--(axis cs:39105500,0.378982610098101)
--(axis cs:35726000,0.38634078297066)
--(axis cs:31683900,0.393475608159638)
--(axis cs:28323300,0.436731190222863)
--(axis cs:24961300,0.487480270647661)
--(axis cs:20931700,0.476384204895614)
--(axis cs:17586100,0.552794478843995)
--(axis cs:13589000,0.632227167164146)
--(axis cs:10275100,0.747494241804816)
--(axis cs:6333000,1.03299047000473)
--(axis cs:3112700,1.24967312968802)
--(axis cs:19400,1.42098593711853)
--cycle;

\addplot [line width=\linewidthothers, \bbrl_color, mark=*, mark size=0, mark options={solid}]
table {%
19400 1.40945127606392
3112700 1.22373530082405
6333000 1.00575792958261
10275100 0.717698366468539
13589000 0.602704574423512
17586100 0.520682390620272
20931700 0.460966806364441
24961300 0.449709878612886
28323300 0.409730292429816
31683900 0.3776094350079
35726000 0.359877120055971
39105500 0.35014828209457
43159900 0.330970792315977
46545200 0.299872443261326
50609500 0.283932630807135
53996700 0.27779348092433
57383400 0.306137400555055
61450800 0.260454850064662
64838500 0.242952031906178
68906600 0.245962673592781
72298600 0.289123441377289
76370300 0.242738367688516
79765700 0.259263679317567
83157500 0.239412062566477
87230500 0.228700483694378
90624600 0.234906958139958
94697100 0.240227678786931
98094600 0.251753424960818
102170100 0.195673520258848
105564600 0.21342218157765
108958800 0.216298700616731
113034000 0.220371797176326
116432700 0.201277769310651
120507600 0.203783155173737
123901200 0.194930513371878
127975600 0.193888418077068
131373700 0.229818391905113
134769900 0.218491905534499
138848400 0.178165850454487
142246700 0.199937776487571
146320600 0.159968495490582
149719000 0.204027994310894
153801300 0.217090249456014
157203500 0.186587557111552
160601700 0.183153317217146
164680900 0.176622979180795
168080600 0.162897545542764
172161200 0.177074085839115
175565900 0.189225897352113
179652200 0.189955912850035
};

\path [draw=\replace_color, fill=\replace_color, opacity=0.2]
(axis cs:11400,0.0966473594307899)
--(axis cs:11400,0.0917117558419704)
--(axis cs:1721400,0.344121853122488)
--(axis cs:3431400,0.354825845082814)
--(axis cs:5483400,0.353931557724849)
--(axis cs:7193400,0.36604523155302)
--(axis cs:9245400,0.374726904453027)
--(axis cs:10955400,0.36859849042783)
--(axis cs:13007400,0.370409794943095)
--(axis cs:14717400,0.34597809243076)
--(axis cs:16427400,0.31957014377417)
--(axis cs:18479400,0.292798005366129)
--(axis cs:20189400,0.266861826861916)
--(axis cs:22241400,0.224573913030725)
--(axis cs:23951400,0.208273667838547)
--(axis cs:26003400,0.19347966284515)
--(axis cs:27713400,0.183979956426484)
--(axis cs:29423400,0.187049493590223)
--(axis cs:31475400,0.167419565318784)
--(axis cs:33185400,0.155673823380409)
--(axis cs:35237400,0.155035560646295)
--(axis cs:36947400,0.154206076119969)
--(axis cs:38999400,0.141788490565117)
--(axis cs:40709400,0.143889163385453)
--(axis cs:42419400,0.149302188694072)
--(axis cs:44471400,0.13712528958166)
--(axis cs:46181400,0.137497100240427)
--(axis cs:48233400,0.138773154601377)
--(axis cs:49943400,0.126599060841649)
--(axis cs:51995400,0.115638254989725)
--(axis cs:53705400,0.120997848686605)
--(axis cs:55415400,0.11511702364371)
--(axis cs:57467400,0.118756682866206)
--(axis cs:59177400,0.119624162769226)
--(axis cs:61229400,0.109490532482318)
--(axis cs:62939400,0.117270192914396)
--(axis cs:64991400,0.104058324133778)
--(axis cs:66701400,0.113387446290815)
--(axis cs:68411400,0.118055452129279)
--(axis cs:70463400,0.0984052278603943)
--(axis cs:72173400,0.105775905124897)
--(axis cs:74225400,0.0945265129694259)
--(axis cs:75935400,0.101668246216759)
--(axis cs:77987400,0.103105441917186)
--(axis cs:79697400,0.111146316316518)
--(axis cs:81407400,0.100732947819738)
--(axis cs:83459400,0.10532561337629)
--(axis cs:85169400,0.104695472662782)
--(axis cs:87221400,0.0975494160818629)
--(axis cs:88931400,0.102222038798119)
--(axis cs:90983400,0.092867100511631)
--(axis cs:90983400,0.102397630844498)
--(axis cs:90983400,0.102397630844498)
--(axis cs:88931400,0.110584919520714)
--(axis cs:87221400,0.111001539907413)
--(axis cs:85169400,0.117545775018701)
--(axis cs:83459400,0.123424146241862)
--(axis cs:81407400,0.11900336195441)
--(axis cs:79697400,0.119503472122423)
--(axis cs:77987400,0.118804949449671)
--(axis cs:75935400,0.119778607759095)
--(axis cs:74225400,0.114411291506904)
--(axis cs:72173400,0.115775900860628)
--(axis cs:70463400,0.126604773093935)
--(axis cs:68411400,0.132666196918219)
--(axis cs:66701400,0.126744966788347)
--(axis cs:64991400,0.115586622084727)
--(axis cs:62939400,0.136111005741617)
--(axis cs:61229400,0.127184809168346)
--(axis cs:59177400,0.140735116679653)
--(axis cs:57467400,0.1362145619787)
--(axis cs:55415400,0.123012298394239)
--(axis cs:53705400,0.131416990231831)
--(axis cs:51995400,0.13074183384623)
--(axis cs:49943400,0.145126355013577)
--(axis cs:48233400,0.161945557721613)
--(axis cs:46181400,0.160555343158527)
--(axis cs:44471400,0.157835684188462)
--(axis cs:42419400,0.167350727449497)
--(axis cs:40709400,0.175183589899013)
--(axis cs:38999400,0.171626665823502)
--(axis cs:36947400,0.185774979304656)
--(axis cs:35237400,0.184223034686236)
--(axis cs:33185400,0.200842158179527)
--(axis cs:31475400,0.192180762547989)
--(axis cs:29423400,0.221479140244326)
--(axis cs:27713400,0.21919561253998)
--(axis cs:26003400,0.227271856459004)
--(axis cs:23951400,0.247372778013264)
--(axis cs:22241400,0.24666626219073)
--(axis cs:20189400,0.2988261181804)
--(axis cs:18479400,0.319734581940662)
--(axis cs:16427400,0.361532249135507)
--(axis cs:14717400,0.371668142704689)
--(axis cs:13007400,0.396248749759432)
--(axis cs:10955400,0.417842774015798)
--(axis cs:9245400,0.427389840890377)
--(axis cs:7193400,0.423634894200736)
--(axis cs:5483400,0.428324052480065)
--(axis cs:3431400,0.430330482902718)
--(axis cs:1721400,0.392026411078405)
--(axis cs:11400,0.0966473594307899)
--cycle;

\addplot [line width=\linewidthothers, \replace_color, mark=*, mark size=0, mark options={solid}]
table {%
11400 0.0945014171302319
1721400 0.367588471854106
3431400 0.392282643075305
5483400 0.385118062002135
7193400 0.396959244999136
9245400 0.401812965978447
10955400 0.390956680837623
13007400 0.386447870374508
14717400 0.360953111224863
16427400 0.338897864525638
18479400 0.303324549376424
20189400 0.278931919933269
22241400 0.236413937234771
23951400 0.232816024459088
26003400 0.211328704356461
27713400 0.200590483862869
29423400 0.210645251025577
31475400 0.182503644111906
33185400 0.180484862517613
35237400 0.171129909863263
36947400 0.169697208248173
38999400 0.156462823623568
40709400 0.161248100825658
42419400 0.157679540539534
44471400 0.145703598427015
46181400 0.151895262790813
48233400 0.154802852761604
49943400 0.134448058484289
51995400 0.123335722858628
53705400 0.125835543063852
55415400 0.119350819992852
57467400 0.130169190230749
59177400 0.129109385488313
61229400 0.119407185280464
62939400 0.128618239368904
64991400 0.109335536852536
66701400 0.120670601568449
68411400 0.126862193761498
70463400 0.114525519295683
72173400 0.110787163363367
74225400 0.10214883655005
75935400 0.114151945392578
77987400 0.112305326715966
79697400 0.114993877857956
81407400 0.10414895779203
83459400 0.113636726211612
85169400 0.110118616283174
87221400 0.104219201204055
88931400 0.107676621839222
90983400 0.0965989092121809
};

\path [draw=\res_color, fill=\res_color, opacity=0.2]
(axis cs:11400,0.0970339626073837)
--(axis cs:11400,0.0923925619572401)
--(axis cs:1721400,0.0929617423098534)
--(axis cs:3431400,0.093040444939561)
--(axis cs:5483400,0.0775663404169507)
--(axis cs:7193400,0.0745434428311729)
--(axis cs:9245400,0.0703896692910457)
--(axis cs:10955400,0.0674125304492598)
--(axis cs:13007400,0.0760173990458162)
--(axis cs:14717400,0.0730694941775818)
--(axis cs:16427400,0.0652158500153422)
--(axis cs:18479400,0.0627661718386256)
--(axis cs:20189400,0.0674776947772288)
--(axis cs:22241400,0.0520261162334912)
--(axis cs:23951400,0.0602427952705047)
--(axis cs:26003400,0.055091527484495)
--(axis cs:27713400,0.0591179746114609)
--(axis cs:29423400,0.0573304346962338)
--(axis cs:31475400,0.0604579927686013)
--(axis cs:33185400,0.059113283885013)
--(axis cs:35237400,0.0554789498167607)
--(axis cs:36947400,0.0545337939829583)
--(axis cs:38999400,0.0513192426839)
--(axis cs:40709400,0.0574526069146582)
--(axis cs:42419400,0.0577809587157603)
--(axis cs:44471400,0.058800993685669)
--(axis cs:46181400,0.0540564618405561)
--(axis cs:48233400,0.0546430088407123)
--(axis cs:49943400,0.0545955334258029)
--(axis cs:51995400,0.0492110766356506)
--(axis cs:53705400,0.0559849135848593)
--(axis cs:55415400,0.0483832179745135)
--(axis cs:57467400,0.058012912340655)
--(axis cs:59177400,0.0617748441106635)
--(axis cs:61229400,0.0495293438599412)
--(axis cs:62939400,0.0548524494027217)
--(axis cs:64991400,0.0501697087958872)
--(axis cs:66701400,0.0522716243808271)
--(axis cs:68411400,0.0502830436319349)
--(axis cs:70463400,0.0498105035629586)
--(axis cs:72173400,0.0485791056879919)
--(axis cs:74225400,0.0543635848286389)
--(axis cs:75935400,0.0559153975000104)
--(axis cs:77987400,0.0502350040884965)
--(axis cs:79697400,0.0532031415578397)
--(axis cs:81407400,0.0533458871714772)
--(axis cs:83459400,0.0513302729585418)
--(axis cs:85169400,0.0551371867079848)
--(axis cs:87221400,0.0544102316553366)
--(axis cs:88931400,0.0516982066045615)
--(axis cs:90983400,0.0478010882329183)
--(axis cs:90983400,0.0565957861320242)
--(axis cs:90983400,0.0565957861320242)
--(axis cs:88931400,0.0596904120652545)
--(axis cs:87221400,0.0602255573452487)
--(axis cs:85169400,0.0587526804888061)
--(axis cs:83459400,0.0558745568537049)
--(axis cs:81407400,0.0576961401724822)
--(axis cs:79697400,0.0598847346159035)
--(axis cs:77987400,0.0559918596610796)
--(axis cs:75935400,0.06039619382016)
--(axis cs:74225400,0.0641268075890805)
--(axis cs:72173400,0.055017702235865)
--(axis cs:70463400,0.0577405111902577)
--(axis cs:68411400,0.0542579208730738)
--(axis cs:66701400,0.0597354383133552)
--(axis cs:64991400,0.0560722061503443)
--(axis cs:62939400,0.060149859264087)
--(axis cs:61229400,0.0614556888744222)
--(axis cs:59177400,0.0712625249844921)
--(axis cs:57467400,0.0679591090155912)
--(axis cs:55415400,0.053734527933385)
--(axis cs:53705400,0.0611321559262156)
--(axis cs:51995400,0.0538078950850078)
--(axis cs:49943400,0.0630684187093808)
--(axis cs:48233400,0.0584750168233827)
--(axis cs:46181400,0.0597891829186626)
--(axis cs:44471400,0.0635125383060144)
--(axis cs:42419400,0.0604299280846713)
--(axis cs:40709400,0.0654056827210493)
--(axis cs:38999400,0.0595728180789677)
--(axis cs:36947400,0.0564872898357013)
--(axis cs:35237400,0.0577608667051464)
--(axis cs:33185400,0.0678781926935981)
--(axis cs:31475400,0.0711415875436587)
--(axis cs:29423400,0.0652248911528725)
--(axis cs:27713400,0.0693141178414674)
--(axis cs:26003400,0.0609647147000694)
--(axis cs:23951400,0.0691334639955045)
--(axis cs:22241400,0.0637600907733967)
--(axis cs:20189400,0.0727266115854888)
--(axis cs:18479400,0.0712022983589306)
--(axis cs:16427400,0.0742889237076426)
--(axis cs:14717400,0.0786888976495806)
--(axis cs:13007400,0.0861173152092984)
--(axis cs:10955400,0.0751036001748145)
--(axis cs:9245400,0.0787241750212842)
--(axis cs:7193400,0.0816954477995058)
--(axis cs:5483400,0.089936964645176)
--(axis cs:3431400,0.0997416254822383)
--(axis cs:1721400,0.0953094972646795)
--(axis cs:11400,0.0970339626073837)
--cycle;

\addplot [line width=\linewidthtop, \res_color, mark=*, mark size=0, mark options={solid}]
table {%
11400 0.0950231980532408
1721400 0.0944614995387383
3431400 0.0972083468877827
5483400 0.0839399606166182
7193400 0.0786447824661094
9245400 0.0745492861885317
10955400 0.0717574220669398
13007400 0.0801132077761595
14717400 0.0765365235262445
16427400 0.0707331815574687
18479400 0.0659538860627024
20189400 0.0703217106797276
22241400 0.0562754913005161
23951400 0.0660254935022117
26003400 0.0576366497378234
27713400 0.0650628534539051
29423400 0.0623710107535599
31475400 0.0646139202777583
33185400 0.0634168942611946
35237400 0.05650128232658
36947400 0.0549151238484846
38999400 0.0549420691207018
40709400 0.0612033736629155
42419400 0.0584126483131143
44471400 0.0606307850604781
46181400 0.056341191670759
48233400 0.0573908579552539
49943400 0.0589534533146131
51995400 0.0520615379951258
53705400 0.0577534149929934
55415400 0.0506753600101871
57467400 0.062077055932027
59177400 0.0648899250655476
61229400 0.0574816177885315
62939400 0.0565622213597022
64991400 0.0535292121111002
66701400 0.0566363051783993
68411400 0.0517911410836158
70463400 0.0547231579207279
72173400 0.0513248773770739
74225400 0.0598120544159971
75935400 0.0585210363119315
77987400 0.0542489881706274
79697400 0.0561231792584666
81407400 0.0557036724216099
83459400 0.0532158345019488
85169400 0.0560020876996301
87221400 0.0559479929660897
88931400 0.0535868486465448
90983400 0.0521769011618897
};

\path [draw=\whole_color, fill=\whole_color, opacity=0.2]
(axis cs:11400,0.094001017510891)
--(axis cs:11400,0.0876605380326509)
--(axis cs:1721400,0.0777549350750633)
--(axis cs:3431400,0.0769375741037948)
--(axis cs:5483400,0.0681751752854325)
--(axis cs:7193400,0.0733666786710545)
--(axis cs:9245400,0.0663278802975978)
--(axis cs:10955400,0.0651390728381921)
--(axis cs:13007400,0.0708376089252864)
--(axis cs:14717400,0.0681976701728818)
--(axis cs:16427400,0.0602484308836577)
--(axis cs:18479400,0.0562346247813027)
--(axis cs:20189400,0.0593817821882431)
--(axis cs:22241400,0.0511858209665815)
--(axis cs:23951400,0.0541555021962748)
--(axis cs:26003400,0.0529392807102242)
--(axis cs:27713400,0.0580075407226238)
--(axis cs:29423400,0.0553909394891897)
--(axis cs:31475400,0.0564057971897367)
--(axis cs:33185400,0.0568261085686025)
--(axis cs:35237400,0.0529132185417121)
--(axis cs:36947400,0.0548754503622821)
--(axis cs:38999400,0.0505802417168108)
--(axis cs:40709400,0.0553815692286098)
--(axis cs:42419400,0.0570076002051506)
--(axis cs:44471400,0.0549987703753058)
--(axis cs:46181400,0.0523621853093259)
--(axis cs:48233400,0.0538166247318674)
--(axis cs:49943400,0.0593402078780883)
--(axis cs:51995400,0.0524610948721035)
--(axis cs:53705400,0.0570308568896535)
--(axis cs:55415400,0.0514541575228697)
--(axis cs:57467400,0.0602566853623183)
--(axis cs:59177400,0.0629118495264744)
--(axis cs:61229400,0.0538815365197028)
--(axis cs:62939400,0.056772511773553)
--(axis cs:64991400,0.0487137497644417)
--(axis cs:66701400,0.0546514831762818)
--(axis cs:68411400,0.0520060088570109)
--(axis cs:70463400,0.0526525601808872)
--(axis cs:72173400,0.0513894274504648)
--(axis cs:74225400,0.0558258896909393)
--(axis cs:75935400,0.0595450364552065)
--(axis cs:77987400,0.0557367049822681)
--(axis cs:79697400,0.0532089615192543)
--(axis cs:81407400,0.0537020462789435)
--(axis cs:83459400,0.057051246718243)
--(axis cs:85169400,0.0600492815624171)
--(axis cs:87221400,0.0543451737696756)
--(axis cs:88931400,0.0523811580305584)
--(axis cs:90983400,0.0516649848696276)
--(axis cs:90983400,0.0595140422855504)
--(axis cs:90983400,0.0595140422855504)
--(axis cs:88931400,0.0625202296447975)
--(axis cs:87221400,0.0634527102595358)
--(axis cs:85169400,0.0686204970455846)
--(axis cs:83459400,0.0617127657721525)
--(axis cs:81407400,0.061395807567287)
--(axis cs:79697400,0.059403042582129)
--(axis cs:77987400,0.0635653473649669)
--(axis cs:75935400,0.0624084031314166)
--(axis cs:74225400,0.0609822086113584)
--(axis cs:72173400,0.0572490853093435)
--(axis cs:70463400,0.0561981805479708)
--(axis cs:68411400,0.0603994865340719)
--(axis cs:66701400,0.0592947509974428)
--(axis cs:64991400,0.05790703470437)
--(axis cs:62939400,0.0634446410762196)
--(axis cs:61229400,0.0606538689543915)
--(axis cs:59177400,0.0694633432142283)
--(axis cs:57467400,0.0681178951447915)
--(axis cs:55415400,0.0561669978527669)
--(axis cs:53705400,0.0657526959522909)
--(axis cs:51995400,0.0568438214313961)
--(axis cs:49943400,0.0645307658638622)
--(axis cs:48233400,0.0596078489684575)
--(axis cs:46181400,0.0608001329153553)
--(axis cs:44471400,0.0619349498358851)
--(axis cs:42419400,0.0618002528386605)
--(axis cs:40709400,0.0604653386445138)
--(axis cs:38999400,0.0536732015173173)
--(axis cs:36947400,0.0621394037344134)
--(axis cs:35237400,0.0595063450333714)
--(axis cs:33185400,0.0660118250318186)
--(axis cs:31475400,0.0636095897400273)
--(axis cs:29423400,0.0597154988879471)
--(axis cs:27713400,0.0636671712405328)
--(axis cs:26003400,0.0591942113771827)
--(axis cs:23951400,0.0671477102702489)
--(axis cs:22241400,0.0589743734193009)
--(axis cs:20189400,0.0684659633915631)
--(axis cs:18479400,0.0699970754487398)
--(axis cs:16427400,0.0691181478265909)
--(axis cs:14717400,0.073929248955261)
--(axis cs:13007400,0.0796489780126178)
--(axis cs:10955400,0.0745352060649023)
--(axis cs:9245400,0.0748737617160902)
--(axis cs:7193400,0.0783818054101113)
--(axis cs:5483400,0.0739348104705186)
--(axis cs:3431400,0.0826560448149394)
--(axis cs:1721400,0.0847112917690538)
--(axis cs:11400,0.094001017510891)
--cycle;

\addplot [line width=\linewidthothers, \whole_color, mark=*, mark size=0, mark options={solid}]
table {%
11400 0.0911441240459681
1721400 0.0797384614706971
3431400 0.0796103098673484
5483400 0.0715373532376589
7193400 0.0759447903742467
9245400 0.0710715692587375
10955400 0.0674305738160062
13007400 0.0758046026512373
14717400 0.0705698265753299
16427400 0.0651186781764455
18479400 0.0632648490970899
20189400 0.0639887806903218
22241400 0.0561061791587972
23951400 0.0627549325993393
26003400 0.0572911854158342
27713400 0.0615201101052666
29423400 0.0576364755015244
31475400 0.0600070092355927
33185400 0.0609346730751501
35237400 0.055150122159912
36947400 0.0569580046945872
38999400 0.051440830623034
40709400 0.0588613137307555
42419400 0.0596406960558812
44471400 0.0576737766351969
46181400 0.0559525019747378
48233400 0.0571615334790643
49943400 0.0620334761670641
51995400 0.0547764461191428
53705400 0.0609955527025026
55415400 0.0542117464969935
57467400 0.0643843962513747
59177400 0.0665438827211874
61229400 0.0561760266110358
62939400 0.0605263336426637
64991400 0.0531454498683627
66701400 0.056729811785877
68411400 0.0565130615161312
70463400 0.053574914459018
72173400 0.0541365508213648
74225400 0.058118685562219
75935400 0.0611071850061842
77987400 0.0605772686300359
79697400 0.0557609475692509
81407400 0.0577082677874585
83459400 0.058484637558523
85169400 0.0630232763192309
87221400 0.0589087256458947
88931400 0.0580957732312952
90983400 0.0548186117249188
};

\path [draw=C2, fill=C2, opacity=0.2]
(axis cs:121600,3.31307426291009)
--(axis cs:121600,3.23396869771502)
--(axis cs:3283200,0.66145536273881)
--(axis cs:6444800,0.482279917258168)
--(axis cs:9606400,0.414758343348431)
--(axis cs:12889600,0.374556941264256)
--(axis cs:16051200,0.347080080321684)
--(axis cs:19212800,0.31849794457431)
--(axis cs:22374400,0.304027810025748)
--(axis cs:25657600,0.317124735533014)
--(axis cs:28819200,0.300688160607683)
--(axis cs:31980800,0.294886948442324)
--(axis cs:35142400,0.295617171726979)
--(axis cs:38425600,0.291413868079511)
--(axis cs:41587200,0.291681883562047)
--(axis cs:44748800,0.274694893426434)
--(axis cs:47910400,0.283764581659308)
--(axis cs:51193600,0.28338226400085)
--(axis cs:54355200,0.284617510960467)
--(axis cs:57516800,0.275316496863153)
--(axis cs:60800000,0.287685871435175)
--(axis cs:60800000,0.304973137783661)
--(axis cs:60800000,0.304973137783661)
--(axis cs:57516800,0.303448532123261)
--(axis cs:54355200,0.310474546079669)
--(axis cs:51193600,0.304076275690927)
--(axis cs:47910400,0.309423927324723)
--(axis cs:44748800,0.298933180699858)
--(axis cs:41587200,0.308146011862599)
--(axis cs:38425600,0.314266642725254)
--(axis cs:35142400,0.332880162814942)
--(axis cs:31980800,0.312518407159956)
--(axis cs:28819200,0.332709080660318)
--(axis cs:25657600,0.338196980196453)
--(axis cs:22374400,0.346429545652563)
--(axis cs:19212800,0.355909768061683)
--(axis cs:16051200,0.367748103772748)
--(axis cs:12889600,0.414792662857922)
--(axis cs:9606400,0.448583640475658)
--(axis cs:6444800,0.555575150242222)
--(axis cs:3283200,0.731670889489024)
--(axis cs:121600,3.31307426291009)
--cycle;

\addplot [line width=\linewidthothers, C2, mark=*, mark size=0, mark options={solid}]
table {%
121600 3.27442622118208
3283200 0.70999060741039
6444800 0.512174919127436
9606400 0.43662764620366
12889600 0.396384969502455
16051200 0.357210015599047
19212800 0.328665601441399
22374400 0.329283126990835
25657600 0.329533902681455
28819200 0.316698456272075
31980800 0.303824565732707
35142400 0.314684396615084
38425600 0.304042064578755
41587200 0.296887083808321
44748800 0.286468793638824
47910400 0.293337102834736
51193600 0.292186785404168
54355200 0.294972912775599
57516800 0.287992990943314
60800000 0.296036680290817
};

\path [draw=C5, fill=C5, opacity=0.2]
(axis cs:121600,0.494902496879497)
--(axis cs:121600,0.35032233865722)
--(axis cs:3283200,0.66245133071644)
--(axis cs:6444800,0.48374816567222)
--(axis cs:9606400,0.342530835485403)
--(axis cs:12889600,0.313886673610771)
--(axis cs:16051200,0.337384870681799)
--(axis cs:19212800,0.2845215353565)
--(axis cs:22374400,0.267909481178638)
--(axis cs:25657600,0.28005489171358)
--(axis cs:28819200,0.26571402189943)
--(axis cs:31980800,0.263566739738541)
--(axis cs:35142400,0.284590094706918)
--(axis cs:38425600,0.265368705220359)
--(axis cs:41587200,0.267655049929992)
--(axis cs:44748800,0.263281784220218)
--(axis cs:47910400,0.261770171729913)
--(axis cs:51193600,0.261380853573787)
--(axis cs:54355200,0.276631346190398)
--(axis cs:57516800,0.272179565559027)
--(axis cs:60800000,0.270207418683054)
--(axis cs:60800000,0.285919823707169)
--(axis cs:60800000,0.285919823707169)
--(axis cs:57516800,0.296526424061044)
--(axis cs:54355200,0.299141343776833)
--(axis cs:51193600,0.293831947265776)
--(axis cs:47910400,0.303811375596621)
--(axis cs:44748800,0.300351210575764)
--(axis cs:41587200,0.295302195360779)
--(axis cs:38425600,0.305275729319835)
--(axis cs:35142400,0.337824650151424)
--(axis cs:31980800,0.285467590359894)
--(axis cs:28819200,0.327017826705975)
--(axis cs:25657600,0.322751127880173)
--(axis cs:22374400,0.341953649934239)
--(axis cs:19212800,0.315768804165074)
--(axis cs:16051200,0.37111457969735)
--(axis cs:12889600,0.333570639604248)
--(axis cs:9606400,0.380764560848978)
--(axis cs:6444800,0.52320229491741)
--(axis cs:3283200,0.7155459228793)
--(axis cs:121600,0.494902496879497)
--cycle;

\addplot [line width=\linewidthothers, C5, mark=*, mark size=0, mark options={solid}]
table {%
121600 0.399309330014924
3283200 0.682469094125854
6444800 0.508860700202587
9606400 0.359533659292621
12889600 0.325283000353864
16051200 0.350903790913079
19212800 0.298989876406741
22374400 0.307905321801384
25657600 0.303462875466831
28819200 0.294194955834331
31980800 0.274121045489041
35142400 0.30858785024898
38425600 0.28021168316489
41587200 0.281862560264559
44748800 0.280127597820282
47910400 0.272776986666546
51193600 0.273332389743201
54355200 0.288126718525547
57516800 0.284912095304057
60800000 0.276441281949897
};

\path [draw=C4, fill=C4, opacity=0.2]
(axis cs:10400,6.09317413611061)
--(axis cs:10400,5.04598428344287)
--(axis cs:416400,4.45749812533818)
--(axis cs:833200,4.61534937773269)
--(axis cs:1240800,4.54278657318234)
--(axis cs:1645200,4.07073121699001)
--(axis cs:2037200,3.5481438629167)
--(axis cs:2449600,3.14091205155517)
--(axis cs:2872400,2.48493992604083)
--(axis cs:3272000,1.82137915824142)
--(axis cs:3672400,1.44961009043546)
--(axis cs:4081200,1.46840042208528)
--(axis cs:4482400,1.35369491024889)
--(axis cs:4892800,1.34382649016765)
--(axis cs:5296400,1.2821606696049)
--(axis cs:5721200,1.29134840850599)
--(axis cs:6139200,1.30986035817158)
--(axis cs:6531200,1.37680881293885)
--(axis cs:6947200,1.32572515665387)
--(axis cs:7351600,1.35765636485967)
--(axis cs:7776400,1.29992857380452)
--(axis cs:7776400,1.40837291424882)
--(axis cs:7776400,1.40837291424882)
--(axis cs:7351600,1.55133218716266)
--(axis cs:6947200,1.47048213264634)
--(axis cs:6531200,1.56601632833348)
--(axis cs:6139200,1.55467305405391)
--(axis cs:5721200,1.6604597081159)
--(axis cs:5296400,1.56927729755028)
--(axis cs:4892800,1.57767619432465)
--(axis cs:4482400,1.58667814079756)
--(axis cs:4081200,1.70264871335517)
--(axis cs:3672400,1.84925070027784)
--(axis cs:3272000,2.09794626969868)
--(axis cs:2872400,2.87356395632076)
--(axis cs:2449600,3.50727986215698)
--(axis cs:2037200,4.12549743280681)
--(axis cs:1645200,4.40578639464815)
--(axis cs:1240800,4.94535499936277)
--(axis cs:833200,4.89760080561753)
--(axis cs:416400,4.91809817109449)
--(axis cs:10400,6.09317413611061)
--cycle;

\addplot [line width=\linewidthothers, C4, mark=*, mark size=0, mark options={solid}]
table {%
10400 5.741683609985
416400 4.67048104998244
833200 4.68353532675207
1240800 4.80632551935158
1645200 4.2752825361872
2037200 3.85311466648744
2449600 3.36786842446484
2872400 2.64527749777845
3272000 2.00997893610717
3672400 1.63698850187763
4081200 1.57758761701015
4482400 1.46437962265797
4892800 1.42450414717302
5296400 1.40234352895501
5721200 1.51380987166798
6139200 1.45708392294034
6531200 1.50784001439342
6947200 1.37912786128799
7351600 1.43048725821421
7776400 1.35426728747449
};

\path [draw=C6, fill=C6, opacity=0.2]
(axis cs:10400,5.06275907770196)
--(axis cs:10400,3.66919733102453)
--(axis cs:397600,4.06349652869322)
--(axis cs:802800,4.02468639904787)
--(axis cs:1204800,4.12751051217924)
--(axis cs:1610800,4.03771398531675)
--(axis cs:1994000,3.66234780797552)
--(axis cs:2397200,2.94602021230361)
--(axis cs:2798000,2.16776041192557)
--(axis cs:3194800,1.51034222640256)
--(axis cs:3606800,1.41839689502085)
--(axis cs:3992400,1.2789029444176)
--(axis cs:4410800,1.24053817395277)
--(axis cs:4816000,1.20499120867603)
--(axis cs:5220000,1.2572551430639)
--(axis cs:5627200,1.28914150127903)
--(axis cs:6026800,1.22243037845859)
--(axis cs:6442000,1.23486487870281)
--(axis cs:6892400,1.31144829218535)
--(axis cs:7280400,1.25077253881908)
--(axis cs:7691200,1.38847380656542)
--(axis cs:7691200,1.55091497039197)
--(axis cs:7691200,1.55091497039197)
--(axis cs:7280400,1.49686879981897)
--(axis cs:6892400,1.47149316044855)
--(axis cs:6442000,1.41848309554638)
--(axis cs:6026800,1.43752189369267)
--(axis cs:5627200,1.52375684622975)
--(axis cs:5220000,1.40676135018684)
--(axis cs:4816000,1.40545118604615)
--(axis cs:4410800,1.50753486198572)
--(axis cs:3992400,1.56919502787327)
--(axis cs:3606800,1.63225887681691)
--(axis cs:3194800,2.03088763430676)
--(axis cs:2798000,2.4828926171837)
--(axis cs:2397200,3.26559803439035)
--(axis cs:1994000,4.05460030961974)
--(axis cs:1610800,4.58329380490534)
--(axis cs:1204800,4.80083242771044)
--(axis cs:802800,4.40281390446511)
--(axis cs:397600,4.52513838694727)
--(axis cs:10400,5.06275907770196)
--cycle;

\addplot [line width=\linewidthothers, C6, mark=*, mark size=0, mark options={solid}]
table {%
10400 4.1861025467552
397600 4.28771186080092
802800 4.27434443991353
1204800 4.40834948267909
1610800 4.28004467162464
1994000 3.86686661592239
2397200 3.10067240109769
2798000 2.33184980955623
3194800 1.83744853286155
3606800 1.5535740720615
3992400 1.44409429762117
4410800 1.41354422449375
4816000 1.30438808128196
5220000 1.32139099712921
5627200 1.37520730282147
6026800 1.29573658224734
6442000 1.31837766530555
6892400 1.43141393315238
7280400 1.40014160453042
7691200 1.48099850411189
};

\end{axis}

\end{tikzpicture}

%% file: images/tex/dual_arm_nm_success.tex
%
%

\begin{tikzpicture}
\def\res_color{C0}
\def\replace_color{C1}
\def\whole_color{C3}
\def\bbrl_color{C9}
\def\ppo_color{C2}
\def\sac_color{C4}
\def\linewidthtop{1mm}
\def\linewidthothers{0.5mm}

\begin{axis}[
legend cell align={left},
legend style={fill opacity=0.8, draw opacity=1, text opacity=1, draw=lightgray204, at={(0.03,0.03)},  anchor=north west},
tick align=outside,
tick pos=left,
x grid style={white},
scaled x ticks=false,
xticklabels={,0,1,2,3,4},
xlabel={Environment Interactions ($\times 10^7$)},
xmajorgrids,
xmin=-1500000, xmax=41000000.05,
xtick style={color=black},
y grid style={white},
ylabel={Success Rate, Non-Markovian, IQM},
ymajorgrids,
ymin=-0.05, ymax=0.9,
ytick style={color=black},
axis background/.style={fill=plot_background},
label style={font=\large},
tick label style={font=\large},
x axis line style={draw=none},
y axis line style={draw=none},
]
\path [draw=\bbrl_color, fill=\bbrl_color, opacity=0.2]
(axis cs:19300,0)
--(axis cs:19300,0)
--(axis cs:2430700,0.000205592106794938)
--(axis cs:4960400,0.00422009922890539)
--(axis cs:7548700,0.00340110372155777)
--(axis cs:10161900,0.00796599129971298)
--(axis cs:12791200,0.00873365982462748)
--(axis cs:15436300,0.0103450894482359)
--(axis cs:18095500,0.015552987126798)
--(axis cs:20761400,0.0196311679686167)
--(axis cs:23438300,0.02080490111998)
--(axis cs:26121000,0.0245721010017004)
--(axis cs:28808500,0.0208994561466741)
--(axis cs:31499200,0.0238438479009914)
--(axis cs:34190600,0.033545612289028)
--(axis cs:36886000,0.03132010121705)
--(axis cs:39583800,0.0352207285524472)
--(axis cs:42290900,0.0432000843525627)
--(axis cs:45671800,0.0547930808021018)
--(axis cs:48377100,0.0427884890926663)
--(axis cs:51088600,0.0644121687550673)
--(axis cs:53795700,0.0579730456506418)
--(axis cs:56501800,0.0683802526252586)
--(axis cs:59206600,0.0918379778105777)
--(axis cs:61911900,0.0863695275736531)
--(axis cs:64618400,0.0800866367461497)
--(axis cs:67325200,0.0787675590587404)
--(axis cs:70036700,0.0922077657543093)
--(axis cs:72746500,0.0907984852464627)
--(axis cs:75455300,0.0907902632097485)
--(axis cs:78164400,0.0903085867692849)
--(axis cs:80877400,0.0932432649737067)
--(axis cs:83590700,0.0851398860136964)
--(axis cs:86305800,0.0992643296229019)
--(axis cs:89697300,0.104950877886664)
--(axis cs:92415500,0.103885856800742)
--(axis cs:95129700,0.100151155752562)
--(axis cs:97842900,0.110362840260895)
--(axis cs:100559400,0.11116232403601)
--(axis cs:103278000,0.125077549443603)
--(axis cs:105994800,0.131552246422357)
--(axis cs:108711500,0.120715916109514)
--(axis cs:111431500,0.120604995125147)
--(axis cs:114150500,0.121223435419718)
--(axis cs:116864900,0.135731282473943)
--(axis cs:119579300,0.132939277729907)
--(axis cs:122294200,0.112081617361855)
--(axis cs:125012000,0.133598440924135)
--(axis cs:127726500,0.12906729725907)
--(axis cs:130444800,0.119718741597397)
--(axis cs:133843100,0.136073776273936)
--(axis cs:133843100,0.158585389824829)
--(axis cs:133843100,0.158585389824829)
--(axis cs:130444800,0.146504541005817)
--(axis cs:127726500,0.15545150135762)
--(axis cs:125012000,0.15259684759437)
--(axis cs:122294200,0.138176693695439)
--(axis cs:119579300,0.158730739245727)
--(axis cs:116864900,0.161402336993444)
--(axis cs:114150500,0.141651995944197)
--(axis cs:111431500,0.14159636970723)
--(axis cs:108711500,0.138247815797985)
--(axis cs:105994800,0.153504458040957)
--(axis cs:103278000,0.138259719939278)
--(axis cs:100559400,0.13451877805706)
--(axis cs:97842900,0.130096504289497)
--(axis cs:95129700,0.115459723967525)
--(axis cs:92415500,0.122084117868806)
--(axis cs:89697300,0.119462579357514)
--(axis cs:86305800,0.116378849260777)
--(axis cs:83590700,0.102012736404147)
--(axis cs:80877400,0.113793261165846)
--(axis cs:78164400,0.104306070635424)
--(axis cs:75455300,0.104108694495899)
--(axis cs:72746500,0.100927181060977)
--(axis cs:70036700,0.0954187838754898)
--(axis cs:67325200,0.0903466745180843)
--(axis cs:64618400,0.0935302165803212)
--(axis cs:61911900,0.0941654188104995)
--(axis cs:59206600,0.10267872189185)
--(axis cs:56501800,0.0738127685745149)
--(axis cs:53795700,0.0776044996345954)
--(axis cs:51088600,0.0772858397284908)
--(axis cs:48377100,0.054378570430085)
--(axis cs:45671800,0.0633257274181721)
--(axis cs:42290900,0.0545270489290482)
--(axis cs:39583800,0.0429721608156205)
--(axis cs:36886000,0.0513564633677194)
--(axis cs:34190600,0.0444828327405731)
--(axis cs:31499200,0.0344572138840579)
--(axis cs:28808500,0.0256191006430312)
--(axis cs:26121000,0.0310396660234836)
--(axis cs:23438300,0.030241585917917)
--(axis cs:20761400,0.0249022565898767)
--(axis cs:18095500,0.0185659422961631)
--(axis cs:15436300,0.0151844591885551)
--(axis cs:12791200,0.0158235086164611)
--(axis cs:10161900,0.0118505328408141)
--(axis cs:7548700,0.00752900752252117)
--(axis cs:4960400,0.0084999483451611)
--(axis cs:2430700,0.00246710528153926)
--(axis cs:19300,0)
--cycle;

\addplot [line width=\linewidthothers, \bbrl_color, mark=*, mark size=0, mark options={solid}]
table {%
19300 0
2430700 0.00113075658737216
4960400 0.00592362243151001
7548700 0.0060195921244599
10161900 0.00962490782616143
12791200 0.0117429304899179
15436300 0.014322407051043
18095500 0.0175614011204368
20761400 0.0226630765548036
23438300 0.0247942625660328
26121000 0.0264379590493112
28808500 0.023650942652022
31499200 0.026630336670822
34190600 0.0374502323931289
36886000 0.0415537893603596
39583800 0.0392137575103588
42290900 0.0486423192345342
45671800 0.0570112661396692
48377100 0.0493074211714646
51088600 0.0687281186472442
53795700 0.0670631063132276
56501800 0.0718619117374213
59206600 0.0951992648351922
61911900 0.0899971325976125
64618400 0.0885182238035026
67325200 0.0814196499476443
70036700 0.0934919793110546
72746500 0.0959436538478918
75455300 0.0973236103955044
78164400 0.0938764641142273
80877400 0.100717279987218
83590700 0.089739657833908
86305800 0.104910017862344
89697300 0.111050380540037
92415500 0.114271405395635
95129700 0.106923435635912
97842900 0.118159302488549
100559400 0.123393755572821
103278000 0.130186340397296
105994800 0.142819858047402
108711500 0.130065294483809
111431500 0.132288733022196
114150500 0.131811610226837
116864900 0.150030046250796
119579300 0.149345691335462
122294200 0.130890438741834
125012000 0.14260381501822
127726500 0.140718976838504
130444800 0.132500310119319
133843100 0.146529532093813
};

\path [draw=\replace_color, fill=\replace_color, opacity=0.2]
(axis cs:11400,0.338815778493881)
--(axis cs:11400,0.320723682641983)
--(axis cs:1379400,0.212068253895268)
--(axis cs:2747400,0.233854595193407)
--(axis cs:4115400,0.227670772278543)
--(axis cs:5483400,0.248858062383988)
--(axis cs:6851400,0.247346518864799)
--(axis cs:8219400,0.245368276698523)
--(axis cs:9587400,0.239798564935036)
--(axis cs:10955400,0.222041248723056)
--(axis cs:12323400,0.225378207617928)
--(axis cs:13691400,0.238421444434762)
--(axis cs:15059400,0.21813573965775)
--(axis cs:16427400,0.213253369880143)
--(axis cs:17795400,0.205381449345128)
--(axis cs:19163400,0.205934449224218)
--(axis cs:20531400,0.200074256567323)
--(axis cs:21899400,0.222111387112743)
--(axis cs:23609400,0.240561709062045)
--(axis cs:24977400,0.221722897560218)
--(axis cs:26345400,0.219793255807774)
--(axis cs:27713400,0.226219373151578)
--(axis cs:29081400,0.222194186009506)
--(axis cs:30449400,0.241174932652588)
--(axis cs:31817400,0.24583806770596)
--(axis cs:33185400,0.223024372465841)
--(axis cs:34553400,0.234861865322965)
--(axis cs:35921400,0.232852040376715)
--(axis cs:37289400,0.225341710997826)
--(axis cs:38657400,0.202373210992964)
--(axis cs:40025400,0.227285556392491)
--(axis cs:41393400,0.214252297075425)
--(axis cs:42761400,0.227298643087347)
--(axis cs:44129400,0.237379203369137)
--(axis cs:45839400,0.23846834692752)
--(axis cs:47207400,0.230992017571885)
--(axis cs:48575400,0.224421674713967)
--(axis cs:49943400,0.236801405578678)
--(axis cs:51311400,0.251379060974989)
--(axis cs:52679400,0.235254296922845)
--(axis cs:54047400,0.247512787567984)
--(axis cs:55415400,0.219940891035934)
--(axis cs:56783400,0.242692607179709)
--(axis cs:58151400,0.259021883229538)
--(axis cs:59519400,0.242475235788448)
--(axis cs:60887400,0.262592582101658)
--(axis cs:62255400,0.219227656187349)
--(axis cs:63623400,0.232595169253465)
--(axis cs:64991400,0.220524680312165)
--(axis cs:66359400,0.244156194795157)
--(axis cs:68069400,0.254295591533585)
--(axis cs:68069400,0.274558287391642)
--(axis cs:68069400,0.274558287391642)
--(axis cs:66359400,0.269504401414261)
--(axis cs:64991400,0.241606855814375)
--(axis cs:63623400,0.249730890643188)
--(axis cs:62255400,0.234172818369725)
--(axis cs:60887400,0.277109262655278)
--(axis cs:59519400,0.256733873559862)
--(axis cs:58151400,0.26895056622604)
--(axis cs:56783400,0.255730451030935)
--(axis cs:55415400,0.236501928440246)
--(axis cs:54047400,0.2588004767903)
--(axis cs:52679400,0.249554303619656)
--(axis cs:51311400,0.265885314234538)
--(axis cs:49943400,0.257454507514925)
--(axis cs:48575400,0.235933658018748)
--(axis cs:47207400,0.237595574678882)
--(axis cs:45839400,0.251976330949692)
--(axis cs:44129400,0.248189027555202)
--(axis cs:42761400,0.236490218003206)
--(axis cs:41393400,0.223664341446914)
--(axis cs:40025400,0.245992430395209)
--(axis cs:38657400,0.216735637238198)
--(axis cs:37289400,0.234121646626307)
--(axis cs:35921400,0.24506924393346)
--(axis cs:34553400,0.246459338097284)
--(axis cs:33185400,0.240413083817974)
--(axis cs:31817400,0.2596528037834)
--(axis cs:30449400,0.248580453649634)
--(axis cs:29081400,0.23102214848091)
--(axis cs:27713400,0.232927391361134)
--(axis cs:26345400,0.226213044317736)
--(axis cs:24977400,0.229819427096918)
--(axis cs:23609400,0.250179312858421)
--(axis cs:21899400,0.23299665839094)
--(axis cs:20531400,0.209242277019179)
--(axis cs:19163400,0.223382570091153)
--(axis cs:17795400,0.21861106108621)
--(axis cs:16427400,0.230778524277753)
--(axis cs:15059400,0.228243814890706)
--(axis cs:13691400,0.250013968712063)
--(axis cs:12323400,0.245735018217111)
--(axis cs:10955400,0.23502048579453)
--(axis cs:9587400,0.252491874711201)
--(axis cs:8219400,0.26038945908894)
--(axis cs:6851400,0.26308872625453)
--(axis cs:5483400,0.262153398109376)
--(axis cs:4115400,0.242740832745767)
--(axis cs:2747400,0.253479010891169)
--(axis cs:1379400,0.227487665601075)
--(axis cs:11400,0.338815778493881)
--cycle;

\addplot [line width=\linewidthothers, \replace_color, mark=*, mark size=0, mark options={solid}]
table {%
11400 0.329769738018513
1379400 0.216488488251343
2747400 0.242476617044304
4115400 0.236265884738714
5483400 0.256234542817595
6851400 0.254479872923401
8219400 0.25119723874946
9587400 0.248208625052146
10955400 0.229519040518961
12323400 0.238261861406795
13691400 0.246024901170082
15059400 0.222819810270368
16427400 0.222538873749641
17795400 0.211779077741011
19163400 0.214372485379996
20531400 0.204624285269658
21899400 0.227786960155944
23609400 0.245652635911814
24977400 0.225243705589538
26345400 0.22351346356084
27713400 0.2300058868808
29081400 0.226340830355774
30449400 0.242984454353896
31817400 0.253536186566855
33185400 0.234853533263594
34553400 0.239558606904158
35921400 0.240316103600705
37289400 0.228094544010608
38657400 0.211714873649897
40025400 0.236750394658518
41393400 0.219531880770311
42761400 0.232562412230992
44129400 0.244428940944834
45839400 0.24563110871114
47207400 0.233840213759752
48575400 0.228469559710395
49943400 0.244846523224006
51311400 0.259382168333296
52679400 0.24111912789021
54047400 0.252180475110193
55415400 0.229785002830008
56783400 0.252363581026987
58151400 0.262715101811728
59519400 0.247876168558778
60887400 0.268175504114985
62255400 0.22474738647182
63623400 0.243163856443916
64991400 0.228886926712526
66359400 0.254844460382348
68069400 0.261919547205187
};

\path [draw=\res_color, fill=\res_color, opacity=0.2]
(axis cs:11400,0.340460523962975)
--(axis cs:11400,0.319078952074051)
--(axis cs:1379400,0.225791529519483)
--(axis cs:2747400,0.19808799243765)
--(axis cs:4115400,0.190760161589424)
--(axis cs:5483400,0.220900195819979)
--(axis cs:6851400,0.195496284771661)
--(axis cs:8219400,0.215200604121965)
--(axis cs:9587400,0.247751876310135)
--(axis cs:10955400,0.264353770769495)
--(axis cs:12323400,0.324661986090931)
--(axis cs:13691400,0.390071358283597)
--(axis cs:15059400,0.418212502375798)
--(axis cs:16427400,0.467673509819256)
--(axis cs:17795400,0.496201204852614)
--(axis cs:19163400,0.537065306897355)
--(axis cs:20531400,0.588266801094836)
--(axis cs:21899400,0.606509973553131)
--(axis cs:23609400,0.646439970712318)
--(axis cs:24977400,0.604501472794651)
--(axis cs:26345400,0.652168127316999)
--(axis cs:27713400,0.642735370860885)
--(axis cs:29081400,0.663320635064591)
--(axis cs:30449400,0.702663078481747)
--(axis cs:31817400,0.718331245628493)
--(axis cs:33185400,0.699091534012386)
--(axis cs:34553400,0.696065142919041)
--(axis cs:35921400,0.70406929767195)
--(axis cs:37289400,0.708987413868782)
--(axis cs:38657400,0.700398000821535)
--(axis cs:40025400,0.745682003211515)
--(axis cs:41393400,0.681517199687095)
--(axis cs:42761400,0.723480005076931)
--(axis cs:44129400,0.734675145797022)
--(axis cs:45839400,0.727021050491138)
--(axis cs:47207400,0.734434166679029)
--(axis cs:48575400,0.739981951251516)
--(axis cs:49943400,0.740341134047097)
--(axis cs:51311400,0.763298179025424)
--(axis cs:52679400,0.750033844342283)
--(axis cs:54047400,0.744141414303006)
--(axis cs:55415400,0.71952619617486)
--(axis cs:56783400,0.737962962691337)
--(axis cs:58151400,0.7501855386608)
--(axis cs:59519400,0.737973537985798)
--(axis cs:60887400,0.757199190275043)
--(axis cs:62255400,0.746306917365073)
--(axis cs:63623400,0.750017691473016)
--(axis cs:64991400,0.739966926324874)
--(axis cs:66359400,0.76147350244875)
--(axis cs:68069400,0.749578449115536)
--(axis cs:68069400,0.764028117244318)
--(axis cs:68069400,0.764028117244318)
--(axis cs:66359400,0.774044779334622)
--(axis cs:64991400,0.764678633793703)
--(axis cs:63623400,0.771222117785593)
--(axis cs:62255400,0.760171210220566)
--(axis cs:60887400,0.774690801478251)
--(axis cs:59519400,0.756427243756139)
--(axis cs:58151400,0.769105257731504)
--(axis cs:56783400,0.757514660891742)
--(axis cs:55415400,0.740550025038327)
--(axis cs:54047400,0.751952036137375)
--(axis cs:52679400,0.767759067529475)
--(axis cs:51311400,0.77575654160192)
--(axis cs:49943400,0.750994989235547)
--(axis cs:48575400,0.761565351689195)
--(axis cs:47207400,0.7529378512042)
--(axis cs:45839400,0.743587986926809)
--(axis cs:44129400,0.744023232127115)
--(axis cs:42761400,0.742227901453628)
--(axis cs:41393400,0.694443211572097)
--(axis cs:40025400,0.772582482891823)
--(axis cs:38657400,0.71706267604388)
--(axis cs:37289400,0.729234137360615)
--(axis cs:35921400,0.717580931171508)
--(axis cs:34553400,0.728010779338946)
--(axis cs:33185400,0.725790562992343)
--(axis cs:31817400,0.732977536054425)
--(axis cs:30449400,0.72297961705726)
--(axis cs:29081400,0.685597046466088)
--(axis cs:27713400,0.661026547442661)
--(axis cs:26345400,0.672463165725096)
--(axis cs:24977400,0.63286068757091)
--(axis cs:23609400,0.670098510127157)
--(axis cs:21899400,0.638204550459921)
--(axis cs:20531400,0.621382707013672)
--(axis cs:19163400,0.580227607819668)
--(axis cs:17795400,0.547427716375854)
--(axis cs:16427400,0.519439015620268)
--(axis cs:15059400,0.473030896319023)
--(axis cs:13691400,0.43473975328841)
--(axis cs:12323400,0.348538352462989)
--(axis cs:10955400,0.289870058219792)
--(axis cs:9587400,0.27646403911397)
--(axis cs:8219400,0.236761090682666)
--(axis cs:6851400,0.223336703487817)
--(axis cs:5483400,0.234594614901327)
--(axis cs:4115400,0.208147954890592)
--(axis cs:2747400,0.224024720999296)
--(axis cs:1379400,0.249177628196776)
--(axis cs:11400,0.340460523962975)
--cycle;

\addplot [line width=\linewidthtop, \res_color, mark=*, mark size=0, mark options={solid}]
table {%
11400 0.329769738018513
1379400 0.242084703641012
2747400 0.214487172735971
4115400 0.198522066771147
5483400 0.225902945722055
6851400 0.214944541614521
8219400 0.226897799690792
9587400 0.266534336222148
10955400 0.268294768925151
12323400 0.337699803664475
13691400 0.411405978824075
15059400 0.44553194975638
16427400 0.489329687204069
17795400 0.518552703097187
19163400 0.557090229668355
20531400 0.608420103284104
21899400 0.618705340316562
23609400 0.660270043351719
24977400 0.613605308268115
26345400 0.660976458080629
27713400 0.653099099834621
29081400 0.67657864261846
30449400 0.713551738688948
31817400 0.725448681097877
33185400 0.718624514514083
34553400 0.710675565005738
35921400 0.713220289253955
37289400 0.717206677432215
38657400 0.707938536968462
40025400 0.762039271311264
41393400 0.689918314098153
42761400 0.732655073499852
44129400 0.740452749137407
45839400 0.733164449363095
47207400 0.744200617842714
48575400 0.751493854766451
49943400 0.74641355966886
51311400 0.766764219403599
52679400 0.757053950795943
54047400 0.747815808337243
55415400 0.730940909594976
56783400 0.74857154014624
58151400 0.758965273670809
59519400 0.746909122298576
60887400 0.768306643514878
62255400 0.757018229889254
63623400 0.755720911860528
64991400 0.760673231878036
66359400 0.7665673315654
68069400 0.757286164268894
};

\path [draw=\whole_color, fill=\whole_color, opacity=0.2]
(axis cs:11400,0.310032896697521)
--(axis cs:11400,0.29440788179636)
--(axis cs:1379400,0.150030839489773)
--(axis cs:2747400,0.171524851175491)
--(axis cs:4115400,0.181449888359566)
--(axis cs:5483400,0.242711743740244)
--(axis cs:6851400,0.221764488270551)
--(axis cs:8219400,0.23380503244425)
--(axis cs:9587400,0.271005055290172)
--(axis cs:10955400,0.287662223416158)
--(axis cs:12323400,0.326843760444863)
--(axis cs:13691400,0.367152720163088)
--(axis cs:15059400,0.374120514223857)
--(axis cs:16427400,0.393477191619042)
--(axis cs:17795400,0.399905475385258)
--(axis cs:19163400,0.433719384808382)
--(axis cs:20531400,0.440878359188565)
--(axis cs:21899400,0.44946496461007)
--(axis cs:23609400,0.479742997889141)
--(axis cs:24977400,0.451802233240893)
--(axis cs:26345400,0.493645010286367)
--(axis cs:27713400,0.491417103016205)
--(axis cs:29081400,0.493929923187462)
--(axis cs:30449400,0.547225121901949)
--(axis cs:31817400,0.55169645520059)
--(axis cs:33185400,0.558656851892947)
--(axis cs:34553400,0.557171402216285)
--(axis cs:35921400,0.548957674630007)
--(axis cs:37289400,0.552915933475365)
--(axis cs:38657400,0.58070482502916)
--(axis cs:40025400,0.602027591254884)
--(axis cs:41393400,0.564702539508181)
--(axis cs:42761400,0.579149660831518)
--(axis cs:44129400,0.623646558607248)
--(axis cs:45839400,0.598495805762329)
--(axis cs:47207400,0.608180549352119)
--(axis cs:48575400,0.629177786538725)
--(axis cs:49943400,0.613839344662753)
--(axis cs:51311400,0.634116356529517)
--(axis cs:52679400,0.630270433408968)
--(axis cs:54047400,0.63853755026771)
--(axis cs:55415400,0.632060643470172)
--(axis cs:56783400,0.646633274992289)
--(axis cs:58151400,0.653528600848834)
--(axis cs:59519400,0.644114019371688)
--(axis cs:60887400,0.650112285971412)
--(axis cs:62255400,0.621340755697497)
--(axis cs:63623400,0.643475528848061)
--(axis cs:64991400,0.653116911754325)
--(axis cs:66359400,0.674818404874669)
--(axis cs:68069400,0.655446396802034)
--(axis cs:68069400,0.675462990036079)
--(axis cs:68069400,0.675462990036079)
--(axis cs:66359400,0.690083743212879)
--(axis cs:64991400,0.665038841583224)
--(axis cs:63623400,0.672520105839223)
--(axis cs:62255400,0.636455039362514)
--(axis cs:60887400,0.662949327257466)
--(axis cs:59519400,0.66295756895108)
--(axis cs:58151400,0.672914641767294)
--(axis cs:56783400,0.665182034484085)
--(axis cs:55415400,0.651047438068197)
--(axis cs:54047400,0.659116097391924)
--(axis cs:52679400,0.665933657804281)
--(axis cs:51311400,0.663721790493224)
--(axis cs:49943400,0.637803609198445)
--(axis cs:48575400,0.66510397723881)
--(axis cs:47207400,0.633624558903491)
--(axis cs:45839400,0.631989306590476)
--(axis cs:44129400,0.670374592765486)
--(axis cs:42761400,0.613040767006059)
--(axis cs:41393400,0.5998280373158)
--(axis cs:40025400,0.648227493410189)
--(axis cs:38657400,0.610255553475761)
--(axis cs:37289400,0.602504039302424)
--(axis cs:35921400,0.579668515077509)
--(axis cs:34553400,0.592130522350905)
--(axis cs:33185400,0.588986185509328)
--(axis cs:31817400,0.582530077906187)
--(axis cs:30449400,0.591229715382746)
--(axis cs:29081400,0.539679918460804)
--(axis cs:27713400,0.534023944782266)
--(axis cs:26345400,0.525157273980783)
--(axis cs:24977400,0.492872963344744)
--(axis cs:23609400,0.524033575033293)
--(axis cs:21899400,0.493480893165025)
--(axis cs:20531400,0.489359788732967)
--(axis cs:19163400,0.482424305410623)
--(axis cs:17795400,0.433999804291962)
--(axis cs:16427400,0.426488520043085)
--(axis cs:15059400,0.39639442529257)
--(axis cs:13691400,0.384957097799712)
--(axis cs:12323400,0.345961795559288)
--(axis cs:10955400,0.324992526465885)
--(axis cs:9587400,0.29446454922969)
--(axis cs:8219400,0.25301964835001)
--(axis cs:6851400,0.236231566202171)
--(axis cs:5483400,0.253094881816537)
--(axis cs:4115400,0.193440086123701)
--(axis cs:2747400,0.185145328025101)
--(axis cs:1379400,0.163188732927665)
--(axis cs:11400,0.310032896697521)
--cycle;

\addplot [line width=\linewidthothers, \whole_color, mark=*, mark size=0, mark options={solid}]
table {%
11400 0.304276309907436
1379400 0.154451069654897
2747400 0.179398386695539
4115400 0.190547941349905
5483400 0.249062147071982
6851400 0.228041448774999
8219400 0.242723079860857
9587400 0.284430811479045
10955400 0.304135892984999
12323400 0.332815147693728
13691400 0.37010072864806
15059400 0.377023959851822
16427400 0.404460930967526
17795400 0.416506481818823
19163400 0.463081823310209
20531400 0.463783025955405
21899400 0.469026953138407
23609400 0.504222027265246
24977400 0.477219971479222
26345400 0.513374396309296
27713400 0.520382754024955
29081400 0.515641904642282
30449400 0.571290122765433
31817400 0.568497465139327
33185400 0.572941908480558
34553400 0.574028999122493
35921400 0.567925014768924
37289400 0.581930425169747
38657400 0.593703731753533
40025400 0.634305056170571
41393400 0.590556717826434
42761400 0.599571679383927
44129400 0.645735033975586
45839400 0.61818298530315
47207400 0.628410478654867
48575400 0.648447798138717
49943400 0.622185822833977
51311400 0.647576107232036
52679400 0.645952889980969
54047400 0.649793562918001
55415400 0.641904587080577
56783400 0.66092651011971
58151400 0.661833697999375
59519400 0.656131267011398
60887400 0.656253929515944
62255400 0.631076484796271
63623400 0.658530962460569
64991400 0.660411597258861
66359400 0.682018239841517
68069400 0.664571278951749
};

\path [draw=C2, fill=C2, opacity=0.2]
(axis cs:121600,0)
--(axis cs:121600,0)
--(axis cs:3283200,0)
--(axis cs:6444800,0)
--(axis cs:9606400,0)
--(axis cs:12889600,0)
--(axis cs:16051200,0)
--(axis cs:19212800,0)
--(axis cs:22374400,0)
--(axis cs:25657600,0)
--(axis cs:28819200,0)
--(axis cs:31980800,0)
--(axis cs:35142400,0.0197368421052632)
--(axis cs:38425600,0.0723684210526316)
--(axis cs:41587200,0.0855263157894737)
--(axis cs:44748800,0.118421052631579)
--(axis cs:47910400,0.105263157894737)
--(axis cs:51193600,0.144736842105263)
--(axis cs:54355200,0.144736842105263)
--(axis cs:57516800,0.203947368421053)
--(axis cs:60800000,0.217105263157895)
--(axis cs:60800000,0.361842105263158)
--(axis cs:60800000,0.361842105263158)
--(axis cs:57516800,0.394736842105263)
--(axis cs:54355200,0.355263157894737)
--(axis cs:51193600,0.342105263157895)
--(axis cs:47910400,0.322368421052632)
--(axis cs:44748800,0.447368421052632)
--(axis cs:41587200,0.368421052631579)
--(axis cs:38425600,0.296052631578947)
--(axis cs:35142400,0.177631578947368)
--(axis cs:31980800,0.144736842105263)
--(axis cs:28819200,0.111842105263158)
--(axis cs:25657600,0.0592105263157895)
--(axis cs:22374400,0.0460526315789474)
--(axis cs:19212800,0.0394736842105263)
--(axis cs:16051200,0.0131578947368421)
--(axis cs:12889600,0)
--(axis cs:9606400,0)
--(axis cs:6444800,0)
--(axis cs:3283200,0)
--(axis cs:121600,0)
--cycle;

\addplot [line width=\linewidthothers, C2, mark=*, mark size=0, mark options={solid}]
table {%
121600 0
3283200 0
6444800 0
9606400 0
12889600 0
16051200 0
19212800 0
22374400 0
25657600 0
28819200 0.0131578947368421
31980800 0.0328947368421053
35142400 0.105263157894737
38425600 0.243421052631579
41587200 0.328947368421053
44748800 0.342105263157895
47910400 0.25
51193600 0.263157894736842
54355200 0.269736842105263
57516800 0.302631578947368
60800000 0.328947368421053
};

\path [draw=C5, fill=C5, opacity=0.2]
(axis cs:121600,0.118421052631579)
--(axis cs:121600,0.0328947368421053)
--(axis cs:3283200,0)
--(axis cs:6444800,0)
--(axis cs:9606400,0)
--(axis cs:12889600,0)
--(axis cs:16051200,0)
--(axis cs:19212800,0.0131578947368421)
--(axis cs:22374400,0.00657894736842105)
--(axis cs:25657600,0.0723684210526316)
--(axis cs:28819200,0.0921052631578947)
--(axis cs:31980800,0.111842105263158)
--(axis cs:35142400,0.0592105263157895)
--(axis cs:38425600,0.105263157894737)
--(axis cs:41587200,0.164473684210526)
--(axis cs:44748800,0.223519736842106)
--(axis cs:47910400,0.171052631578947)
--(axis cs:51193600,0.184210526315789)
--(axis cs:54355200,0.217105263157895)
--(axis cs:57516800,0.296052631578947)
--(axis cs:60800000,0.269736842105263)
--(axis cs:60800000,0.361842105263158)
--(axis cs:60800000,0.361842105263158)
--(axis cs:57516800,0.388157894736842)
--(axis cs:54355200,0.296052631578947)
--(axis cs:51193600,0.315789473684211)
--(axis cs:47910400,0.263157894736842)
--(axis cs:44748800,0.381578947368421)
--(axis cs:41587200,0.276315789473684)
--(axis cs:38425600,0.256743421052629)
--(axis cs:35142400,0.197368421052632)
--(axis cs:31980800,0.217105263157895)
--(axis cs:28819200,0.243421052631579)
--(axis cs:25657600,0.171052631578947)
--(axis cs:22374400,0.0789473684210526)
--(axis cs:19212800,0.0789473684210526)
--(axis cs:16051200,0.00657894736842105)
--(axis cs:12889600,0)
--(axis cs:9606400,0)
--(axis cs:6444800,0)
--(axis cs:3283200,0)
--(axis cs:121600,0.118421052631579)
--cycle;

\addplot [line width=\linewidthothers, C5, mark=*, mark size=0, mark options={solid}]
table {%
121600 0.0723684210526316
3283200 0
6444800 0
9606400 0
12889600 0
16051200 0
19212800 0.0526315789473684
22374400 0.0394736842105263
25657600 0.131578947368421
28819200 0.151315789473684
31980800 0.157894736842105
35142400 0.125
38425600 0.164473684210526
41587200 0.210526315789474
44748800 0.269736842105263
47910400 0.197368421052632
51193600 0.236842105263158
54355200 0.25
57516800 0.348684210526316
60800000 0.322368421052632
};

\path [draw=C4, fill=C4, opacity=0.2]
(axis cs:10400,0)
--(axis cs:10400,0)
--(axis cs:285600,0)
--(axis cs:562400,0)
--(axis cs:836000,0)
--(axis cs:1107600,0)
--(axis cs:1384800,0)
--(axis cs:1658800,0)
--(axis cs:1938400,0)
--(axis cs:2212000,0)
--(axis cs:2487200,0)
--(axis cs:2759200,0)
--(axis cs:3034000,0)
--(axis cs:3310000,0)
--(axis cs:3584400,0)
--(axis cs:3863600,0)
--(axis cs:4140400,0)
--(axis cs:4414800,0)
--(axis cs:4685600,0)
--(axis cs:4963200,0)
--(axis cs:5247200,0)
--(axis cs:5247200,0.0375)
--(axis cs:5247200,0.0375)
--(axis cs:4963200,0.0125)
--(axis cs:4685600,0.0125)
--(axis cs:4414800,0)
--(axis cs:4140400,0.025)
--(axis cs:3863600,0.0125)
--(axis cs:3584400,0)
--(axis cs:3310000,0)
--(axis cs:3034000,0)
--(axis cs:2759200,0)
--(axis cs:2487200,0)
--(axis cs:2212000,0)
--(axis cs:1938400,0)
--(axis cs:1658800,0)
--(axis cs:1384800,0)
--(axis cs:1107600,0)
--(axis cs:836000,0)
--(axis cs:562400,0)
--(axis cs:285600,0)
--(axis cs:10400,0)
--cycle;

\addplot [line width=\linewidthothers, C4, mark=*, mark size=0, mark options={solid}]
table {%
10400 0
285600 0
562400 0
836000 0
1107600 0
1384800 0
1658800 0
1938400 0
2212000 0
2487200 0
2759200 0
3034000 0
3310000 0
3584400 0
3863600 0
4140400 0
4414800 0
4685600 0
4963200 0
5247200 0
};

\path [draw=C4, fill=C4, opacity=0.2]
(axis cs:10400,0)
--(axis cs:10400,0)
--(axis cs:290000,0)
--(axis cs:564000,0)
--(axis cs:840800,0)
--(axis cs:1112800,0)
--(axis cs:1385200,0)
--(axis cs:1654800,0)
--(axis cs:1930400,0)
--(axis cs:2208400,0)
--(axis cs:2486800,0)
--(axis cs:2760000,0)
--(axis cs:3034800,0)
--(axis cs:3312400,0)
--(axis cs:3588000,0)
--(axis cs:3861200,0)
--(axis cs:4143600,0)
--(axis cs:4424000,0)
--(axis cs:4704800,0)
--(axis cs:4974000,0)
--(axis cs:5248000,0)
--(axis cs:5248000,0)
--(axis cs:5248000,0)
--(axis cs:4974000,0)
--(axis cs:4704800,0)
--(axis cs:4424000,0)
--(axis cs:4143600,0)
--(axis cs:3861200,0)
--(axis cs:3588000,0)
--(axis cs:3312400,0)
--(axis cs:3034800,0)
--(axis cs:2760000,0.0125)
--(axis cs:2486800,0.0125)
--(axis cs:2208400,0.0125)
--(axis cs:1930400,0)
--(axis cs:1654800,0)
--(axis cs:1385200,0)
--(axis cs:1112800,0)
--(axis cs:840800,0)
--(axis cs:564000,0)
--(axis cs:290000,0)
--(axis cs:10400,0)
--cycle;

\addplot [line width=\linewidthothers, C4, mark=*, mark size=0, mark options={solid}]
table {%
10400 0
290000 0
564000 0
840800 0
1112800 0
1385200 0
1654800 0
1930400 0
2208400 0
2486800 0
2760000 0
3034800 0
3312400 0
3588000 0
3861200 0
4143600 0
4424000 0
4704800 0
4974000 0
5248000 0
};
\end{axis}

\end{tikzpicture}

%% file: images/tex/dual_arm_m_success.tex
%
%

\begin{tikzpicture}
\def\res_color{C0}
\def\replace_color{C1}
\def\whole_color{C3}
\def\bbrl_color{C9}
\def\ppo_color{C2}
\def\sac_color{C4}
\def\linewidthtop{1mm}
\def\linewidthothers{0.5mm}

\begin{axis}[
legend cell align={left},
legend style={fill opacity=0.8, draw opacity=1, text opacity=1, draw=lightgray204, at={(0.03,0.03)},  anchor=north west},
tick align=outside,
tick pos=left,
x grid style={white},
scaled x ticks=false,
xticklabels={,0,1,2,3,4},
xlabel={Environment Interactions ($\times 10^7$)},
xmajorgrids,
xmin=-1500000, xmax=41000000.05,
xtick style={color=black},
y grid style={white},
ylabel={Success Rate, Markovian, IQM},
ymajorgrids,
ymin=-0.05, ymax=0.9,
ytick style={color=black},
axis background/.style={fill=plot_background},
label style={font=\large},
tick label style={font=\large},
x axis line style={draw=none},
y axis line style={draw=none},
]
\path [draw=\bbrl_color, fill=\bbrl_color, opacity=0.2]
(axis cs:20300,0)
--(axis cs:20300,0)
--(axis cs:3077900,0.000927734388824319)
--(axis cs:6205500,0.0033592224767176)
--(axis cs:9975000,0.00318107609788854)
--(axis cs:13130500,0.00154174308004165)
--(axis cs:16933300,0.00402451255652757)
--(axis cs:20117100,0.00709206541905898)
--(axis cs:23908900,0.0049311505544603)
--(axis cs:27066900,0.0041343202164004)
--(axis cs:30224100,0.00430227655380794)
--(axis cs:34019300,0.0080799708010582)
--(axis cs:37169200,0.00302032841810821)
--(axis cs:40969500,0.00517904378848373)
--(axis cs:44145300,0.00720920342486491)
--(axis cs:47949000,0.0060027003050685)
--(axis cs:51128000,0.00524070758331277)
--(axis cs:54318700,0.00473074760181278)
--(axis cs:58161900,0.00896989322699006)
--(axis cs:61380000,0.0094332690529374)
--(axis cs:65254400,0.0147410309704044)
--(axis cs:68506700,0.0177970532596165)
--(axis cs:72415100,0.0274699767349984)
--(axis cs:75694300,0.0216967102075709)
--(axis cs:78992100,0.0194909048942046)
--(axis cs:82968800,0.033555906196307)
--(axis cs:86293700,0.0305410587745285)
--(axis cs:90303100,0.0369678696784118)
--(axis cs:93662000,0.046659535729108)
--(axis cs:97693900,0.0525462964151474)
--(axis cs:101082900,0.0606225768021017)
--(axis cs:104468800,0.0444871995410397)
--(axis cs:108555900,0.0464712947612175)
--(axis cs:111970200,0.0611263064923968)
--(axis cs:116081800,0.0500693061278606)
--(axis cs:119514500,0.0626740884208666)
--(axis cs:123641800,0.0681836615162864)
--(axis cs:127092000,0.0581846893907551)
--(axis cs:130544100,0.0738724935982007)
--(axis cs:134697700,0.0699529492317407)
--(axis cs:138166500,0.0773172977875376)
--(axis cs:142334300,0.0869471350747713)
--(axis cs:145810700,0.0882252328684414)
--(axis cs:149984400,0.0938869669704295)
--(axis cs:153481300,0.0946732147001659)
--(axis cs:156972400,0.0926745767656699)
--(axis cs:161160400,0.0836683150761176)
--(axis cs:164649400,0.0945031322952469)
--(axis cs:168837100,0.0956818649861598)
--(axis cs:172329400,0.0992685561694506)
--(axis cs:176519000,0.105473320817023)
--(axis cs:176519000,0.119024493479717)
--(axis cs:176519000,0.119024493479717)
--(axis cs:172329400,0.108782373369068)
--(axis cs:168837100,0.117531258195478)
--(axis cs:164649400,0.11026022706987)
--(axis cs:161160400,0.0956802034478307)
--(axis cs:156972400,0.112156175359182)
--(axis cs:153481300,0.113251268634909)
--(axis cs:149984400,0.109547755906926)
--(axis cs:145810700,0.11080723880773)
--(axis cs:142334300,0.10489513816559)
--(axis cs:138166500,0.0991745219674471)
--(axis cs:134697700,0.0940340521153007)
--(axis cs:130544100,0.0996284043951768)
--(axis cs:127092000,0.0873383364364714)
--(axis cs:123641800,0.089620223535622)
--(axis cs:119514500,0.0865260976535547)
--(axis cs:116081800,0.0704528593819204)
--(axis cs:111970200,0.0781029726660005)
--(axis cs:108555900,0.0677208728923802)
--(axis cs:104468800,0.0658265112687524)
--(axis cs:101082900,0.0815381344506166)
--(axis cs:97693900,0.0676246538812376)
--(axis cs:93662000,0.0598109017900387)
--(axis cs:90303100,0.0506456136428782)
--(axis cs:86293700,0.0473748990468025)
--(axis cs:82968800,0.0463603224372757)
--(axis cs:78992100,0.0336600690963181)
--(axis cs:75694300,0.0321393237041063)
--(axis cs:72415100,0.0414994932168209)
--(axis cs:68506700,0.0277626359443366)
--(axis cs:65254400,0.0290336977721909)
--(axis cs:61380000,0.0203786366806373)
--(axis cs:58161900,0.0164354272651025)
--(axis cs:54318700,0.0148082898667964)
--(axis cs:51128000,0.00963428229030899)
--(axis cs:47949000,0.0110069498660997)
--(axis cs:44145300,0.0143794938273514)
--(axis cs:40969500,0.00969708516384125)
--(axis cs:37169200,0.0079449985151152)
--(axis cs:34019300,0.0103084170419275)
--(axis cs:30224100,0.00714541203022819)
--(axis cs:27066900,0.00712541211203844)
--(axis cs:23908900,0.00748077785432533)
--(axis cs:20117100,0.0120352824329306)
--(axis cs:16933300,0.00749580905578408)
--(axis cs:13130500,0.00358353560820901)
--(axis cs:9975000,0.0069073441017462)
--(axis cs:6205500,0.0044190788927537)
--(axis cs:3077900,0.00336914067520411)
--(axis cs:20300,0)
--cycle;

\addplot [line width=\linewidthothers, \bbrl_color, mark=*, mark size=0, mark options={solid}]
table {%
20300 0
3077900 0.00256347660069878
6205500 0.00394210823486674
9975000 0.00478276023637658
13130500 0.00265090283576971
16933300 0.00607984758356511
20117100 0.0103213993472205
23908900 0.0063093459918966
27066900 0.00532556223407315
30224100 0.0061161957729322
34019300 0.00897059050888332
37169200 0.00588868742781337
40969500 0.00741115532783645
44145300 0.0122137129443557
47949000 0.00868407986772353
51128000 0.00695587740999483
54318700 0.00977000945923963
58161900 0.0124697422431914
61380000 0.0128652655357586
65254400 0.0207070687883902
68506700 0.0217875888250012
72415100 0.0335557635533844
75694300 0.0277870236706893
78992100 0.0251655208655971
82968800 0.0399515407395846
86293700 0.0401516580525213
90303100 0.0427859139980843
93662000 0.0527566773236098
97693900 0.0595401431919546
101082900 0.0687130864106639
104468800 0.0516033976197706
108555900 0.0566209935631996
111970200 0.0687723154576123
116081800 0.0588431435671535
119514500 0.073232666857518
123641800 0.0768400616112627
127092000 0.0729156402678413
130544100 0.0853060570461009
134697700 0.0812043501903551
138166500 0.0888832817587586
142334300 0.0976046211883477
145810700 0.100698789775314
149984400 0.101137517026466
153481300 0.104625390202471
156972400 0.103840565360602
161160400 0.0904185946398287
164649400 0.10368877956719
168837100 0.107638202277315
172329400 0.105920202075288
176519000 0.112092124716421
};

\path [draw=\replace_color, fill=\replace_color, opacity=0.2]
(axis cs:11400,0.340460523962975)
--(axis cs:11400,0.320723682641983)
--(axis cs:1379400,0.215923110023141)
--(axis cs:3089400,0.220405979809584)
--(axis cs:4799400,0.200421131468829)
--(axis cs:6509400,0.20967650957337)
--(axis cs:7877400,0.224224745217069)
--(axis cs:9587400,0.243379071000821)
--(axis cs:11297400,0.253148738964796)
--(axis cs:13007400,0.254006444766543)
--(axis cs:14375400,0.246644380942284)
--(axis cs:16085400,0.271967127873908)
--(axis cs:17795400,0.257873573229114)
--(axis cs:19505400,0.245848428746401)
--(axis cs:20873400,0.246746881397109)
--(axis cs:22583400,0.260897388404549)
--(axis cs:24293400,0.262376655840054)
--(axis cs:26003400,0.243421034597977)
--(axis cs:27371400,0.267250637694275)
--(axis cs:29081400,0.268093780335078)
--(axis cs:30791400,0.287897209036642)
--(axis cs:32501400,0.268459398157208)
--(axis cs:33869400,0.24768237152603)
--(axis cs:35579400,0.273278046954611)
--(axis cs:37289400,0.264462054207255)
--(axis cs:38999400,0.244142138714352)
--(axis cs:40367400,0.271780373096383)
--(axis cs:42077400,0.261224591972633)
--(axis cs:43787400,0.285502468729028)
--(axis cs:45497400,0.251848656289986)
--(axis cs:46865400,0.273818975598138)
--(axis cs:48575400,0.270632276539852)
--(axis cs:50285400,0.282062450212173)
--(axis cs:51995400,0.284355292967983)
--(axis cs:53363400,0.276100883591321)
--(axis cs:55073400,0.254475769099295)
--(axis cs:56783400,0.281185030393699)
--(axis cs:58493400,0.279370303082715)
--(axis cs:59861400,0.275984085789254)
--(axis cs:61571400,0.289068135383004)
--(axis cs:63281400,0.286555109433008)
--(axis cs:64991400,0.253307652569193)
--(axis cs:66359400,0.268236329338571)
--(axis cs:68069400,0.273299456154962)
--(axis cs:69779400,0.29626080046013)
--(axis cs:71489400,0.270437246884165)
--(axis cs:72857400,0.279377535932648)
--(axis cs:74567400,0.298806274244703)
--(axis cs:76277400,0.266118800397861)
--(axis cs:77987400,0.26745860231908)
--(axis cs:79697400,0.274835409888227)
--(axis cs:79697400,0.281981952469267)
--(axis cs:79697400,0.281981952469267)
--(axis cs:77987400,0.278980351076494)
--(axis cs:76277400,0.273429172441022)
--(axis cs:74567400,0.30694904600766)
--(axis cs:72857400,0.28692934034527)
--(axis cs:71489400,0.280894324532819)
--(axis cs:69779400,0.304998348127644)
--(axis cs:68069400,0.28872962692033)
--(axis cs:66359400,0.292037965996223)
--(axis cs:64991400,0.260639420053255)
--(axis cs:63281400,0.294393531887694)
--(axis cs:61571400,0.299399794246077)
--(axis cs:59861400,0.285880313553431)
--(axis cs:58493400,0.290294384354963)
--(axis cs:56783400,0.290370885060861)
--(axis cs:55073400,0.264594469610279)
--(axis cs:53363400,0.287555370816581)
--(axis cs:51995400,0.286884253749827)
--(axis cs:50285400,0.288151163898559)
--(axis cs:48575400,0.280384076322458)
--(axis cs:46865400,0.291913269036245)
--(axis cs:45497400,0.263982492254923)
--(axis cs:43787400,0.295907302994868)
--(axis cs:42077400,0.266958927698879)
--(axis cs:40367400,0.281635524836164)
--(axis cs:38999400,0.252531978831202)
--(axis cs:37289400,0.273418056238434)
--(axis cs:35579400,0.281900810192927)
--(axis cs:33869400,0.258639647724409)
--(axis cs:32501400,0.273679281645933)
--(axis cs:30791400,0.296688422548036)
--(axis cs:29081400,0.271709328483146)
--(axis cs:27371400,0.275771490938962)
--(axis cs:26003400,0.255375328316102)
--(axis cs:24293400,0.272961371857323)
--(axis cs:22583400,0.271481592849903)
--(axis cs:20873400,0.266048358836129)
--(axis cs:19505400,0.262372029062738)
--(axis cs:17795400,0.269393427235261)
--(axis cs:16085400,0.291330788890653)
--(axis cs:14375400,0.259874747990926)
--(axis cs:13007400,0.278064589966306)
--(axis cs:11297400,0.282655527235012)
--(axis cs:9587400,0.260583818378687)
--(axis cs:7877400,0.257219985340747)
--(axis cs:6509400,0.238662538422901)
--(axis cs:4799400,0.21777624637798)
--(axis cs:3089400,0.24074032244971)
--(axis cs:1379400,0.235300160944462)
--(axis cs:11400,0.340460523962975)
--cycle;

\addplot [line width=\linewidthothers, \replace_color, mark=*, mark size=0, mark options={solid}]
table {%
11400 0.330592103302479
1379400 0.226202714256942
3089400 0.230002953030635
4799400 0.209962239590368
6509400 0.223053671655876
7877400 0.240275388550974
9587400 0.248693708379026
11297400 0.269519730888456
13007400 0.268961965724164
14375400 0.255267948079541
16085400 0.285306848684103
17795400 0.26459448234825
19505400 0.259516657390113
20873400 0.259351318708888
22583400 0.265878777847255
24293400 0.267110419505985
26003400 0.250200611918983
27371400 0.271091783168938
29081400 0.269576484768389
30791400 0.292971559231663
32501400 0.271639603127251
33869400 0.254656724469694
35579400 0.278148955481688
37289400 0.268497403590727
38999400 0.248695617804067
40367400 0.277416940732993
42077400 0.263774967118009
43787400 0.291858760853108
45497400 0.258832411730394
46865400 0.281005049739235
48575400 0.27726268043706
50285400 0.284987320068314
51995400 0.286277690044783
53363400 0.279639832977716
55073400 0.25824759685902
56783400 0.286383114735372
58493400 0.285396552786551
59861400 0.279844443332916
61571400 0.296687966683527
63281400 0.291770370542647
64991400 0.257459373683444
66359400 0.279827161567354
68069400 0.281719516929765
69779400 0.299705089265029
71489400 0.275429929272876
72857400 0.28258747110215
74567400 0.304858616769119
76277400 0.268766114863706
77987400 0.271844259730717
79697400 0.277898528122523
};

\path [draw=\res_color, fill=\res_color, opacity=0.2]
(axis cs:11400,0.335526317358017)
--(axis cs:11400,0.319901317358017)
--(axis cs:1379400,0.207545229699463)
--(axis cs:3089400,0.178881193191046)
--(axis cs:4799400,0.191944023219548)
--(axis cs:6509400,0.207182473639222)
--(axis cs:7877400,0.228357734025523)
--(axis cs:9587400,0.246314045275626)
--(axis cs:11297400,0.272066456912931)
--(axis cs:13007400,0.315950576553483)
--(axis cs:14375400,0.340094775925102)
--(axis cs:16085400,0.431996793053194)
--(axis cs:17795400,0.446039871706022)
--(axis cs:19505400,0.502088627779223)
--(axis cs:20873400,0.507286611520653)
--(axis cs:22583400,0.543318907230399)
--(axis cs:24293400,0.540989484395665)
--(axis cs:26003400,0.566407360033168)
--(axis cs:27371400,0.552190632925571)
--(axis cs:29081400,0.548785437516262)
--(axis cs:30791400,0.606641958283249)
--(axis cs:32501400,0.584155400191424)
--(axis cs:33869400,0.561512522928628)
--(axis cs:35579400,0.575223355581446)
--(axis cs:37289400,0.584739153654815)
--(axis cs:38999400,0.599473124660685)
--(axis cs:40367400,0.591222435401897)
--(axis cs:42077400,0.566787358025924)
--(axis cs:43787400,0.599055009312022)
--(axis cs:45497400,0.591749224558921)
--(axis cs:46865400,0.577202166188784)
--(axis cs:48575400,0.593546823214323)
--(axis cs:50285400,0.587925672787838)
--(axis cs:51995400,0.599135671039015)
--(axis cs:53363400,0.579026977778697)
--(axis cs:55073400,0.59905262862471)
--(axis cs:56783400,0.571488859461754)
--(axis cs:58493400,0.604130046600957)
--(axis cs:59861400,0.570783601947628)
--(axis cs:61571400,0.583101237516178)
--(axis cs:63281400,0.564955340786327)
--(axis cs:64991400,0.572776203755215)
--(axis cs:66359400,0.601254522060448)
--(axis cs:68069400,0.570086294804663)
--(axis cs:69779400,0.596964954522502)
--(axis cs:71489400,0.548688308705821)
--(axis cs:72857400,0.61216374450434)
--(axis cs:74567400,0.552947326787667)
--(axis cs:76277400,0.581205297363484)
--(axis cs:77987400,0.606515665643067)
--(axis cs:79697400,0.575800978198437)
--(axis cs:79697400,0.615172029694304)
--(axis cs:79697400,0.615172029694304)
--(axis cs:77987400,0.634619304045721)
--(axis cs:76277400,0.608676024931723)
--(axis cs:74567400,0.601746635288893)
--(axis cs:72857400,0.643627122199306)
--(axis cs:71489400,0.585812277007582)
--(axis cs:69779400,0.626343250068537)
--(axis cs:68069400,0.610199959125106)
--(axis cs:66359400,0.628118985127115)
--(axis cs:64991400,0.602816799146561)
--(axis cs:63281400,0.597270020857429)
--(axis cs:61571400,0.610937350962033)
--(axis cs:59861400,0.598389710960475)
--(axis cs:58493400,0.636220773196601)
--(axis cs:56783400,0.604746655649544)
--(axis cs:55073400,0.61620213079536)
--(axis cs:53363400,0.604783800680858)
--(axis cs:51995400,0.625392752580984)
--(axis cs:50285400,0.614214699019977)
--(axis cs:48575400,0.627299777331298)
--(axis cs:46865400,0.606375846997346)
--(axis cs:45497400,0.618826899730412)
--(axis cs:43787400,0.640846917355636)
--(axis cs:42077400,0.602439041686054)
--(axis cs:40367400,0.630701233161331)
--(axis cs:38999400,0.635985853699464)
--(axis cs:37289400,0.602741977872965)
--(axis cs:35579400,0.603221326170899)
--(axis cs:33869400,0.579727760704512)
--(axis cs:32501400,0.601351737499004)
--(axis cs:30791400,0.631898615106097)
--(axis cs:29081400,0.580477539307219)
--(axis cs:27371400,0.575782366795464)
--(axis cs:26003400,0.597210068922046)
--(axis cs:24293400,0.576871765218562)
--(axis cs:22583400,0.579231700505172)
--(axis cs:20873400,0.53923805718316)
--(axis cs:19505400,0.520594342929755)
--(axis cs:17795400,0.474514556226324)
--(axis cs:16085400,0.462854552193764)
--(axis cs:14375400,0.385268309367696)
--(axis cs:13007400,0.342999743791554)
--(axis cs:11297400,0.294755019850162)
--(axis cs:9587400,0.268417002284778)
--(axis cs:7877400,0.255660769945432)
--(axis cs:6509400,0.224492620889016)
--(axis cs:4799400,0.218320794866941)
--(axis cs:3089400,0.202402216906194)
--(axis cs:1379400,0.235454356996343)
--(axis cs:11400,0.335526317358017)
--cycle;

\addplot [line width=\linewidthtop, \res_color, mark=*, mark size=0, mark options={solid}]
table {%
11400 0.329769738018513
1379400 0.225328946951777
3089400 0.194363243419502
4799400 0.208410965680059
6509400 0.216481063507736
7877400 0.246296764006979
9587400 0.260524913838874
11297400 0.28458741466664
13007400 0.326921992382595
14375400 0.366884374420905
16085400 0.446335763561079
17795400 0.455742898739471
19505400 0.510956113771899
20873400 0.52302809844858
22583400 0.565396371400032
24293400 0.556708465562639
26003400 0.582891783941805
27371400 0.566160165602732
29081400 0.561633759632719
30791400 0.61636605553783
32501400 0.589293287346395
33869400 0.568930428721415
35579400 0.589175163024371
37289400 0.592582380964883
38999400 0.609838941277145
40367400 0.607356535647364
42077400 0.581078433635568
43787400 0.615326448648987
45497400 0.601437907496691
46865400 0.589398217951019
48575400 0.602329945970175
50285400 0.59825252358378
51995400 0.605610298884105
53363400 0.587238655689812
55073400 0.606873558521231
56783400 0.588517700059647
58493400 0.613956526687653
59861400 0.579576356142974
61571400 0.592778632083904
63281400 0.581539685055806
64991400 0.589437897207727
66359400 0.618040822410149
68069400 0.593860853138876
69779400 0.613504951856122
71489400 0.574965729206862
72857400 0.628174717131964
74567400 0.576514907595081
76277400 0.595597823621507
77987400 0.622724065800631
79697400 0.591798675426167
};

\path [draw=\whole_color, fill=\whole_color, opacity=0.2]
(axis cs:12000,0.324999988079071)
--(axis cs:12000,0.298437505960464)
--(axis cs:1812000,0.150659179664217)
--(axis cs:3612000,0.181738281586149)
--(axis cs:5772000,0.219369810634086)
--(axis cs:7572000,0.188473769281078)
--(axis cs:9732000,0.193290709285951)
--(axis cs:11532000,0.204725742189477)
--(axis cs:13692000,0.239609939543806)
--(axis cs:15492000,0.260805094258013)
--(axis cs:17292000,0.268402453384177)
--(axis cs:19452000,0.26586711040157)
--(axis cs:21252000,0.304250396360518)
--(axis cs:23412000,0.29890111509663)
--(axis cs:25212000,0.307992951270748)
--(axis cs:27372000,0.320076989972759)
--(axis cs:29172000,0.32305342368348)
--(axis cs:30972000,0.36004286471831)
--(axis cs:33132000,0.362403282418709)
--(axis cs:34932000,0.373970550379176)
--(axis cs:37092000,0.357749859161904)
--(axis cs:38892000,0.379078751441466)
--(axis cs:41052000,0.404032522733684)
--(axis cs:42852000,0.416759710805852)
--(axis cs:44652000,0.405654158149027)
--(axis cs:46812000,0.4468979212461)
--(axis cs:48612000,0.423559259220556)
--(axis cs:50772000,0.451133472846751)
--(axis cs:52572000,0.432873342209938)
--(axis cs:54732000,0.442751265247809)
--(axis cs:56532000,0.466204136118918)
--(axis cs:58332000,0.425573852292896)
--(axis cs:60492000,0.447112256085405)
--(axis cs:62292000,0.450034441876543)
--(axis cs:64452000,0.458344645071731)
--(axis cs:66252000,0.443351592380203)
--(axis cs:68412000,0.462507029741714)
--(axis cs:70212000,0.464979248768343)
--(axis cs:72012000,0.474175983893069)
--(axis cs:74172000,0.494312744287029)
--(axis cs:75972000,0.472102330800584)
--(axis cs:78132000,0.477390863304379)
--(axis cs:79932000,0.456801589230822)
--(axis cs:82092000,0.503065124579882)
--(axis cs:83892000,0.469864588438713)
--(axis cs:85692000,0.47192300891837)
--(axis cs:87852000,0.49864652920511)
--(axis cs:89652000,0.503520700815582)
--(axis cs:91812000,0.503579229099326)
--(axis cs:93612000,0.500920479089709)
--(axis cs:95772000,0.501545085184958)
--(axis cs:95772000,0.538791598289396)
--(axis cs:95772000,0.538791598289396)
--(axis cs:93612000,0.529909943740124)
--(axis cs:91812000,0.533928447835496)
--(axis cs:89652000,0.531123641809797)
--(axis cs:87852000,0.522138800596628)
--(axis cs:85692000,0.505712620241837)
--(axis cs:83892000,0.507811291897097)
--(axis cs:82092000,0.533518092104612)
--(axis cs:79932000,0.486127302103765)
--(axis cs:78132000,0.514000174514201)
--(axis cs:75972000,0.504666412388108)
--(axis cs:74172000,0.511755596529201)
--(axis cs:72012000,0.504111361282158)
--(axis cs:70212000,0.505717948794159)
--(axis cs:68412000,0.492876526043833)
--(axis cs:66252000,0.465277494217278)
--(axis cs:64452000,0.477428343677281)
--(axis cs:62292000,0.485169605502269)
--(axis cs:60492000,0.485734193898797)
--(axis cs:58332000,0.465959363400662)
--(axis cs:56532000,0.486619504912191)
--(axis cs:54732000,0.471755916968008)
--(axis cs:52572000,0.457319050662092)
--(axis cs:50772000,0.47237784238103)
--(axis cs:48612000,0.453329143987737)
--(axis cs:46812000,0.471064556572896)
--(axis cs:44652000,0.4353356669784)
--(axis cs:42852000,0.436602208804358)
--(axis cs:41052000,0.433044355144253)
--(axis cs:38892000,0.402902847361728)
--(axis cs:37092000,0.383516087263094)
--(axis cs:34932000,0.390465921102913)
--(axis cs:33132000,0.387328713019082)
--(axis cs:30972000,0.375737861612642)
--(axis cs:29172000,0.346801352803198)
--(axis cs:27372000,0.337895140396137)
--(axis cs:25212000,0.32234390012787)
--(axis cs:23412000,0.322215936809394)
--(axis cs:21252000,0.329837153216668)
--(axis cs:19452000,0.284085420988416)
--(axis cs:17292000,0.283389833793392)
--(axis cs:15492000,0.267715397000722)
--(axis cs:13692000,0.25145961027267)
--(axis cs:11532000,0.225872380086093)
--(axis cs:9732000,0.20528391482903)
--(axis cs:7572000,0.198215479897685)
--(axis cs:5772000,0.230151601224347)
--(axis cs:3612000,0.193743134677788)
--(axis cs:1812000,0.161572266602889)
--(axis cs:12000,0.324999988079071)
--cycle;

\addplot [line width=\linewidthothers, \whole_color, mark=*, mark size=0, mark options={solid}]
table {%
1812000 0.154565431759693
3612000 0.188639068193879
5772000 0.225593892575773
7572000 0.192313290172562
9732000 0.199406544475402
11532000 0.217326462540461
13692000 0.245250928224292
15492000 0.264163737537711
17292000 0.274682014477988
19452000 0.272321312950199
21252000 0.318353962611947
23412000 0.308953681018646
25212000 0.313804083428329
27372000 0.325461130216851
29172000 0.335085064285471
30972000 0.368047944300397
33132000 0.377491470994319
34932000 0.378264322903242
37092000 0.373640733592503
38892000 0.393367030332027
41052000 0.420230236701106
42852000 0.425994306948564
44652000 0.42155167434189
46812000 0.456561085291788
48612000 0.44095900994931
50772000 0.463922891320284
52572000 0.450350124570781
54732000 0.465020122184
56532000 0.479805057686895
58332000 0.449755476286539
60492000 0.47304095487057
62292000 0.475307145537013
64452000 0.469642258676745
66252000 0.455739514541054
68412000 0.472657120971643
70212000 0.487231472250967
72012000 0.487979888966219
74172000 0.504619864260246
75972000 0.492682239771986
78132000 0.496620979374086
79932000 0.46908386017399
82092000 0.518474452657861
83892000 0.490639652589665
85692000 0.489756306346791
87852000 0.516682448200676
89652000 0.521337243734227
91812000 0.52428358634652
93612000 0.517580145920422
95772000 0.51710911605645
};

\path [draw=C2, fill=C2, opacity=0.2]
(axis cs:121600,0)
--(axis cs:121600,0)
--(axis cs:3283200,0)
--(axis cs:6444800,0)
--(axis cs:9606400,0)
--(axis cs:12889600,0)
--(axis cs:16051200,0)
--(axis cs:19212800,0)
--(axis cs:22374400,0.0460526315789474)
--(axis cs:25657600,0.0657894736842105)
--(axis cs:28819200,0.0263157894736842)
--(axis cs:31980800,0.111842105263158)
--(axis cs:35142400,0.171052631578947)
--(axis cs:38425600,0.131578947368421)
--(axis cs:41587200,0.0921052631578947)
--(axis cs:44748800,0.171052631578947)
--(axis cs:47910400,0.184210526315789)
--(axis cs:51193600,0.210526315789474)
--(axis cs:54355200,0.157894736842105)
--(axis cs:57516800,0.131578947368421)
--(axis cs:60800000,0.184210526315789)
--(axis cs:60800000,0.230263157894737)
--(axis cs:60800000,0.230263157894737)
--(axis cs:57516800,0.25)
--(axis cs:54355200,0.210526315789474)
--(axis cs:51193600,0.302631578947368)
--(axis cs:47910400,0.282894736842105)
--(axis cs:44748800,0.256578947368421)
--(axis cs:41587200,0.157894736842105)
--(axis cs:38425600,0.210526315789474)
--(axis cs:35142400,0.276315789473684)
--(axis cs:31980800,0.164638157894734)
--(axis cs:28819200,0.0986842105263158)
--(axis cs:25657600,0.157894736842105)
--(axis cs:22374400,0.105263157894737)
--(axis cs:19212800,0.0197368421052632)
--(axis cs:16051200,0)
--(axis cs:12889600,0.0197368421052632)
--(axis cs:9606400,0)
--(axis cs:6444800,0)
--(axis cs:3283200,0)
--(axis cs:121600,0)
--cycle;

\addplot [line width=\linewidthothers, C2, mark=*, mark size=0, mark options={solid}]
table {%
121600 0
3283200 0
6444800 0
9606400 0
12889600 0
16051200 0
19212800 0
22374400 0.0855263157894737
25657600 0.111842105263158
28819200 0.0526315789473684
31980800 0.131578947368421
35142400 0.217105263157895
38425600 0.177631578947368
41587200 0.125
44748800 0.217105263157895
47910400 0.210526315789474
51193600 0.256578947368421
54355200 0.184210526315789
57516800 0.210526315789474
60800000 0.210526315789474
};

\path [draw=C5, fill=C5, opacity=0.2]
(axis cs:121600,0.105263157894737)
--(axis cs:121600,0.0460526315789474)
--(axis cs:3283200,0)
--(axis cs:6444800,0)
--(axis cs:9606400,0)
--(axis cs:12889600,0.0129934210526323)
--(axis cs:16051200,0.0526315789473684)
--(axis cs:19212800,0.0526315789473684)
--(axis cs:22374400,0.105263157894737)
--(axis cs:25657600,0.0855263157894737)
--(axis cs:28819200,0.0394736842105263)
--(axis cs:31980800,0.111842105263158)
--(axis cs:35142400,0.144736842105263)
--(axis cs:38425600,0.0921052631578947)
--(axis cs:41587200,0.105263157894737)
--(axis cs:44748800,0.157894736842105)
--(axis cs:47910400,0.203947368421053)
--(axis cs:51193600,0.25)
--(axis cs:54355200,0.131578947368421)
--(axis cs:57516800,0.0986842105263158)
--(axis cs:60800000,0.125)
--(axis cs:60800000,0.236842105263158)
--(axis cs:60800000,0.236842105263158)
--(axis cs:57516800,0.236842105263158)
--(axis cs:54355200,0.177631578947368)
--(axis cs:51193600,0.375)
--(axis cs:47910400,0.302796052631577)
--(axis cs:44748800,0.289473684210526)
--(axis cs:41587200,0.164473684210526)
--(axis cs:38425600,0.184210526315789)
--(axis cs:35142400,0.269736842105263)
--(axis cs:31980800,0.177631578947368)
--(axis cs:28819200,0.105263157894737)
--(axis cs:25657600,0.151315789473684)
--(axis cs:22374400,0.171052631578947)
--(axis cs:19212800,0.111842105263158)
--(axis cs:16051200,0.171052631578947)
--(axis cs:12889600,0.0592105263157895)
--(axis cs:9606400,0)
--(axis cs:6444800,0)
--(axis cs:3283200,0)
--(axis cs:121600,0.105263157894737)
--cycle;

\addplot [line width=\linewidthothers, C5, mark=*, mark size=0, mark options={solid}]
table {%
121600 0.0657894736842105
3283200 0
6444800 0
9606400 0
12889600 0.0328947368421053
16051200 0.0986842105263158
19212800 0.0855263157894737
22374400 0.144736842105263
25657600 0.118421052631579
28819200 0.0657894736842105
31980800 0.144736842105263
35142400 0.236842105263158
38425600 0.138157894736842
41587200 0.125
44748800 0.223684210526316
47910400 0.25
51193600 0.315789473684211
54355200 0.151315789473684
57516800 0.171052631578947
60800000 0.164473684210526
};

\path [draw=C4, fill=C4, opacity=0.2]
(axis cs:10400,0)
--(axis cs:10400,0)
--(axis cs:278000,0)
--(axis cs:547200,0)
--(axis cs:814800,0)
--(axis cs:1088400,0)
--(axis cs:1358400,0)
--(axis cs:1631200,0)
--(axis cs:1902400,0)
--(axis cs:2177200,0)
--(axis cs:2448400,0)
--(axis cs:2720000,0)
--(axis cs:2994400,0)
--(axis cs:3265600,0)
--(axis cs:3531600,0)
--(axis cs:3808000,0)
--(axis cs:4082800,0)
--(axis cs:4352000,0)
--(axis cs:4630800,0)
--(axis cs:4904000,0)
--(axis cs:5174400,0)
--(axis cs:5174400,0.0375)
--(axis cs:5174400,0.0375)
--(axis cs:4904000,0)
--(axis cs:4630800,0.0125)
--(axis cs:4352000,0.0125)
--(axis cs:4082800,0)
--(axis cs:3808000,0)
--(axis cs:3531600,0)
--(axis cs:3265600,0)
--(axis cs:2994400,0.0125)
--(axis cs:2720000,0)
--(axis cs:2448400,0)
--(axis cs:2177200,0)
--(axis cs:1902400,0)
--(axis cs:1631200,0)
--(axis cs:1358400,0)
--(axis cs:1088400,0)
--(axis cs:814800,0)
--(axis cs:547200,0)
--(axis cs:278000,0)
--(axis cs:10400,0)
--cycle;

\addplot [line width=\linewidthothers, C4, mark=*, mark size=0, mark options={solid}]
table {%
10400 0
278000 0
547200 0
814800 0
1088400 0
1358400 0
1631200 0
1902400 0
2177200 0
2448400 0
2720000 0
2994400 0
3265600 0
3531600 0
3808000 0
4082800 0
4352000 0
4630800 0
4904000 0
5174400 0
};

\path [draw=C4, fill=C4, opacity=0.2]
(axis cs:10400,0)
--(axis cs:10400,0)
--(axis cs:281200,0)
--(axis cs:550800,0)
--(axis cs:820400,0)
--(axis cs:1088800,0)
--(axis cs:1360800,0)
--(axis cs:1627600,0)
--(axis cs:1898400,0)
--(axis cs:2171200,0)
--(axis cs:2438000,0)
--(axis cs:2704800,0)
--(axis cs:2975200,0)
--(axis cs:3249600,0)
--(axis cs:3522000,0)
--(axis cs:3795600,0)
--(axis cs:4064800,0)
--(axis cs:4336400,0)
--(axis cs:4605600,0)
--(axis cs:4876800,0)
--(axis cs:5149600,0)
--(axis cs:5149600,0.0375)
--(axis cs:5149600,0.0375)
--(axis cs:4876800,0.0125)
--(axis cs:4605600,0)
--(axis cs:4336400,0)
--(axis cs:4064800,0)
--(axis cs:3795600,0.0125)
--(axis cs:3522000,0)
--(axis cs:3249600,0)
--(axis cs:2975200,0)
--(axis cs:2704800,0.0125)
--(axis cs:2438000,0)
--(axis cs:2171200,0)
--(axis cs:1898400,0)
--(axis cs:1627600,0)
--(axis cs:1360800,0)
--(axis cs:1088800,0)
--(axis cs:820400,0)
--(axis cs:550800,0)
--(axis cs:281200,0)
--(axis cs:10400,0)
--cycle;

\addplot [line width=\linewidthothers, C4, mark=*, mark size=0, mark options={solid}]
table {%
10400 0
281200 0
550800 0
820400 0
1088800 0
1360800 0
1627600 0
1898400 0
2171200 0
2438000 0
2704800 0
2975200 0
3249600 0
3522000 0
3795600 0
4064800 0
4336400 0
4605600 0
4876800 0
5149600 0
};
\end{axis}

\end{tikzpicture}

%% file: images/tex/dual_arm_nm_collision_counts.tex
%
%

\begin{tikzpicture}
\def\res_color{C0}
\def\replace_color{C1}
\def\whole_color{C3}
\def\bbrl_color{C9}
\def\ppo_color{C2}
\def\sac_color{C4}
\def\linewidthtop{1mm}
\def\linewidthothers{0.5mm}

\begin{axis}[
legend cell align={left},
legend style={fill opacity=0.8, draw opacity=1, text opacity=1, draw=lightgray204, at={(0.03,0.03)},  anchor=north west},
tick align=outside,
tick pos=left,
x grid style={white},
scaled x ticks=false,
xticklabels={,0,1,2,3,4},
xlabel={Environment Interactions ($\times 10^7$)},
xmajorgrids,
xmin=-1500000, xmax=41000000.05,
xtick style={color=black},
y grid style={white},
ylabel={Collision Counts, IQM},
ymajorgrids,
ymin=0., ymax=40.0,
ytick style={color=black},
axis background/.style={fill=plot_background},
label style={font=\large},
tick label style={font=\large},
x axis line style={draw=none},
y axis line style={draw=none},
]
\path [draw=\bbrl_color, fill=\bbrl_color, opacity=0.2]
(axis cs:19300,35.6940793991089)
--(axis cs:19300,34.5016441345215)
--(axis cs:2430700,38.4147818088531)
--(axis cs:4960400,35.6636493930593)
--(axis cs:7548700,34.0068439370953)
--(axis cs:10161900,31.0299669099477)
--(axis cs:12791200,29.0874104691648)
--(axis cs:15436300,28.9765953986804)
--(axis cs:18095500,25.5349196438496)
--(axis cs:20761400,23.8293785558747)
--(axis cs:23438300,22.5000266700151)
--(axis cs:26121000,21.8779041782589)
--(axis cs:28808500,20.8921058909438)
--(axis cs:31499200,20.146616057803)
--(axis cs:34190600,19.8061419037369)
--(axis cs:36886000,18.17761160736)
--(axis cs:39583800,17.9067545587515)
--(axis cs:42290900,16.3884629819908)
--(axis cs:45671800,16.8997941111179)
--(axis cs:48377100,17.7495577012679)
--(axis cs:51088600,16.1878209555303)
--(axis cs:53795700,16.2843882220064)
--(axis cs:56501800,15.662875298013)
--(axis cs:59206600,14.6488551334743)
--(axis cs:61911900,14.3656805565513)
--(axis cs:64618400,13.4787107926992)
--(axis cs:67325200,15.1716990383697)
--(axis cs:70036700,14.9443252629051)
--(axis cs:72746500,14.9474898595022)
--(axis cs:75455300,13.8838713210353)
--(axis cs:78164400,13.1745693219734)
--(axis cs:80877400,14.9665678403295)
--(axis cs:83590700,13.3789681636435)
--(axis cs:86305800,13.7053054385776)
--(axis cs:89697300,13.2218489843099)
--(axis cs:92415500,13.2466547716923)
--(axis cs:95129700,12.9331207068663)
--(axis cs:97842900,13.0090490863346)
--(axis cs:100559400,12.8246959946882)
--(axis cs:103278000,12.3191251161573)
--(axis cs:105994800,11.5338340427621)
--(axis cs:108711500,12.606440428883)
--(axis cs:111431500,12.5675089405194)
--(axis cs:114150500,11.8603096397806)
--(axis cs:116864900,12.4655613298564)
--(axis cs:119579300,12.2023895128571)
--(axis cs:122294200,12.4134944253584)
--(axis cs:125012000,11.0037324601908)
--(axis cs:127726500,12.0255930161431)
--(axis cs:130444800,11.4778176203858)
--(axis cs:133843100,12.2221170015637)
--(axis cs:133843100,12.9375662734369)
--(axis cs:133843100,12.9375662734369)
--(axis cs:130444800,12.395902208304)
--(axis cs:127726500,13.1125121014189)
--(axis cs:125012000,11.7724797884599)
--(axis cs:122294200,13.2795342445268)
--(axis cs:119579300,13.3716626505219)
--(axis cs:116864900,13.2239100991954)
--(axis cs:114150500,12.8677584550238)
--(axis cs:111431500,13.1889160257072)
--(axis cs:108711500,13.2879807354949)
--(axis cs:105994800,12.7957607931992)
--(axis cs:103278000,12.9697912998703)
--(axis cs:100559400,13.3540590846659)
--(axis cs:97842900,13.7211225717473)
--(axis cs:95129700,13.6342647550472)
--(axis cs:92415500,14.2119088474561)
--(axis cs:89697300,13.7357393835292)
--(axis cs:86305800,14.5357687164706)
--(axis cs:83590700,14.1339145429715)
--(axis cs:80877400,15.7088184488112)
--(axis cs:78164400,13.7456115166469)
--(axis cs:75455300,14.6933293547598)
--(axis cs:72746500,15.7330733241519)
--(axis cs:70036700,15.7099019421846)
--(axis cs:67325200,16.1373988788931)
--(axis cs:64618400,14.8561639277463)
--(axis cs:61911900,15.2646459741622)
--(axis cs:59206600,15.4769387476701)
--(axis cs:56501800,16.3407993011938)
--(axis cs:53795700,17.3621695921626)
--(axis cs:51088600,17.4495335370087)
--(axis cs:48377100,19.1709808740092)
--(axis cs:45671800,17.7706327187835)
--(axis cs:42290900,17.6659873460088)
--(axis cs:39583800,19.2785721305139)
--(axis cs:36886000,19.4244121705088)
--(axis cs:34190600,21.0313205530826)
--(axis cs:31499200,21.8008346603828)
--(axis cs:28808500,22.0107367385754)
--(axis cs:26121000,23.4892249891429)
--(axis cs:23438300,24.456036522638)
--(axis cs:20761400,25.5277777958725)
--(axis cs:18095500,26.707279392167)
--(axis cs:15436300,30.3420870131258)
--(axis cs:12791200,30.5531967558818)
--(axis cs:10161900,31.9190534541995)
--(axis cs:7548700,35.2962125227787)
--(axis cs:4960400,37.0838918983936)
--(axis cs:2430700,38.9637134075165)
--(axis cs:19300,35.6940793991089)
--cycle;

\addplot [line width=\linewidthothers, \bbrl_color, mark=*, mark size=0, mark options={solid}]
table {%
19300 35.050986289978
2430700 38.742855489254
4960400 36.5305049680173
7548700 34.6189386255573
10161900 31.4576881050307
12791200 29.4929668202331
15436300 29.421800683432
18095500 26.1413122843107
20761400 24.7037805883313
23438300 23.6360590810283
26121000 22.7344698565965
28808500 21.4389083647707
31499200 21.2491658327515
34190600 20.5054706049464
36886000 18.6386027317722
39583800 18.6128247548243
42290900 17.2415326943856
45671800 17.4857354847473
48377100 18.4993705093527
51088600 16.8366179127225
53795700 16.8050640800268
56501800 15.9605663228405
59206600 15.1502967317308
61911900 14.9053471272942
64618400 14.2558035851598
67325200 15.6258253008194
70036700 15.379317477528
72746500 15.4363884511986
75455300 14.1706747237804
78164400 13.4024230193593
80877400 15.3514774364202
83590700 13.7259383939845
86305800 14.0879758149121
89697300 13.4414410521631
92415500 13.7174542094894
95129700 13.2387657323581
97842900 13.3154983381375
100559400 12.9763962820658
103278000 12.6443886387582
105994800 12.1189838335815
108711500 12.8477704383853
111431500 12.8423644687055
114150500 12.2094002349401
116864900 12.8680612703101
119579300 12.6476816293259
122294200 12.6494158560953
125012000 11.3674138495671
127726500 12.6114268521991
130444800 11.9115128831581
133843100 12.5008408572021
};

\path [draw=\replace_color, fill=\replace_color, opacity=0.2]
(axis cs:11400,16.265625)
--(axis cs:11400,15.6652960777283)
--(axis cs:1379400,20.8491467535496)
--(axis cs:2747400,19.3878426793963)
--(axis cs:4115400,19.7294995791744)
--(axis cs:5483400,18.4398919642626)
--(axis cs:6851400,19.3189327528248)
--(axis cs:8219400,19.8600983177107)
--(axis cs:9587400,20.0306287777016)
--(axis cs:10955400,19.6713675297654)
--(axis cs:12323400,18.285458444775)
--(axis cs:13691400,18.0073163652217)
--(axis cs:15059400,17.1922193635034)
--(axis cs:16427400,16.4637717536381)
--(axis cs:17795400,16.5006485390692)
--(axis cs:19163400,15.7473939945023)
--(axis cs:20531400,16.0741114609818)
--(axis cs:21899400,14.5371244679066)
--(axis cs:23609400,13.9289643019287)
--(axis cs:24977400,13.9642563968562)
--(axis cs:26345400,14.2477614432216)
--(axis cs:27713400,13.6824821231235)
--(axis cs:29081400,13.159731851969)
--(axis cs:30449400,11.6563956667264)
--(axis cs:31817400,10.7790439145118)
--(axis cs:33185400,11.4913299829783)
--(axis cs:34553400,11.4905318468884)
--(axis cs:35921400,10.9573961601353)
--(axis cs:37289400,11.7310843175985)
--(axis cs:38657400,11.5901885146072)
--(axis cs:40025400,10.0380041530705)
--(axis cs:41393400,10.9840608056853)
--(axis cs:42761400,10.4670130974366)
--(axis cs:44129400,10.4577924351818)
--(axis cs:45839400,10.6508611682307)
--(axis cs:47207400,10.4374118509541)
--(axis cs:48575400,9.72251773590132)
--(axis cs:49943400,9.65856673266378)
--(axis cs:51311400,8.52295363932485)
--(axis cs:52679400,8.75539747255591)
--(axis cs:54047400,8.60948387657802)
--(axis cs:55415400,9.42721644162496)
--(axis cs:56783400,8.93756546241766)
--(axis cs:58151400,8.71177986795433)
--(axis cs:59519400,8.79742403053418)
--(axis cs:60887400,8.17039381459672)
--(axis cs:62255400,9.11288756989891)
--(axis cs:63623400,8.26381731875093)
--(axis cs:64991400,8.64042056034792)
--(axis cs:66359400,8.02833026910457)
--(axis cs:68069400,8.30487367467429)
--(axis cs:68069400,8.93419665406838)
--(axis cs:68069400,8.93419665406838)
--(axis cs:66359400,8.80819682253785)
--(axis cs:64991400,9.40648034785379)
--(axis cs:63623400,8.96665498807013)
--(axis cs:62255400,9.90976729856839)
--(axis cs:60887400,8.51493928326607)
--(axis cs:59519400,9.40146073357989)
--(axis cs:58151400,9.32432274311856)
--(axis cs:56783400,9.66644640796146)
--(axis cs:55415400,10.1889135944043)
--(axis cs:54047400,9.24441472782789)
--(axis cs:52679400,9.7124667595081)
--(axis cs:51311400,9.56266940256446)
--(axis cs:49943400,10.4713208421704)
--(axis cs:48575400,10.2755052859988)
--(axis cs:47207400,11.1017123588168)
--(axis cs:45839400,11.631310542557)
--(axis cs:44129400,11.5700134051831)
--(axis cs:42761400,11.3419348476495)
--(axis cs:41393400,11.8030448557211)
--(axis cs:40025400,11.1419233737371)
--(axis cs:38657400,12.7537095160202)
--(axis cs:37289400,12.5685652516968)
--(axis cs:35921400,11.9242311255821)
--(axis cs:34553400,12.4316139167686)
--(axis cs:33185400,12.2407105912918)
--(axis cs:31817400,11.7922373908978)
--(axis cs:30449400,12.7371952942311)
--(axis cs:29081400,13.9211219799279)
--(axis cs:27713400,14.1715435163992)
--(axis cs:26345400,15.1074258242835)
--(axis cs:24977400,14.7006210488549)
--(axis cs:23609400,14.5702798378899)
--(axis cs:21899400,15.3406001684147)
--(axis cs:20531400,17.1167004617911)
--(axis cs:19163400,17.1278279302686)
--(axis cs:17795400,17.8203549340172)
--(axis cs:16427400,17.868229758939)
--(axis cs:15059400,17.8645416916487)
--(axis cs:13691400,19.0478773134356)
--(axis cs:12323400,19.3583239129598)
--(axis cs:10955400,20.4102069543736)
--(axis cs:9587400,20.5227372352533)
--(axis cs:8219400,20.672887083448)
--(axis cs:6851400,20.4076625874181)
--(axis cs:5483400,19.2782227959324)
--(axis cs:4115400,20.5030889966292)
--(axis cs:2747400,20.265075947158)
--(axis cs:1379400,22.2291322052479)
--(axis cs:11400,16.265625)
--cycle;

\addplot [line width=\linewidthothers, \replace_color, mark=*, mark size=0, mark options={solid}]
table {%
11400 15.976973772049
1379400 21.4941921979189
2747400 19.9322125762701
4115400 19.9549637068412
5483400 18.9146056050704
6851400 19.8893482975329
8219400 20.3660019886772
9587400 20.3116816639254
10955400 20.3283122694994
12323400 18.9510295046273
13691400 18.4902574249413
15059400 17.4089098638338
16427400 17.2892731029155
17795400 17.321122129748
19163400 16.3884820119202
20531400 16.6016032437256
21899400 14.9886967121486
23609400 14.1684165825679
24977400 14.2593987431537
26345400 14.6136863299254
27713400 13.9819704749598
29081400 13.5522172674818
30449400 12.2194835336929
31817400 11.3433402566808
33185400 11.7380797472389
34553400 11.7132152379106
35921400 11.3781551719687
37289400 12.1125333012946
38657400 12.0103091848156
40025400 10.304522892279
41393400 11.2138977887622
42761400 10.9443414422468
44129400 10.9796724127149
45839400 11.2851121871297
47207400 10.7662311722468
48575400 9.94524744187713
49943400 10.0106609780558
51311400 9.06273766719557
52679400 9.28293372067769
54047400 9.0924472020929
55415400 9.77104320629726
56783400 9.35597246939908
58151400 9.07312848009574
59519400 9.17562093888274
60887400 8.34872636741573
62255400 9.58997093570087
63623400 8.64013056611933
64991400 9.18168882567627
66359400 8.49614161503746
68069400 8.67492688456586
};

\path [draw=\res_color, fill=\res_color, opacity=0.2]
--(axis cs:11400,15.6883225440979)
--(axis cs:1379400,18.6024875044823)
--(axis cs:2747400,20.3329406753182)
--(axis cs:4115400,21.5414321882417)
--(axis cs:5483400,19.651241739517)
--(axis cs:6851400,18.7678288950515)
--(axis cs:8219400,18.1304388641076)
--(axis cs:9587400,16.0423356099892)
--(axis cs:10955400,14.5982410542262)
--(axis cs:12323400,12.3588551892313)
--(axis cs:13691400,10.2764221539534)
--(axis cs:15059400,9.04218215726036)
--(axis cs:16427400,8.64869215345277)
--(axis cs:17795400,7.69242528954021)
--(axis cs:19163400,7.15188063829578)
--(axis cs:20531400,6.19721365760754)
--(axis cs:21899400,5.70726883706423)
--(axis cs:23609400,5.54449611654603)
--(axis cs:24977400,5.8414659854456)
--(axis cs:26345400,5.37500917960512)
--(axis cs:27713400,5.33667055699211)
--(axis cs:29081400,4.5139129829417)
--(axis cs:30449400,4.11063652230566)
--(axis cs:31817400,4.00504298499398)
--(axis cs:33185400,3.9404093157894)
--(axis cs:34553400,3.82380693332769)
--(axis cs:35921400,4.19498469244326)
--(axis cs:37289400,3.87911544820609)
--(axis cs:38657400,4.6933413749794)
--(axis cs:40025400,2.8805743789338)
--(axis cs:41393400,4.52977983603528)
--(axis cs:42761400,3.82039598336496)
--(axis cs:44129400,3.88031190912398)
--(axis cs:45839400,3.75997461005943)
--(axis cs:47207400,3.425202773162)
--(axis cs:48575400,3.13955514880922)
--(axis cs:49943400,3.15428453605644)
--(axis cs:51311400,2.64303346879849)
--(axis cs:52679400,3.02365973118801)
--(axis cs:54047400,3.0639409715388)
--(axis cs:55415400,3.50385887074042)
--(axis cs:56783400,3.42529075365201)
--(axis cs:58151400,2.8569947430516)
--(axis cs:59519400,3.02369777947565)
--(axis cs:60887400,2.65975625448561)
--(axis cs:62255400,2.99952596496063)
--(axis cs:63623400,2.52640040533008)
--(axis cs:64991400,2.94372247043408)
--(axis cs:66359400,2.53224365808743)
--(axis cs:68069400,2.85135827378704)
--(axis cs:68069400,3.14955327791438)
--(axis cs:68069400,3.14955327791438)
--(axis cs:66359400,2.82502679032987)
--(axis cs:64991400,3.48015733961543)
--(axis cs:63623400,3.08035470300747)
--(axis cs:62255400,3.48007050074825)
--(axis cs:60887400,3.22810610598195)
--(axis cs:59519400,3.56584587807388)
--(axis cs:58151400,3.21422813501227)
--(axis cs:56783400,3.79591435609614)
--(axis cs:55415400,3.90715583035318)
--(axis cs:54047400,3.4547449274579)
--(axis cs:52679400,3.45246357684829)
--(axis cs:51311400,3.0398067270574)
--(axis cs:49943400,3.6119268652551)
--(axis cs:48575400,3.46311550561594)
--(axis cs:47207400,3.82120075110056)
--(axis cs:45839400,4.30642969390224)
--(axis cs:44129400,4.10690535249118)
--(axis cs:42761400,4.04863303055209)
--(axis cs:41393400,4.97495890363498)
--(axis cs:40025400,3.28346892595486)
--(axis cs:38657400,5.06030544048589)
--(axis cs:37289400,4.55550224168785)
--(axis cs:35921400,4.59605568907547)
--(axis cs:34553400,4.37777586497354)
--(axis cs:33185400,4.48612004974071)
--(axis cs:31817400,4.28785701455887)
--(axis cs:30449400,4.57827996647877)
--(axis cs:29081400,5.21645766072403)
--(axis cs:27713400,6.01139050772179)
--(axis cs:26345400,5.96775668742208)
--(axis cs:24977400,6.47223819849412)
--(axis cs:23609400,5.97700896417422)
--(axis cs:21899400,6.35209279647905)
--(axis cs:20531400,7.31564314610218)
--(axis cs:19163400,8.09392391594393)
--(axis cs:17795400,8.90245420966828)
--(axis cs:16427400,9.98869529528114)
--(axis cs:15059400,10.5898256690649)
--(axis cs:13691400,11.6722698144885)
--(axis cs:12323400,13.5783463568314)
--(axis cs:10955400,15.2036444938267)
--(axis cs:9587400,17.0209303985097)
--(axis cs:8219400,18.8876515322853)
--(axis cs:6851400,19.800366106798)
--(axis cs:5483400,20.2852314858355)
--(axis cs:4115400,22.1058183701825)
--(axis cs:2747400,21.112902129069)
--(axis cs:1379400,19.4262949079275)
--(axis cs:11400,16.2442438602448)
--cycle;

\addplot [line width=\linewidthtop, \res_color, mark=*, mark size=0, mark options={solid}]
table {%
11400 15.9629936218262
1379400 19.0416325032711
2747400 20.9556532446295
4115400 21.8402075503254
5483400 19.9679361847811
6851400 19.1386002539657
8219400 18.410326228132
9587400 16.4836382399326
10955400 14.9600744412704
12323400 13.0547427771862
13691400 11.0124170873425
15059400 9.78895109985427
16427400 9.2633307986546
17795400 8.44940596990067
19163400 7.69803728947613
20531400 6.71724669550394
21899400 6.1175665494264
23609400 5.71897036466682
24977400 6.15665862198733
26345400 5.69568463121971
27713400 5.69500806937609
29081400 4.79849391205635
30449400 4.30859211026599
31817400 4.14871040196245
33185400 4.11419385167362
34553400 4.1654580465831
35921400 4.39990647392377
37289400 4.17247247374192
38657400 4.90274422428252
40025400 3.08222312315973
41393400 4.71848170467817
42761400 3.92801490144375
44129400 3.96610557222857
45839400 3.93450503327983
47207400 3.62020705287761
48575400 3.29398492877578
49943400 3.39901997069129
51311400 2.90428884646489
52679400 3.24827181859294
54047400 3.19085702706876
55415400 3.68608683686167
56783400 3.56166098287575
58151400 2.99721364896058
59519400 3.18555552586146
60887400 2.92489722396754
62255400 3.22320762303131
63623400 2.79326739976478
64991400 3.15885980007449
66359400 2.70639054790376
68069400 3.03685222710401
};

\path [draw=\whole_color, fill=\whole_color, opacity=0.2]
(axis cs:11400,16.4810857772827)
--(axis cs:11400,16.0353620052338)
--(axis cs:1379400,24.7221422642469)
--(axis cs:2747400,22.0613219625782)
--(axis cs:4115400,19.5140784143005)
--(axis cs:5483400,16.007981145769)
--(axis cs:6851400,14.2053324503158)
--(axis cs:8219400,13.2256504517325)
--(axis cs:9587400,11.6570699435403)
--(axis cs:10955400,10.2923592835732)
--(axis cs:12323400,9.42501291978114)
--(axis cs:13691400,8.94558473000769)
--(axis cs:15059400,7.96541415312874)
--(axis cs:16427400,7.43297869685978)
--(axis cs:17795400,7.47880027493046)
--(axis cs:19163400,7.0786503316215)
--(axis cs:20531400,6.41713035448626)
--(axis cs:21899400,6.1717549799015)
--(axis cs:23609400,5.80676553039042)
--(axis cs:24977400,6.32249454551779)
--(axis cs:26345400,6.02426781747108)
--(axis cs:27713400,5.90652814986493)
--(axis cs:29081400,5.60858512884804)
--(axis cs:30449400,4.73303322036425)
--(axis cs:31817400,4.93697827079433)
--(axis cs:33185400,4.86784915509209)
--(axis cs:34553400,4.80545090352934)
--(axis cs:35921400,5.36119406565868)
--(axis cs:37289400,5.30945620494756)
--(axis cs:38657400,5.39500814061066)
--(axis cs:40025400,4.02059429722923)
--(axis cs:41393400,4.90179219220604)
--(axis cs:42761400,4.59351247178322)
--(axis cs:44129400,4.27923891490452)
--(axis cs:45839400,4.67259148368428)
--(axis cs:47207400,4.33950318107788)
--(axis cs:48575400,3.9059495009502)
--(axis cs:49943400,4.33037476109911)
--(axis cs:51311400,3.97923015881665)
--(axis cs:52679400,4.21170156819873)
--(axis cs:54047400,4.11289293592719)
--(axis cs:55415400,4.53917711990341)
--(axis cs:56783400,4.29444339103895)
--(axis cs:58151400,3.88501300696884)
--(axis cs:59519400,4.09017753965082)
--(axis cs:60887400,3.81526537047089)
--(axis cs:62255400,4.29474889009718)
--(axis cs:63623400,3.79084642296145)
--(axis cs:64991400,4.24288382482747)
--(axis cs:66359400,3.35498412480219)
--(axis cs:68069400,3.84438526981706)
--(axis cs:68069400,4.09501563864216)
--(axis cs:68069400,4.09501563864216)
--(axis cs:66359400,3.52938118728848)
--(axis cs:64991400,4.54795811248578)
--(axis cs:63623400,4.2106375182314)
--(axis cs:62255400,4.62182329989011)
--(axis cs:60887400,4.0790583358126)
--(axis cs:59519400,4.37347831498285)
--(axis cs:58151400,4.18931605729578)
--(axis cs:56783400,4.75077484758026)
--(axis cs:55415400,4.86730917905385)
--(axis cs:54047400,4.49219076122742)
--(axis cs:52679400,4.64516147475486)
--(axis cs:51311400,4.42707866130618)
--(axis cs:49943400,4.79271237094866)
--(axis cs:48575400,4.36080334450729)
--(axis cs:47207400,4.80176333802885)
--(axis cs:45839400,5.13460363409438)
--(axis cs:44129400,4.82299433377547)
--(axis cs:42761400,5.07291947707656)
--(axis cs:41393400,5.34918123419517)
--(axis cs:40025400,4.4306698975765)
--(axis cs:38657400,5.73579000237141)
--(axis cs:37289400,5.94680007248739)
--(axis cs:35921400,5.74145595962085)
--(axis cs:34553400,5.29204765849504)
--(axis cs:33185400,5.2776359689767)
--(axis cs:31817400,5.4072768975969)
--(axis cs:30449400,5.33862265852261)
--(axis cs:29081400,6.12393579453392)
--(axis cs:27713400,6.42757056765953)
--(axis cs:26345400,6.5157527617051)
--(axis cs:24977400,7.00104129477478)
--(axis cs:23609400,6.50475706141452)
--(axis cs:21899400,6.69433082518481)
--(axis cs:20531400,7.44402037475968)
--(axis cs:19163400,7.64122152881615)
--(axis cs:17795400,8.04521168139822)
--(axis cs:16427400,8.03224690439842)
--(axis cs:15059400,8.34489658396117)
--(axis cs:13691400,9.20045105183042)
--(axis cs:12323400,9.89738190689135)
--(axis cs:10955400,10.7910745262804)
--(axis cs:9587400,12.1555279939246)
--(axis cs:8219400,13.7079201625811)
--(axis cs:6851400,14.9844617462581)
--(axis cs:5483400,16.5210440498464)
--(axis cs:4115400,20.0479233605438)
--(axis cs:2747400,23.0711188828573)
--(axis cs:1379400,25.5928248912096)
--(axis cs:11400,16.4810857772827)
--cycle;

\addplot [line width=\linewidthothers, \whole_color, mark=*, mark size=0, mark options={solid}]
table {%
11400 16.2458882331848
1379400 24.9631988555193
2747400 22.534545937553
4115400 19.5852394137764
5483400 16.2782177133631
6851400 14.686508655187
8219400 13.3505640644095
9587400 11.9154254478173
10955400 10.5830167342564
12323400 9.66374670650711
13691400 9.02286604964688
15059400 8.21323382721863
16427400 7.743678861383
17795400 7.78814622872671
19163400 7.33494657566223
20531400 7.07652190166674
21899400 6.49409885629334
23609400 6.22434791035463
24977400 6.69389948463734
26345400 6.21675085454348
27713400 6.20601279716077
29081400 5.86414597715391
30449400 5.17967734379412
31817400 5.19096869828977
33185400 5.02537679955502
34553400 5.04911504485697
35921400 5.54066880069697
37289400 5.63878648123639
38657400 5.54447108159887
40025400 4.19203875844007
41393400 5.08204518427961
42761400 4.72798965788789
44129400 4.58203956135144
45839400 4.90765574430411
47207400 4.52375054960238
48575400 4.0668585635681
49943400 4.58027693794593
51311400 4.15190010359159
52679400 4.3129228103272
54047400 4.26827276341687
55415400 4.72956365202143
56783400 4.48929443900798
58151400 3.98514505348886
59519400 4.22041310807122
60887400 3.91642355576787
62255400 4.43615305263361
63623400 3.98135122045523
64991400 4.35560615000467
66359400 3.47306397825548
68069400 3.96392675444307
};

\path [draw=C5, fill=C5, opacity=0.2]
(axis cs:121600,29.0131578947368)
--(axis cs:121600,25.6578947368421)
--(axis cs:3283200,11.125)
--(axis cs:6444800,3.80098684210527)
--(axis cs:9606400,5.11184210526316)
--(axis cs:12889600,1.82236842105263)
--(axis cs:16051200,1.58552631578947)
--(axis cs:19212800,1.84210526315789)
--(axis cs:22374400,1.97368421052632)
--(axis cs:25657600,1.22368421052632)
--(axis cs:28819200,1.60526315789474)
--(axis cs:31980800,1.13157894736842)
--(axis cs:35142400,0.921052631578947)
--(axis cs:38425600,1.06578947368421)
--(axis cs:41587200,1.10526315789474)
--(axis cs:44748800,0.960526315789474)
--(axis cs:47910400,1.15789473684211)
--(axis cs:51193600,0.980263157894737)
--(axis cs:54355200,1.18421052631579)
--(axis cs:57516800,1.11842105263158)
--(axis cs:60800000,1.13157894736842)
--(axis cs:60800000,1.67763157894737)
--(axis cs:60800000,1.67763157894737)
--(axis cs:57516800,1.43421052631579)
--(axis cs:54355200,1.73684210526316)
--(axis cs:51193600,1.42763157894737)
--(axis cs:47910400,1.65131578947368)
--(axis cs:44748800,1.76315789473684)
--(axis cs:41587200,1.70411184210526)
--(axis cs:38425600,1.58552631578947)
--(axis cs:35142400,1.27631578947368)
--(axis cs:31980800,1.73684210526316)
--(axis cs:28819200,2.23026315789474)
--(axis cs:25657600,1.5)
--(axis cs:22374400,4.88815789473684)
--(axis cs:19212800,2.19078947368421)
--(axis cs:16051200,2.16447368421053)
--(axis cs:12889600,2.90789473684211)
--(axis cs:9606400,6.42763157894737)
--(axis cs:6444800,7.82236842105263)
--(axis cs:3283200,14.7513157894737)
--(axis cs:121600,29.0131578947368)
--cycle;

\addplot [line width=\linewidthothers, C5, mark=*, mark size=0, mark options={solid}]
table {%
121600 27.1052631578947
3283200 13.2039473684211
6444800 5.34210526315789
9606400 5.86184210526316
12889600 2.28947368421053
16051200 1.76315789473684
19212800 1.94078947368421
22374400 3.24342105263158
25657600 1.35526315789474
28819200 1.98684210526316
31980800 1.375
35142400 1.11184210526316
38425600 1.28289473684211
41587200 1.375
44748800 1.30921052631579
47910400 1.35526315789474
51193600 1.22368421052632
54355200 1.50657894736842
57516800 1.24342105263158
60800000 1.34210526315789
};

\path [draw=C2, fill=C2, opacity=0.2]
(axis cs:121600,13.375)
--(axis cs:121600,12.0460526315789)
--(axis cs:3283200,9.61842105263158)
--(axis cs:6444800,4.28947368421053)
--(axis cs:9606400,2.18421052631579)
--(axis cs:12889600,1.19736842105263)
--(axis cs:16051200,1.40131578947368)
--(axis cs:19212800,1.51315789473684)
--(axis cs:22374400,1.75)
--(axis cs:25657600,1.11842105263158)
--(axis cs:28819200,1.23026315789474)
--(axis cs:31980800,1.01973684210526)
--(axis cs:35142400,1.07894736842105)
--(axis cs:38425600,0.921052631578947)
--(axis cs:41587200,1.28947368421053)
--(axis cs:44748800,0.782894736842105)
--(axis cs:47910400,0.848519736842106)
--(axis cs:51193600,0.763157894736842)
--(axis cs:54355200,0.789473684210526)
--(axis cs:57516800,0.697368421052632)
--(axis cs:60800000,0.848684210526316)
--(axis cs:60800000,1.09210526315789)
--(axis cs:60800000,1.09210526315789)
--(axis cs:57516800,1.05263157894737)
--(axis cs:54355200,0.921052631578947)
--(axis cs:51193600,1.44736842105263)
--(axis cs:47910400,1.57894736842105)
--(axis cs:44748800,1.28947368421053)
--(axis cs:41587200,2.47368421052632)
--(axis cs:38425600,1.15131578947368)
--(axis cs:35142400,1.46710526315789)
--(axis cs:31980800,1.58552631578947)
--(axis cs:28819200,1.48684210526316)
--(axis cs:25657600,1.63815789473684)
--(axis cs:22374400,2.28289473684211)
--(axis cs:19212800,1.84210526315789)
--(axis cs:16051200,1.76315789473684)
--(axis cs:12889600,2.04605263157895)
--(axis cs:9606400,3.97368421052632)
--(axis cs:6444800,7.48684210526316)
--(axis cs:3283200,13.0657894736842)
--(axis cs:121600,13.375)
--cycle;

\addplot [line width=\linewidthothers, C2, mark=*, mark size=0, mark options={solid}]
table {%
121600 12.3157894736842
3283200 11.4539473684211
6444800 6.14473684210526
9606400 3.03947368421053
12889600 1.61842105263158
16051200 1.48026315789474
19212800 1.69078947368421
22374400 2.01973684210526
25657600 1.32236842105263
28819200 1.36842105263158
31980800 1.21052631578947
35142400 1.15131578947368
38425600 1.02631578947368
41587200 1.57236842105263
44748800 0.934210526315789
47910400 1.01315789473684
51193600 1.03289473684211
54355200 0.861842105263158
57516800 0.888157894736842
60800000 1.01315789473684
};

\path [draw=C4, fill=C4, opacity=0.2]
(axis cs:10400,48.15)
--(axis cs:10400,11.925)
--(axis cs:278000,23.45)
--(axis cs:547200,25.475)
--(axis cs:814800,14.8246875)
--(axis cs:1088400,19.925)
--(axis cs:1358400,23.65)
--(axis cs:1631200,25.0625)
--(axis cs:1902400,25.8875)
--(axis cs:2177200,28.15)
--(axis cs:2448400,22.575)
--(axis cs:2720000,13.325)
--(axis cs:2994400,15.725)
--(axis cs:3265600,14.1625)
--(axis cs:3531600,6.8125)
--(axis cs:3808000,6.9)
--(axis cs:4082800,6.04812500000001)
--(axis cs:4352000,5.05)
--(axis cs:4630800,4.5)
--(axis cs:4904000,2.9625)
--(axis cs:5174400,3.725)
--(axis cs:5174400,6.1875)
--(axis cs:5174400,6.1875)
--(axis cs:4904000,5.9375)
--(axis cs:4630800,9.675)
--(axis cs:4352000,8.775)
--(axis cs:4082800,13.975)
--(axis cs:3808000,12.025)
--(axis cs:3531600,16.81375)
--(axis cs:3265600,22.175)
--(axis cs:2994400,24.075)
--(axis cs:2720000,21.7125)
--(axis cs:2448400,32.0375)
--(axis cs:2177200,45.975)
--(axis cs:1902400,34.925)
--(axis cs:1631200,34.2375)
--(axis cs:1358400,35.55)
--(axis cs:1088400,28.3875)
--(axis cs:814800,21.5625)
--(axis cs:547200,33.5125)
--(axis cs:278000,35.1784374999999)
--(axis cs:10400,48.15)
--cycle;

\addplot [line width=\linewidthothers, C4, mark=*, mark size=0, mark options={solid}]
table {%
10400 30.9375
278000 29.0125
547200 30.2375
814800 18.4375
1088400 24.675
1358400 28.7125
1631200 28.0875
1902400 30.55
2177200 38.1625
2448400 27.65
2720000 18.0625
2994400 19.2
3265600 17.3
3531600 11.5375
3808000 9.6625
4082800 9.45
4352000 7.1125
4630800 7.9375
4904000 4.4125
5174400 4.7
};

\path [draw=C6, fill=C6, opacity=0.2]
(axis cs:10400,42.35)
--(axis cs:10400,29.8375)
--(axis cs:281200,0)
--(axis cs:550800,0)
--(axis cs:820400,0)
--(axis cs:1088800,0)
--(axis cs:1360800,0)
--(axis cs:1627600,0)
--(axis cs:1898400,0)
--(axis cs:2171200,0)
--(axis cs:2438000,0)
--(axis cs:2704800,0)
--(axis cs:2975200,0)
--(axis cs:3249600,0)
--(axis cs:3522000,0)
--(axis cs:3795600,0)
--(axis cs:4064800,0)
--(axis cs:4336400,0)
--(axis cs:4605600,0)
--(axis cs:4876800,0)
--(axis cs:5149600,0)
--(axis cs:5149600,3.7)
--(axis cs:5149600,3.7)
--(axis cs:4876800,5.175)
--(axis cs:4605600,4.32562499999999)
--(axis cs:4336400,8.5)
--(axis cs:4064800,7.025)
--(axis cs:3795600,8.1625)
--(axis cs:3522000,5.2625)
--(axis cs:3249600,16.05)
--(axis cs:2975200,16.275)
--(axis cs:2704800,24.5)
--(axis cs:2438000,35.125)
--(axis cs:2171200,28.9)
--(axis cs:1898400,27.425)
--(axis cs:1627600,33.8384375)
--(axis cs:1360800,29.175)
--(axis cs:1088800,28.475)
--(axis cs:820400,34.0625)
--(axis cs:550800,35.0125)
--(axis cs:281200,37.1125)
--(axis cs:10400,42.35)
--cycle;

\addplot [line width=\linewidthothers, C6, mark=*, mark size=0, mark options={solid}]
table {%
10400 34.6375
281200 16.35
550800 13.7
820400 13.975
1088800 13.1625
1360800 13.125
1627600 14.6625
1898400 11.775
2171200 13.45
2438000 12.1375
2704800 10.6625
2975200 6.7875
3249600 5.4625
3522000 2.075
3795600 3.1125
4064800 2.7125
4336400 3.725
4605600 1.4375
4876800 2.35
5149600 1.45
};

\end{axis}

\end{tikzpicture}

%% file: images/tex/dual_arm_nm_dist2target.tex
%
%

\begin{tikzpicture}
\def\res_color{C0}
\def\replace_color{C1}
\def\whole_color{C3}
\def\bbrl_color{C9}
\def\ppo_color{C2}
\def\sac_color{C4}
\def\linewidthtop{1mm}
\def\linewidthothers{0.5mm}

\begin{axis}[
legend cell align={left},
legend style={fill opacity=0.8, draw opacity=1, text opacity=1, draw=lightgray204, at={(0.03,0.03)},  anchor=north west},
tick align=outside,
tick pos=left,
x grid style={white},
scaled x ticks=false,
xticklabels={,0,1,2,3,4},
xlabel={Environment Interactions ($\times 10^7$)},
xmajorgrids,
xmin=-1500000, xmax=41000000.05,
xtick style={color=black},
y grid style={white},
ylabel={Distance to Target, IQM},
ymajorgrids,
ymin=-0.15, ymax=1.5,
ytick style={color=black},
axis background/.style={fill=plot_background},
label style={font=\large},
tick label style={font=\large},
x axis line style={draw=none},
y axis line style={draw=none},
]
\path [draw=\bbrl_color, fill=\bbrl_color, opacity=0.2]
(axis cs:19300,0.886815249919891)
--(axis cs:19300,0.851892203092575)
--(axis cs:2430700,0.858034610748291)
--(axis cs:4960400,0.840493335151405)
--(axis cs:7548700,0.843841913199867)
--(axis cs:10161900,0.78428846520228)
--(axis cs:12791200,0.779681398443955)
--(axis cs:15436300,0.766998512716977)
--(axis cs:18095500,0.747593424491595)
--(axis cs:20761400,0.737841291706481)
--(axis cs:23438300,0.700087242447125)
--(axis cs:26121000,0.69847617964234)
--(axis cs:28808500,0.704596781122365)
--(axis cs:31499200,0.685984078429252)
--(axis cs:34190600,0.641279276916962)
--(axis cs:36886000,0.646667211620689)
--(axis cs:39583800,0.642600265789385)
--(axis cs:42290900,0.605611819833971)
--(axis cs:45671800,0.594959365841848)
--(axis cs:48377100,0.613166911869181)
--(axis cs:51088600,0.579364546967716)
--(axis cs:53795700,0.577383893799544)
--(axis cs:56501800,0.572801835869257)
--(axis cs:59206600,0.599918761607678)
--(axis cs:61911900,0.58349889783935)
--(axis cs:64618400,0.577880553560603)
--(axis cs:67325200,0.618339523407452)
--(axis cs:70036700,0.59503797818802)
--(axis cs:72746500,0.578396202118471)
--(axis cs:75455300,0.554486206229995)
--(axis cs:78164400,0.558760921152118)
--(axis cs:80877400,0.591641931148174)
--(axis cs:83590700,0.560121923181902)
--(axis cs:86305800,0.568885566696891)
--(axis cs:89697300,0.546768234853703)
--(axis cs:92415500,0.552800794373004)
--(axis cs:95129700,0.540400351978681)
--(axis cs:97842900,0.520325034887152)
--(axis cs:100559400,0.543596816518545)
--(axis cs:103278000,0.55115070448919)
--(axis cs:105994800,0.504107751473624)
--(axis cs:108711500,0.538019848883563)
--(axis cs:111431500,0.520778891287973)
--(axis cs:114150500,0.519216693593671)
--(axis cs:116864900,0.541195091903213)
--(axis cs:119579300,0.547351763727895)
--(axis cs:122294200,0.555383351513234)
--(axis cs:125012000,0.507230483800817)
--(axis cs:127726500,0.527472055997296)
--(axis cs:130444800,0.521636217505391)
--(axis cs:133843100,0.529052698508857)
--(axis cs:133843100,0.595152553931063)
--(axis cs:133843100,0.595152553931063)
--(axis cs:130444800,0.59398863928739)
--(axis cs:127726500,0.587050721772253)
--(axis cs:125012000,0.579213549467376)
--(axis cs:122294200,0.618155537381363)
--(axis cs:119579300,0.614707467380726)
--(axis cs:116864900,0.598725699281083)
--(axis cs:114150500,0.577025990203488)
--(axis cs:111431500,0.592925400121155)
--(axis cs:108711500,0.571383452807648)
--(axis cs:105994800,0.568462521537289)
--(axis cs:103278000,0.574122639459269)
--(axis cs:100559400,0.623699826053538)
--(axis cs:97842900,0.588333721554823)
--(axis cs:95129700,0.611535575996088)
--(axis cs:92415500,0.61412917770029)
--(axis cs:89697300,0.570471493486103)
--(axis cs:86305800,0.622710285094869)
--(axis cs:83590700,0.608546552886303)
--(axis cs:80877400,0.62596023830338)
--(axis cs:78164400,0.603711845232058)
--(axis cs:75455300,0.589804394288356)
--(axis cs:72746500,0.61596289497545)
--(axis cs:70036700,0.62984394878575)
--(axis cs:67325200,0.658304391070911)
--(axis cs:64618400,0.637674740848068)
--(axis cs:61911900,0.621490880803429)
--(axis cs:59206600,0.647161659657753)
--(axis cs:56501800,0.615813594683131)
--(axis cs:53795700,0.600945226266608)
--(axis cs:51088600,0.61021898861555)
--(axis cs:48377100,0.656669542518811)
--(axis cs:45671800,0.619934447919596)
--(axis cs:42290900,0.641041030386014)
--(axis cs:39583800,0.662087320150656)
--(axis cs:36886000,0.679820226462339)
--(axis cs:34190600,0.679619036659827)
--(axis cs:31499200,0.725422235067242)
--(axis cs:28808500,0.751501838488966)
--(axis cs:26121000,0.742651624042532)
--(axis cs:23438300,0.726055732966729)
--(axis cs:20761400,0.769093553577835)
--(axis cs:18095500,0.787696631394281)
--(axis cs:15436300,0.797621260363866)
--(axis cs:12791200,0.814285855652429)
--(axis cs:10161900,0.819126042992821)
--(axis cs:7548700,0.874098725500517)
--(axis cs:4960400,0.868553121166769)
--(axis cs:2430700,0.875745131634176)
--(axis cs:19300,0.886815249919891)
--cycle;

\addplot [line width=\linewidthothers, \bbrl_color, mark=*, mark size=0, mark options={solid}]
table {%
19300 0.872543349862099
2430700 0.863451066426933
4960400 0.852722718263976
7548700 0.854301290357398
10161900 0.803934302261723
12791200 0.797575957924721
15436300 0.779886922694898
18095500 0.758664230235379
20761400 0.754278479027989
23438300 0.71032639657826
26121000 0.717529620550182
28808500 0.728340097496784
31499200 0.708215382565405
34190600 0.661540705352869
36886000 0.655675435049833
39583800 0.648542970130412
42290900 0.624619937125578
45671800 0.604101961961219
48377100 0.641938244644909
51088600 0.599381602306958
53795700 0.589755517914321
56501800 0.601036229987375
59206600 0.626116980442121
61911900 0.599930596374107
64618400 0.609487812083835
67325200 0.645129921659915
70036700 0.608681704991634
72746500 0.601331475071507
75455300 0.573798216740987
78164400 0.579410633431951
80877400 0.610354174653724
83590700 0.587182251812397
86305800 0.59404871693716
89697300 0.563803723356827
92415500 0.602229924482118
95129700 0.583225818442926
97842900 0.563558628947071
100559400 0.603301115689709
103278000 0.564101025978704
105994800 0.537653351735156
108711500 0.557211305625348
111431500 0.563058875142172
114150500 0.557282939457748
116864900 0.576204769117461
119579300 0.583227385422347
122294200 0.588611638854098
125012000 0.548756147976469
127726500 0.565748366784577
130444800 0.5617800679703
133843100 0.570352787147099
};

\path [draw=\replace_color, fill=\replace_color, opacity=0.2]
(axis cs:11400,0.138345684856176)
--(axis cs:11400,0.134064530022442)
--(axis cs:1379400,0.339471691288054)
--(axis cs:2747400,0.375442225602455)
--(axis cs:4115400,0.405983118765107)
--(axis cs:5483400,0.408740319597314)
--(axis cs:6851400,0.421268195038181)
--(axis cs:8219400,0.434045176715506)
--(axis cs:9587400,0.446152769597757)
--(axis cs:10955400,0.468248471860812)
--(axis cs:12323400,0.432438909684768)
--(axis cs:13691400,0.437132495867754)
--(axis cs:15059400,0.448685353800344)
--(axis cs:16427400,0.436290268447067)
--(axis cs:17795400,0.434688547386728)
--(axis cs:19163400,0.447200838284464)
--(axis cs:20531400,0.469629564661818)
--(axis cs:21899400,0.429017398155276)
--(axis cs:23609400,0.413115442461336)
--(axis cs:24977400,0.408162083601432)
--(axis cs:26345400,0.417906194825062)
--(axis cs:27713400,0.404118665249724)
--(axis cs:29081400,0.415588863190334)
--(axis cs:30449400,0.431272886456103)
--(axis cs:31817400,0.409698904709614)
--(axis cs:33185400,0.402536266234573)
--(axis cs:34553400,0.412651891156177)
--(axis cs:35921400,0.405306183573991)
--(axis cs:37289400,0.397797155451466)
--(axis cs:38657400,0.415108267709746)
--(axis cs:40025400,0.405295666652736)
--(axis cs:41393400,0.402369895376895)
--(axis cs:42761400,0.413351863281978)
--(axis cs:44129400,0.415151829178539)
--(axis cs:45839400,0.401821128210948)
--(axis cs:47207400,0.412229081341279)
--(axis cs:48575400,0.410254962492479)
--(axis cs:49943400,0.388712662686183)
--(axis cs:51311400,0.389706330229753)
--(axis cs:52679400,0.391750296120471)
--(axis cs:54047400,0.386029591108359)
--(axis cs:55415400,0.388462113041235)
--(axis cs:56783400,0.381869550414252)
--(axis cs:58151400,0.384878932147856)
--(axis cs:59519400,0.389509521670016)
--(axis cs:60887400,0.38877421909433)
--(axis cs:62255400,0.402160382462173)
--(axis cs:63623400,0.388422631179254)
--(axis cs:64991400,0.387100541325595)
--(axis cs:66359400,0.381322371063111)
--(axis cs:68069400,0.385584400566217)
--(axis cs:68069400,0.400101287391014)
--(axis cs:68069400,0.400101287391014)
--(axis cs:66359400,0.417052863056205)
--(axis cs:64991400,0.408932341076482)
--(axis cs:63623400,0.408989403228929)
--(axis cs:62255400,0.419417908163942)
--(axis cs:60887400,0.410067585307117)
--(axis cs:59519400,0.414929796770461)
--(axis cs:58151400,0.400383603814238)
--(axis cs:56783400,0.40424207060716)
--(axis cs:55415400,0.402378365924072)
--(axis cs:54047400,0.398296474470884)
--(axis cs:52679400,0.408971269761539)
--(axis cs:51311400,0.404899602262563)
--(axis cs:49943400,0.403749526948423)
--(axis cs:48575400,0.422792149090706)
--(axis cs:47207400,0.426983106662162)
--(axis cs:45839400,0.421080062728153)
--(axis cs:44129400,0.43011719528771)
--(axis cs:42761400,0.428912564872382)
--(axis cs:41393400,0.412524507767234)
--(axis cs:40025400,0.416252833647113)
--(axis cs:38657400,0.433171527086496)
--(axis cs:37289400,0.412727584039348)
--(axis cs:35921400,0.424279250005011)
--(axis cs:34553400,0.421970275659382)
--(axis cs:33185400,0.419321177189303)
--(axis cs:31817400,0.420630674980935)
--(axis cs:30449400,0.443501419650851)
--(axis cs:29081400,0.430700243857759)
--(axis cs:27713400,0.414913451565772)
--(axis cs:26345400,0.428351805190705)
--(axis cs:24977400,0.421980925995727)
--(axis cs:23609400,0.4285515451396)
--(axis cs:21899400,0.445313194303729)
--(axis cs:20531400,0.486689756842464)
--(axis cs:19163400,0.465970950258202)
--(axis cs:17795400,0.459118041067772)
--(axis cs:16427400,0.461101317185891)
--(axis cs:15059400,0.466365184643034)
--(axis cs:13691400,0.461593393791552)
--(axis cs:12323400,0.46580310877183)
--(axis cs:10955400,0.498331863615652)
--(axis cs:9587400,0.462983931835768)
--(axis cs:8219400,0.449131676787541)
--(axis cs:6851400,0.445630382485149)
--(axis cs:5483400,0.441352582063246)
--(axis cs:4115400,0.44839694975235)
--(axis cs:2747400,0.416694960367749)
--(axis cs:1379400,0.366786936298013)
--(axis cs:11400,0.138345684856176)
--cycle;

\addplot [line width=\linewidthothers, \replace_color, mark=*, mark size=0, mark options={solid}]
table {%
11400 0.135752603411674
1379400 0.351485156686977
2747400 0.405650620275992
4115400 0.425077569208042
5483400 0.423905021294615
6851400 0.432145251413093
8219400 0.43895409716587
9587400 0.455086109547841
10955400 0.48213314525423
12323400 0.447986806905603
13691400 0.445147743271523
15059400 0.453945846230658
16427400 0.446733193925472
17795400 0.439409263700344
19163400 0.45247021082848
20531400 0.47662126509365
21899400 0.436774730207492
23609400 0.422233299809054
24977400 0.415058245807526
26345400 0.421818741170111
27713400 0.411921627228057
29081400 0.424650932626516
30449400 0.436978116881766
31817400 0.414647309213904
33185400 0.412897698801887
34553400 0.418499249849691
35921400 0.415736888583841
37289400 0.404248157955524
38657400 0.424571960839571
40025400 0.412398875437573
41393400 0.406988845788668
42761400 0.417509831346121
44129400 0.419605304575326
45839400 0.410294590630592
47207400 0.417427286642726
48575400 0.415390594244524
49943400 0.396889831524988
51311400 0.394083103275394
52679400 0.401427731510112
54047400 0.391205005594259
55415400 0.395035066752422
56783400 0.392661012106802
58151400 0.390486649858159
59519400 0.39926456447339
60887400 0.398173702389366
62255400 0.411444187224257
63623400 0.397600322736761
64991400 0.397568923290357
66359400 0.398325617336953
68069400 0.393122199863229
};

\path [draw=\res_color, fill=\res_color, opacity=0.2]
(axis cs:11400,0.138463359326124)
--(axis cs:11400,0.133799348026514)
--(axis cs:1379400,0.167023566551507)
--(axis cs:2747400,0.215369153782376)
--(axis cs:4115400,0.257376016948911)
--(axis cs:5483400,0.244544933273573)
--(axis cs:6851400,0.257226021078715)
--(axis cs:8219400,0.251158682844286)
--(axis cs:9587400,0.225773771393788)
--(axis cs:10955400,0.227854678001461)
--(axis cs:12323400,0.204529378188554)
--(axis cs:13691400,0.162040158961642)
--(axis cs:15059400,0.145405165637995)
--(axis cs:16427400,0.136133968857419)
--(axis cs:17795400,0.113639302482217)
--(axis cs:19163400,0.108021364556338)
--(axis cs:20531400,0.105390692276431)
--(axis cs:21899400,0.0988688096981432)
--(axis cs:23609400,0.0865636425120341)
--(axis cs:24977400,0.0951871475142944)
--(axis cs:26345400,0.0864228895103071)
--(axis cs:27713400,0.0871445504498427)
--(axis cs:29081400,0.0790965463318348)
--(axis cs:30449400,0.08551371189337)
--(axis cs:31817400,0.0757625361147341)
--(axis cs:33185400,0.0746951683657347)
--(axis cs:34553400,0.0739726330421435)
--(axis cs:35921400,0.0766708699393641)
--(axis cs:37289400,0.0719881431910201)
--(axis cs:38657400,0.0791835497103675)
--(axis cs:40025400,0.0669456321415139)
--(axis cs:41393400,0.077110978042888)
--(axis cs:42761400,0.0787361722801975)
--(axis cs:44129400,0.0732912719857475)
--(axis cs:45839400,0.0755025336885936)
--(axis cs:47207400,0.0688075987034865)
--(axis cs:48575400,0.0662769507468339)
--(axis cs:49943400,0.069219994421428)
--(axis cs:51311400,0.0646999666906906)
--(axis cs:52679400,0.069495718812789)
--(axis cs:54047400,0.0701197771593033)
--(axis cs:55415400,0.0724411558216946)
--(axis cs:56783400,0.0750026994615206)
--(axis cs:58151400,0.0685463736790958)
--(axis cs:59519400,0.0720490889014404)
--(axis cs:60887400,0.0649275812503903)
--(axis cs:62255400,0.0692705951565538)
--(axis cs:63623400,0.0653712580861544)
--(axis cs:64991400,0.0678200949271514)
--(axis cs:66359400,0.0661125363211373)
--(axis cs:68069400,0.0686158437501062)
--(axis cs:68069400,0.0723060755704838)
--(axis cs:68069400,0.0723060755704838)
--(axis cs:66359400,0.0723614874135347)
--(axis cs:64991400,0.0727053146811119)
--(axis cs:63623400,0.070238315328449)
--(axis cs:62255400,0.0741922252568144)
--(axis cs:60887400,0.0684909971149757)
--(axis cs:59519400,0.0816037636158279)
--(axis cs:58151400,0.0762058345853644)
--(axis cs:56783400,0.0790375239401303)
--(axis cs:55415400,0.0752212111295959)
--(axis cs:54047400,0.0770549040207463)
--(axis cs:52679400,0.0725487598326894)
--(axis cs:51311400,0.0692061042403315)
--(axis cs:49943400,0.0713654698653647)
--(axis cs:48575400,0.0724415655329914)
--(axis cs:47207400,0.0747760398768738)
--(axis cs:45839400,0.0818356719240299)
--(axis cs:44129400,0.0844770787490102)
--(axis cs:42761400,0.0833102740024111)
--(axis cs:41393400,0.0866140439603258)
--(axis cs:40025400,0.0717182035677251)
--(axis cs:38657400,0.0821204912572845)
--(axis cs:37289400,0.0805619340741642)
--(axis cs:35921400,0.0823698047359585)
--(axis cs:34553400,0.08363671448109)
--(axis cs:33185400,0.0785949666331326)
--(axis cs:31817400,0.0812779383548945)
--(axis cs:30449400,0.0904746009187106)
--(axis cs:29081400,0.0828692940073772)
--(axis cs:27713400,0.094685293234591)
--(axis cs:26345400,0.0922735519436945)
--(axis cs:24977400,0.102495069340787)
--(axis cs:23609400,0.0938483365495189)
--(axis cs:21899400,0.108548992886275)
--(axis cs:20531400,0.119107915265944)
--(axis cs:19163400,0.120958642151732)
--(axis cs:17795400,0.131865780682861)
--(axis cs:16427400,0.161339633833701)
--(axis cs:15059400,0.171327205802457)
--(axis cs:13691400,0.187086492168212)
--(axis cs:12323400,0.213693086297351)
--(axis cs:10955400,0.244776148316959)
--(axis cs:9587400,0.240417282341315)
--(axis cs:8219400,0.260634951567559)
--(axis cs:6851400,0.274693949725211)
--(axis cs:5483400,0.254958903160002)
--(axis cs:4115400,0.268835837728807)
--(axis cs:2747400,0.230675058759516)
--(axis cs:1379400,0.176543056964874)
--(axis cs:11400,0.138463359326124)
--cycle;

\addplot [line width=\linewidthtop, \res_color, mark=*, mark size=0, mark options={solid}]
table {%
11400 0.135679725557566
1379400 0.17210090556182
2747400 0.226816737369518
4115400 0.264074323759814
5483400 0.249774361713037
6851400 0.265490217814317
8219400 0.256687537579008
9587400 0.231358082491034
10955400 0.238279907194636
12323400 0.208891911811884
13691400 0.175355775338191
15059400 0.156360826023788
16427400 0.142912965716098
17795400 0.119752355295464
19163400 0.114593683352106
20531400 0.109354453825238
21899400 0.103497799327547
23609400 0.0897009014951825
24977400 0.0977235122080298
26345400 0.0890438993743107
27713400 0.0901659710382416
29081400 0.0800995478721793
30449400 0.0869701514915586
31817400 0.0783100035054003
33185400 0.0762241386734655
34553400 0.0779241031976636
35921400 0.0796457453632544
37289400 0.0750487748711528
38657400 0.0806199754750937
40025400 0.0694331100164413
41393400 0.081932458440973
42761400 0.0807378241927727
44129400 0.077746062332594
45839400 0.0788404480736361
47207400 0.0711580222778254
48575400 0.069061316311349
49943400 0.0698974530348057
51311400 0.0669822010406391
52679400 0.0711120954203018
54047400 0.0745684616588207
55415400 0.0741205052401796
56783400 0.0766699453372977
58151400 0.0720431214087204
59519400 0.0775232713977505
60887400 0.0669027780824003
62255400 0.072021835243456
63623400 0.0667764234992394
64991400 0.0697150518719738
66359400 0.0690504020278232
68069400 0.0703897947206465
};

\path [draw=\whole_color, fill=\whole_color, opacity=0.2]
(axis cs:11400,0.145469665061682)
--(axis cs:11400,0.13837069645524)
--(axis cs:1379400,0.268766339402646)
--(axis cs:2747400,0.279969036392868)
--(axis cs:4115400,0.287892997788731)
--(axis cs:5483400,0.256821322931955)
--(axis cs:6851400,0.26057866755086)
--(axis cs:8219400,0.236058717018935)
--(axis cs:9587400,0.219518645449767)
--(axis cs:10955400,0.212282330389946)
--(axis cs:12323400,0.203707408009908)
--(axis cs:13691400,0.187717692104109)
--(axis cs:15059400,0.183498514244272)
--(axis cs:16427400,0.179899077285163)
--(axis cs:17795400,0.163931967076083)
--(axis cs:19163400,0.158101147299437)
--(axis cs:20531400,0.155304078905128)
--(axis cs:21899400,0.159020381493718)
--(axis cs:23609400,0.137800028987003)
--(axis cs:24977400,0.14162766911959)
--(axis cs:26345400,0.139928453822861)
--(axis cs:27713400,0.132533923019685)
--(axis cs:29081400,0.138823614737769)
--(axis cs:30449400,0.128992706163116)
--(axis cs:31817400,0.126334337599529)
--(axis cs:33185400,0.119718323859736)
--(axis cs:34553400,0.123924563595779)
--(axis cs:35921400,0.118635113300531)
--(axis cs:37289400,0.12176157163838)
--(axis cs:38657400,0.115292871049834)
--(axis cs:40025400,0.109380447891918)
--(axis cs:41393400,0.108492399355672)
--(axis cs:42761400,0.115294245587726)
--(axis cs:44129400,0.0966520377797653)
--(axis cs:45839400,0.11853714785889)
--(axis cs:47207400,0.107536550326466)
--(axis cs:48575400,0.100199272073278)
--(axis cs:49943400,0.104988629293399)
--(axis cs:51311400,0.107762821842317)
--(axis cs:52679400,0.102939517726707)
--(axis cs:54047400,0.101716803475943)
--(axis cs:55415400,0.0982875731903059)
--(axis cs:56783400,0.0992026828228974)
--(axis cs:58151400,0.0993677602474292)
--(axis cs:59519400,0.0993231465160344)
--(axis cs:60887400,0.0981504698951155)
--(axis cs:62255400,0.106318476250111)
--(axis cs:63623400,0.098705594686658)
--(axis cs:64991400,0.0984262866646718)
--(axis cs:66359400,0.0972750559035762)
--(axis cs:68069400,0.0973925673069151)
--(axis cs:68069400,0.105292421942656)
--(axis cs:68069400,0.105292421942656)
--(axis cs:66359400,0.101869772880787)
--(axis cs:64991400,0.105811985573738)
--(axis cs:63623400,0.107043381358772)
--(axis cs:62255400,0.114639869457629)
--(axis cs:60887400,0.10420288889801)
--(axis cs:59519400,0.107630541935569)
--(axis cs:58151400,0.106752591932229)
--(axis cs:56783400,0.107090915390372)
--(axis cs:55415400,0.106135786064626)
--(axis cs:54047400,0.108168090222756)
--(axis cs:52679400,0.113777589524213)
--(axis cs:51311400,0.115856265307122)
--(axis cs:49943400,0.113880045549962)
--(axis cs:48575400,0.112115794070796)
--(axis cs:47207400,0.117410560714877)
--(axis cs:45839400,0.136829979335829)
--(axis cs:44129400,0.111763938210227)
--(axis cs:42761400,0.125904837507326)
--(axis cs:41393400,0.119488485799692)
--(axis cs:40025400,0.122061367634167)
--(axis cs:38657400,0.126170534147872)
--(axis cs:37289400,0.139113496592754)
--(axis cs:35921400,0.135950639513878)
--(axis cs:34553400,0.138706887723706)
--(axis cs:33185400,0.134156550887758)
--(axis cs:31817400,0.13972200409877)
--(axis cs:30449400,0.144739227750261)
--(axis cs:29081400,0.155420682445576)
--(axis cs:27713400,0.152500828024008)
--(axis cs:26345400,0.151801994457379)
--(axis cs:24977400,0.159233610078229)
--(axis cs:23609400,0.162127283307557)
--(axis cs:21899400,0.176319220337588)
--(axis cs:20531400,0.177237859892925)
--(axis cs:19163400,0.172538064881834)
--(axis cs:17795400,0.181691377052237)
--(axis cs:16427400,0.190900287870297)
--(axis cs:15059400,0.19546635487936)
--(axis cs:13691400,0.20053012095325)
--(axis cs:12323400,0.222190226919579)
--(axis cs:10955400,0.230835365929107)
--(axis cs:9587400,0.234087563508611)
--(axis cs:8219400,0.24859229567228)
--(axis cs:6851400,0.2743132700665)
--(axis cs:5483400,0.264706144431557)
--(axis cs:4115400,0.307761862273765)
--(axis cs:2747400,0.294229553488549)
--(axis cs:1379400,0.290576712228358)
--(axis cs:11400,0.145469665061682)
--cycle;

\addplot [line width=\linewidthothers, \whole_color, mark=*, mark size=0, mark options={solid}]
table {%
11400 0.140550941228867
1379400 0.280338161624968
2747400 0.285960365552455
4115400 0.301493925413524
5483400 0.260033844792133
6851400 0.266997364358275
8219400 0.242245220913276
9587400 0.224748875574642
10955400 0.223363445075747
12323400 0.213334033589356
13691400 0.196578003163287
15059400 0.190843354103865
16427400 0.187625504892167
17795400 0.172683100338799
19163400 0.167605024165284
20531400 0.164534531446021
21899400 0.169165414653062
23609400 0.150770898418703
24977400 0.149848575823911
26345400 0.145704719714861
27713400 0.142158829113005
29081400 0.148873673779279
30449400 0.138284288478273
31817400 0.132921317716898
33185400 0.126541834367948
34553400 0.130678707137609
35921400 0.126593378335232
37289400 0.131295207012444
38657400 0.119840951920983
40025400 0.112924500425097
41393400 0.111869992786149
42761400 0.119866720179055
44129400 0.104488730131094
45839400 0.126793666856809
47207400 0.109932140157087
48575400 0.104899571849305
49943400 0.109409396299477
51311400 0.111557829275981
52679400 0.106669740019225
54047400 0.104179584944661
55415400 0.100963387455631
56783400 0.102228402150109
58151400 0.101706524421661
59519400 0.102988521723767
60887400 0.100777579594676
62255400 0.1096287748163
63623400 0.102919734215131
64991400 0.102499635524621
66359400 0.0993402624987614
68069400 0.101751957629541
};

\path [draw=C2, fill=C2, opacity=0.2]
(axis cs:121600,3.80479203492427)
--(axis cs:121600,3.73156168365168)
--(axis cs:3283200,1.36978278353375)
--(axis cs:6444800,1.24700571458556)
--(axis cs:9606400,1.18337839655179)
--(axis cs:12889600,1.05431270780992)
--(axis cs:16051200,0.946338556711977)
--(axis cs:19212800,0.851413890051967)
--(axis cs:22374400,0.789144473175007)
--(axis cs:25657600,0.712467260331928)
--(axis cs:28819200,0.709725356915701)
--(axis cs:31980800,0.689623752227108)
--(axis cs:35142400,0.696171847303096)
--(axis cs:38425600,0.632410840679038)
--(axis cs:41587200,0.666663908187537)
--(axis cs:44748800,0.642123502746392)
--(axis cs:47910400,0.620822416085954)
--(axis cs:51193600,0.639560929146605)
--(axis cs:54355200,0.642495969015483)
--(axis cs:57516800,0.619975590152109)
--(axis cs:60800000,0.62808898037164)
--(axis cs:60800000,0.648106241339883)
--(axis cs:60800000,0.648106241339883)
--(axis cs:57516800,0.646328214827404)
--(axis cs:54355200,0.68936110154583)
--(axis cs:51193600,0.676068586517031)
--(axis cs:47910400,0.679408708270591)
--(axis cs:44748800,0.702033456813185)
--(axis cs:41587200,0.719923288336524)
--(axis cs:38425600,0.679686836546842)
--(axis cs:35142400,0.722544043669933)
--(axis cs:31980800,0.739446097984533)
--(axis cs:28819200,0.75619188536339)
--(axis cs:25657600,0.736244073815469)
--(axis cs:22374400,0.886302142226173)
--(axis cs:19212800,0.955059456599464)
--(axis cs:16051200,1.02520591841763)
--(axis cs:12889600,1.10204177598866)
--(axis cs:9606400,1.27880323891182)
--(axis cs:6444800,1.35487028125581)
--(axis cs:3283200,1.40251555538413)
--(axis cs:121600,3.80479203492427)
--cycle;

\addplot [line width=\linewidthothers, C2, mark=*, mark size=0, mark options={solid}]
table {%
121600 3.76470880939086
3283200 1.3836675407802
6444800 1.31314716406973
9606400 1.24444409306389
12889600 1.08420458041948
16051200 0.972295956475843
19212800 0.904955239296596
22374400 0.829339712058862
25657600 0.725394603228816
28819200 0.738271310438506
31980800 0.710075729616143
35142400 0.704034490000183
38425600 0.655718239603786
41587200 0.69260917057653
44748800 0.671657895257552
47910400 0.652241530704013
51193600 0.65737792634785
54355200 0.66780550950412
57516800 0.636656342954189
60800000 0.633445286042257
};

\path [draw=C5, fill=C5, opacity=0.2]
(axis cs:121600,0.344350426037314)
--(axis cs:121600,0.297275635454194)
--(axis cs:3283200,1.27484038116736)
--(axis cs:6444800,1.08311029481099)
--(axis cs:9606400,1.02082448733535)
--(axis cs:12889600,0.784172969137915)
--(axis cs:16051200,0.735356247193364)
--(axis cs:19212800,0.748358487521906)
--(axis cs:22374400,0.760952934667502)
--(axis cs:25657600,0.754282117466948)
--(axis cs:28819200,0.809970310639194)
--(axis cs:31980800,0.758343068781358)
--(axis cs:35142400,0.715621907579401)
--(axis cs:38425600,0.721600282565527)
--(axis cs:41587200,0.726578300946542)
--(axis cs:44748800,0.693618674780634)
--(axis cs:47910400,0.698326921412585)
--(axis cs:51193600,0.682975361003402)
--(axis cs:54355200,0.730863862920885)
--(axis cs:57516800,0.701623362947403)
--(axis cs:60800000,0.70015669482454)
--(axis cs:60800000,0.747075158885397)
--(axis cs:60800000,0.747075158885397)
--(axis cs:57516800,0.725645974900151)
--(axis cs:54355200,0.764886079166742)
--(axis cs:51193600,0.714083159475416)
--(axis cs:47910400,0.72445840611412)
--(axis cs:44748800,0.720485509355368)
--(axis cs:41587200,0.751397825661161)
--(axis cs:38425600,0.792660197605171)
--(axis cs:35142400,0.751885924503485)
--(axis cs:31980800,0.780947555580853)
--(axis cs:28819200,0.841779097972152)
--(axis cs:25657600,0.777629535681245)
--(axis cs:22374400,0.870635353391609)
--(axis cs:19212800,0.775164138349347)
--(axis cs:16051200,0.787651426960189)
--(axis cs:12889600,0.856663935982636)
--(axis cs:9606400,1.06761936491467)
--(axis cs:6444800,1.18923294108925)
--(axis cs:3283200,1.30364515307917)
--(axis cs:121600,0.344350426037314)
--cycle;

\addplot [line width=\linewidthothers, C5, mark=*, mark size=0, mark options={solid}]
table {%
121600 0.323304273372556
3283200 1.28549662743218
6444800 1.12679720552519
9606400 1.04393161335805
12889600 0.812814711115254
16051200 0.762089540816543
19212800 0.759411596369283
22374400 0.813394184923458
25657600 0.760411477676518
28819200 0.833003781942093
31980800 0.767429277841967
35142400 0.730538436510034
38425600 0.761927461809941
41587200 0.737427085859526
44748800 0.704824185704949
47910400 0.716781313037383
51193600 0.696047027228099
54355200 0.750406216628118
57516800 0.71415027015131
60800000 0.716444737899403
};

\path [draw=C4, fill=C4, opacity=0.2]
(axis cs:10400,6.0855532844842)
--(axis cs:10400,5.20282836847211)
--(axis cs:278000,6.2791590187658)
--(axis cs:547200,4.54735510734075)
--(axis cs:814800,3.49398819096885)
--(axis cs:1088400,2.64196516720372)
--(axis cs:1358400,2.31847262972878)
--(axis cs:1631200,2.07024326463656)
--(axis cs:1902400,2.01708391437781)
--(axis cs:2177200,1.82513399276728)
--(axis cs:2448400,1.73894748472487)
--(axis cs:2720000,1.64006477235506)
--(axis cs:2994400,1.55240585332913)
--(axis cs:3265600,1.40868231656464)
--(axis cs:3531600,1.33672097509778)
--(axis cs:3808000,1.30499625670822)
--(axis cs:4082800,1.34774307226453)
--(axis cs:4352000,1.19531787962533)
--(axis cs:4630800,1.06309967324508)
--(axis cs:4904000,1.05058876208434)
--(axis cs:5174400,1.04001871426135)
--(axis cs:5174400,1.17626256603007)
--(axis cs:5174400,1.17626256603007)
--(axis cs:4904000,1.17597911024906)
--(axis cs:4630800,1.23232026647126)
--(axis cs:4352000,1.31674449711493)
--(axis cs:4082800,1.4788223810207)
--(axis cs:3808000,1.44885316755701)
--(axis cs:3531600,1.54841250209959)
--(axis cs:3265600,1.55401137061881)
--(axis cs:2994400,1.69850010917702)
--(axis cs:2720000,1.79315949357586)
--(axis cs:2448400,1.86125780975537)
--(axis cs:2177200,2.07062377577343)
--(axis cs:1902400,2.17751908238719)
--(axis cs:1631200,2.35787468993753)
--(axis cs:1358400,2.52458518303483)
--(axis cs:1088400,3.12647216321417)
--(axis cs:814800,4.12910444775735)
--(axis cs:547200,5.5490253115944)
--(axis cs:278000,7.19405673535946)
--(axis cs:10400,6.0855532844842)
--cycle;

\addplot [line width=\linewidthothers, C4, mark=*, mark size=0, mark options={solid}]
table {%
10400 5.68909849173136
278000 6.70313429663551
547200 4.93865756302761
814800 3.98263101621425
1088400 2.92454994115677
1358400 2.40765976854585
1631200 2.19337607625283
1902400 2.07714974956149
2177200 1.9165557279179
2448400 1.81755415983014
2720000 1.70285661832264
2994400 1.61457971944511
3265600 1.47776651900986
3531600 1.47442737368691
3808000 1.3539815460538
4082800 1.40906115173018
4352000 1.26605364253762
4630800 1.11280398586644
4904000 1.09705731343121
5174400 1.10868351912468
};

\path [draw=C6, fill=C6, opacity=0.2]
(axis cs:10400,4.0647241344737)
--(axis cs:10400,3.26272092695599)
--(axis cs:281200,0)
--(axis cs:550800,0)
--(axis cs:820400,0)
--(axis cs:1088800,0)
--(axis cs:1360800,0)
--(axis cs:1627600,0)
--(axis cs:1898400,0)
--(axis cs:2171200,0)
--(axis cs:2438000,0)
--(axis cs:2704800,0)
--(axis cs:2975200,0)
--(axis cs:3249600,0)
--(axis cs:3522000,0)
--(axis cs:3795600,0)
--(axis cs:4064800,0)
--(axis cs:4336400,0)
--(axis cs:4605600,0)
--(axis cs:4876800,0)
--(axis cs:5149600,0)
--(axis cs:5149600,1.06039745086933)
--(axis cs:5149600,1.06039745086933)
--(axis cs:4876800,1.06425359566498)
--(axis cs:4605600,1.08250901372014)
--(axis cs:4336400,1.15641025493892)
--(axis cs:4064800,1.33580139309371)
--(axis cs:3795600,1.41620909394837)
--(axis cs:3522000,1.52942249466859)
--(axis cs:3249600,1.59864184384365)
--(axis cs:2975200,1.78861930955214)
--(axis cs:2704800,1.88508899034952)
--(axis cs:2438000,1.99754262403389)
--(axis cs:2171200,1.98863032104384)
--(axis cs:1898400,2.13528534421819)
--(axis cs:1627600,2.42744109010823)
--(axis cs:1360800,2.89434500051319)
--(axis cs:1088800,3.28842133296633)
--(axis cs:820400,3.50251363604348)
--(axis cs:550800,4.68940662513395)
--(axis cs:281200,5.85455284578309)
--(axis cs:10400,4.0647241344737)
--cycle;

\addplot [line width=\linewidthothers, C6, mark=*, mark size=0, mark options={solid}]
table {%
10400 3.65557215491732
281200 2.59223842526701
550800 2.27257991621537
820400 1.64950773137747
1088800 1.51400103640485
1360800 1.40326248528962
1627600 1.20080485415425
1898400 1.02996857469306
2171200 0.960668498518009
2438000 0.940401395653559
2704800 0.908777742460967
2975200 0.835410254670242
3249600 0.753425469836354
3522000 0.736375116964539
3795600 0.699115821951433
4064800 0.637380356839033
4336400 0.546942596237761
4605600 0.512597274965434
4876800 0.488853738810463
5149600 0.51172225634096
};

\end{axis}

\end{tikzpicture}

%% file: images/tex/0_result_table.tex
\begin{table}[!]
\centering
\renewcommand{\arraystretch}{1.5}
\setlength{\aboverulesep}{0pt}
\setlength{\belowrulesep}{0pt}

\resizebox{0.9\linewidth}{!} {
\begin{tabular}{l|cc|cc}
\toprule

\multirow{3}{*}{\textbf{Methods}} & \multicolumn{2}{c|}{\textbf{Multi-box}}  & \multicolumn{2}{c}{\textbf{Dual-arm}} \\[-3pt]
  & PT [s] & Success  & PT [s] & Success    \\
  & $\bm{\downarrow}$ & $\bm{\uparrow}$ & $\bm{\downarrow}$  &  $\bm{\uparrow}$       \\

\hline
ST-RRT* & 1.0  & \textbf{0.928} & 5.0 & 0.392 \\ 
RRT-Connect & 0.609  &0.587 & 4.229 & 0.204\\
\hline
\replaced{ERL + DMPs}         & \multirow{2}{3em}{\centering \added{0.011}}  & \added{0.092} & \multirow{2}{3em}{\centering \added{0.010}} & \added{0.002}\\
\replaced{ERL + DMPs + Residuals} &   & \added{0.584} &  & \added{0.168}\\
\hline
BBRL & \multirow{4}{3em}{\centering \textbf{0.010}} & 0.731 & \multirow{4}{3em}{\centering \textbf{0.0098}}  & 0.133 \\
Full Residual&  & 0.881  &  &  0.674\\
Partial Replacement &  & 0.812 &  & 0.299\\
MoRe-ERL (ours) &  & 0.889 &  & \textbf{0.767}\\
\bottomrule
\end{tabular}
}
\caption{Results of MoRe-ERL and baseline methods in both environments, evaluated by planning time (PT) and success rate. The arrow $\bm{\downarrow}$ indicates lower is better, and $\bm{\uparrow}$ indicates higher is better. \replaced{An episode is successful if the robot reaches the goal region (radius 0.1) without collisions. Sampling-based results are averaged over 10 seeds, RL results over 8 seeds.}
}
\label{table:multi-box}
\end{table}

%% file: tex/5_conclusions.tex
\section{Limitations and Future Works}
We proposed MoRe-ERL, a general residual learning framework tailored for episodic reinforcement learning. MoRe-ERL can be built on top of any method within the ERL category and demonstrates significant improvements over learning from scratch. By identifying the crucial intervals within the reference trajectory and applying residual learning to these segments, MoRe-ERL enhances both learning efficiency and task performance.
However, although MoRe-ERL is capable of refining the trajectories and achieved a high success rate, its performance highly depends on the quality of reference trajectories. In case of poor quality, MoRe-ERL has to make extra effort to escape from the pattern provided by the reference trajectory and can not demonstrate advantages. Furthermore, while MoRe-ERL offers a novel solution to refine trajectories, the timing to trigger and request such refinements is also crucial. \replaced{An additional high-level decision-making layer is needed to trigger refinement and select best suited refinement strategies.}

